\documentclass[10pt,twocolumn,letterpaper]{article}

 \usepackage[pagenumbers]{cvpr} 

\definecolor{cvprblue}{rgb}{0.21,0.49,0.74}
\usepackage[pagebackref,breaklinks,colorlinks,allcolors=cvprblue]{hyperref}
\usepackage{bm}
\usepackage{multirow}
\usepackage[table]{xcolor} 
\usepackage{colortbl}

\usepackage{amsmath}
\usepackage{wrapfig}
\usepackage{algorithm}
\usepackage{algcompatible}

\usepackage{multirow}
\usepackage{multicol}
\usepackage{threeparttable}
\usepackage{booktabs}

\usepackage{longtable}   
\usepackage{array}       
\usepackage{siunitx}     
\usepackage{paracol}

\usepackage{adjustbox}
\usepackage{amsmath}
\newcommand{\best}[1]{\textbf{#1}}
\newcommand{\second}[1]{\underline{#1}}
\newcommand{\one}[1]{\cellcolor{blue!15}{\strut #1}} 
\newcommand{\two}[1]{\cellcolor{green!10}{\strut #1}}

\newtheorem{lemma}{Lemma}

\newcommand{\mymethod}{DDTime}

\definecolor{americanrose}{rgb}{1.0, 0.01, 0.24}
\usepackage{pifont}

\def\paperID{22532} 
\def\confName{CVPR}
\def\confYear{2026}

\title{DDTime: Dataset Distillation with Spectral Alignment and Information Bottleneck for Time-Series Forecasting}

\author{
Yuqi Li\textsuperscript{1}\thanks{Equal Contribution.} \qquad
Kuiye Ding\textsuperscript{1}\footnotemark[1] \qquad
Chuanguang Yang\textsuperscript{2}\thanks{Corresponding Author.} \qquad
Hao Wang\textsuperscript{3}\\
Haoxuan Wang\textsuperscript{4} \qquad
Huiran Duan\textsuperscript{1} \qquad
Junming Liu\textsuperscript{5} \qquad
Yingli Tian\textsuperscript{1}\footnotemark[2]
\\
\textsuperscript{1}The City University of New York, CUNY\\
\textsuperscript{2}Institute of Computing Technology, Chinese Academy of Sciences \\
\textsuperscript{3}Zhejiang University\\
\textsuperscript{4}University of Illinois at Chicago\\
\textsuperscript{5}Tongji University\\
}

\begin{document}
\maketitle

\begin{abstract}
Time-series forecasting is fundamental across many domains, yet training accurate models often requires large-scale datasets and substantial computational resources. 
Dataset distillation offers a promising alternative by synthesizing compact datasets that preserve the learning behavior of full data. 
However, extending dataset distillation to time-series forecasting is non-trivial due to two fundamental challenges: 
\ding{182} \emph{temporal bias from strong autocorrelation}, which leads to distorted value-term alignment between teacher and student models; and 
\ding{183} \emph{insufficient diversity among synthetic samples}, arising from the absence of explicit categorical priors to regularize trajectory variety. 

In this work, we propose \emph{\mymethod }, a lightweight and plug-in distillation framework built upon first-order condensation decomposition. 
To tackle \textbf{Challenge \ding{182}}, it revisits value-term alignment through temporal statistics and introduces a \emph{frequency-domain alignment} mechanism to mitigate autocorrelation-induced bias, ensuring spectral consistency and temporal fidelity. 
To address \textbf{Challenge \ding{183}}, we further design an \emph{inter-sample regularization} inspired by the information bottleneck principle, which enhances diversity and maximizes information density across synthetic trajectories. 
The combined objective is theoretically compatible with a wide range of condensation paradigms and supports stable first-order optimization. 
Extensive experiments on 20 benchmark datasets and diverse forecasting architectures demonstrate that \emph{\mymethod} consistently outperforms existing distillation methods, achieving about 30\% relative accuracy gains while introducing about 2.49\% computational overhead. 
All code and distilled datasets will be released.
\end{abstract}
    
\section{Introduction}
\label{sec:intro}

\begin{figure}[t]
  \centering
  \includegraphics[width=1.0\linewidth]{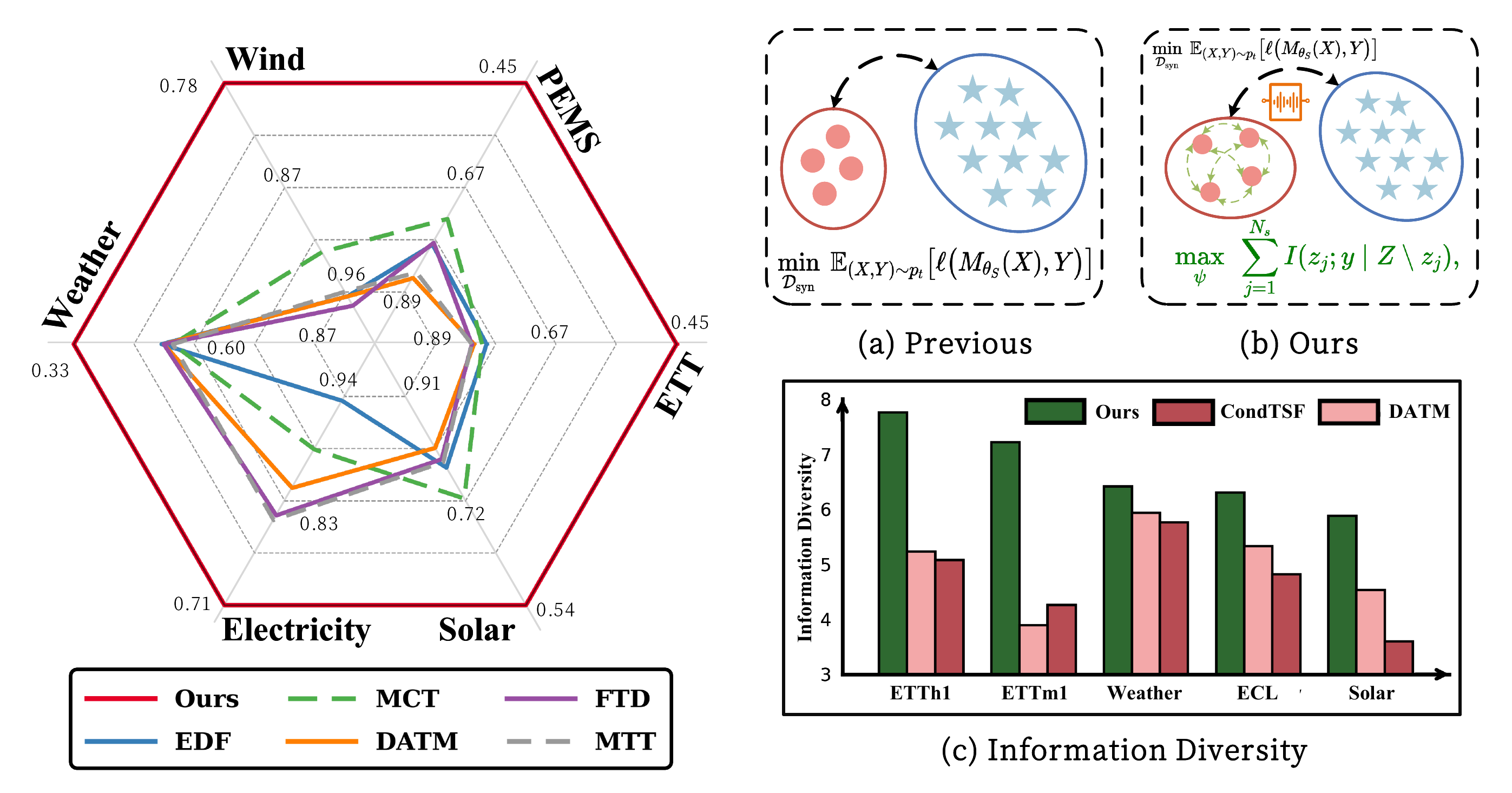}
  \caption{Left: Performance of \mymethod. Average results (MSE) are
reported following CondTSF~\cite{ding2024condtsf}, see details in Table~\ref{tab:average}. Right: (a)\&(b): Unlike previous distillation frameworks that operate purely in the temporal domain, our method introduces frequency-domain alignment and diversity regularization to enrich synthetic samples. (c) Quantitative comparison of data diversity shows that our method achieves consistently higher diversity scores than CondTSF ~\cite{ding2024condtsf} and DATM~\cite{guo2024datm}.}
  \label{fig:idea}
\end{figure}

Recent advances in time-series forecasting (TSF)~\cite{ansari2024chronos, goswami2024moment, timesfm, shi2024timemoe, liu2025sundial, aksu2024giftevalbenchmarkgeneraltime, shen2025visiontspp, ding2025kairosunifiedtraininguniversal} have been accompanied by a rapid growth in dataset scale, making model training and adaptation increasingly computationally demanding. Dataset distillation (DD)~\cite{wang2020datasetdistillation} provides an alternative approach by synthesizing compact datasets that preserve the learning behavior of models trained on full data, substantially reducing training time and memory consumption~\cite{maclaurin2015gradient}. Although DD has demonstrated success in computer vision~\cite{cui2022dc, wu2025robust, du2023ftd, wang2025edf} and graph~\cite{gao2025graph, zhang2024geom} domain, extending it to TSF is non-trivial~\cite{timeclass, miao2025timedc, ding2024condtsf} because time series data exhibit autocorrelation across time steps~\cite{wu2021autoformer, ding2025dualsg, ding2025timemosaic}, temporal autocorrelation~\cite{wang2025fredf}, and multi-scale temporal structures~\cite{wang2023timemixer, wang2024timemixer++, hu2025adaptive}. Recent progress in foundation models for TSF further highlights the need for specialized condensation frameworks that account for temporal statistics instead of treating sequences as independent samples.

A principled development in time-series data distillation is the first-order decomposition introduced by CondTSF~\cite{ding2024condtsf}, which reformulates the condensation objective into two complementary components: a parameter term that aligns the student and teacher in parameter space, and a value term that enforces consistency between their predictions. The core challenge, however, lies in the statistical structure of the labels. The temporal domain value term performs pointwise alignment between the student and teacher predictions. Such alignment mixes trend, seasonal, and high-frequency components within a single loss, which over-penalizes dependencies across forecasting horizons and consequently magnifies the bias in teacher–student consistency objectives~\cite{wang2025fredf}. In addition, unlike classification tasks where class labels inherently ensure diversity among synthetic examples~\cite{cui2025optical, wang2025NCFM, yu2023review, geng2023survey, lei2023survey}, TSF lacks categorical priors, causing the synthetic trajectories to become redundant and thereby reducing their overall information density.

To address these challenges, we propose \textit{\mymethod}, a lightweight and plug-and-play framework for data distillation in TSF. As illustrated in Figure~\ref{fig:idea}, \textit{\mymethod} incorporates frequency-domain alignment and diversity regularization to synthesize more informative and diverse trajectories. Building upon the first-order condensation decomposition established in prior work~\cite{ding2024condtsf}, \textit{\mymethod} revisits the value term from the perspective of label statistics. Specifically, we adapt frequency-domain alignment, which was originally employed in forecasting models to decorrelate horizon-wise dependencies for teacher–student alignment in data distillation. This design ensures spectral consistency while preserving fine-grained temporal fidelity~\cite{wang2025fredf}.

Furthermore, motivated by the information bottleneck principle~\cite{tishby2000informationbottleneckmethod}, \textit{\mymethod} introduces an inter-sample regularization that encourages diversity among synthetic trajectories. This design enhances the information density of the distilled dataset and compensates for the absence of explicit categorical priors in TSF~\cite{ding2024condtsf, miao2025timedc}. Because \textit{\mymethod} operates directly on the value-term optimization shared across different formulations, it is theoretically compatible with a wide range of dataset distillation paradigms. Within our design, the gradient term is instantiated through trajectory experts, making \textit{\mymethod} particularly effective when integrated into trajectory-matching based condensation methods where both terms are jointly optimized. While it can also be incorporated into other distillation paradigms that focus primarily on the value term, the performance gains are less pronounced, reflecting the complementary nature of the two-term formulation in time-series distillation.

Our contributions are as follows:
\begin{enumerate}[leftmargin=10pt, topsep=0pt, itemsep=1pt, partopsep=1pt, parsep=1pt]
\item We refine the value term of time-series dataset distillation by introducing frequency-domain alignment, which mitigates horizon dependency bias and improves spectral consistency between teacher and student predictions.  
\item We propose an inter-sample regularization strategy inspired by the information bottleneck principle to promote data diversity and increase information density within the synthetic dataset, compensating for the lack of categorical priors in time-series data.
\item We conduct comprehensive experiments across state-of-the-art TSF models on 20 benchmark datasets, demonstrating that our method consistently achieves strong performance and generalization. Although the theoretical decomposition is derived under simplified assumptions, our framework consistently exhibits robust empirical effectiveness and practical applicability across diverse architectures and datasets.
\end{enumerate}

\section{Related Work}
\label{sec:related}

\paragraph{Time-Series Forecasting (TSF).}
TSF aims to predict future values from historical temporal observations. Recent advances can be broadly categorized into three families: (1) \textit{Transformer-based models}, which adapt attention mechanisms from NLP to temporal dependencies~\cite{wu2021autoformer, patchtst, ding2025timemosaic,chen2024pathformer, ansari2024chronos}; 
(2) \textit{MLP-based models}, such as DLinear~\cite{dlinear} and TimeMixer~\cite{wang2023timemixer}, which achieve competitive accuracy with simpler architectures; and 
(3) \textit{Patch-based models}, which represent sequences as temporal patches for cross-patch interaction learning~\cite{patchmlp,liu2025sundial,woo2024moirai,patchwise,stitsyuk2025xpatch, ding2025dualsg}. Our framework intentionally accounts for the diversity of forecasting architectures and is designed to remain compatible with a wide range of TSF frameworks.

\paragraph{Dataset Distillation (DD).}
DD aims to synthesize a compact synthetic dataset that preserves the learning behavior of the original data. 
Existing methods can be grouped into three main classes. 
(1) \textit{Coreset selection}~\cite{bachem2017practical,chen2012super,sener2017active,tsang2005core} selects representative subsets directly from real data. 
(2) \textit{Matching-based approaches} minimize specific metrics between real and synthetic data, including gradient~\cite{dc2021,kim2022dataset,zhang2023accelerating}, feature or distribution~\cite{wang2022cafe,zhao2023dataset,zhao2023idm, wang2025NCFM}, and trajectory matching~\cite{cazenavette2022mtt,cui2023tesla,du2023ftd,guo2024datm,du2024sequential, zhong2025mct}. 
(3) \textit{Kernel-based methods}~\cite{lee2022dataset,kip} provide closed-form solutions via kernel ridge regression, simplifying the bilevel optimization problem.

\paragraph{Dataset Distillation for Time-Series Forecasting.}
Most existing DD studies focus on static domains~\cite{sachdeva2023datadistillationsurvey, timeclass}. 
In contrast, TSF involves continuous-valued outputs and temporal dependencies, making direct extensions of image-based DD non-trivial. 
\textit{CondTSF}~\cite{ding2024condtsf} bridges this gap by introducing a one-line plugin that jointly enforces \textit{value-term} and \textit{gradient-term} alignment between real and synthetic sequences, and can be integrated into parameter-matching-based distillation frameworks~\cite{wang2025edf, cazenavette2022mtt, guo2024datm}. Another recent study~\cite{miao2025timedc} further investigates time-series distillation by aligning decomposed temporal patterns and matching long-term training trajectories through an expert buffer.
Building on this direction, our work further explores the trajectory-matching paradigm to better capture the temporal optimization dynamics inherent in forecasting models.
\begin{figure*}[t]
  \centering
  \includegraphics[width=1.0\linewidth]{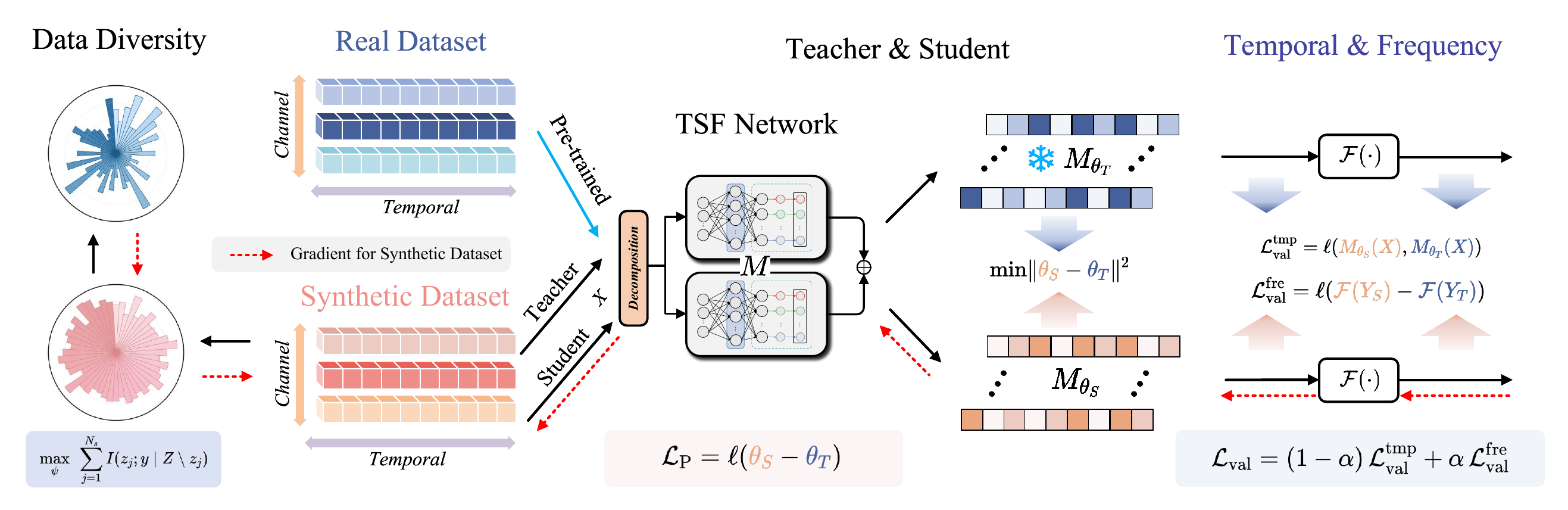}
\caption{Overall framework of the proposed \textit{\mymethod}. The framework consists of three key components: (\textit{i}) a {Data Diversity} module that enhances inter-sample diversity by increasing the pairwise KL divergence among synthetic sequences, the rose plots on the left visualize the KL divergence magnitude of each sample as a sector, (\textit{ii}) a {Teacher–Student} paradigm that enforces parameter matching between learners trained on real and synthetic data, and (\textit{iii}) a joint {Temporal–Frequency Alignment} loss that balances time-domain and spectral-domain supervision via the weighting coefficient $\alpha$. The TSF network illustrated is the sDLinear architecture~\cite{dlinear}. Here, $\mathcal{F}(\cdot)$ denotes the discrete Fourier transform (DFT).}
  \label{fig:framework}
\end{figure*}

\section{Method}

\subsection{Problem Definition.}
\label{sc:pb}

Let the real training set be
$\mathcal{D}_{\mathrm{real}}=\{(X_i,Y_i)\}_{i=1}^{N_r}$,
where $X_i\in\mathbb{R}^{L\times d}$ is an input window
(with context length $L$ and $d$ variables) and
$Y_i\in\mathbb{R}^{T\times d}$ is the corresponding target segment
(with prediction horizon $T$). We seek a compact synthetic set
$\mathcal{D}_{\mathrm{syn}}=\{(X_j^{(s)},Y_j^{(s)})\}_{j=1}^{N_s}$ with $N_s\!\ll\!N_r$,
and define the condensation ratio $r=N_s/N_r$.
We use $\mathbb{E}[\cdot]$ to denote expectation w.r.t. the specified distribution and
$\ell(\cdot,\cdot)$ to denote a per-sample forecasting loss.

Given a forecasting model $M_\theta:\mathbb{R}^{L\times d}\!\to\!\mathbb{R}^{T\times d}$,
let the \emph{teacher} parameters be obtained by training on $\mathcal{D}_{\mathrm{real}}$:
\begin{equation}
\theta_T \;=\; \arg\min_\theta~ \mathbb{E}_{(X,Y)\sim p_r}\big[\ell\big(M_\theta(X),Y\big)\big],
\end{equation}
and the \emph{student} parameters by training on $\mathcal{D}_{\mathrm{syn}}$:
\begin{equation}
\theta_S \;=\; \arg\min_\theta~ \mathbb{E}_{(X^{(s)},Y^{(s)})\sim p_s}\big[\ell\big(M_\theta(X^{(s)}),Y^{(s)}\big)\big],
\end{equation}
where $p_r$ and $p_s$ denote the distributions of real and synthetic data, respectively.
Our ultimate objective is \emph{test performance}: we want the student to perform well on the test distribution $p_t$,

\begin{equation}
\min_{\mathcal{D}_{\mathrm{syn}}}~ \mathbb{E}_{(X,Y)\sim p_t}\big[\ell\big(M_{\theta_S}(X),Y\big)\big].
\label{eq:test-risk}
\end{equation}
However, $p_t$ (and thus Eq.~\eqref{eq:test-risk}) is not directly optimizable during distillation.

\noindent\begin{lemma}[First-order condensation decomposition~\cite{ding2024condtsf}]
\label{thm:firstorder}
Under a first-order Taylor approximation of $M_\theta$ around $\theta_T$, and assuming Lipschitz continuity of $\ell$,
the intractable test objective in Eq.~\eqref{eq:test-risk} can be upper-bounded by two \emph{optimizable} terms that depend only on quantities available during condensation:
\begin{equation}
\begin{aligned}
\mathbb{E}_{(X,Y)\sim p_t}\!
\big[\ell(M_{\theta_S}(X),Y)\big]
\;\lesssim\;
&\underbrace{\|\theta_S - \theta_T\|^2}_{\textbf{Parameter Term}} \\[3pt]
&\hspace{-3.4em}+\;
\underbrace{
\mathbb{E}_{X\sim p_s}\!
\big[\ell(M_{\theta_S}(X^{s}),\,M_{\theta_T}(X^{s}))\big]
}_{\textbf{Value Term}}.
\end{aligned}
\label{eq:two-term}
\end{equation}

\noindent
Where $\|\cdot\|$ is a suitable norm and the hidden constants depend on the local Lipschitz smoothness of $M_\theta$ and $\ell$.
\end{lemma}

The distillation objective in Eq.~\eqref{eq:test-risk} is intractable to optimize directly, since $p_t$ is inaccessible during condensation,
\textbf{Lemma~\ref{thm:firstorder}} resolves this intractability via a first-order linearization of $M_\theta$, 
transforming the implicit test-level goal into two explicit, optimizable terms:
\begin{itemize}[leftmargin=12pt, topsep=2pt]
    \item[(1)] \textbf{Parameter Term}: aligns the student and teacher parameters, encouraging similar optimization end points.
    \item[(2)] \textbf{Value Term}: enforces prediction consistency on synthetize-data inputs $p_s$.
\end{itemize}
\noindent\textit{Remark.}
Following \cite{ding2024condtsf}, the originally introduced \emph{Gradient Term} can be upper-bounded by the parameter discrepancy $|\theta_S - \theta_T|^2$ under a local linearity assumption; thus, our \emph{Parameter Term} serves as its practical surrogate.

Together, these terms replace the unavailable test objective in Eq.~\eqref{eq:test-risk}, 
enabling optimization within the training process while approximating test-level generalization.

\textbf{Implementation.}
We refer to $M_{\theta_T}$ trained on $\mathcal{D}_{\mathrm{real}}$ as the \emph{teacher} and
$M_{\theta_S}$ trained on $\mathcal{D}_{\mathrm{syn}}$ as the \emph{student}.
In practice, the \textit{Parameter Term} can be optimized by aligning trajectories or final weights against a fixed teacher,
while the \textit{Value Term} is optimized by matching teacher–student predictions on inputs drawn from $p_r$ (or minibatches thereof). The \textit{Parameter Term} follows the implementation of CondTSF~\cite{ding2024condtsf}, which adopts a parameter trajectory alignment strategy, formulated as
\begin{equation}
\mathcal{L}_{\mathrm{P}}
=\min_{\text{expert}}\!
\frac{\|\theta_S-\theta_T^{(e)}\|_2^2}
     {\|\theta_T^{(e)}-\theta_T^{(0)}\|_2^2},
\end{equation}
where $\theta_S$ denotes the student parameters trained on
the synthetic dataset $\mathcal{D}_{\mathrm{syn}}$,
$\theta_T^{(0)}$ and $\theta_T^{(e)}$ represent the initial
and final parameters of a teacher model trained on real data,
respectively, and \emph{expert} indicates the selection among
multiple stored teacher trajectories. Our method serves as a lightweight plugin that can be plugged into various trajectory matching frameworks~\cite{cazenavette2022mtt,cui2023tesla,zhong2025mct}.
The detailed procedure is provided in Appendix~\ref{app:pt}.

\subsection{Revisiting the Value Term from Frequency}
\label{sc:fre}

The \textit{Value Term} in Lemma~\ref{thm:firstorder} enforces output consistency
between the student and the teacher on real inputs:
\begin{equation}
\mathcal{L}_{\mathrm{val}}^{\mathrm{tmp}}
=\mathbb{E}_{X\sim p_s}\big[\ell(M_{\theta_S}(X),M_{\theta_T}(X))\big].
\end{equation}
While this formulation aligns predictions in the temporal domain, it overlooks the fact that temporal signals are composed of multiple oscillatory components. As a result, purely time-domain matching may blur periodic structures and amplify the bias of direct forecasting (DF), especially when future steps are autocorrelated.

This phenomenon, termed \emph{label autocorrelation bias}~\cite{sun2025frele, wang2025fredf},
arises when future labels exhibit strong temporal correlation, causing the learning objective of direct forecasting to be biased
toward redundant or low-frequency components. Such bias limits the model’s ability to represent diverse temporal dynamics, motivating
our frequency domain reformulation of the value term. Forecasting in the frequency domain has been shown to alleviate such bias by decorrelating future-step dependencies as the frequency representation exposes periodic and trend components that are difficult to model directly in the temporal domain.

\begin{lemma}[Decorrelation Property of FFT~\cite{wang2025fredf}]
\label{lm:fft-white}
Let $Y$ be a zero-mean wide-sense stationary process and $\mathcal{F}(Y)$ its discrete Fourier transform.
As the sequence length $\mathrm{T}\!\to\!\infty$, different frequency components become asymptotically uncorrelated:
\begin{equation*}
\mathbb{E}[F_k F^*_{k'}] =
\begin{cases}
S_Y(f_k), & \text{if } k=k',\\
0, & \text{if } k\neq k',
\end{cases}
\end{equation*}
where $S_Y(f_k)$ denotes the power spectral density of $Y$. This whitening effect implies that FFT serves as an orthogonal transformation that diagonalizes the label covariance, thereby removing temporal dependencies among future steps.
\end{lemma}

Leveraging this property, we define the \emph{frequency-domain value term} by aligning teacher and student predictions in the decorrelated spectral space:
\begin{equation}
\mathcal{L}_{\mathrm{val}}^{\mathrm{fre}}
=\mathbb{E}_{X\sim p_s}\big[\|\mathcal{F}(Y_S)-\mathcal{F}(Y_T)\|_1\big],
\end{equation}
where $Y_S=M_{\theta_S}(X)$ and $Y_T=M_{\theta_T}(X)$. The FFT acts as a whitening transform that mitigates label autocorrelation bias, yielding a more balanced objective across frequency components.

Since the FFT is differentiable and can be optimized via gradient descent~\cite{zhou2022fedformer},
we combine the temporal and frequency domain terms using a trade-off coefficient
$\alpha\in[0,1]$:
\begin{equation}
\mathcal{L}_{\mathrm{val}}
=\mathbb{E}_{X\sim p_s}\Big[
(1-\alpha)\|Y_S-Y_T\|_2^2
+\alpha\|\mathcal{F}(Y_S)-\mathcal{F}(Y_T)\|_1
\Big].
\end{equation}
This formulation preserves temporal fidelity while enhancing spectral consistency,
thereby mitigating DF bias and stabilizing training.

\subsection{Maximizing Information Density via Inter-Sample Information Bottleneck (ISIB)}

The existing distillation method~\cite{ding2024condtsf} for time-series forecasting focuses on aligning the student with the teacher but overlooks the relationships among synthetic samples. This omission leads to redundant trajectories that share similar temporal or spectral patterns, thereby reducing the overall information density, that is, the amount of non-redundant and task-relevant information encoded per synthetic sample. A low-density synthetic dataset covers only limited temporal modes
and weakens the student’s generalization.

\textbf{Motivation.}
From an information-theoretic view~\cite{tishby2000informationbottleneckmethod, pmlr-v97-poole19a},
an effective synthetic dataset should contain sufficient information about the real distribution $p_r$
while minimizing internal redundancy.
Let $I(X_i^{(s)}; X_j^{(s)})$ denote the mutual information between two synthetic samples.
When this value is large, the dataset encodes repetitive content.
We therefore seek to maximize the overall task-relevant information
while minimizing pairwise redundancy, following the sufficiency–orthogonality principle in~\cite{ba2024exposing}.

\textbf{Dataset-level Information Bottleneck.}
We extend the Local–Global Information Loss framework~\cite{ba2024exposing}
from intra-sample features to inter-sample relations within $\mathcal{D}_{\mathrm{syn}}$.
Let $z_j = f_\psi(X_j^{(s)}, Y_j^{(s)})$ be the latent representation of each synthetic pair,
and $Z = \bigoplus_{j=1}^{N_s} z_j$ the joint synthetic representation.
Each $z_j$ should be both sufficient for the forecasting task and complementary to the others.
The theoretical objective can be expressed as
\begin{equation}
\max_{\psi}\;\sum_{j=1}^{N_s} I(z_j; y \mid Z \setminus z_j),
\label{eq:is-ib}
\end{equation}
where $y$ denotes the target variable from the real distribution.
To approximate this objective, we compute a smooth inverse weighting of the symmetric
Kullback–Leibler divergence~\cite{klrethinking} between the normalized probability representations of synthetic samples.
For each synthetic sample $i$, we obtain a temperature-scaled softmax distribution
$p_i = \mathrm{softmax}(x_i / \tau)$, and define
\begin{equation}
\mathcal{L}_{\mathrm{IS}}
=\frac{1}{S(S-1)}
\sum_{i<j}
\frac{1}{2}\Big(
\mathrm{KL}[p_i\|p_j]+\mathrm{KL}[p_j\|p_i]
\Big).
\end{equation}
This symmetric KL divergence penalizes redundant information
and promotes diversity among synthetic samples.
Minimizing $\mathcal{L}_{\mathrm{IS}}$ encourages each sample to contribute unique information
and increases the overall information density of $\mathcal{D}_{\mathrm{syn}}$. Eq. \eqref{eq:is-ib} measures the marginal contribution of each synthetic sample to the overall task-relevant information, conditioned on the others. We normalize each synthetic sample via a temperature-scaled softmax, ensuring comparable probability mass functions across samples and avoiding scale sensitivity.


\noindent\textit{Implementation.}
For numerical stability and smoother gradients, we use a monotone surrogate that
\emph{decreases} with divergence, minimizing
$\tfrac{1}{S(S-1)}\sum_{i<j}\exp\!\big(-\lambda_{\mathrm{div}}\cdot
\tfrac{1}{2}(\mathrm{KL}[p_i\|p_j]+\mathrm{KL}[p_j\|p_i])\big)$ in practice,
where $\lambda_{\mathrm{div}}>0$ is a tunable coefficient. The value of $\lambda_{\mathrm{div}}$ is set to 0.5.

\textbf{Global Bottleneck.}
Finally, a global aggregation module $G=g_\phi(Z)$ acts as an information bottleneck,
retaining task-relevant content while suppressing redundancy:
\begin{equation}
\mathcal{L}_{\mathrm{G}}
=\mathbb{E}_{G\sim E_\phi(G|Z)}[\mathrm{KL}[P_Z\|P_G]].
\end{equation}
In practice $\mathcal{L}_{\mathrm{G}}$ is optimized implicitly via shared parameters between $f\psi$ rather than added as an explicit regularizer.

\subsection{Overall Training Objective}

The overall optimization pipeline is illustrated in Figure~\ref{fig:framework}. Our distillation framework integrates the previously defined Parameter Term, frequency-aware Value Term,
and Inter-Sample Information Bottleneck into a unified objective:
\begin{equation}
\mathcal{L}_{\mathrm{total}}
=\mathcal{L}_{\mathrm{P}}
+ (1-\alpha)\,\mathcal{L}_{\mathrm{val}}^{\mathrm{tmp}}
+\alpha\,\mathcal{L}_{\mathrm{val}}^{\mathrm{fre}}
+\lambda_{\mathrm{IS}}\mathcal{L}_{\mathrm{IS}}
+\mathcal{L}_{\mathrm{G}}.
\end{equation}

Here, $\lambda_{\mathrm{IS}}$, and $\alpha$
are global balancing coefficients,
while $\alpha$ controls the trade-off between temporal fidelity
and frequency consistency as introduced in Sec.~\ref{sc:fre}. The $\mathcal{L}_{\mathrm{IS}}$ enhance inter-sample diversity. 


\begin{table*}[ht]
    \centering
    \vspace{-4pt}
    \caption{Results of \mymethod{} on five datasets. All values are averaged over five runs; lower MSE indicate better performance.}
    \label{tab:dd_main_exp}
    \vspace{-8pt}
    \setlength{\tabcolsep}{3mm}{\resizebox{0.98\textwidth}{!}{
    \begin{tabular}{l|ccc|ccc|ccc|ccc|ccc}
    \toprule
     Dataset & \multicolumn{3}{c|}{Weather}& \multicolumn{3}{c|}{Electricity }& \multicolumn{3}{c}{Solar-Energy} & \multicolumn{3}{c}{ExchangeRate} & \multicolumn{3}{c}{PEMS03}\\ 
    SynTS Count & 3 & 5 & 20 & 3 & 5 & 20 & 3 & 5 & 20 & 3 & 5 & 20 & 3 & 5 & 20 \\ 
    Ratio (\%) & 0.1 & 0.2 & 0.65 & 0.2 & 0.3 & 1.3 & 0.1 & 0.2 & 0.65 & 0.68 & 1.10  & 4.6 & 0.2 & 0.3  & 1.3 \\\midrule
    Random & 0.75{\scriptsize$\pm$0.07} & 0.76{\scriptsize$\pm$0.10} & 0.70{\scriptsize$\pm$0.03} & 1.11{\scriptsize$\pm$0.07} & 1.21{\scriptsize$\pm$0.15} & 1.15{\scriptsize$\pm$0.03} & 0.95{\scriptsize$\pm$0.03} & 0.94{\scriptsize$\pm$0.04} & 1.04{\scriptsize$\pm$0.05} & 1.31{\scriptsize$\pm$0.16} & 1.27{\scriptsize$\pm$0.13} & 1.34{\scriptsize$\pm$0.20} & 0.99{\scriptsize$\pm$0.08} & 0.98{\scriptsize$\pm$0.05} & 0.88{\scriptsize$\pm$0.09} \\ 
    \midrule
    DC   & 0.63{\scriptsize$\pm$0.07} & 0.61{\scriptsize$\pm$0.04} & 0.63{\scriptsize$\pm$0.02} &
            1.08{\scriptsize$\pm$0.14} & 1.07{\scriptsize$\pm$0.10} & 1.14{\scriptsize$\pm$0.06} &
            0.93{\scriptsize$\pm$0.08} & 0.85{\scriptsize$\pm$0.04} & 0.93{\scriptsize$\pm$0.02} &
            1.28{\scriptsize$\pm$0.09} & 1.20{\scriptsize$\pm$0.18} & 1.05{\scriptsize$\pm$0.20} &
            0.91{\scriptsize$\pm$0.09} & 0.83{\scriptsize$\pm$0.07} & 0.84{\scriptsize$\pm$0.05} \\
    KIP  & 0.73{\scriptsize$\pm$0.06} & 0.68{\scriptsize$\pm$0.05} & 0.68{\scriptsize$\pm$0.04} &
            1.04{\scriptsize$\pm$0.05} & 1.19{\scriptsize$\pm$0.11} & 1.11{\scriptsize$\pm$0.10} &
            0.85{\scriptsize$\pm$0.06} & 0.90{\scriptsize$\pm$0.04} & 0.87{\scriptsize$\pm$0.06} &
            0.95{\scriptsize$\pm$0.06} & 0.92{\scriptsize$\pm$0.09} & 1.45{\scriptsize$\pm$0.12} &
            0.89{\scriptsize$\pm$0.07} & 1.08{\scriptsize$\pm$0.08} & 0.81{\scriptsize$\pm$0.05} \\
    MTT  & 0.56{\scriptsize$\pm$0.01} & 0.55{\scriptsize$\pm$0.02} & 0.50{\scriptsize$\pm$0.01} &
            0.80{\scriptsize$\pm$0.01} & 0.82{\scriptsize$\pm$0.01} & 0.81{\scriptsize$\pm$0.02} &
            0.73{\scriptsize$\pm$0.02} & 0.79{\scriptsize$\pm$0.02} & 0.80{\scriptsize$\pm$0.01} &
            0.66{\scriptsize$\pm$0.05} & 0.60{\scriptsize$\pm$0.02} & 0.70{\scriptsize$\pm$0.09} &
            0.91{\scriptsize$\pm$0.01} & 0.91{\scriptsize$\pm$0.01} & 0.79{\scriptsize$\pm$0.03} \\
    FRePo& 0.54{\scriptsize$\pm$0.02} & 0.55{\scriptsize$\pm$0.02} & 0.53{\scriptsize$\pm$0.02} &
            1.02{\scriptsize$\pm$0.06} & 1.02{\scriptsize$\pm$0.05} & 0.92{\scriptsize$\pm$0.04} &
            0.69{\scriptsize$\pm$0.03} & 0.67{\scriptsize$\pm$0.04} & 0.78{\scriptsize$\pm$0.03} &
            1.10{\scriptsize$\pm$0.24} & 1.02{\scriptsize$\pm$0.07} & 0.90{\scriptsize$\pm$0.07} &
            0.69{\scriptsize$\pm$0.02} & 0.84{\scriptsize$\pm$0.05} & 0.64{\scriptsize$\pm$0.04} \\
    IDM  & 0.69{\scriptsize$\pm$0.04} & 0.63{\scriptsize$\pm$0.03} & 0.67{\scriptsize$\pm$0.07} &
            1.07{\scriptsize$\pm$0.08} & 1.04{\scriptsize$\pm$0.06} & 1.08{\scriptsize$\pm$0.13} &
            1.01{\scriptsize$\pm$0.05} & 0.94{\scriptsize$\pm$0.04} & 0.90{\scriptsize$\pm$0.05} &
            1.20{\scriptsize$\pm$0.08} & 1.28{\scriptsize$\pm$0.17} & 1.37{\scriptsize$\pm$0.05} &
            0.91{\scriptsize$\pm$0.07} & 0.95{\scriptsize$\pm$0.11} & 0.82{\scriptsize$\pm$0.04} \\
    TESLA& 0.52{\scriptsize$\pm$0.02} & 0.56{\scriptsize$\pm$0.01} & 0.57{\scriptsize$\pm$0.02} &
            0.98{\scriptsize$\pm$0.05} & 1.07{\scriptsize$\pm$0.01} & 1.07{\scriptsize$\pm$0.02} &
            0.80{\scriptsize$\pm$0.01} & 0.74{\scriptsize$\pm$0.01} & 0.82{\scriptsize$\pm$0.06} &
            0.91{\scriptsize$\pm$0.15} & 0.89{\scriptsize$\pm$0.12} & 0.97{\scriptsize$\pm$0.06} &
            0.78{\scriptsize$\pm$0.04} & 0.74{\scriptsize$\pm$0.02} & 0.73{\scriptsize$\pm$0.03} \\
    PP   & 0.49{\scriptsize$\pm$0.01} & 0.57{\scriptsize$\pm$0.00} & 0.52{\scriptsize$\pm$0.01} &
            0.88{\scriptsize$\pm$0.02} & 0.81{\scriptsize$\pm$0.03} & 0.84{\scriptsize$\pm$0.03} &
            0.86{\scriptsize$\pm$0.01} & 0.80{\scriptsize$\pm$0.01} & 0.80{\scriptsize$\pm$0.01} &
            0.72{\scriptsize$\pm$0.03} & 1.00{\scriptsize$\pm$0.08} & 0.70{\scriptsize$\pm$0.06} &
            0.86{\scriptsize$\pm$0.01} & 0.87{\scriptsize$\pm$0.02} & 0.82{\scriptsize$\pm$0.02} \\
    FTD  & 0.53{\scriptsize$\pm$0.02} & 0.55{\scriptsize$\pm$0.01} & 0.52{\scriptsize$\pm$0.01} &
            0.81{\scriptsize$\pm$0.02} & 0.81{\scriptsize$\pm$0.02} & 0.82{\scriptsize$\pm$0.01} &
            0.74{\scriptsize$\pm$0.02} & 0.84{\scriptsize$\pm$0.01} & 0.77{\scriptsize$\pm$0.02} &
            0.98{\scriptsize$\pm$0.06} & 0.86{\scriptsize$\pm$0.05} & 0.75{\scriptsize$\pm$0.06} &
            0.78{\scriptsize$\pm$0.02} & 0.75{\scriptsize$\pm$0.02} & 0.76{\scriptsize$\pm$0.02} \\
    DATM & 0.51{\scriptsize$\pm$0.01} & 0.48{\scriptsize$\pm$0.02} & 0.56{\scriptsize$\pm$0.01} &
            0.86{\scriptsize$\pm$0.02} & 0.81{\scriptsize$\pm$0.02} & 0.86{\scriptsize$\pm$0.02} &
            0.84{\scriptsize$\pm$0.01} & 0.81{\scriptsize$\pm$0.02} & 0.79{\scriptsize$\pm$0.01} &
            0.83{\scriptsize$\pm$0.03} & 1.04{\scriptsize$\pm$0.08} & 0.71{\scriptsize$\pm$0.05} &
            0.84{\scriptsize$\pm$0.04} & 0.86{\scriptsize$\pm$0.05} & 0.74{\scriptsize$\pm$0.05} \\
    MCT  & 0.64{\scriptsize$\pm$0.05} & 0.51{\scriptsize$\pm$0.03} & 0.48{\scriptsize$\pm$0.01} &
            0.90{\scriptsize$\pm$0.03} & 0.84{\scriptsize$\pm$0.06} & 0.90{\scriptsize$\pm$0.03} &
            0.73{\scriptsize$\pm$0.02} & 0.72{\scriptsize$\pm$0.01} & 0.74{\scriptsize$\pm$0.01} &
            1.01{\scriptsize$\pm$0.11} & 0.93{\scriptsize$\pm$0.12} & 0.99{\scriptsize$\pm$0.12} &
            0.72{\scriptsize$\pm$0.05} & 0.70{\scriptsize$\pm$0.04} & 0.80{\scriptsize$\pm$0.08} \\
    EDF  & 0.50{\scriptsize$\pm$0.01} & 0.59{\scriptsize$\pm$0.01} & 0.50{\scriptsize$\pm$0.01} &
            0.86{\scriptsize$\pm$0.04} & 0.86{\scriptsize$\pm$0.04} & 1.01{\scriptsize$\pm$0.03} &
            0.77{\scriptsize$\pm$0.04} & 0.78{\scriptsize$\pm$0.01} & 0.79{\scriptsize$\pm$0.03} &
            0.90{\scriptsize$\pm$0.05} & 0.96{\scriptsize$\pm$0.03} & 0.71{\scriptsize$\pm$0.05} &
            0.72{\scriptsize$\pm$0.02} & 0.69{\scriptsize$\pm$0.03} & 0.69{\scriptsize$\pm$0.01} \\
            
    \rowcolor{green!8}\textbf{\mymethod{} (Ours)} &
    \best{0.33{\scriptsize$\pm$0.01}} & \best{0.33{\scriptsize$\pm$0.01}} & \best{0.33{\scriptsize$\pm$0.01}} &
    \best{0.70{\scriptsize$\pm$0.02}} & \best{0.73{\scriptsize$\pm$0.02}} & \best{0.72{\scriptsize$\pm$0.02}} &
    \best{0.50{\scriptsize$\pm$0.01}} & \best{0.49{\scriptsize$\pm$0.01}} & \best{0.58{\scriptsize$\pm$0.00}} &
    \best{0.39{\scriptsize$\pm$0.04}} & \best{0.56{\scriptsize$\pm$0.04}} & \best{0.38{\scriptsize$\pm$0.05}} &
    \best{0.28{\scriptsize$\pm$0.01}} & \best{0.34{\scriptsize$\pm$0.01}} & \best{0.37{\scriptsize$\pm$0.00}} \\
    
    \midrule
     Whole Dataset & \multicolumn{3}{c|}{\textbf{0.21{\scriptsize$\pm$0.00}}}& \multicolumn{3}{c|}{\textbf{0.60{\scriptsize$\pm$0.02}}}& \multicolumn{3}{c}{\textbf{0.45{\scriptsize$\pm$0.01}}} & \multicolumn{3}{c}{\textbf{0.50{\scriptsize$\pm$0.07}}} & \multicolumn{3}{c}{\textbf{0.21{\scriptsize$\pm$0.00}}} \\ 
    \bottomrule
    \end{tabular}}
    }
    \vspace{-5pt}
\end{table*}

\section{Experiment}
\label{sec:exp}

\subsection{Setup}
\noindent\textbf{Datasets.}
We conduct experiments on twenty benchmark datasets, including ETT (ETTh1, ETTh2, ETTm1, ETTm2)~\cite{informer}, Weather~\cite{informer}, Traffic~\cite{pems_traffic}, Electricity~\cite{electricityloaddiagrams20112014_321}, Exchange-rate~\cite{lai2018modelinglongshorttermtemporal}, Solar-energy~\cite{wang2023timemixer},  PEMS (03, 04, 07, 08)~\cite{pems_traffic},  ILI~\cite{cdc_illness}, Wind (L1–L4)~\cite{ding2025timemosaic},Wike2000~\cite{qiu2024tfb}, and GlobalTemp~\cite{shi2024timemoe}. The dataset loading, preprocessing, and evaluation benchmarks follow the experimental settings of CondTSF~\cite{ding2024condtsf}. 

\noindent\textbf{Baseline.}
We evaluate \mymethod\ with two groups of baselines. \textit{Dataset distillation frameworks:} We include representative distillation methods originally designed for image classification, including EDF~\cite{wang2025edf}, MCT~\cite{zhong2025mct}, DATM~\cite{guo2024datm}, PP~\cite{li2023pp}, IDM~\cite{zhao2023idm}, FRePo~\cite{FRePo}, KIP~\cite{kip}, {DC}~\cite{dc2021}, {DSA}~\cite{zhao2021dsa}, 
{MTT}~\cite{cazenavette2022mtt}, {TESLA}~\cite{cui2023tesla}, 
and {FTD}~\cite{du2023ftd}, 
as well as {CondTSF}~\cite{ding2024condtsf}, which is specifically developed for time-series forecasting. \textit{Time-series forecasting models:} We conduct experiments across diverse TSF architectures, including SimpleTM~\cite{chen2025simpletm}, xPatch~\cite{stitsyuk2025xpatch}, FreTS~\cite{yi2023frequencydomain}, {DLinear}~\cite{dlinear}, {PatchTST}~\cite{patchtst}, 
{iTransformer}~\cite{liu2023itransformer}, CNN~\cite{ding2024condtsf}, MLP~\cite{ding2024condtsf} and LSTM~\cite{ding2024condtsf}. These models collectively cover common TSF paradigms such as decomposition, multi-scale learning, wavelet and frequency-domain analysis, as well as patch-based modeling, encompassing a wide spectrum of architectures including Transformers, CNNs, MLPs, and LSTMs.

\noindent\textbf{Metrics.}
We use mean squared error (MSE) and mean absolute error (MAE) as evaluation metrics for TSF, following CondTS~\cite{ding2024condtsf} and iTransformer~\cite{liu2023itransformer}.

\noindent\textbf{Implementation Details.}
All methods are implemented in PyTorch~\cite{paszke2019pytorch} and trained on a single NVIDIA RTX 5090 GPU. All time-series datasets are standardized and split into training, validation, and test sets with a ratio of 70\%, 15\%, and 15\%, respectively. We follow the unified sliding-window preprocessing adopted in prior work~\cite{ding2024condtsf}, where the input and prediction lengths are set to $T_{\text{in}}=24$ and $T_{\text{out}}=24$. Each synthetic sequence $\mathbf{X}_{\text{syn}}$ is optimized via differentiable distillation. For a fair comparison with the original~\cite{ding2024condtsf}, we adopt its best-reported hyperparameter settings, where the distillation interval is set to $5$ and the conditional update coefficient to $0.01$.

The learning rate for student parameters is set to $3\times10^{-4}$, and the learning rate for synthetic data updates is $0.1$. 
During distillation, expert trajectories are reconstructed from the teacher network’s replay buffers, and random trajectory segments are sampled for imitation. The student model is updated for twenty gradient steps on the synthetic data at each iteration. Model performance is evaluated every fifty iterations and averaged over 5 random seeds to ensure stability.

\begin{table}[h] 
    \setlength{\tabcolsep}{1pt}
    \scriptsize
    \caption{Results of \mymethod{} on 15 datasets, including ETTh1, ETTh2, ETTm1, ETTm2, PEMS03, PEMS04, PEMS07, PEMS08, WindL1, WindL2, WindL3, WindL4, Weather, Electricity, Solar dataset. Lower MSE/MAE indicate better performance.}
    \centering
    \begin{threeparttable}
    \resizebox{1\linewidth}{!}{
        \begin{tabular}{l*{6}{cc}@{}} 
             \toprule
                 \multicolumn{1}{c}{\multirow{2}{*}{Models}} & \multicolumn{2}{c}{\textbf{\mymethod}} & \multicolumn{2}{c}{EDF} & \multicolumn{2}{c}{MCT} & \multicolumn{2}{c}{DATM} & \multicolumn{2}{c}{FTD} & \multicolumn{2}{c}{MTT}\\
            & \multicolumn{2}{c}{\scalebox{0.8}{\textbf{Ours}}} &
            \multicolumn{2}{c}{\scalebox{0.8}{\citeyearpar{wang2025edf}}} & 
            \multicolumn{2}{c}{\scalebox{0.8}{\citeyearpar{zhong2025mct}}} & 
            \multicolumn{2}{c}{\scalebox{0.8}{\citeyearpar{guo2024datm}}}  & 
            \multicolumn{2}{c}{\scalebox{0.8}{\citeyearpar{du2023ftd}}}  & 
            \multicolumn{2}{c}{\scalebox{0.8}{\citeyearpar{cazenavette2022mtt}}}\\
            \cmidrule(lr){2-3} \cmidrule(lr){4-5} \cmidrule(lr){6-7} \cmidrule(lr){8-9} \cmidrule(lr){10-11} \cmidrule(lr){12-13} 
            \multicolumn{1}{c}{Metric}& MSE & MAE & MSE & MAE & MSE & MAE & MSE & MAE & MSE & MAE & MSE & MAE\\
            \toprule
            \multicolumn{1}{c}{ETT (all)} 
                & \best{0.448} & \best{0.470}
                & \second{0.796} & \second{0.686}
                & 0.804 & 0.691
                & 0.819 & 0.705
                & 0.823 & 0.702
                & 0.823 & 0.701 \\
            \multicolumn{1}{c}{PEMS (all)} 
                & \best{0.454} & \best{0.531}
                & 0.792 & 0.736
                & \second{0.738} & \second{0.709}
                & 0.862 & 0.773
                & 0.788 & 0.739
                & 0.850 & 0.774 \\
            \multicolumn{1}{c}{Wind (all)} 
                & \best{0.776} & \best{0.698}
                & 0.959 & 0.800
                & \second{0.921} & \second{0.779}
                & 0.959 & 0.804
                & 0.967 & 0.808
                & 0.955 & 0.803 \\
            \multicolumn{1}{c}{Weather} 
                & \best{0.330} & \best{0.373}
                & \second{0.525} & \second{0.510}
                & 0.550 & 0.518
                & 0.530 & 0.510
                & 0.530 & 0.518
                & 0.550 & 0.528 \\
            \multicolumn{1}{c}{Electricity} 
                & \best{0.715} & \best{0.683}
                & 0.938 & 0.775
                & 0.885 & 0.758
                & 0.843 & 0.748
                & 0.813 & \second{0.730}
                & \second{0.808} & 0.730 \\
            \multicolumn{1}{c}{Solar} 
                & \best{0.535} & \best{0.598}
                & 0.780 & 0.733
                & \second{0.725} & \second{0.698}
                & 0.815 & 0.765
                & 0.795 & 0.750
                & 0.790 & 0.745 \\
            \bottomrule
        \end{tabular}
    }
    \end{threeparttable}
    \label{tab:average}
\end{table}

\subsection{Main Results}

Although our method is implemented as a plugin rather than an independent distillation framework, it can be flexibly integrated into different trajectory matching paradigms.
Specifically, for datasets including ETT (ETTh1, ETTh2, ETTm1, ETTm2)~\cite{informer}, Wind~\cite{ding2025timemosaic}, Wike2000~\cite{qiu2024tfb}, Weather~\cite{informer}, Traffic~\cite{pems_traffic}, ILI~\cite{cdc_illness}, PEMS03~\cite{pems_traffic}, and PEMS07~\cite{pems_traffic}, we adopt DATM~\cite{guo2024datm} as the base distillation method.
For Electricity, Solar, ExchangeRate, and PEMS04, we use FTD, and for GlobalTemp and PEMS08, we employ MTT.
All three frameworks, namely DATM, FTD, and MTT, belong to the trajectory matching family, which aligns optimization trajectories between real and synthetic datasets.
This flexible configuration allows our plugin to be consistently evaluated under a unified trajectory based condensation paradigm while adapting to the characteristics of each dataset.
The comparison results with other models in terms of average performance are shown in Table~\ref{tab:dd_main_exp} and Table~\ref{tab:average}, and more detailed results with additional distillation methods are provided in Appendix Table~\ref{tab::long-term}.

\textbf{Result Analysis.}
As shown in Table~\ref{tab:dd_main_exp} and Table~\ref{tab:average}, our method consistently achieves superior results compared to existing dataset distillation approaches, using DLinear~\cite{dlinear} as the forecasting backbone.
Unlike image classification distillation, where the target is to reach $100\%$ classification accuracy, TSF distillation has no well-defined notion of “optimal” performance, since achieving zero prediction error is practically impossible due to the intrinsic stochasticity and temporal uncertainty of real-world signals.

Interestingly, as reported in Appendix Table~\ref{tab::long-term}, several datasets such as {ExchangeRate}, {ILI}, {ETTh2}, {PEMS04}, and {PEMS07} even allow distilled subsets to surpass the models trained on their full datasets when the synthetic samples are carefully optimized.
This observation aligns with recent findings~\cite{blast} that modern time-series benchmarks contain substantial redundancy and noise.
Consequently, a compact synthetic set that preserves diverse temporal modes can sometimes outperform the full training set by providing cleaner, more representative supervision signals. 


\textbf{Runtime and Memory Efficiency.}
We further evaluate the computational overhead of our framework, and detailed results are reported in Appendix~\ref{app:runtime_memory}.
Overall, our method is on average {2.59\% faster} than the baseline, with GPU memory usage increasing by only {2.49\%}.
The acceleration mainly stems from faster convergence, as the introduced frequency-domain alignment stabilizes optimization and the diversity regularization provides more informative synthetic samples, enabling efficient learning with minimal additional cost.

\begin{figure}[t]
  \centering
  \begin{subfigure}[t]{0.495\linewidth}
    \centering
    \includegraphics[width=\linewidth]{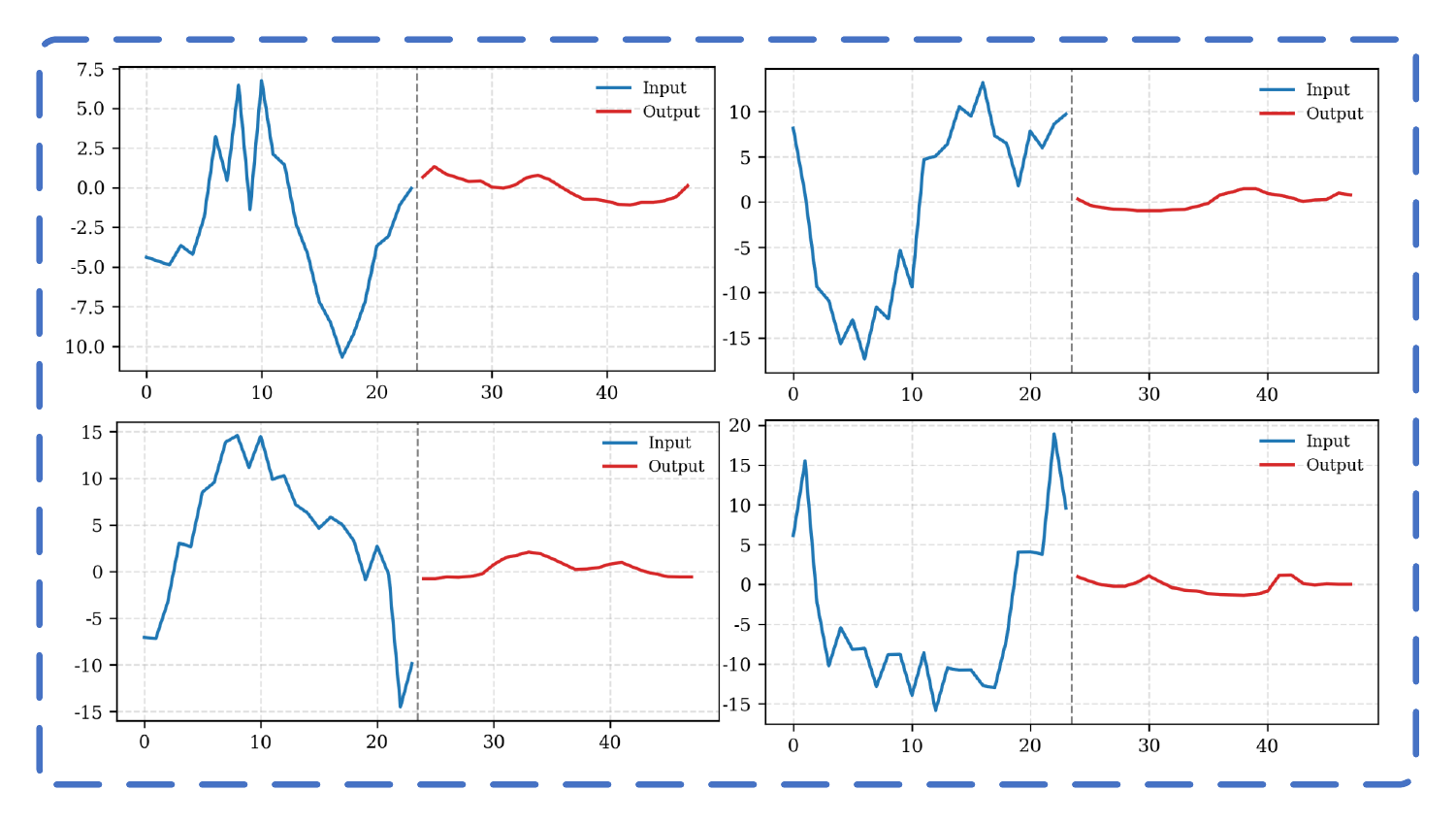}
    \caption{DATM}
    \label{fig:datm}
  \end{subfigure}
  \hfill
  \begin{subfigure}[t]{0.495\linewidth}
    \centering
    \includegraphics[width=\linewidth]{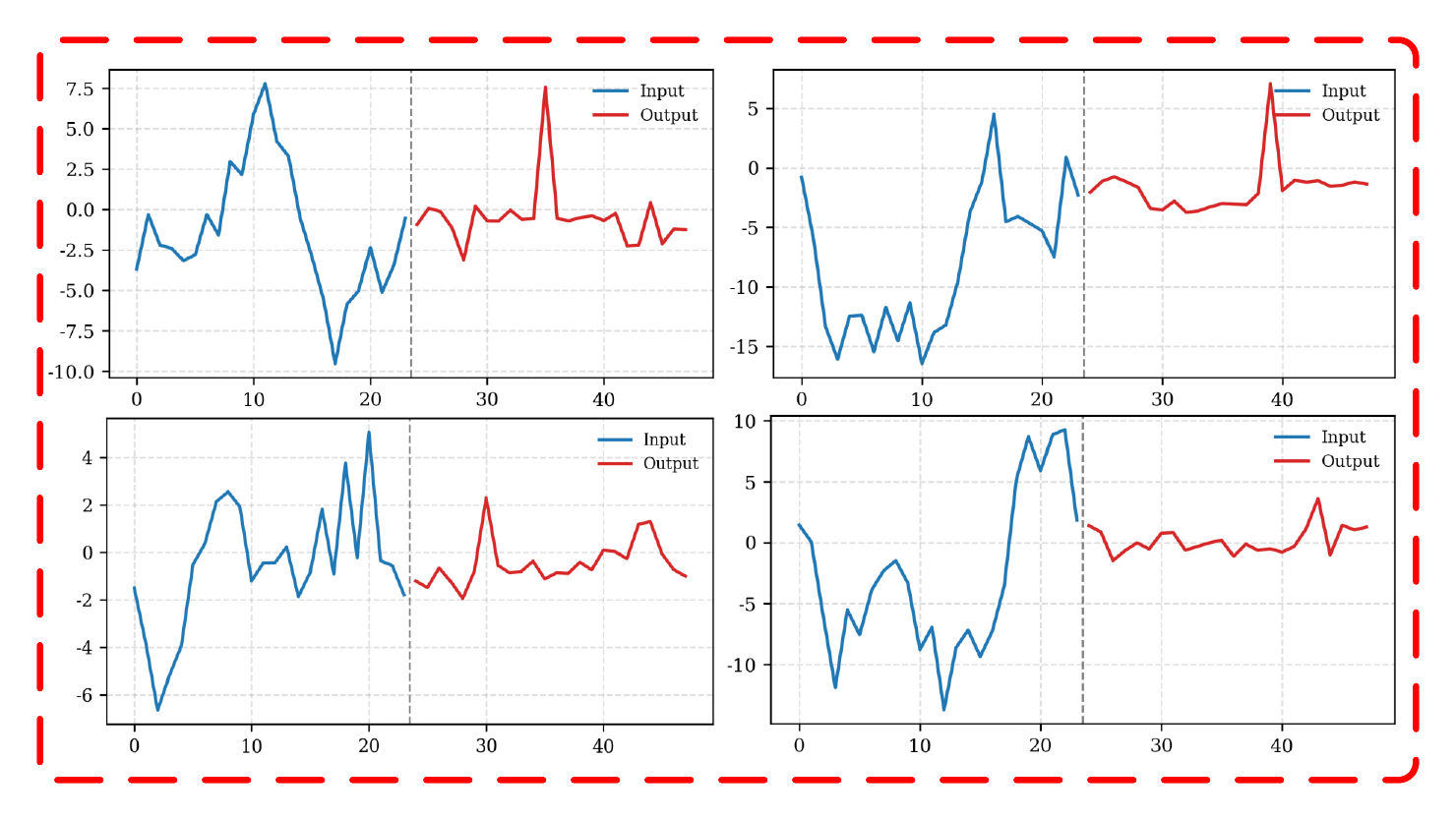}
    \caption{DATM+Ours}
    \label{fig:datm_ours}
  \end{subfigure}
  \caption{
    Visualization of synthetic time-series samples generated by DATM and DATM+Ours on the {Electricity} dataset.
    The horizontal-axis shows time step, the vertical-axis shows normalized time-series values; blue and red lines indicate inputs and outputs. DATM+Ours produces label segments with richer periodic structures and more natural local oscillations.}
    \vspace{-8pt}
  \label{fig:datm_compare}
\end{figure}

\textbf{Visualization of Samples.}
Figure~\ref{fig:datm_compare} visualizes the synthetic time-series samples generated by DATM and DATM+Ours. The baseline DATM exhibits a strong smoothness bias in the label trajectories, producing sequences with suppressed temporal variance and limited high-frequency components,
which leads to a loss of realistic temporal dynamics. In contrast, DATM+Ours generates label segments with richer periodic structures and more natural local oscillations, closely resembling the frequency and amplitude patterns observed in real data. Furthermore, the synthetic samples from DATM+Ours demonstrate higher distributional diversity across sequences, indicating improved fidelity and generalization potential of the distilled dataset. See more details in Appendix Table~\ref{fig:vis_syn_full}.


    
    


\begin{table}[t]
    \setlength{\tabcolsep}{4pt}
    \renewcommand{\arraystretch}{1.15}
    \scriptsize
    \centering
    \caption{Average MSE/MAE performance across four datasets (\textit{Weather}, \textit{Electricity}, \textit{Solar}, and \textit{Exchange}), averaged over four synthetic sample settings (3, 5, 10, and 20). Detailed results are provided in Appendix Table~\ref{tab::abla_big}.}
    \begin{threeparttable}
    \resizebox{0.95\linewidth}{!}{
        \begin{tabular}{c|c c|c c|c c|c c}
            \toprule
            \multirow{2}{*}{\scalebox{1.1}{Models}} 
            & \multicolumn{2}{c}{DATM} 
            & \multicolumn{2}{c}{FTD} 
            & \multicolumn{2}{c}{PP} 
            & \multicolumn{2}{c}{MTT} \\
            
            & \multicolumn{2}{c}{\scalebox{0.8}{\citeyearpar{guo2024datm}}} 
            & \multicolumn{2}{c}{\scalebox{0.8}{\citeyearpar{du2023ftd}}} 
            & \multicolumn{2}{c}{\scalebox{0.8}{\citeyearpar{li2023pp}}}
            & \multicolumn{2}{c}{\scalebox{0.8}{\citeyearpar{cazenavette2022mtt}}} \\
            \cmidrule(lr){2-3} \cmidrule(lr){4-5} \cmidrule(lr){6-7} \cmidrule(lr){8-9} 
            Metric 
            & MSE & MAE & MSE & MAE & MSE & MAE & MSE & MAE \\
            \toprule

\textbf{Baseline} 
& 0.75 & 0.70 & 0.74 & 0.69 & 0.75 & 0.70 & 0.75 & 0.69 \\

\midrule
\textbf{+TFA} 
& 0.63 & 0.63 & 0.62 & 0.63 & 0.65 & 0.64 & 0.64 & 0.63 \\

\midrule
\textbf{+TFA \& +ISIB} 
& 0.49 & 0.54 & 0.50 & 0.55 & 0.53 & 0.57 & 0.53 & 0.57 \\

\bottomrule
        \end{tabular}
        }
    \end{threeparttable}
    \label{tab::abla_final_avg_mae}
\end{table}

\begin{figure*}[t]
    \centering
    \begin{subfigure}[b]{0.19\textwidth}
        \includegraphics[width=\linewidth]{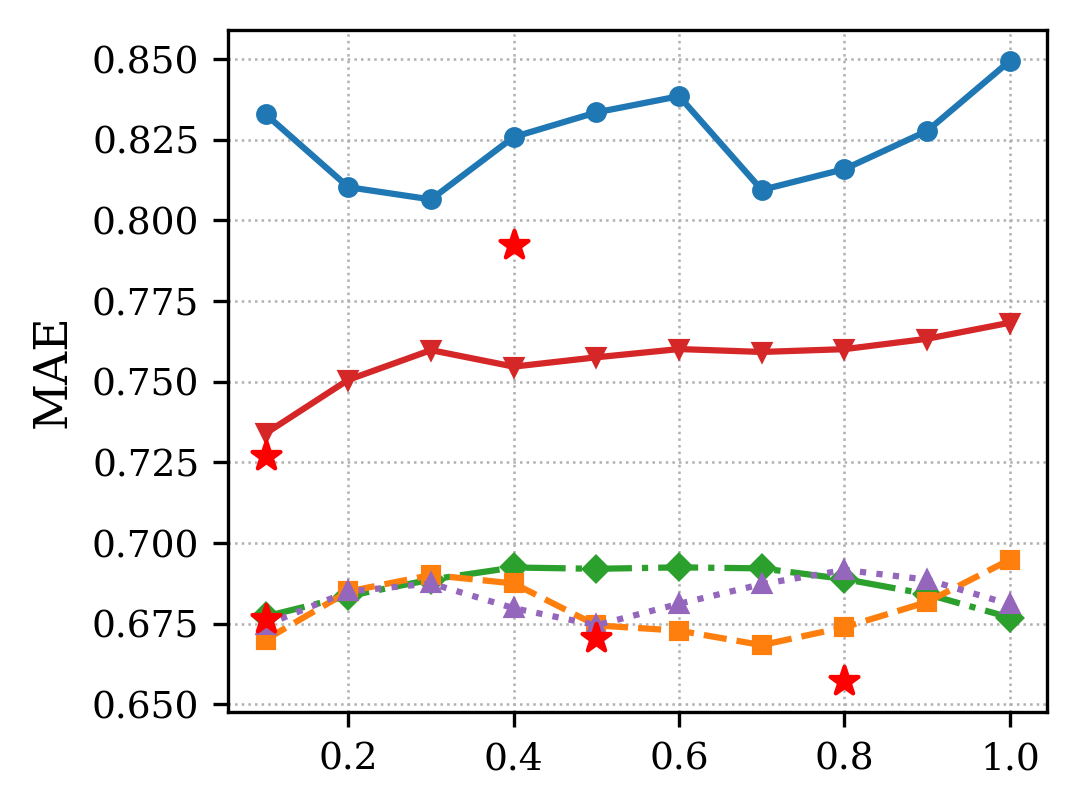}
        \caption{Electricity}
    \end{subfigure}
    \begin{subfigure}[b]{0.19\textwidth}
        \includegraphics[width=\linewidth]{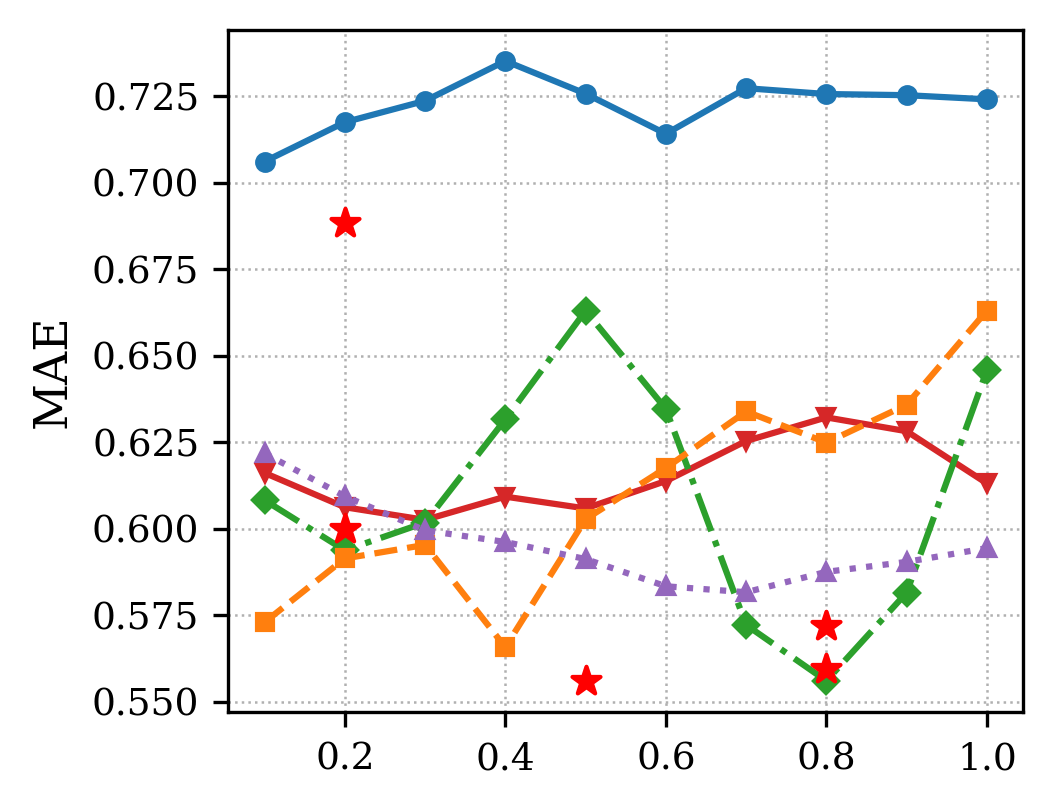}
        \caption{Solar}
    \end{subfigure}
    \begin{subfigure}[b]{0.19\textwidth}
        \includegraphics[width=\linewidth]{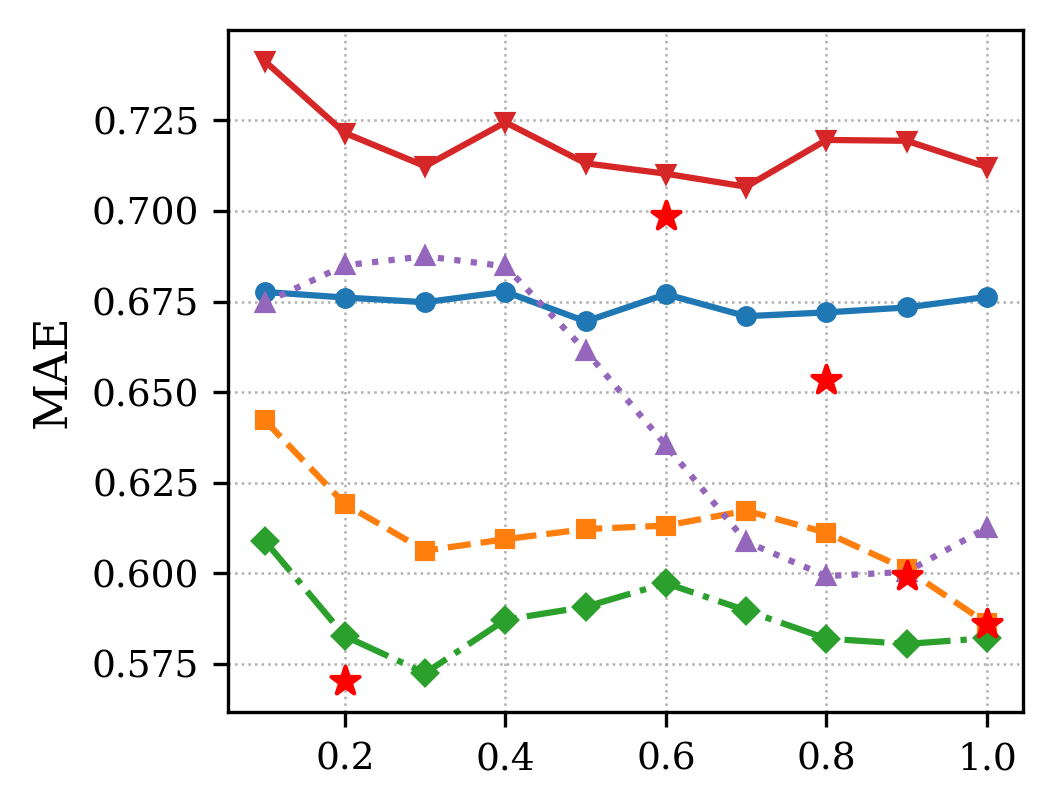}
        \caption{ETTh1}
    \end{subfigure}
    \begin{subfigure}[b]{0.19\textwidth}
        \includegraphics[width=\linewidth]{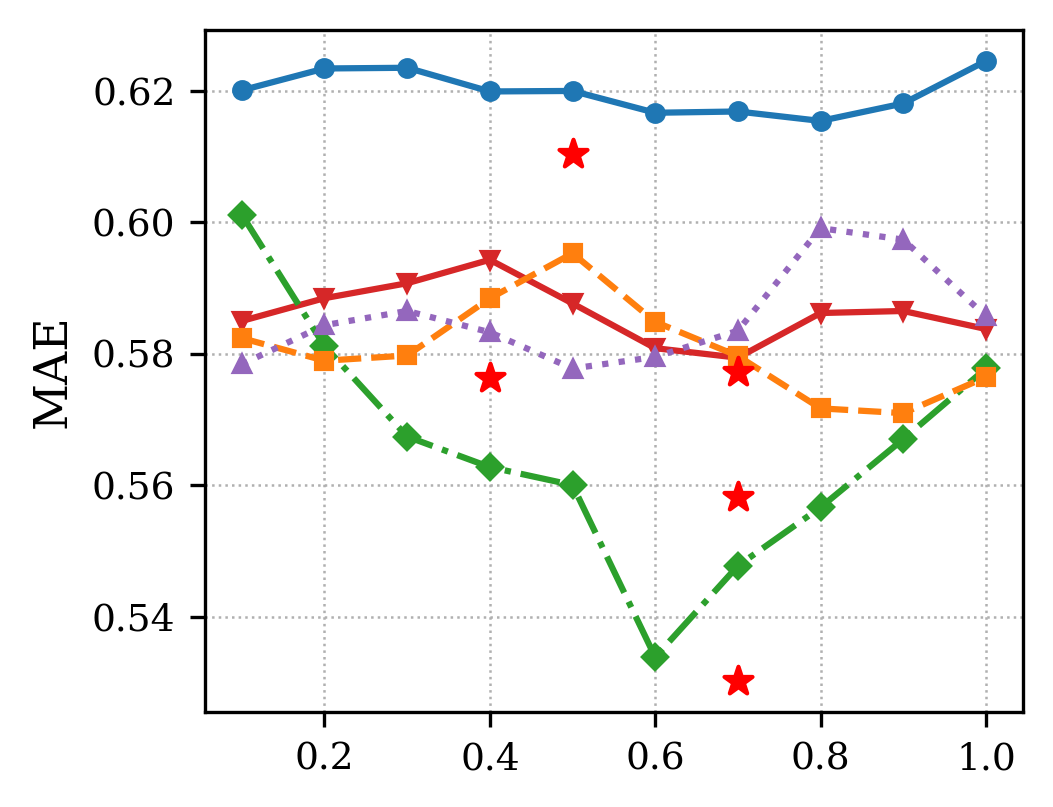}
        \caption{ETTm1}
    \end{subfigure}
    \begin{subfigure}[b]{0.19\textwidth}
        \includegraphics[width=\linewidth]{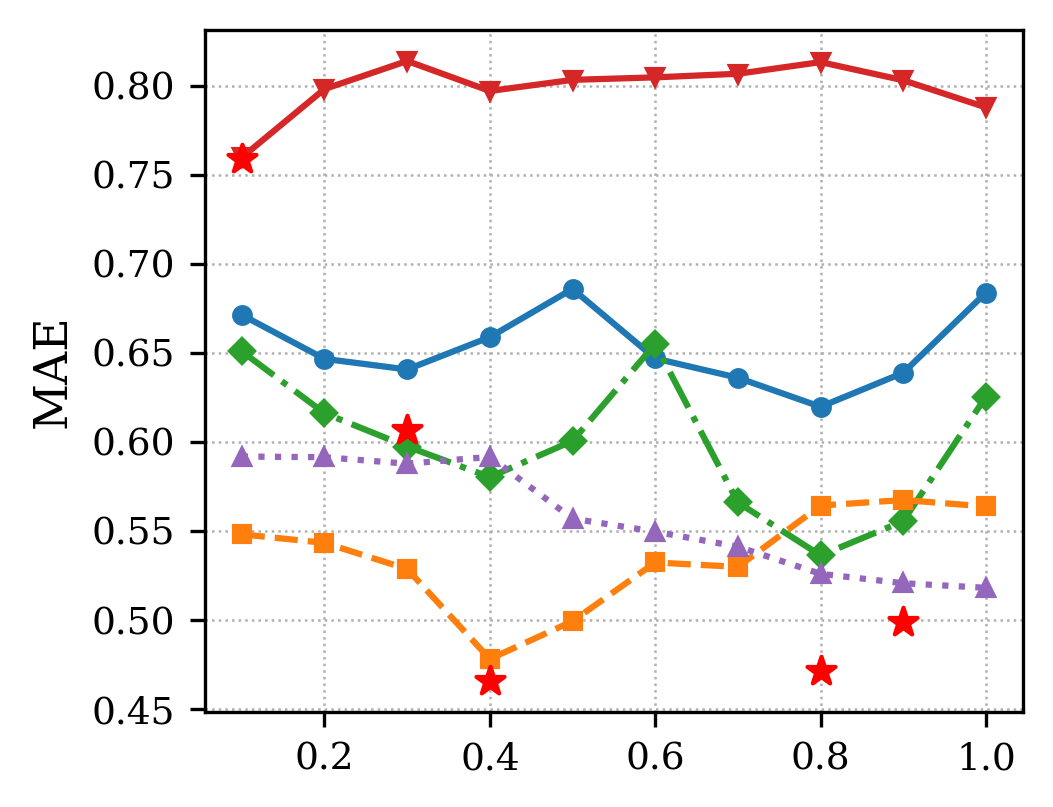}
        \caption{ExchangeRate}
    \end{subfigure}

    
    \includegraphics[width=0.5\textwidth]{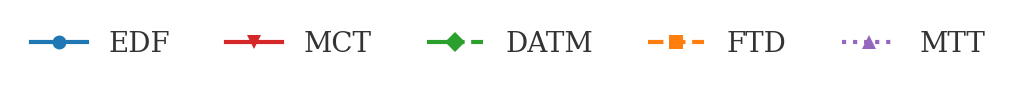}
    \vspace{-10pt}

\caption{
Parameter sensitivity analysis of different distillation baselines integrated with our ISIB-regularization. 
The coefficient $\lambda_{\mathrm{IS}}$ controls the strength of the data diversity loss in our method. 
Each curve shows how the forecasting error (MAE) changes with varying $\lambda_{\mathrm{IS}}$; 
curves are smoothed for clarity, and the optimal points are highlighted with red stars.
}

    \label{fig:lambda_sensitivity}
\end{figure*}

\subsection{Ablation Experiment}
\label{sec:ablation}



Our method is composed of two modules: the {Temporal–Frequency Alignment (TFA)} module and the {Inter-Sample Information Bottleneck (ISIB)} module. As summarized in Table~\ref{tab::abla_final_avg_mae}, incorporating either TFA or ISIB into different distillation frameworks consistently enhances the forecasting performance compared with the baseline.

Specifically, TFA aligns temporal and frequency representations to stabilize trajectory-level consistency between synthetic and real data, while ISIB encourages inter-sample diversity by suppressing redundant correlations among synthesized sequences. In the reported results, the hyperparameters are fixed as $\alpha=0.8$ for TFA and $\lambda_{\mathrm{IS}}=0.6$ for ISIB. We deliberately keep $\alpha$ and $\lambda_{\mathrm{IS}}$ constant in the main text to ensure experimental consistency and avoid cherry picking. Even under this conservative setting, the combined use of TFA and ISIB achieves the lowest overall MAE across most frameworks, confirming that the two modules complement each other in enhancing the representational quality and diversity of distilled time-series data.

\begin{figure}[t]
  \centering
  \includegraphics[width=1\linewidth]{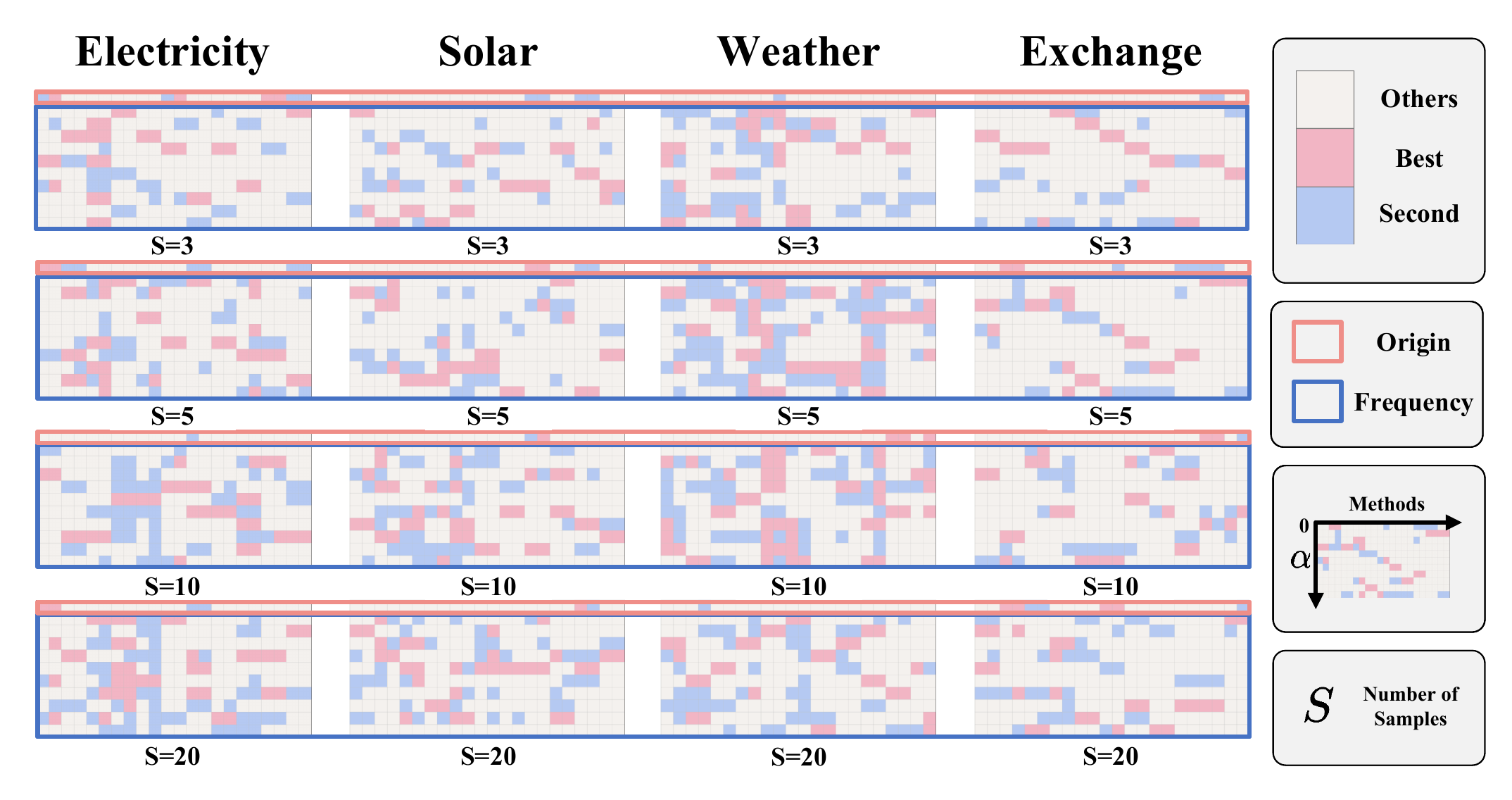}
  \caption{Visualization of Best (pink) and Second Best (blue) results across four representative datasets. Each block represents a specific sample size setting ($S = 3, 5, 10, 20$). Red-bordered regions correspond to the original method, while blue-bordered regions show the results of our frequency domain alignment variant. Gray cells indicate remaining cases.
  }
  \label{fig:freq_heatmap}
\end{figure}

\subsection{Qualitative Analysis}
\label{sec:analysis}

\textbf{Qualitative Analysis of Frequency Domain Label Loss.}
To illustrate the impact of frequency-domain alignment, Figure~\ref{fig:freq_heatmap} visualizes the distribution of the Best and Second Best results across different datasets and sample sizes. In all four datasets, the frequency-aligned variant consistently occupies a larger proportion of the Best and Second Best cells than the original method. This pattern holds under various sample scales ($S=3,5,10,20$), demonstrating that incorporating frequency domain label loss promotes more stable and globally superior performance through improved teacher–student consistency in the spectral space. See more details in Appendix~\ref{app:freq_alignment}.

\textbf{Qualitative Analysis of Data Diversity Loss.}
Figure~\ref{fig:lambda_sensitivity} illustrates the sensitivity of different distillation baselines
after integrating our ISIB-regularization. 
The coefficient $\lambda_{\mathrm{IS}}$ controls the strength of the data diversity loss, balancing inter-sample dissimilarity and forecasting performance during distillation.
For clarity, the curves are smoothed using the Savitzky--Golay filter~\cite{savity},
while the raw (unsmoothed) results are provided in Appendix Figure~\ref{fig:param_sensitivity}.
As $\lambda_{\mathrm{IS}}$ increases from 0.1 to 1.0, all methods show a clear trade-off:
a small value causes redundant synthetic samples,
whereas a large value induces excessive dispersion and temporal incoherence. Across datasets, most methods achieve stable performance within $\lambda_{\mathrm{IS}}\in[0.4,0.8]$, 
indicating that our method functions as a plug-in module that can be directly inserted into
existing distillation frameworks without extensive tuning of $\lambda_{\mathrm{IS}}$.
We fix $\lambda_{\mathrm{IS}}=0.6$ in other experiments.

\begin{figure}[!t]
    \centering
    \begin{subfigure}[t]{0.48\linewidth}
        \centering
        \includegraphics[width=\linewidth]{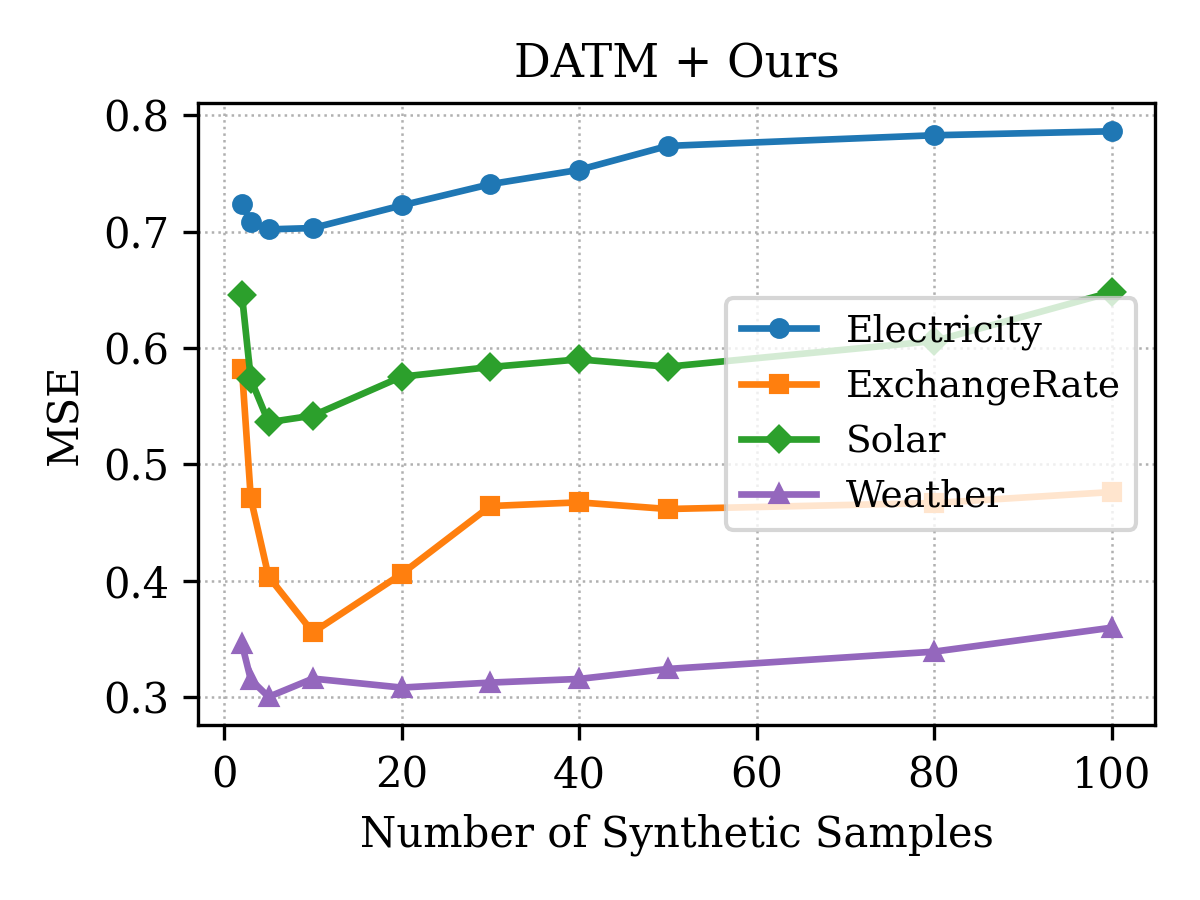}
        \caption{DATM + Ours}
    \end{subfigure}
    \hfill
    \begin{subfigure}[t]{0.48\linewidth}
        \centering
        \includegraphics[width=\linewidth]{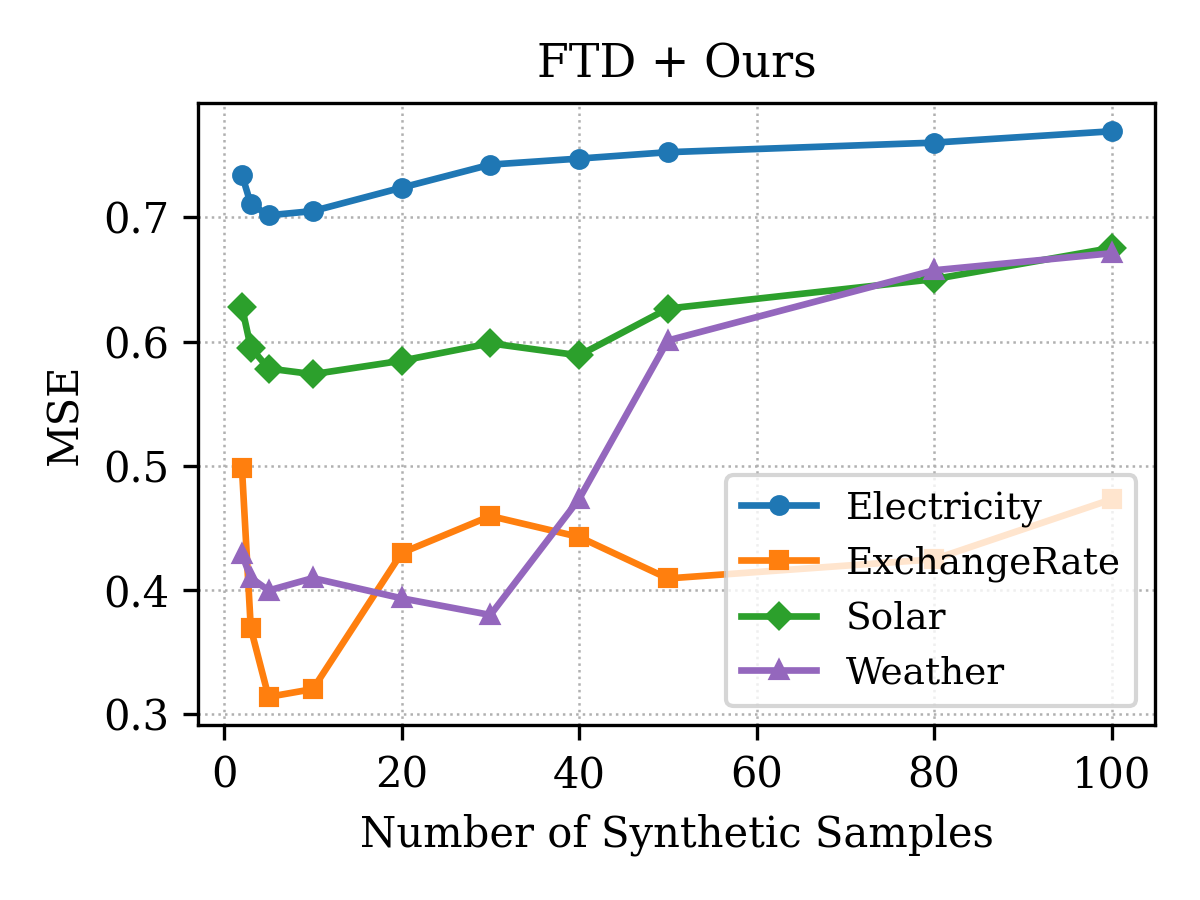}
        \caption{FTD + Ours}
    \end{subfigure}
    \vspace{-3pt}
    \caption{
    Sensitivity analysis of model performance with respect to the number of synthetic samples. Each curve corresponds to a dataset (Electricity, ExchangeRate, Solar, and Weather), and illustrates how forecasting MSE changes as the sample numbers increases. See detail in Appendix Table~\ref{tab::sample_num}.
    }
    \label{fig:num_samples_sensitivity}
\end{figure}

\begin{table}[t] 
    \setlength{\tabcolsep}{1pt}
    \scriptsize
    \caption{Extended ablation average results across four benchmark datasets(ETTh1, ETTh2, ETTm1, ETTm2) under different synthetic sample sizes. \textcolor{green!40!black}{Green rows} denote results of each method augmented with \textit{CondTSF}, and the \textcolor{blue!50!black}{blue rows} denote results augmented with our proposed method. All values are averaged over five runs; lower MSE/MAE indicate better performance.}
    \centering
    \begin{threeparttable}
    \resizebox{0.98\linewidth}{!}{
        \begin{tabular}{l*{5}{cc}@{}} 
                         \toprule
            \multicolumn{1}{c}{\multirow{2}{*}{Models}} & \multicolumn{2}{c}{EDF} & \multicolumn{2}{c}{MCT} & \multicolumn{2}{c}{DATM} & \multicolumn{2}{c}{FTD} & \multicolumn{2}{c}{MTT}\\
            &
            \multicolumn{2}{c}{\scalebox{0.8}{\citeyearpar{wang2025edf}}} & 
            \multicolumn{2}{c}{\scalebox{0.8}{\citeyearpar{zhong2025mct}}} & 
            \multicolumn{2}{c}{\scalebox{0.8}{\citeyearpar{guo2024datm}}}  & 
            \multicolumn{2}{c}{\scalebox{0.8}{\citeyearpar{du2023ftd}}}  & 
            \multicolumn{2}{c}{\scalebox{0.8}{\citeyearpar{cazenavette2022mtt}}}\\
            \cmidrule(lr){2-3} \cmidrule(lr){4-5} \cmidrule(lr){6-7} \cmidrule(lr){8-9} \cmidrule(lr){10-11}
            \multicolumn{1}{c}{Metric}& MSE & MAE & MSE & MAE & MSE & MAE & MSE & MAE & MSE & MAE\\
            \toprule
          \multicolumn{1}{c}{ETTh1} & 0.875 & 0.700 & 0.973 & 0.740 & 0.830 & 0.700 & 0.883 & 0.705 & 0.885 & 0.705 \\
\rowcolor{green!8} \multicolumn{1}{c}{CondTSF\cite{ding2024condtsf}} & 0.860 & 0.693 & 0.978 & 0.743 & 0.835 & 0.678 & 0.823 & 0.673 & 0.818 & 0.673 \\
\rowcolor{blue!8}  \multicolumn{1}{c}{\mymethod} & \best{0.838} & \best{0.678} & \best{0.913} & \best{0.713} & \best{0.688} & \best{0.598} & \best{0.723} & \best{0.615} & \best{0.715} & \best{0.608} \\
\multicolumn{1}{c}{$\downarrow~(\%)$} 
                & 4.23\% & 3.14\% 
                & 6.17\% & 3.65\% 
                & 17.11\% & 14.57\% 
                & 18.12\% & 12.77\% 
                & 19.21\% & 13.76\%\\
                
\midrule
          \multicolumn{1}{c}{ETTh2} & 0.665 & 0.648 & 0.793 & 0.705 & 0.780 & 0.703 & 0.750 & 0.695 & 0.713 & 0.673 \\
\rowcolor{green!8} \multicolumn{1}{c}{CondTSF\cite{ding2024condtsf}} & 0.650 & 0.648 & 0.763 & 0.690 & 0.653 & 0.645 & 0.635 & 0.638 & 0.573 & 0.603\\
\rowcolor{blue!8}  \multicolumn{1}{c}{\mymethod} & \best{0.540} & \best{0.583} & \best{0.633} & \best{0.630} & \best{0.260} & \best{0.385} & \best{0.303} & \best{0.425} & \best{0.275} & \best{0.400}\\
\multicolumn{1}{c}{$\downarrow~(\%)$} 
                & 18.80\% & 10.03\% 
                & 20.18\% & 10.64\% 
                & 66.67\% & 45.23\% 
                & 59.60\% & 38.85\% 
                & 61.43\% & 40.56\% \\
                
\midrule
          \multicolumn{1}{c}{ETTm1} & 0.855 & 0.688 & 0.808 & 0.673 & 0.870 & 0.700 & 0.888 & 0.708 & 0.888 & 0.708\\
\rowcolor{green!8} \multicolumn{1}{c}{CondTSF\cite{ding2024condtsf}} & 0.780 & 0.653 & 0.748 & 0.630 & 0.750 & 0.638 & 0.777 & 0.650 & 0.760 & 0.643\\
\rowcolor{blue!8}  \multicolumn{1}{c}{\mymethod} & \best{0.735} & \best{0.625} & \best{0.718} & \best{0.593} & \best{0.650} & \best{0.563} & \best{0.683} & \best{0.585} & \best{0.678} & \best{0.588}\\
\multicolumn{1}{c}{$\downarrow~(\%)$} 
                & 14.04\% & 9.16\% 
                & 11.14\% & 11.89\% 
                & 25.29\% & 19.57\% 
                & 23.09\% & 17.37\% 
                & 23.65\% & 16.95\% \\
                
\midrule
          \multicolumn{1}{c}{ETTm2} & 0.790 & 0.710 & 0.643 & 0.645 & 0.798 & 0.718 & 0.773 & 0.700 & 0.805 & 0.720\\
          
\rowcolor{green!8} \multicolumn{1}{c}{CondTSF\cite{ding2024condtsf}} & 0.505 & 0.573 & 0.385 & 0.498 & 0.493 & 0.565 & 0.513 & 0.575 & 0.488 & 0.563 \\
\rowcolor{blue!8}  \multicolumn{1}{c}{\mymethod} & \best{0.450} & \best{0.538} & \best{0.188} & \best{0.333} & \best{0.193} & \best{0.335} & \best{0.218} & \best{0.365} & \best{0.223} & \best{0.370} \\
\multicolumn{1}{c}{$\downarrow~(\%)$} 
                & 43.04\% & 24.23\% 
                & 70.76\% & 48.37\% 
                & 75.81\% & 53.34\% 
                & 71.80\% & 47.86\% 
                & 72.30\% & 48.61\% \\
                
            \bottomrule

        \end{tabular}
        }
    \end{threeparttable}
    
    \label{tab:average_abla_plugin}
\end{table}

\textbf{Effect of the number of synthetic samples.}
Figure~\ref{fig:num_samples_sensitivity} presents the sensitivity of model performance 
to the number of synthetic samples.  Performance steadily improves as the sample number increases from 2 to about 20, since the larger set captures more diverse and representative temporal patterns. Beyond this point, performance declines: our approach explicitly enlarges inter-sample differences, and this objective becomes difficult to satisfy when the synthetic set grows excessively, making the optimization harder to converge. 
These results suggest that a moderate number of synthetic samples is sufficient to capture 
krey temporal dynamics while ensuring stable and efficient training.

\textbf{Comparison with Plugin-based Methods.}
As shown in Table~\ref{tab:average_abla_plugin}, both \textit{CondTSF} and our method can serve as generic plugins to existing distillation frameworks. Our plugin consistently achieves better performance across most datasets and baselines, validating its superior compatibility and optimization stability. Extended comparisons on broader datasets and methods are reported in Appendix Tables~\ref{tab::pems_abla}, \ref{tab::ett_abla}, \ref{tab::wind_abla}, \ref{tab::other1_abla} and \ref{tab::other2_abla}. For ILI and Wike2000, however, both plugins perform relatively poorly due to the low predictability and weak temporal regularities inherent in these datasets.



\begin{figure}[t]
    \centering
    \includegraphics[width=1\linewidth]{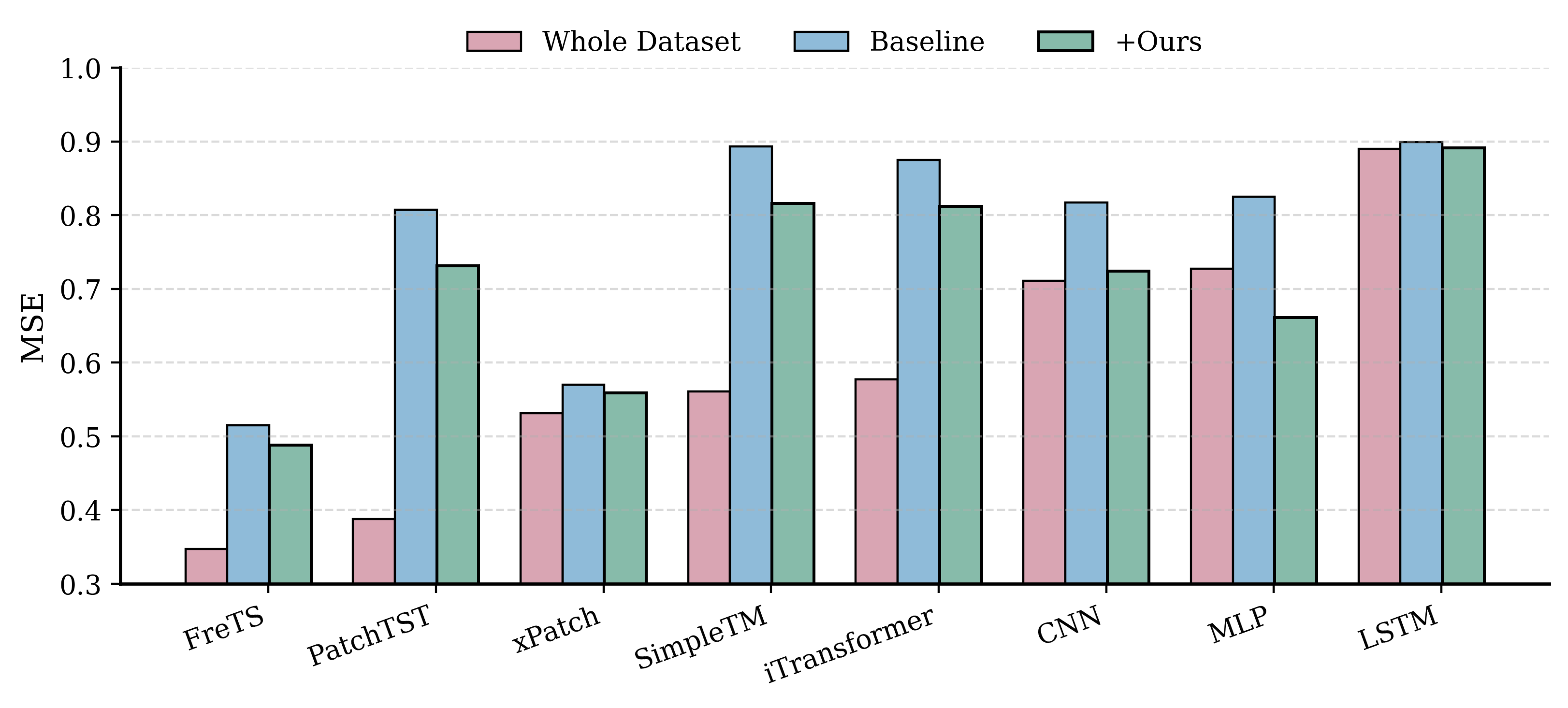}
    \caption{
    Distillation performance across different forecasting architectures.
    See more details in Appendix Sec~\ref{sec:arch}.
    }
    \label{fig:abla_arch}
\end{figure}

\textbf{Different Architectures.}
The theoretical foundation of our framework is derived under a first-order approximation with a shared model architecture and differentiable forecasting dynamics. When the backbone changes, the equality becomes approximate, but the alignment between the value term and gradient term remains practically effective. As shown in Figure~\ref{fig:abla_arch}, our method consistently outperforms baseline methods across diverse architectures, indicating that the proposed objective remains its effectiveness and generality throughout different forecasting frameworks.

\section{Conclusion}
\label{lab:conclusion}


We introduced a dataset distillation framework for time-series forecasting (TSF) that jointly aligns value and gradient terms under a first-order optimization scheme.
The method effectively condenses temporal dynamics while preserving essential forecasting behavior.
Extensive experiments across diverse datasets and architectures demonstrate consistent improvements over baseline methods, confirming the generality and robustness of the proposed objective.
The framework provides a unified perspective for understanding and advancing data condensation in temporal learning, as well as inspires further research into the emerging and impactful area of dataset distillation for TSF.

\clearpage
{
    \small
    \bibliographystyle{ieeenat_fullname}
    \bibliography{main}
}

\clearpage
\appendix
\section{Detailed Dataset Descriptions}

To comprehensively evaluate forecasting performance across diverse temporal dynamics and application domains,
we employ a collection of widely adopted benchmark datasets summarized in Table~\ref{tab:datasets}. Following CondTSF~\cite{ding2024condtsf}, both the input and prediction horizons are set to $T_{\text{in}}=T_{\text{out}}=24$ for all datasets.
Each multivariate time series $\mathbf{X} \in \mathbb{R}^{N \times L}$ is first standardized and segmented into
input--output pairs using a fixed-stride sliding window.
For each window position, the input sequence
$\mathbf{X}^{(i)} \in \mathbb{R}^{N \times T_{\text{in}}}$ 
and its forecasting target
$\mathbf{Y}^{(i)} \in \mathbb{R}^{N \times T_{\text{out}}}$
are defined as
\begin{equation}
\mathbf{X}^{(i)} = \mathbf{X}_{[:,\, i : i + T_{\text{in}}]}, \qquad
\mathbf{Y}^{(i)} = \mathbf{X}_{[:,\, i + T_{\text{in}} : i + T_{\text{in}} + T_{\text{out}}]},
\end{equation}
where the window index $i$ advances by a constant stride $s$ along the temporal axis.
This overlapping-window formulation preserves temporal continuity
while producing sufficient training instances for stable optimization across datasets with different lengths and frequencies.

\subsection{Environmental Domain Datasets}

\textbf{Weather Dataset:} Provides comprehensive meteorological observations including dry and wet bulb temperatures in both Fahrenheit and Celsius, dew point measurements, relative humidity, wind speed and direction, atmospheric pressure readings, and visibility conditions.

\textbf{Global Temp Dataset:} Hourly global temperature dataset comprising 1,000 variables across multiple regions, used to evaluate long-term forecasting performance and model generalization in climate and environmental domains.

\subsection{Energy Domain Datasets}

\textbf{Wind Power Dataset:} High-frequency wind power output records enabling detailed studies on renewable integration dynamics, system stability, and near-term generation forecasting.

\textbf{Solar Power Dataset:} Minute-resolution photovoltaic power generation data supporting fine-grained analysis of solar output dynamics, cloud-induced intermittency, and renewable energy variability.

\textbf{ETT (Electricity Transformer Temperature) Datasets:} Four complementary datasets (ETTh1, ETTh2, ETTm1, ETTm2) recording transformer load and temperature variations at hourly and 15-minute intervals, serving as standard benchmarks for power grid condition monitoring and predictive maintenance research.

\textbf{Electricity Dataset:} Electricity consumption records from 321 individual consumers, providing a detailed representation of distributed usage behavior suitable for demand response evaluation and consumer load modeling.

\subsection{Web Domain Datasets}

\textbf{Wike2000 Dataset:} The Wike2000 dataset belongs to the web domain and consists of daily multivariate time series comprising 2,000 variables with a total sequence length of 792. It captures large-scale web traffic dynamics and is widely used to evaluate forecasting models under high-dimensional, short-horizon, and nonstationary conditions.

\begin{table}[t]
  \centering
  \setlength\tabcolsep{3pt}
  \caption{Statistics of the benchmark datasets used in our study.
    Each dataset is characterized by its variable dimensionality (\textit{Dim}),
    the number of samples in training/validation/testing splits,
    and the original sampling frequency.
    All datasets are processed under a unified normalization and
    fixed-stride $S=12$ windowing protocol to ensure consistent temporal segmentation across experiments, follow the CondTSF~\cite{ding2024condtsf}.}
  \begin{tabular}{c|c|c|c}
    \toprule
    Dataset & Dim & Dataset Size & Frequency \\
    \midrule
    ETTh1, ETTh2 & 7 & (837,357,357) & Hourly \\
    ETTm1, ETTm2 & 7 & (3357,1437,1437) & 15 min \\
    Weather & 21 & (3070,1314,1314) & 10 min \\
    Traffic & 862 & (1020,435,435) & Hourly \\
    Electricity & 321 & (1531,654,654) & Hourly \\
    Exchange-rate & 8 & (439,186,186) & Daily \\
    Solar-energy & 137 & (3063,1311,1311) & 10 min \\
    Wind (L1--L4) & 9 & (2551,1092,1092) & Hourly \\
    PEMS03 & 358 & (1525,652,652) & 5 min \\
    PEMS04 & 307 & (988,421,421) & 5 min \\
    PEMS07 & 883 & (1643,702,702) & 5 min \\
    PEMS08 & 170 & (1038,443,443) & 5 min \\
    Wike2000 & 2000 & (43,16,16) & Daily \\
    ILI & 7 & (53,21,21) & Weekly \\
    Global Temp & 1000 & (201,84,84) & Hourly \\
    \bottomrule
  \end{tabular}
  \label{tab:datasets}
\end{table}

\subsection{Transportation Domain Datasets}

\textbf{Traffic Dataset:} Comprehensive traffic monitoring data collected from 862 highway sensor stations, recording vehicle occupancy and flow dynamics to support research in intelligent transportation systems and congestion management.

\textbf{PEMS Dataset:} The PEMS datasets (PEMS03, PEMS04, PEMS07, PEMS08) originate from the California Department of Transportation Performance Measurement System (Caltrans PeMS) and are widely used benchmarks in traffic flow forecasting research.

\subsection{Financial Domain Datasets}

\textbf{Exchange Rate Dataset:} Daily records of foreign exchange rate dynamics across multiple international currency pairs, supporting quantitative analyses of financial markets, currency volatility, and macroeconomic forecasting.

\subsection{Healthcare Domain Datasets}

\textbf{National Illness Dataset:} Weekly surveillance records capturing influenza-like illness (ILI) prevalence across multiple age groups and healthcare provider networks, serving as a critical resource for epidemiological modeling and public health surveillance.

\section{Implementation Details}

\paragraph{Training Configuration.}
The training process consists of two stages: collecting teacher trajectories and optimizing synthetic sequences with the Synthetic Dataset module. All experiments are implemented in PyTorch and conducted on a single NVIDIA RTX 5090 GPU.

\paragraph{Metrics.}
We use mean squared error (MSE) and mean absolute error (MAE) as evaluation metrics for time-series forecasting. 
These metrics are calculated as follows:
\begin{equation}
\mathrm{MSE} = \frac{1}{H} \sum_{i=1}^{H} (x_i - \hat{x}_i)^2, 
\quad
\mathrm{MAE} = \frac{1}{H} \sum_{i=1}^{H} |x_i - \hat{x}_i|,
\end{equation}
where $x_i, \hat{x}_i \in \mathbb{R}$ denote the ground truth and the predicted values of the $i$-th forecasting step, and $H$ is the forecasting horizon.

\emph{Teacher trajectory collection.}
For each dataset, forty teacher trajectories are generated. Each teacher model is trained using stochastic gradient descent with a learning rate of $5\times10^{-4}$, momentum of 0.9, and no weight decay. Training is performed for 80 epochs with a batch size of 32.  
At every epoch, model parameters are stored as checkpoints to form a complete parameter trajectory. Every five trajectories are saved together in replay buffer files which are later used for parameter-matching during distillation.  
During training, both training and testing mean squared error (MSE) and mean absolute error (MAE) are recorded after each epoch. For all datasets, 40 teacher trajectories are collected (5 per replay buffer, yielding 8 buffers). Each teacher is trained for 80 epochs using SGD, while synthetic samples are optimized with Adam. Distillation alternates between student update and data refinement following CondTSF’s default protocol.

\emph{Synthetic sequence optimization.}
The GCOND module maintains a set of learnable synthetic time-series samples, each represented as a tensor of shape $[S, N, T]$, where $S$ is the number of synthetic samples, $N$ is the number of variables in the dataset, and $T$ equals the sum of the input and prediction lengths ($T_{\text{in}}=24$ and $T_{\text{out}}=24$).  
Synthetic samples are initialized by randomly extracting contiguous windows from the normalized real data to ensure realistic temporal patterns.  
The synthetic parameters are optimized directly by the Adam optimizer with a learning rate of 0.1.  
In each distillation iteration, the student model is trained on the synthetic data for twenty gradient steps, and then the synthetic sequences are refined based on the resulting parameter trajectories.

\emph{distillation protocol.}
The distillation procedure alternates between parameter imitation and data refinement. To ensure fair comparison with the original CondTSF framework, we adopt the same hyperparameter configuration: the distillation interval is set to five iterations, and the conditional update coefficient, which controls how much model predictions are blended back into the synthetic data, is set to 0.01.  
All random seeds are fixed to ensure reproducibility, and deterministic CuDNN settings are enabled.

\emph{Evaluation.}
After distillation, the distilled datasets are evaluated using multiple independently initialized student networks for each forecasting architecture. The results are reported as the mean and standard deviation of MSE and MAE over five runs and across twenty benchmark datasets.  
All experiments use fixed preprocessing and consistent train–validation–test splits for comparability across methods.

\section{Parameter Term}
\label{app:pt}
The Parameter Term $\mathcal{L}_{\mathrm{P}}$ is optimized via a
\textit{parameter trajectory alignment} strategy,
which aligns the student parameters with those of multiple pre-trained teachers.

\textbf{(1) Expert trajectories.}
Several teacher models are trained on real data, and their parameter
snapshots are saved as
$\Theta_T=\{\theta_T^{(0)},\theta_T^{(1)},\dots,\theta_T^{(E)}\}$,
recording the optimization paths at different epochs.

\textbf{(2) Student training.}
The student model $M_{\theta_S}$ is trained on the synthetic dataset
$\mathcal{D}_{\mathrm{syn}}$, with parameters
$\theta_S^{(k)}$ recorded after each iteration.

\textbf{(3) Trajectory matching.}
During distillation, the student is aligned with the most similar teacher
trajectory by minimizing the normalized parameter distance:
\begin{equation}
\mathcal{L}_{\mathrm{P}}
=\min_{\text{expert}}\;
\frac{\|\theta_S-\theta_T^{(E)}\|_2^2}
     {\|\theta_T^{(E)}-\theta_T^{(0)}\|_2^2}.
\end{equation}
This distance is computed using the L2 norm without backpropagation
through teacher models, while gradients are propagated only
through the student network.

We follow prior trajectory matching
using them as backbones for optimization.
Our method is implemented as a \textit{plug-in objective} that replaces
the standard gradient-matching loss with this
parameter-alignment formulation.

\section{Information Diversity Measurement via KL Divergence}
\label{sec:kl_density}

To quantify the diversity and information content of the distilled synthetic datasets, 
we compute the \textit{average symmetric Kullback--Leibler (KL) divergence} among all synthetic samples. 
Given a set of $S$ synthetic time series $\{\mathbf{X}_i \in \mathbb{R}^{N \times T}\}_{i=1}^S$, 
each sample is first normalized and converted into a probability distribution over the temporal dimension via the temperature-controlled softmax:
\begin{equation}
p_i = \mathrm{Softmax}\!\left(\frac{\mathbf{X}_i - \mu_i}{\sigma_i + \epsilon} \bigg/ \tau \right),
\end{equation}
where $\mu_i$ and $\sigma_i$ denote the mean and standard deviation across time steps, 
and $\tau$ is the softmax temperature.

The symmetric KL divergence between two samples $p_i$ and $p_j$ is defined as
\begin{equation}
D_{\mathrm{sym}}(p_i, p_j) = \tfrac{1}{2}\Big(
\mathrm{KL}(p_i \| p_j) + \mathrm{KL}(p_j \| p_i)
\Big),
\end{equation}
and the overall average divergence within the synthetic set is
\begin{equation}
\bar{D}_{\mathrm{KL}} = \frac{2}{S(S-1)} \sum_{i<j} D_{\mathrm{sym}}(p_i, p_j).
\end{equation}

Intuitively, a larger $\bar{D}_{\mathrm{KL}}$ indicates greater dissimilarity among synthetic samples, 
suggesting that the distilled dataset captures richer temporal patterns and covers a broader portion of the underlying data manifold.  
Therefore, we use $\bar{D}_{\mathrm{KL}}$ as a quantitative proxy for the \textit{Information diversity} of synthetic time-series data.

Table~\ref{tab:kl_density} reports the computed average symmetric KL divergences across multiple datasets and sample sizes. 
As shown, our method consistently achieves higher Information diversity than both the DATM and CondTSF baselines, 
demonstrating its ability to generate more diverse and informative synthetic sequences.

\begin{table}[!ht]
\centering
\scriptsize
\setlength{\tabcolsep}{2pt}
\begin{threeparttable}
\caption{
Average symmetric KL divergence under different numbers of synthetic samples ($S$). 
The average KL divergence between synthetic samples is used as a proxy to measure their \textit{Information diversity}. All results are obtained under our method with $\alpha = 0.1$ and $\lambda_{\mathrm{IS}} = 0.2$.
}
\label{tab:kl_density}
\begin{tabular}{c|ccccc}
\toprule
\textbf{Method} & \textbf{Electricity (321)} & \textbf{Solar (137)} & \textbf{Weather (21)} & \textbf{ETTm1 (7)} & \textbf{ETTh1 (7)} \\
\midrule
\multicolumn{6}{c}{\textit{Ours}} \\
\midrule
$S$=3  & 7.9562 & 6.0624 & 6.1783 & 7.5984 & 7.5065 \\
$S$=5  & 5.9573 & 7.1132 & 5.8814 & 7.1882 & 7.8194 \\
$S$=10 & 5.6014 & 4.6531 & 6.7383 & 6.9596 & 7.9601 \\
$S$=20 & 5.7005 & 5.6690 & 6.8403 & 7.1020 & 7.7295 \\
\midrule
\rowcolor{gray!10}Mean & \textbf{6.3038} & \textbf{5.8744} & \textbf{6.4096} & \textbf{7.2121} & \textbf{7.7539} \\
\midrule
\multicolumn{6}{c}{\textit{CondTSF}} \\
\midrule
$S$=3  & 4.7510 & 3.2314 & 7.6868 & 3.8616 & 5.2092 \\
$S$=5  & 5.9471 & 6.4680 & 5.0484 & 3.7866 & 5.8175 \\
$S$=10 & 5.3922 & 5.0312 & 4.9754 & 3.9649 & 4.7749 \\
$S$=20 & 5.2139 & 3.3957 & 6.0091 & 3.9374 & 5.1213 \\
\midrule
\rowcolor{gray!10}Mean & 5.3261 & 4.5316 & 5.9299 & 3.8876 & 5.2307 \\
\midrule
\multicolumn{6}{c}{\textit{DATM}} \\
\midrule
$S$=3  & 2.9747 & 3.0964 & 6.6312 & 5.5428 & 5.1283 \\
$S$=5  & 5.0359 & 3.3059 & 5.9489 & 3.8608 & 4.1114 \\
$S$=10 & 5.8011 & 3.8497 & 5.6042 & 3.6268 & 5.8774 \\
$S$=20 & 5.4619 & 4.1256 & 4.8433 & 4.0031 & 5.1918 \\
\midrule
\rowcolor{gray!10}Mean & 4.8184 & 3.5944 & 5.7569 & 4.2584 & 5.0772 \\
\bottomrule
\end{tabular}
\end{threeparttable}
\end{table}

\section{Ablation Study on Frequency Domain Label Loss Alignment}
\label{app:freq_alignment}

\begin{table}[!t]
\centering
\caption{
Compact comparison of \textbf{Best} and \textbf{Second Best} marks across datasets and sample sizes ($S$). 
Each block compares the baseline and our frequency-domain label-loss alignment method. 
Higher shares indicate stronger consistency across datasets.
}
\scriptsize
\setlength{\tabcolsep}{3pt}
\renewcommand{\arraystretch}{1.05}
\begin{tabular}{c|c|ccc|ccc|ccc}
\toprule
\multirow{2}{*}{\rotatebox[origin=c]{90}{Dataset}} &
\multirow{2}{*}{$S$} &
\multicolumn{3}{c|}{Original} &
\multicolumn{3}{c|}{Ours} &
\multicolumn{3}{c}{Ours Share (\%)} \\
\cmidrule(lr){3-5} \cmidrule(lr){6-8} \cmidrule(lr){9-11}
& & Best & Second & Sum & Best & Second & Sum & Best & Second & Sum \\
\midrule
\multirow{5}{*}{\rotatebox[origin=c]{90}{Exchange}}
 & 3  & 0 & 2 & 2 & 24 & 24 & 48 & 100.0 & 92.3 & 96.0 \\
 & 5  & 2 & 5 & 7 & 24 & 24 & 48 & 92.3 & 82.8 & 87.3 \\
 & 10 & 2 & 1 & 3 & 27 & 28 & 55 & 93.1 & 96.6 & 94.8 \\
 & 20 & 4 & 3 & 7 & 25 & 36 & 61 & 86.2 & 92.3 & 89.7 \\
 & \textbf{Sum} & \textbf{8} & \textbf{11} & \textbf{19} & \textbf{100} & \textbf{112} & \textbf{212} & \textbf{92.9} & \textbf{91.0} & \textbf{91.9} \\
\midrule
\multirow{5}{*}{\rotatebox[origin=c]{90}{ECL}}
 & 3  & 4 & 4 & 8 & 29 & 30 & 59 & 87.9 & 88.2 & 88.1 \\
 & 5  & 2 & 6 & 8 & 36 & 35 & 71 & 94.7 & 85.4 & 89.9 \\
 & 10 & 0 & 1 & 1 & 32 & 51 & 83 & 100.0 & 98.1 & 98.8 \\
 & 20 & 2 & 2 & 4 & 39 & 54 & 93 & 95.1 & 96.4 & 95.9 \\
 & \textbf{Sum} & \textbf{8} & \textbf{13} & \textbf{21} & \textbf{136} & \textbf{170} & \textbf{306} & \textbf{94.4} & \textbf{92.0} & \textbf{93.2} \\
\midrule
\multirow{5}{*}{\rotatebox[origin=c]{90}{Solar}}
 & 3  & 0 & 4 & 4 & 29 & 27 & 56 & 100.0 & 87.1 & 93.3 \\
 & 5  & 2 & 2 & 4 & 29 & 31 & 60 & 93.5 & 93.9 & 93.8 \\
 & 10 & 1 & 1 & 2 & 36 & 46 & 82 & 97.3 & 97.9 & 97.6 \\
 & 20 & 1 & 1 & 2 & 34 & 42 & 76 & 97.1 & 97.7 & 97.4 \\
 & \textbf{Sum} & \textbf{4} & \textbf{8} & \textbf{12} & \textbf{128} & \textbf{146} & \textbf{274} & \textbf{97.0} & \textbf{94.1} & \textbf{95.6} \\
\midrule
\multirow{5}{*}{\rotatebox[origin=c]{90}{Weather}}
 & 3  & 0 & 1 & 1 & 38 & 48 & 86 & 100.0 & 98.0 & 98.9 \\
 & 5  & 0 & 1 & 1 & 50 & 68 &118 & 100.0 & 98.6 & 99.2 \\
 & 10 & 3 & 0 & 3 & 50 & 52 &102 & 94.3 &100.0 & 97.1 \\
 & 20 & 0 & 1 & 1 & 31 & 49 & 80 & 100.0 & 98.0 & 98.8 \\
 & \textbf{Sum} & \textbf{3} & \textbf{3} & \textbf{6} & \textbf{169} & \textbf{217} & \textbf{386} & \textbf{98.6} & \textbf{98.7} & \textbf{98.6} \\
\midrule
\rowcolor{gray!10}
\multicolumn{2}{c|}{\textbf{Total}} &
\textbf{23} & \textbf{35} & \textbf{58} &
\textbf{533} & \textbf{645} & \textbf{1178} &
\textbf{95.9} & \textbf{94.0} & \textbf{94.9} \\
\bottomrule
\end{tabular}
\label{tab:freq_best_second_compact}
\end{table}

To further examine the effectiveness of incorporating frequency domain information into the distillation objective, we conduct an ablation study across four representative datasets: Weather, Electricity, Solar, and ExchangeRate (see Tables~\ref{tab::weather_fre}–\ref{tab::ExchangeRate_fre}).
Each table reports the forecasting results under different values of the alignment coefficient $\alpha \in \{0.1, 0.2, 0.3, 0.4, 0.5, 0.6, 0.7, 0.8, 0.9, 1.0\}$, which controls the relative strength of the frequency domain label loss in the overall training objective.

When $\alpha=0$, the distillation process does not include frequency domain alignment, and the student model tends to overfit to local temporal fluctuations, leading to less stable generalization. Introducing the frequency domain label loss term ($\alpha>0$) consistently improves model stability and reduces forecasting error across all datasets,
demonstrating that aligning teacher and student representations in the spectral space effectively complements time-domain supervision.

However, the performance does not monotonically increase with larger $\alpha$ values. Excessive spectral alignment may overemphasize global frequency matching and dampen short-term temporal adaptation. Empirically, values of $\alpha$ between 0.3 and 0.8 achieve a good balance between temporal fitting and spectral regularization, providing robust improvements over the baseline without frequency alignment.

Table~\ref{tab:freq_best_second_compact} summarizes the aggregated statistics of Best and Second Best results derived from the four dataset-specific tables (\ref{tab::weather_fre}, \ref{tab::ecl_fre}, \ref{tab::solar_fre}, \ref{tab::ExchangeRate_fre}). The aggregated data confirm that introducing frequency domain label loss yields clear overall gains in both forecasting accuracy and consistency across datasets and sample sizes. This supports our core hypothesis that frequency domain alignment serves as an effective inductive bias
for preserving multi-scale temporal–spectral structure during time-series distillation.

\begin{figure*}[t]
    \centering
    \begin{subfigure}[b]{0.19\textwidth}
        \includegraphics[width=\linewidth]{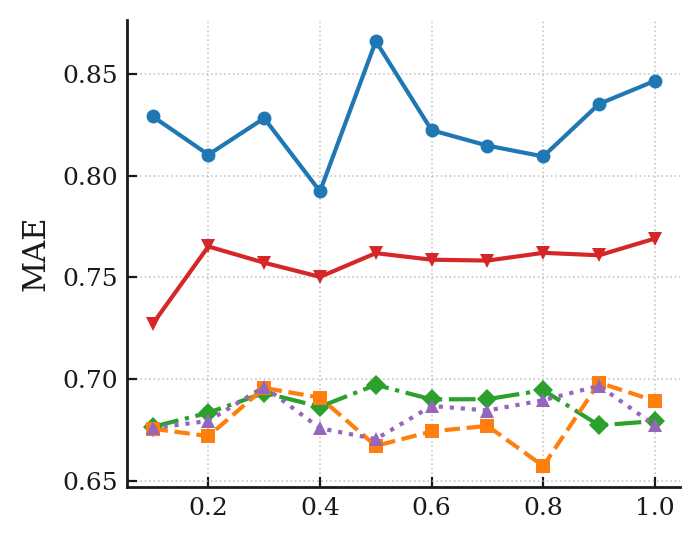}
        \caption{Electricity}
    \end{subfigure}
    \begin{subfigure}[b]{0.19\textwidth}
        \includegraphics[width=\linewidth]{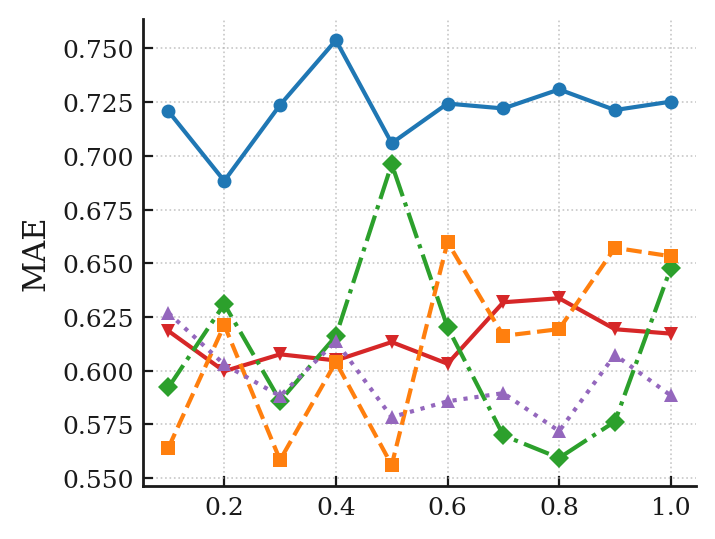}
        \caption{Solar}
    \end{subfigure}
    \begin{subfigure}[b]{0.19\textwidth}
        \includegraphics[width=\linewidth]{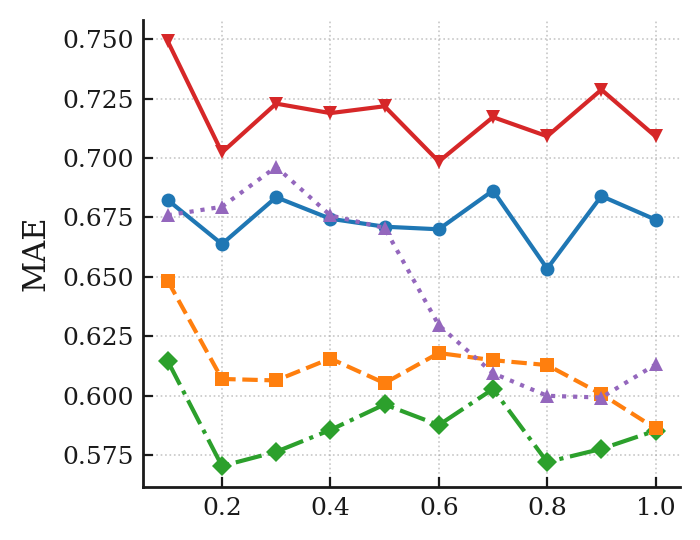}
        \caption{ETTh1}
    \end{subfigure}
    \begin{subfigure}[b]{0.19\textwidth}
        \includegraphics[width=\linewidth]{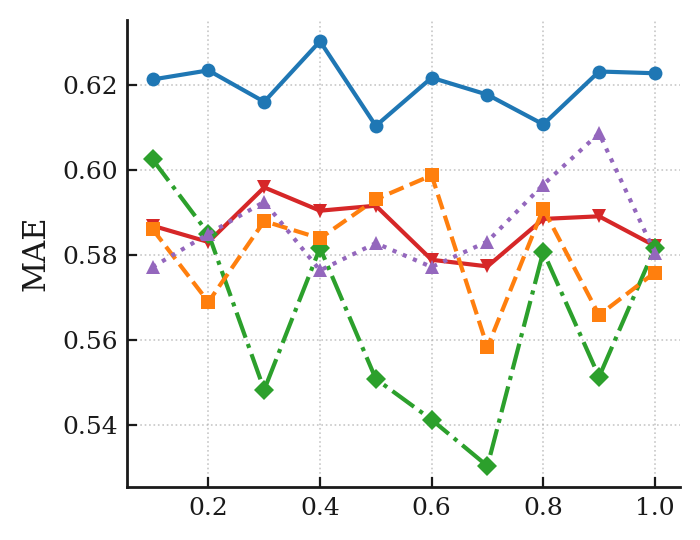}
        \caption{ETTm1}
    \end{subfigure}
    \begin{subfigure}[b]{0.19\textwidth}
        \includegraphics[width=\linewidth]{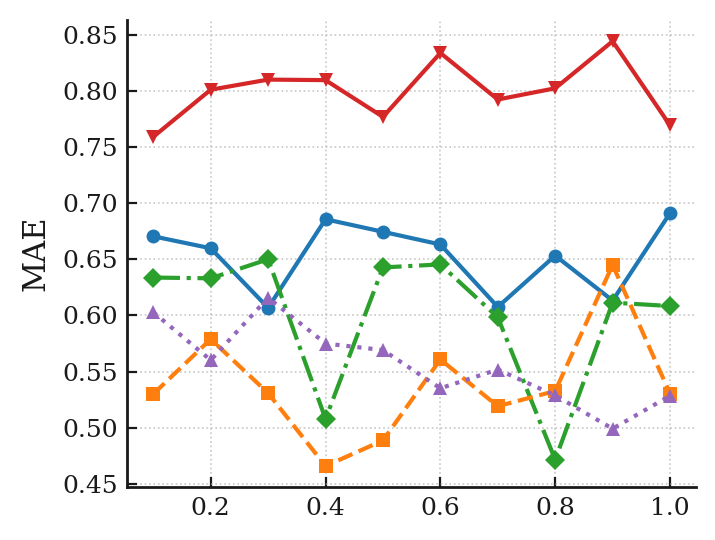}
        \caption{ExchangeRate}
    \end{subfigure}

    \vspace{2pt}
    \includegraphics[width=0.5\textwidth]{Figure/label.png}

    \caption{
        Parameter sensitivity comparison of five dataset condensation methods across datasets.
        Each curve shows MAE under different parameter values (0.1–1.0), and the legend below applies to all subplots.
    }
    \label{fig:param_sensitivity}
\end{figure*}

\begin{figure*}[t]
  \centering
    \caption{
  Comprehensive visualization of synthetic time-series samples generated by four representative dataset distillation frameworks
  (DATM, EDF, MCT, and MTT) and their plug-in enhanced variants (+Ours)
  across eight benchmark datasets: ETTh1, ETTh2, ETTm1, ETTm2, Electricity, ExchangeRate, Solar, Traffic, PEMS03, PEMS04, PEMS07, PEMS08, Wike2000 and Weather. Each dataset row compares the baseline and our plug-in variants across all frameworks. The plug-in method consistently produces more realistic temporal fluctuations and richer label dynamics, highlighting improved fidelity and inter-sample diversity of synthetic sequences, without introducing any additional trainable parameters.
  }
  \scriptsize
  \setlength{\tabcolsep}{2pt}
  \renewcommand{\arraystretch}{1.05}

  \begin{tabular}{c|cccccccc}
    & \multicolumn{2}{c}{\textbf{DATM}} &
      \multicolumn{2}{c}{\textbf{EDF}} &
      \multicolumn{2}{c}{\textbf{MCT}} &
      \multicolumn{2}{c}{\textbf{MTT}} \\[0.2em]
    \textbf{Dataset} &
      \textit{Baseline} & \textit{+Ours} &
      \textit{Baseline} & \textit{+Ours} &
      \textit{Baseline} & \textit{+Ours} &
      \textit{Baseline} & \textit{+Ours} \\[0.3em]
    \hline

    \rotatebox[origin=l]{90}{\textbf{ETTh1}} &
    \includegraphics[width=0.11\linewidth]{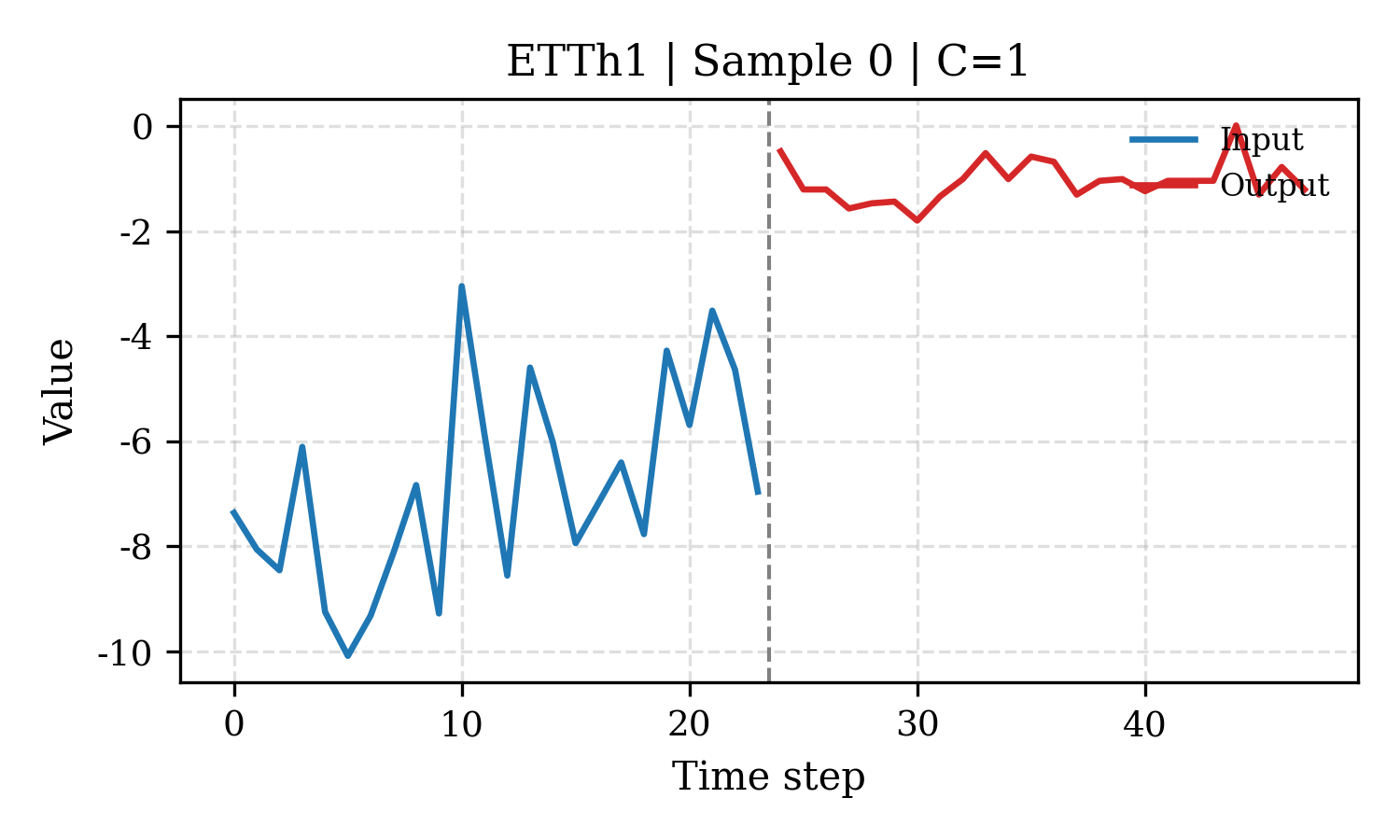} &
    \includegraphics[width=0.11\linewidth]{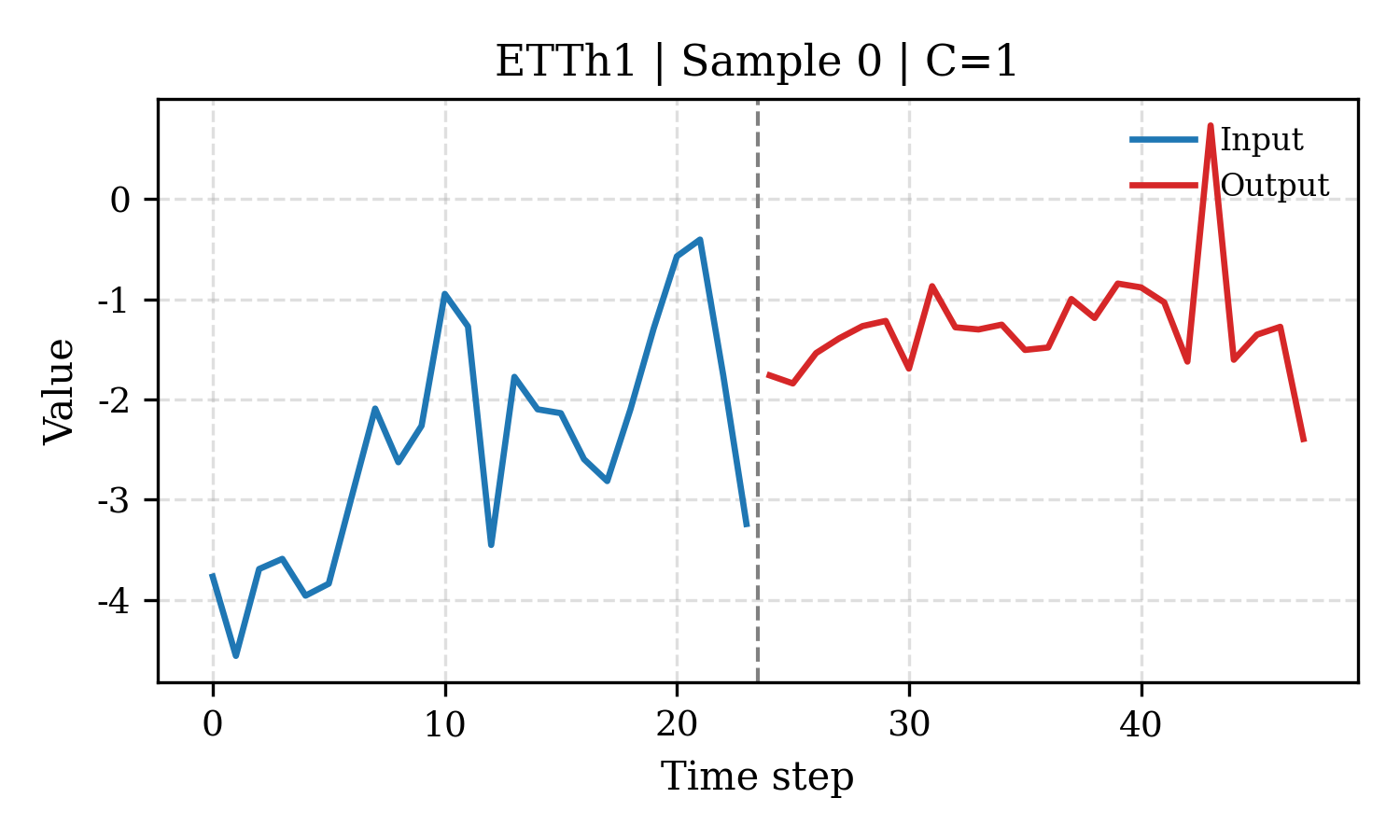} &
    \includegraphics[width=0.11\linewidth]{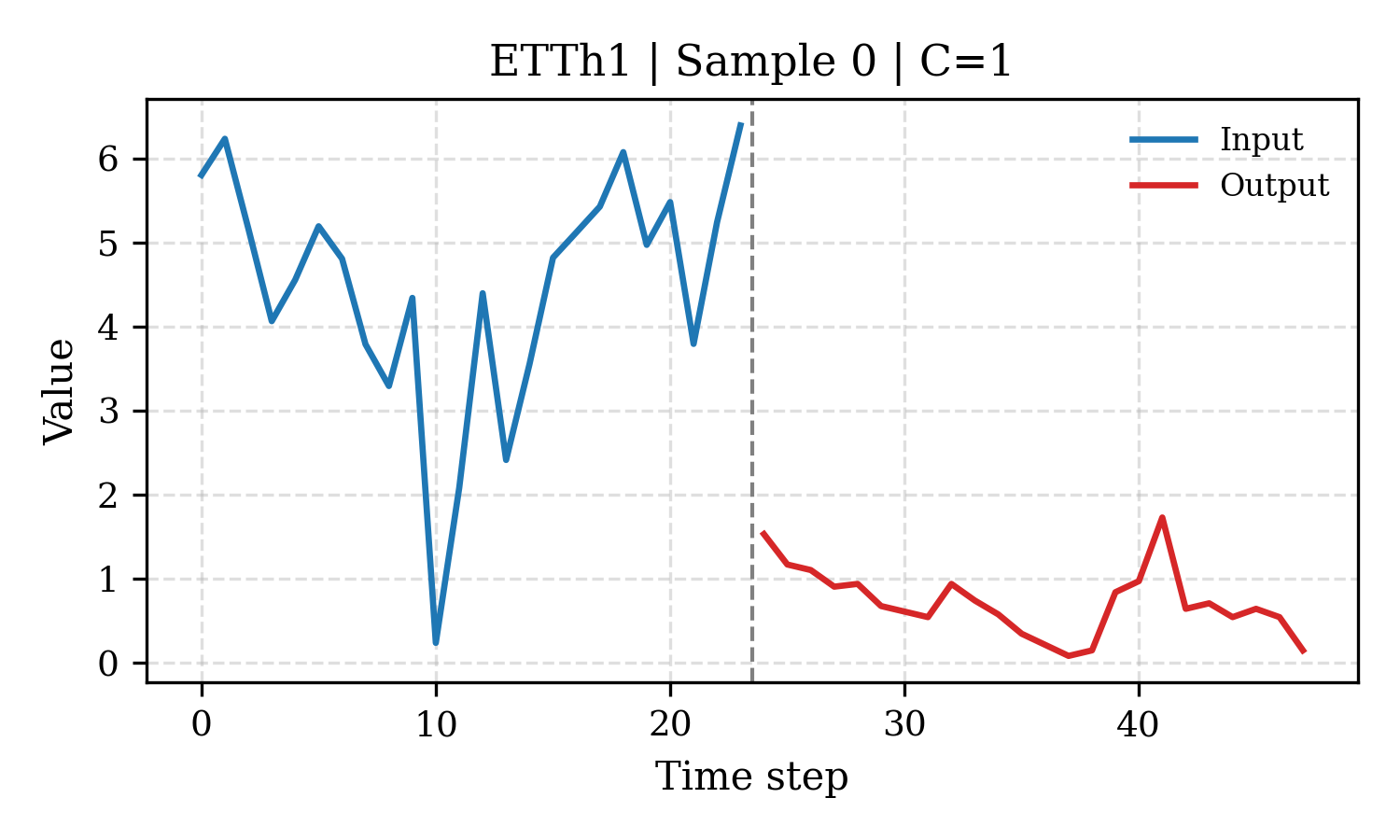} &
    \includegraphics[width=0.11\linewidth]{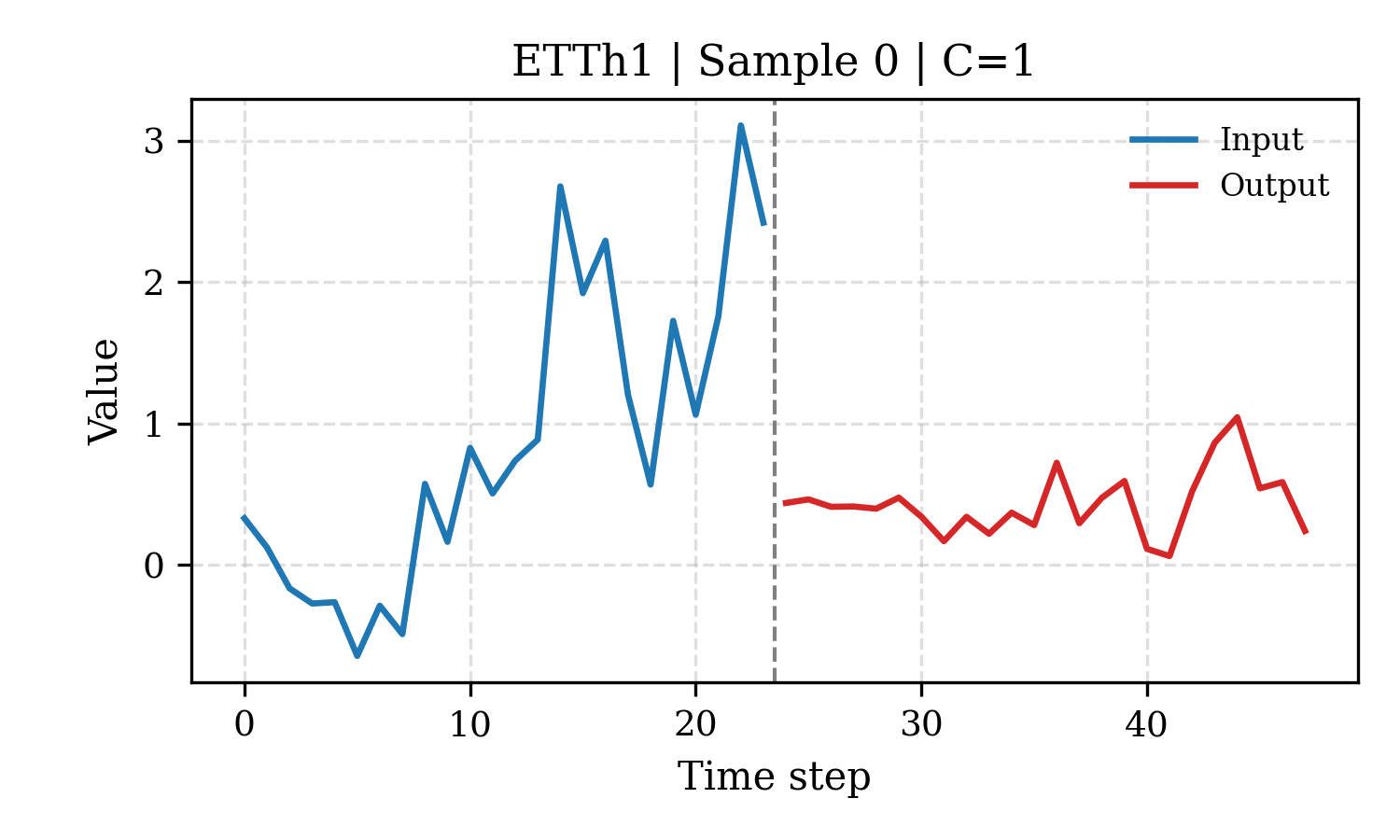} &
    \includegraphics[width=0.11\linewidth]{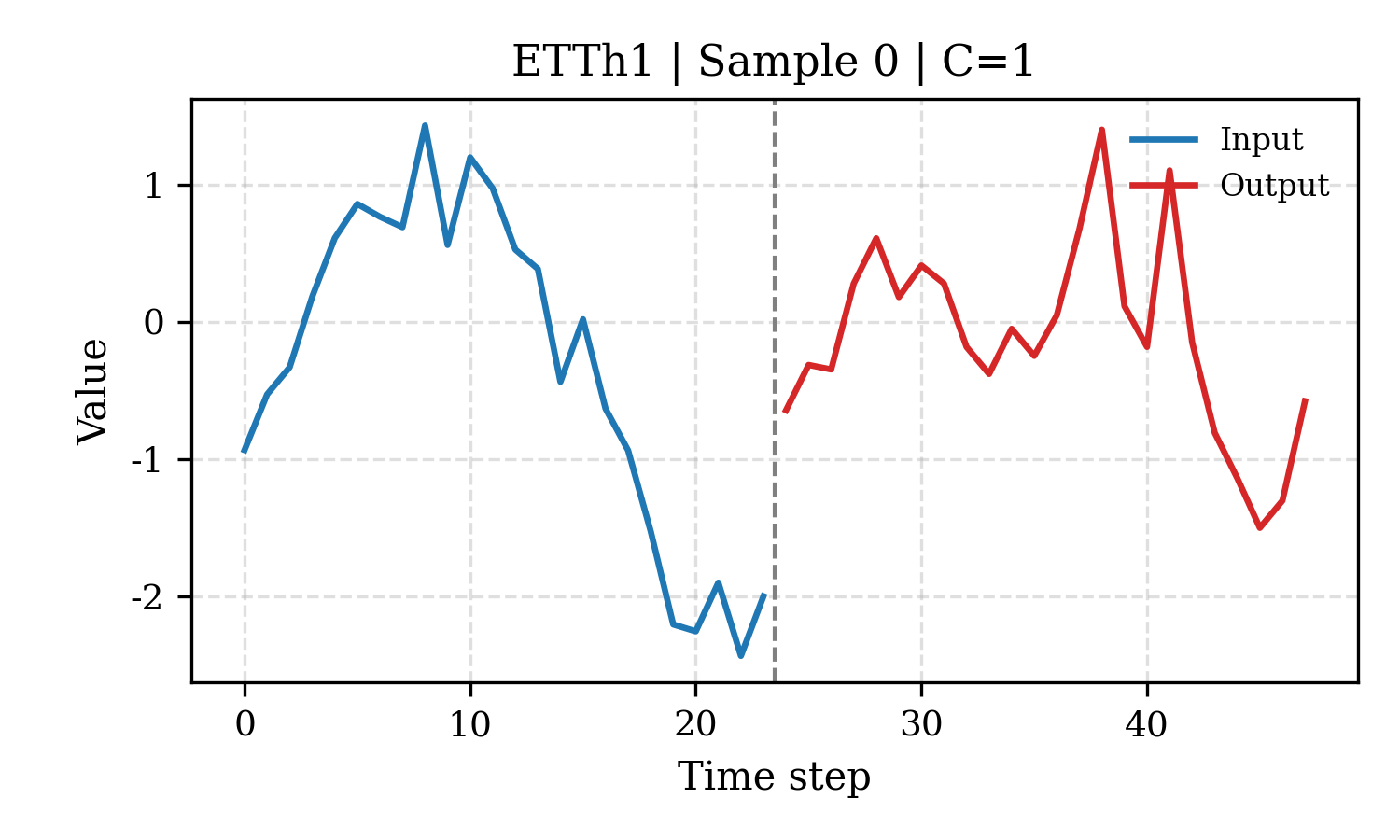} &
    \includegraphics[width=0.11\linewidth]{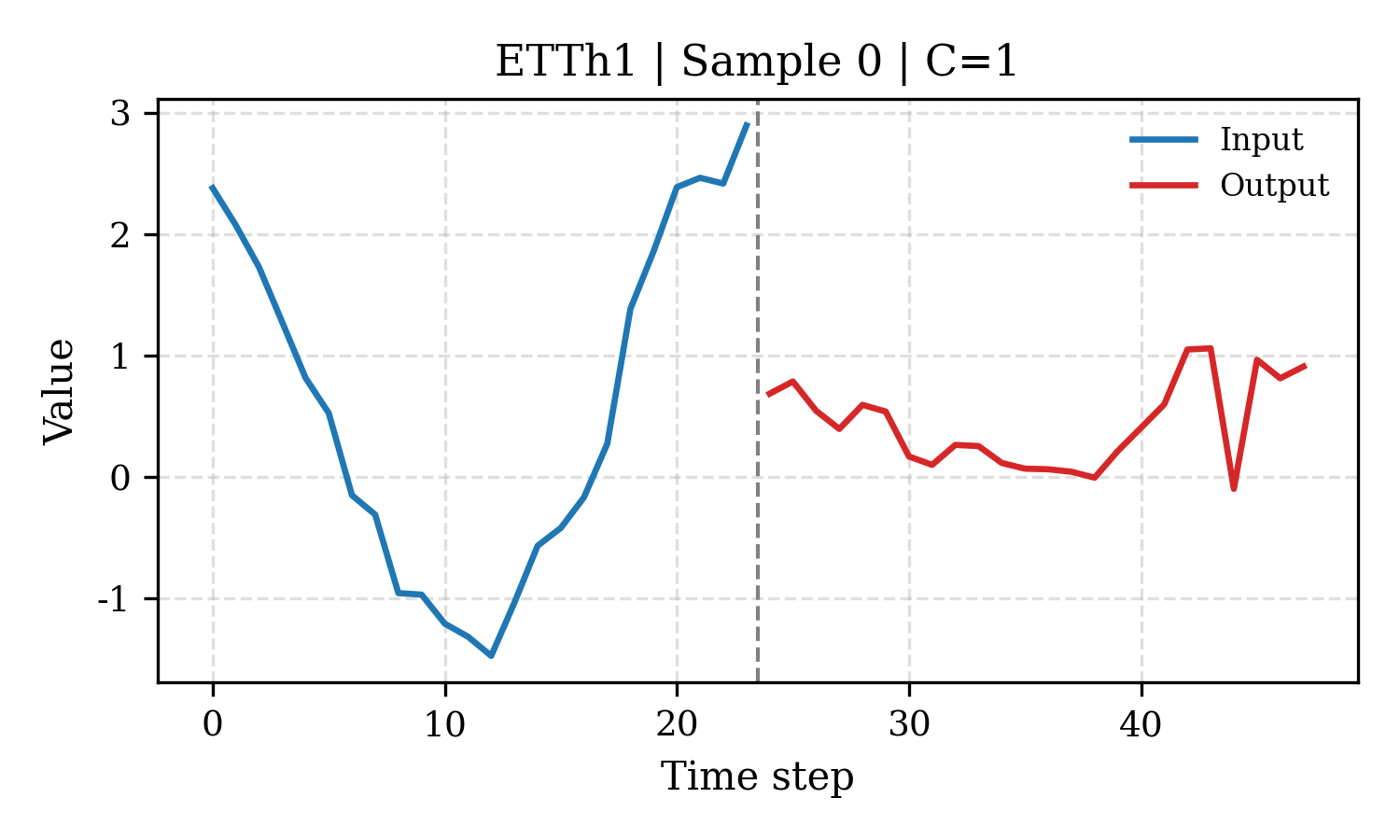} &
    \includegraphics[width=0.11\linewidth]{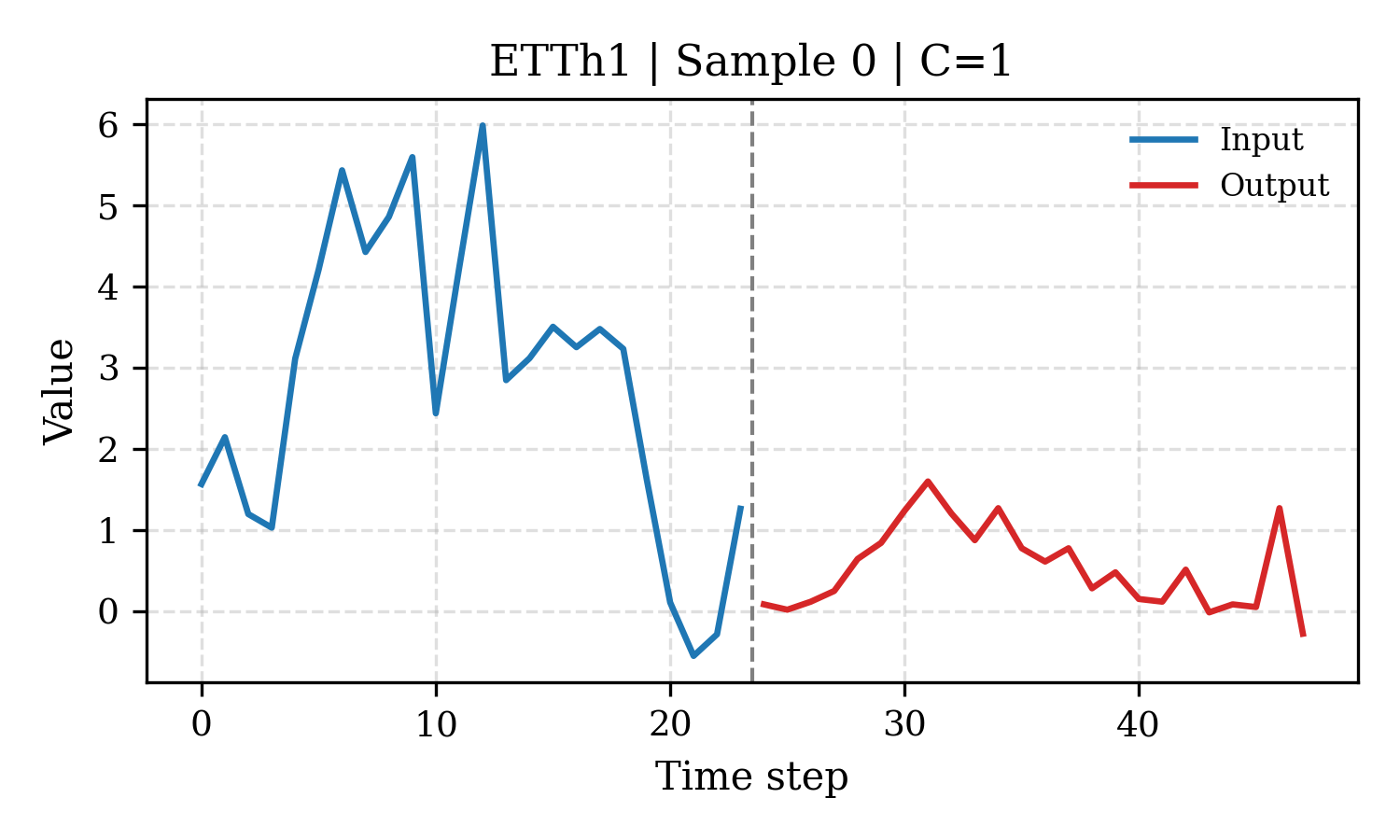} &
    \includegraphics[width=0.11\linewidth]{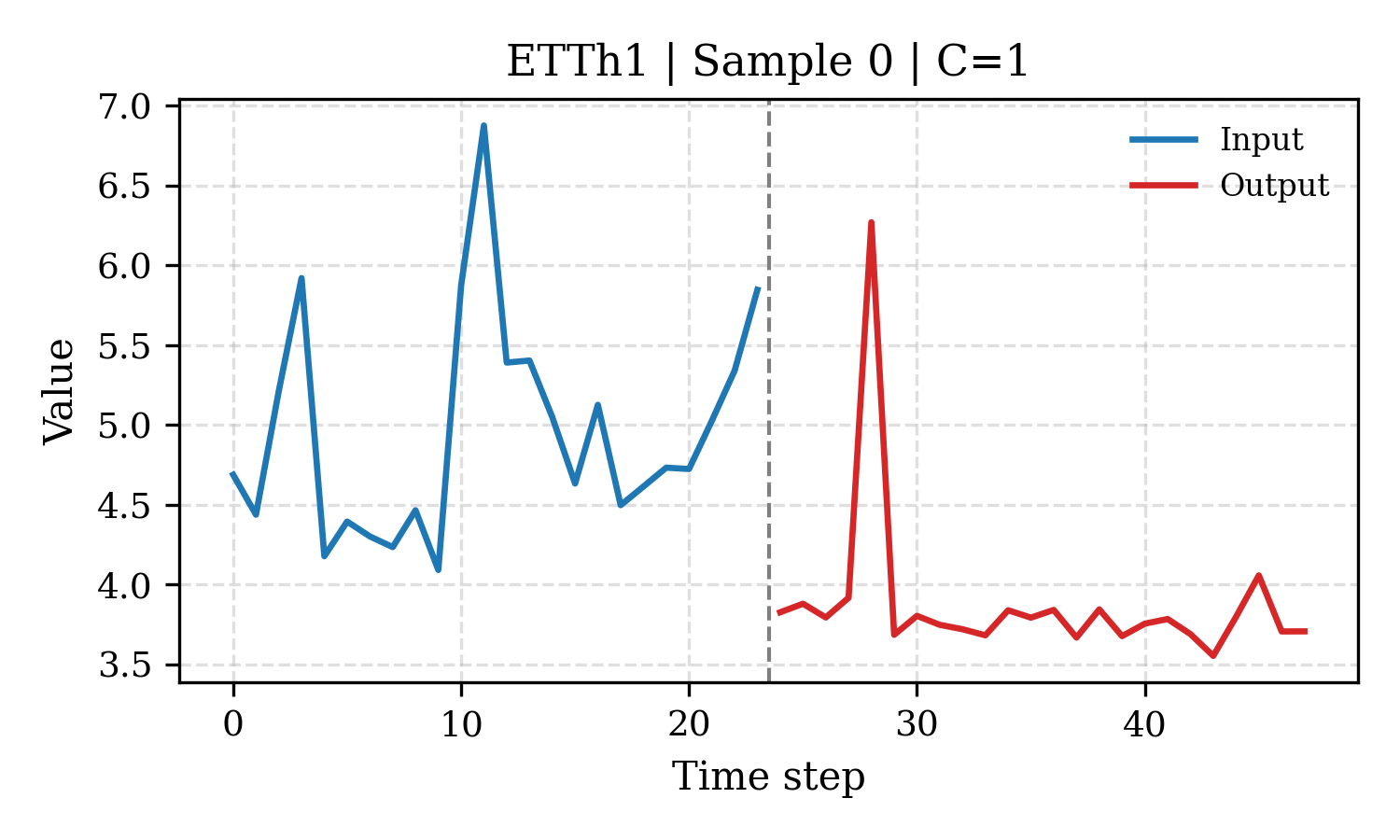} \\[0.3em]

    \rotatebox[origin=l]{90}{\textbf{ETTh2}} &
    \includegraphics[width=0.11\linewidth]{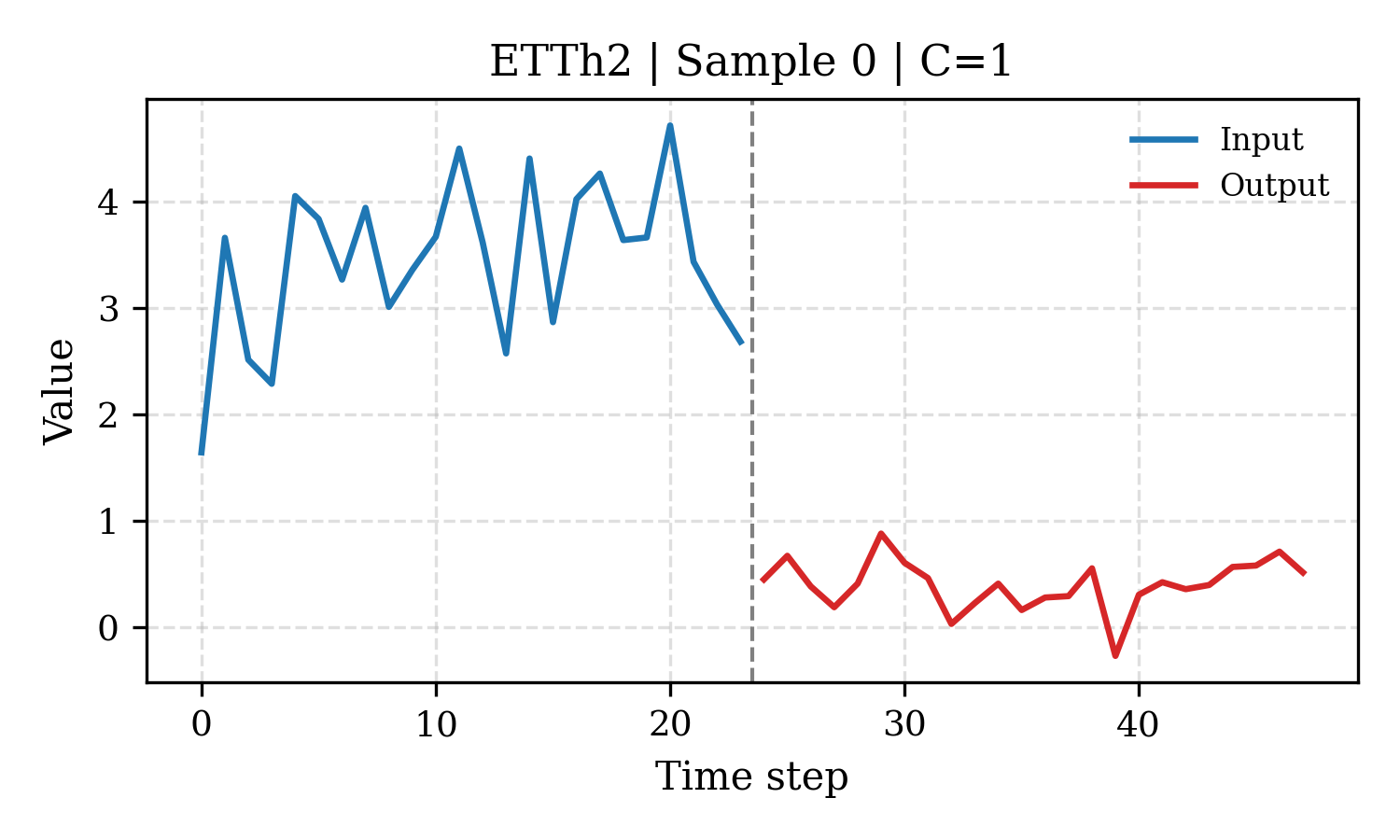} &
    \includegraphics[width=0.11\linewidth]{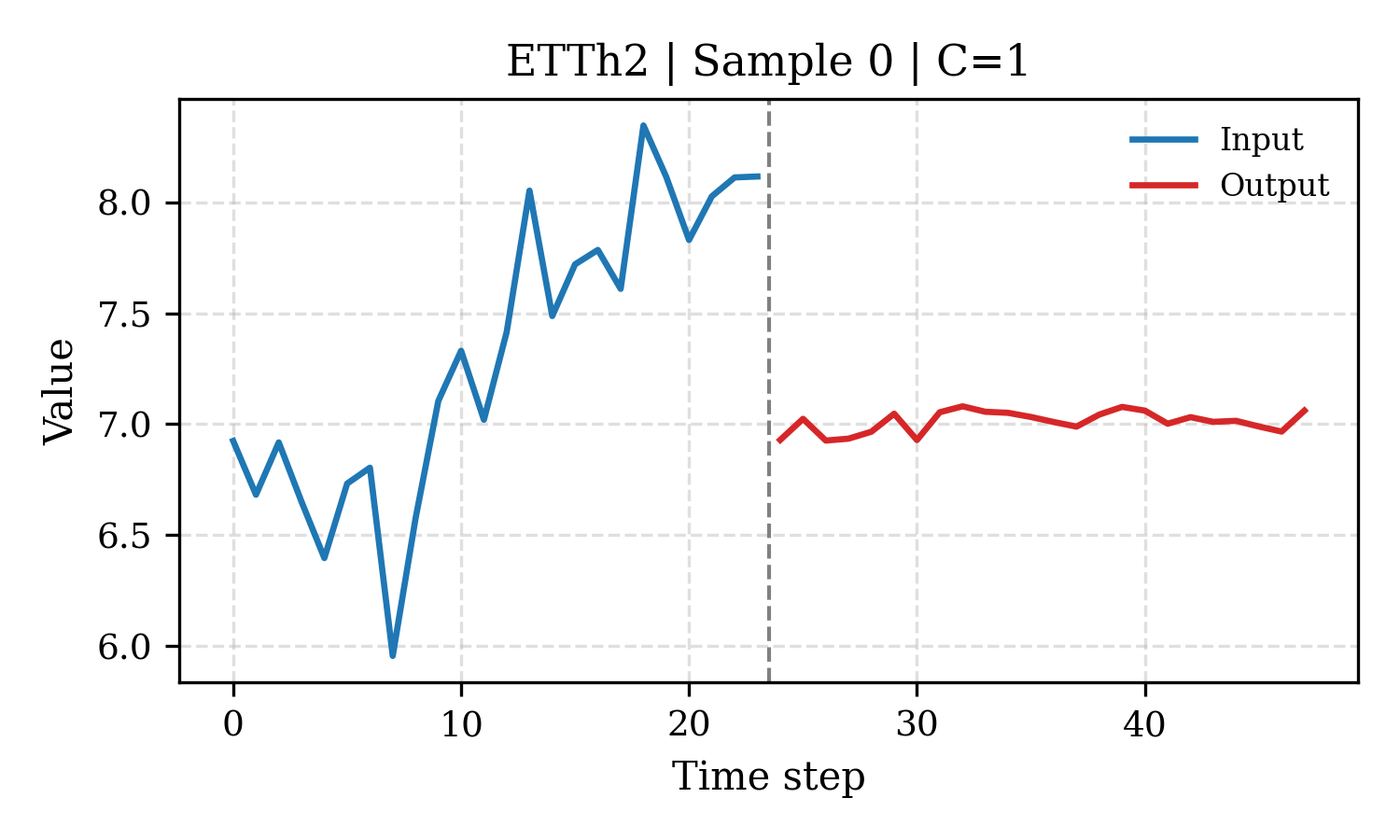} &
    \includegraphics[width=0.11\linewidth]{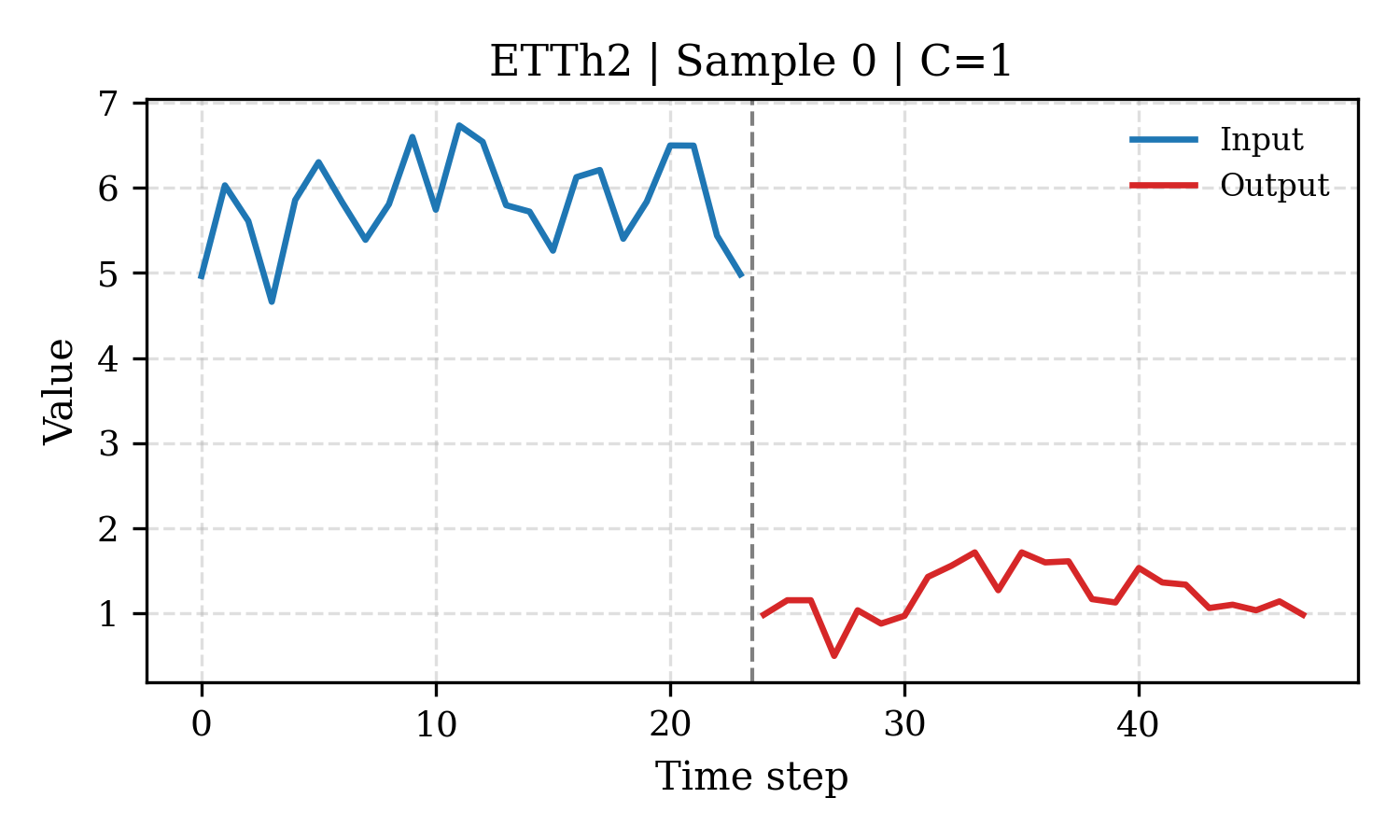} &
    \includegraphics[width=0.11\linewidth]{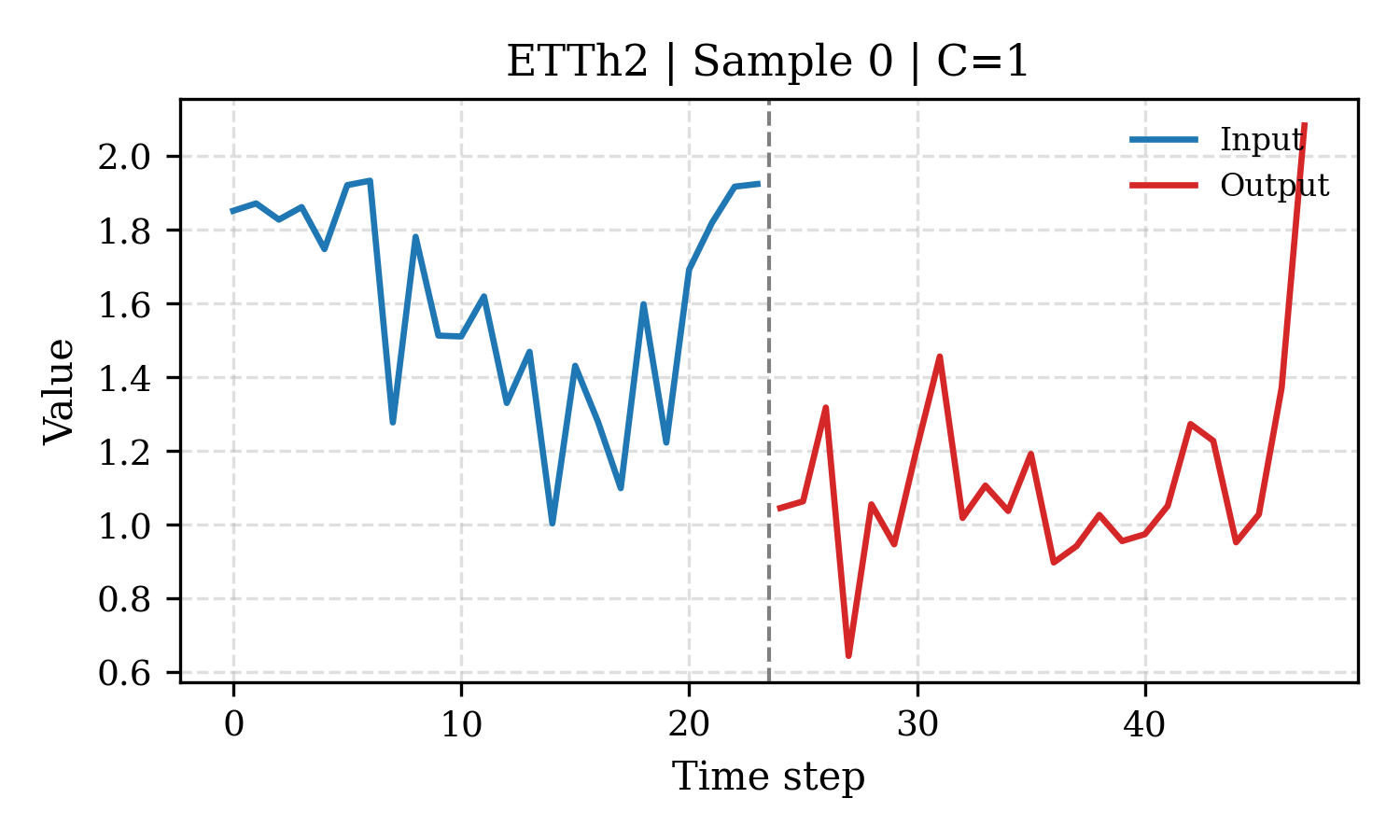} &
    \includegraphics[width=0.11\linewidth]{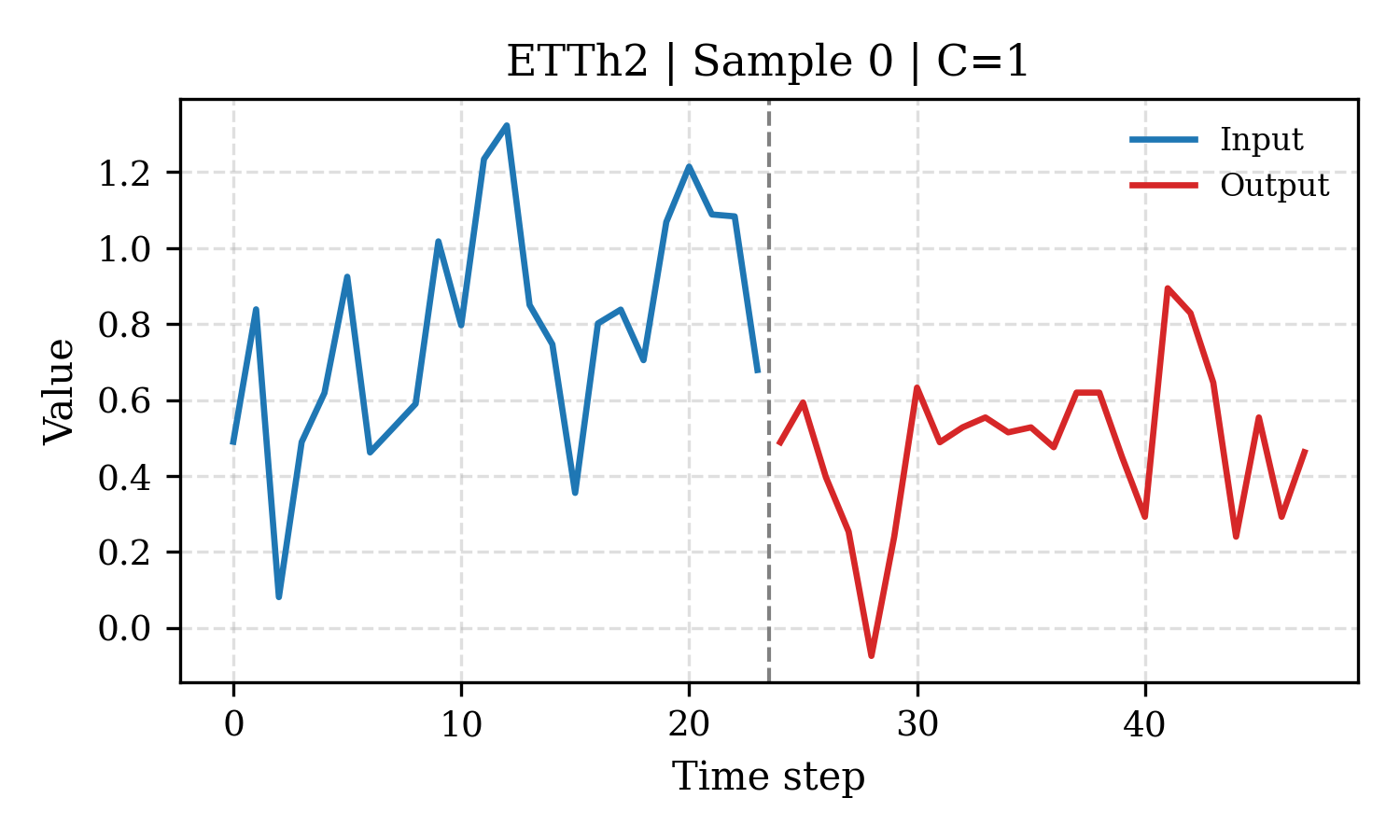} &
    \includegraphics[width=0.11\linewidth]{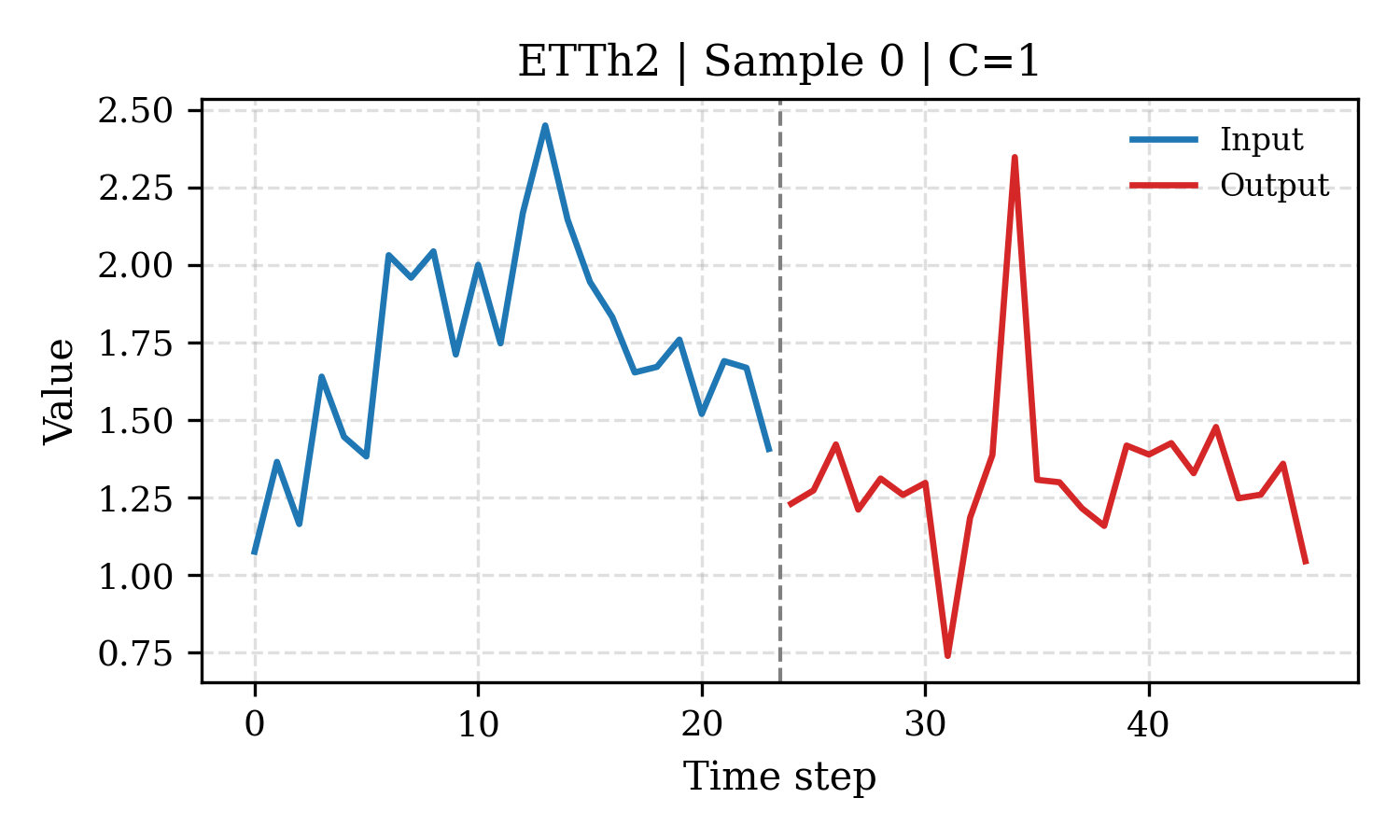} &
    \includegraphics[width=0.11\linewidth]{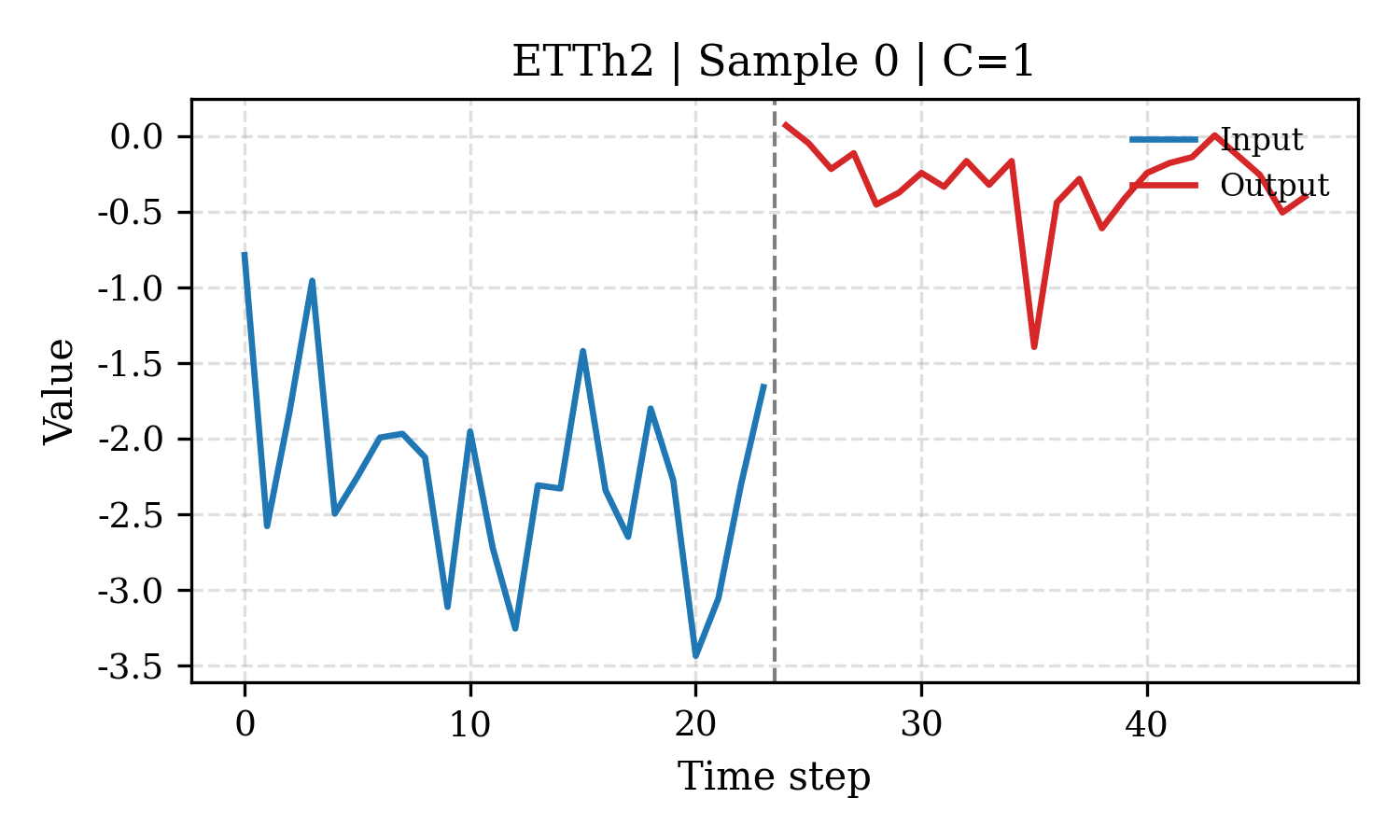} &
    \includegraphics[width=0.11\linewidth]{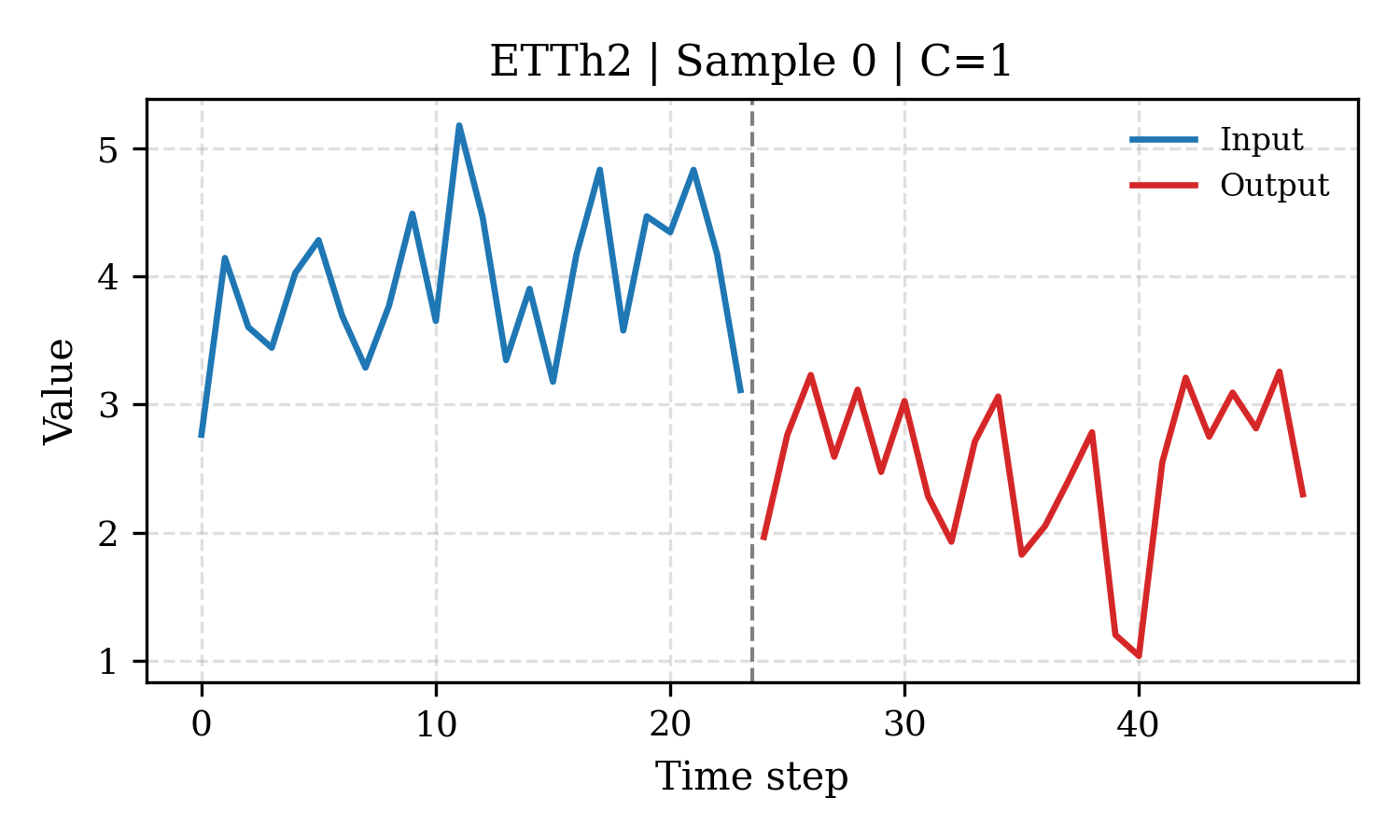} \\[0.3em]

    \rotatebox[origin=l]{90}{\textbf{ETTm1}} &
    \includegraphics[width=0.11\linewidth]{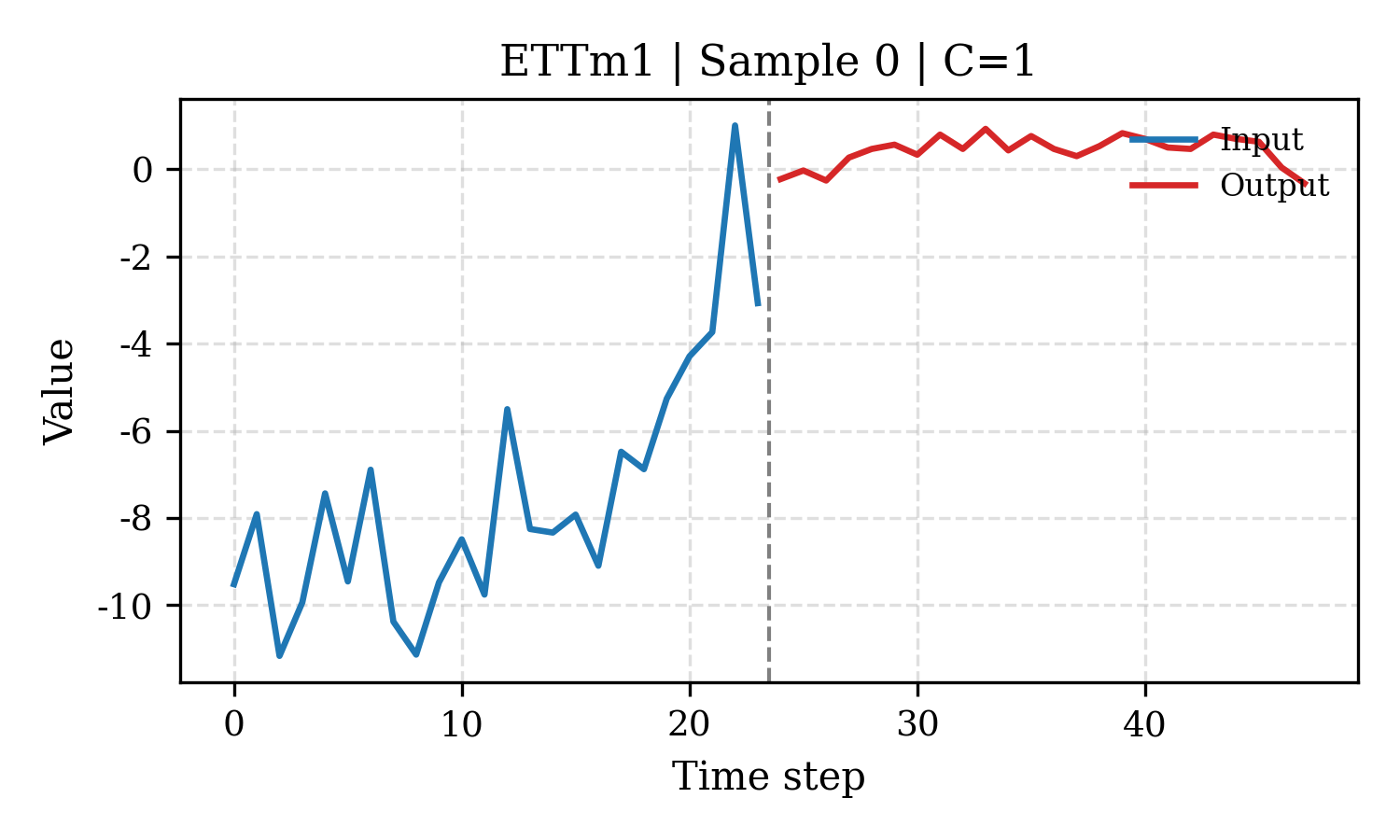} &
    \includegraphics[width=0.11\linewidth]{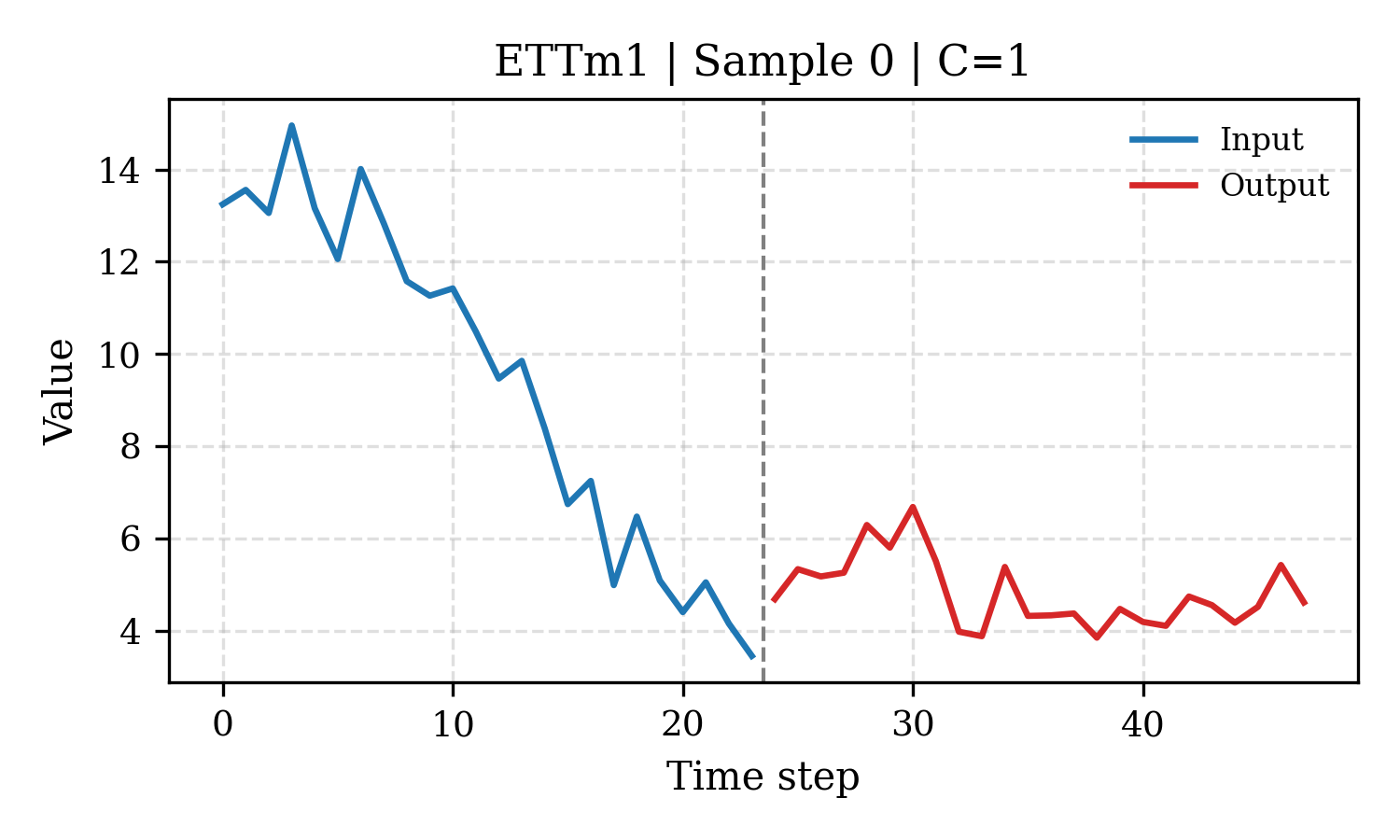} &
    \includegraphics[width=0.11\linewidth]{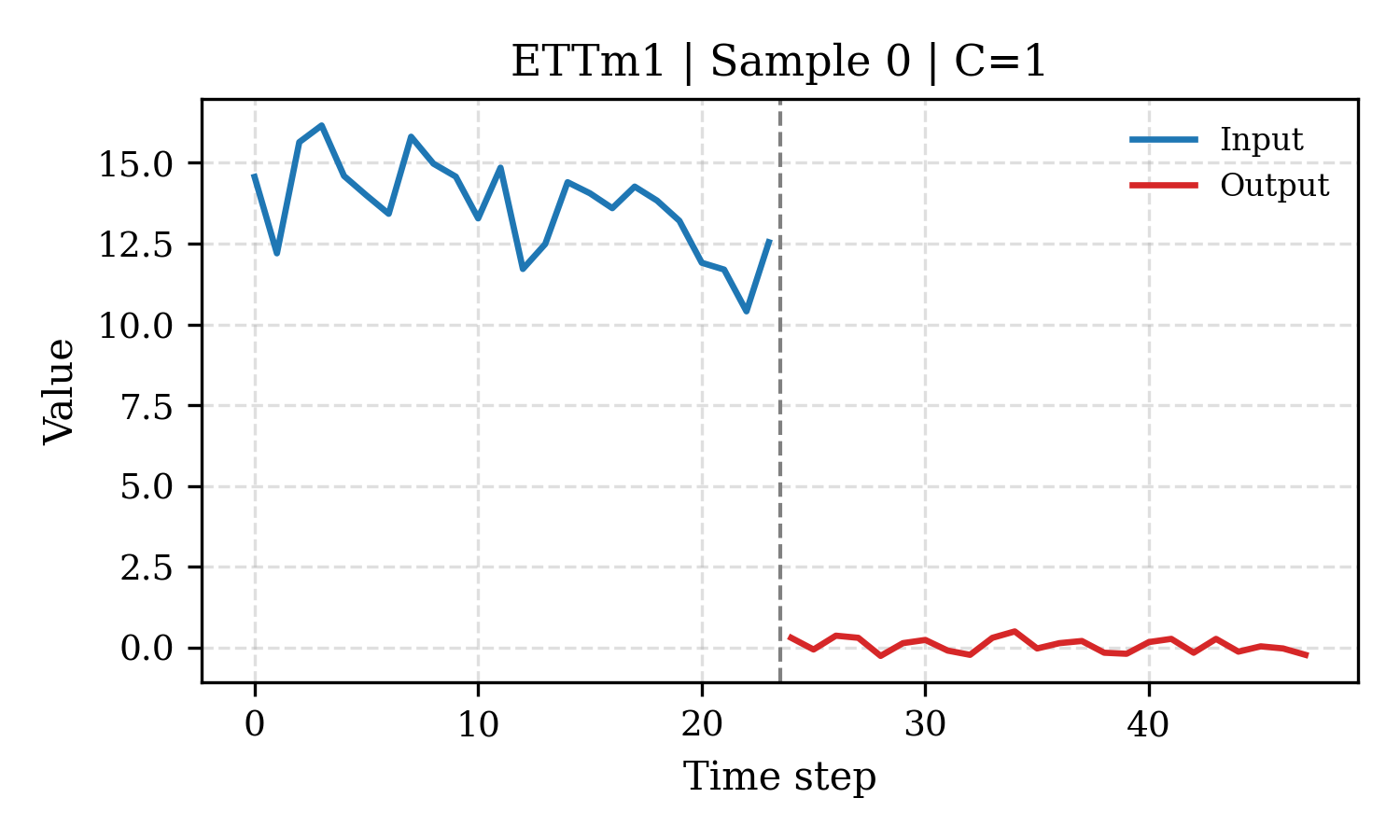} &
    \includegraphics[width=0.11\linewidth]{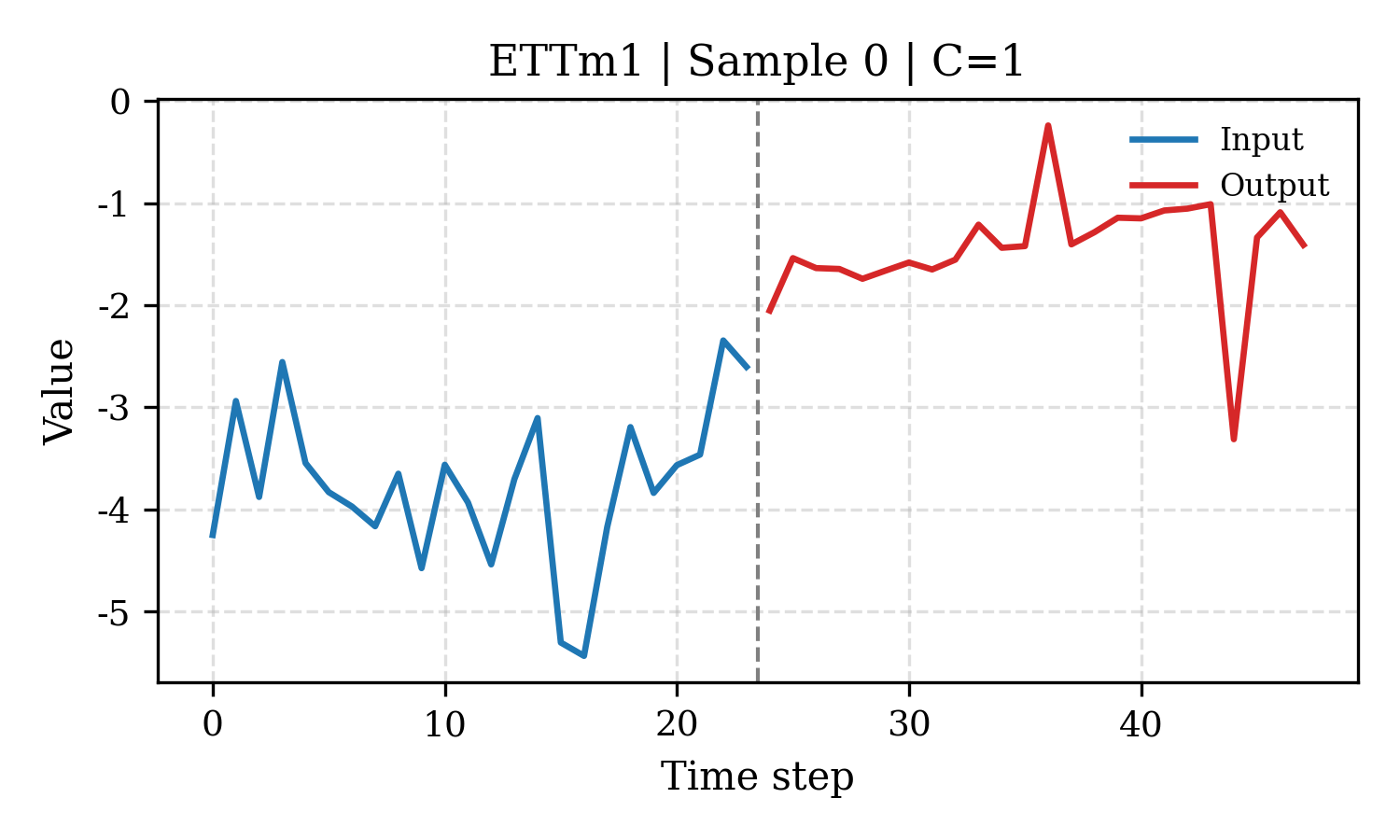} &
    \includegraphics[width=0.11\linewidth]{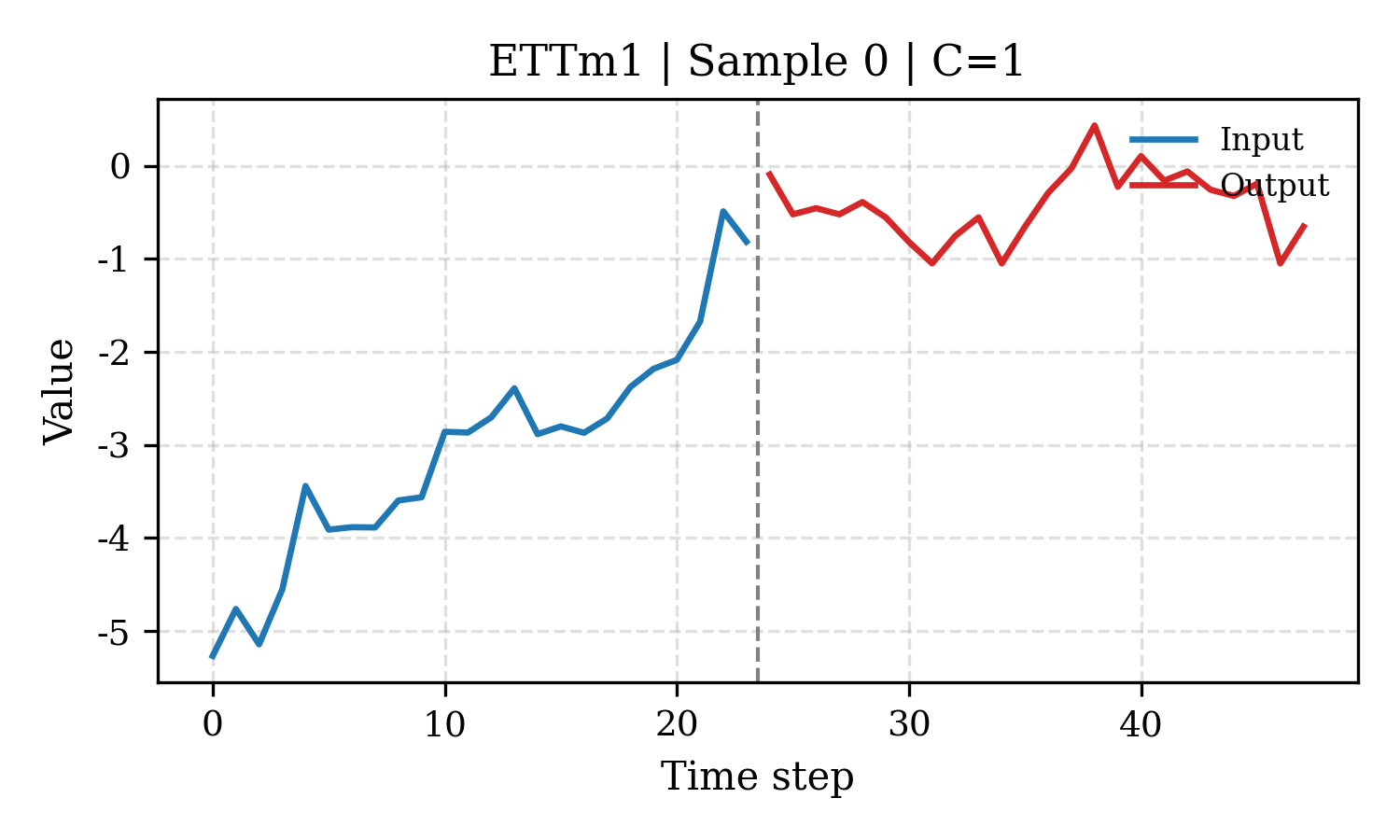} &
    \includegraphics[width=0.11\linewidth]{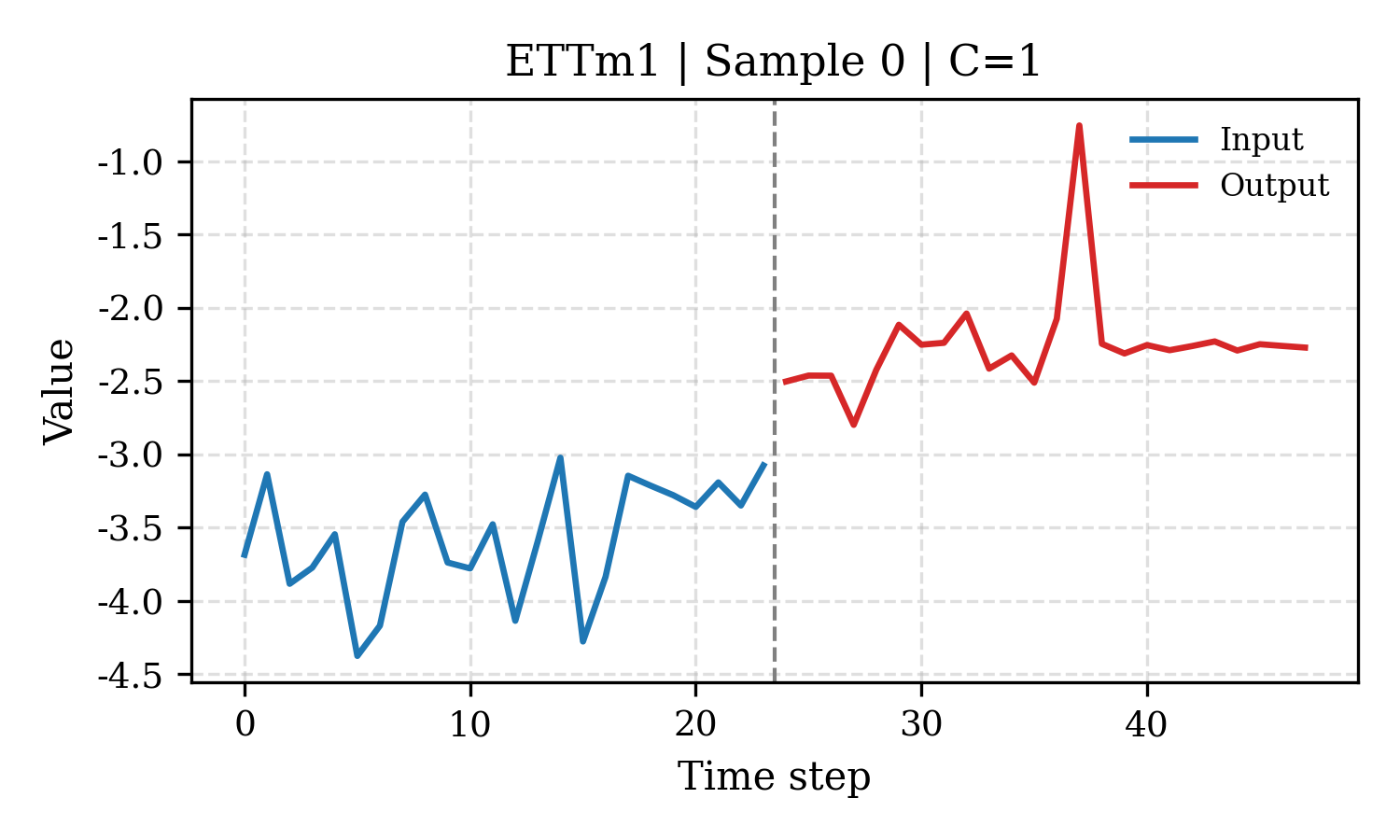} &
    \includegraphics[width=0.11\linewidth]{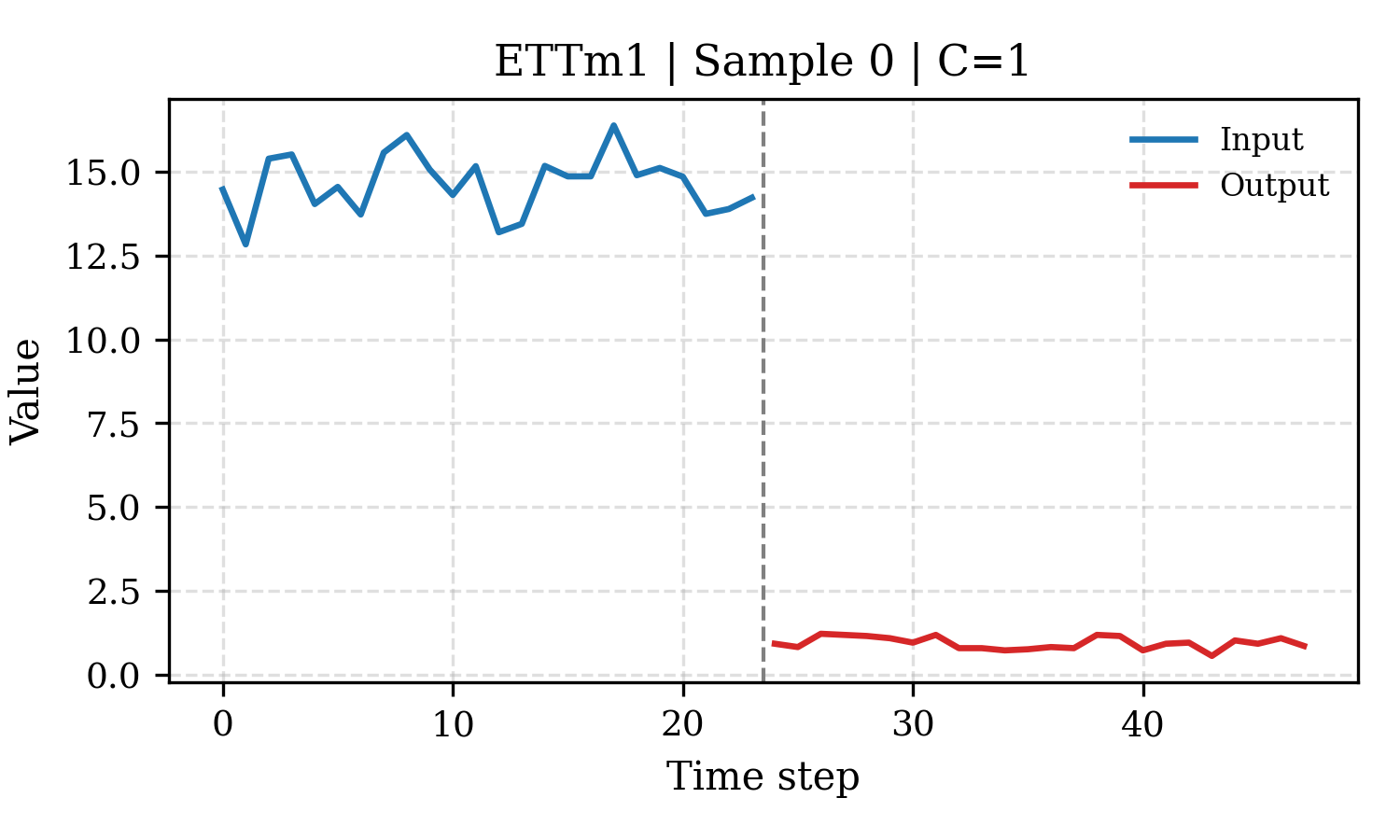} &
    \includegraphics[width=0.11\linewidth]{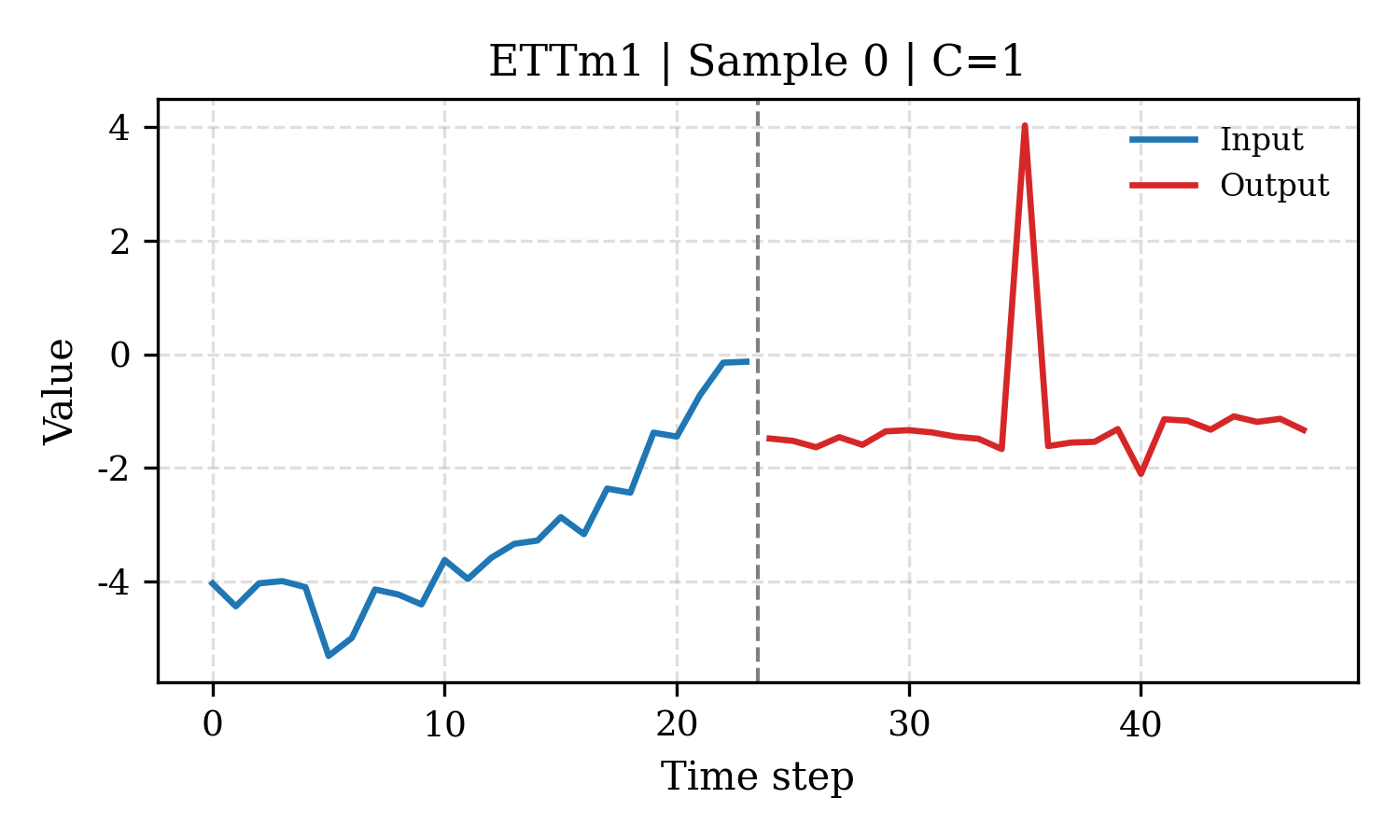} \\[0.3em]

    \rotatebox[origin=l]{90}{\textbf{ETTm2}} &
    \includegraphics[width=0.11\linewidth]{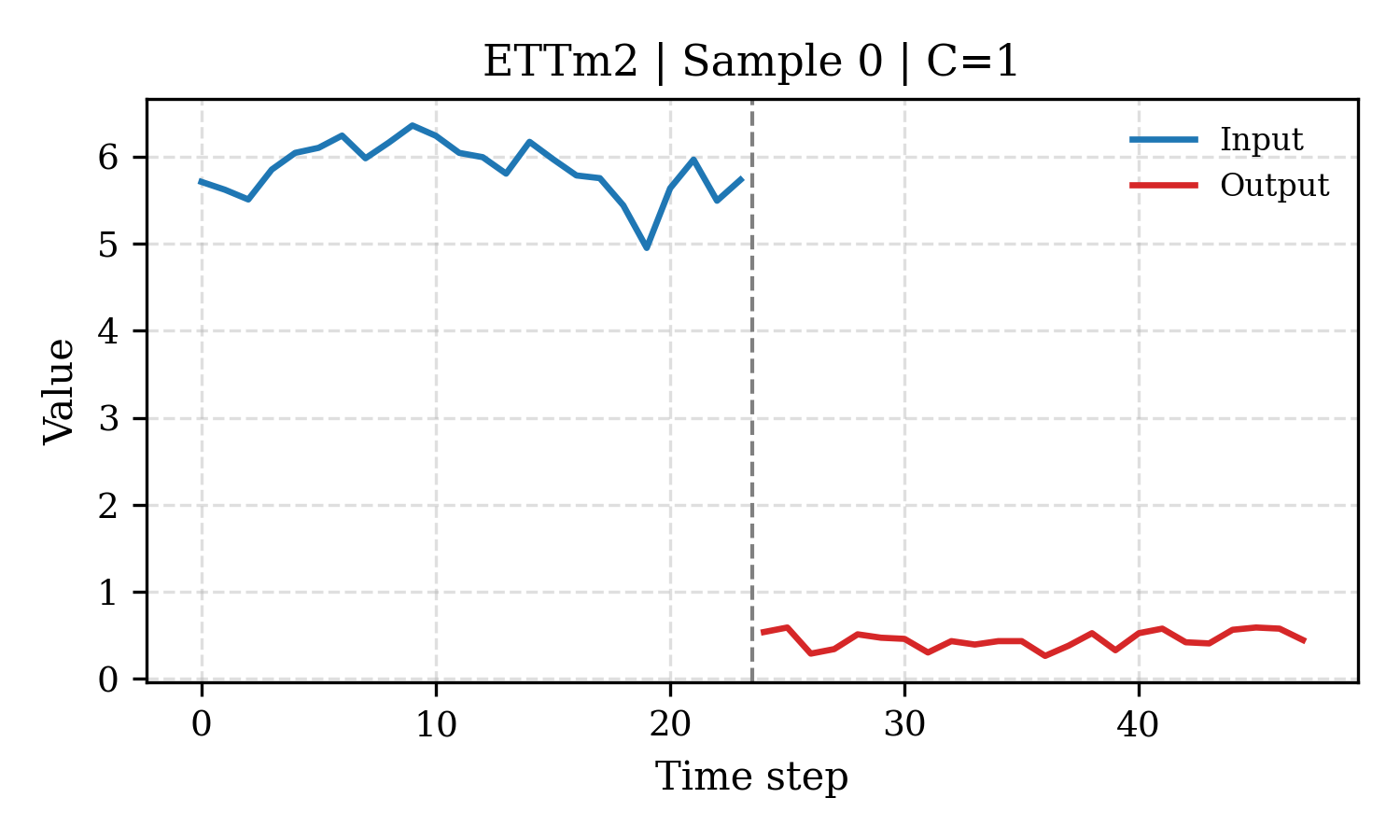} &
    \includegraphics[width=0.11\linewidth]{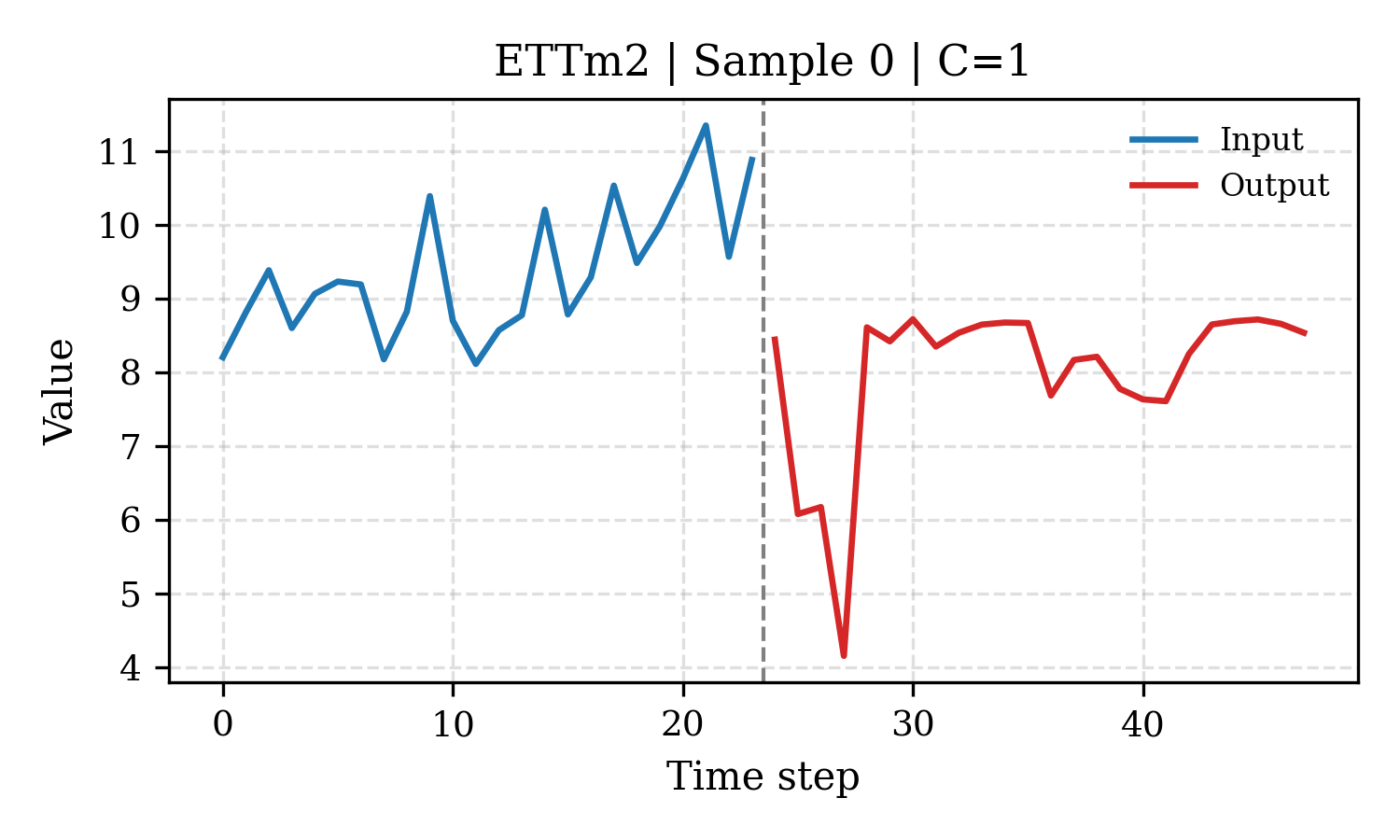} &
    \includegraphics[width=0.11\linewidth]{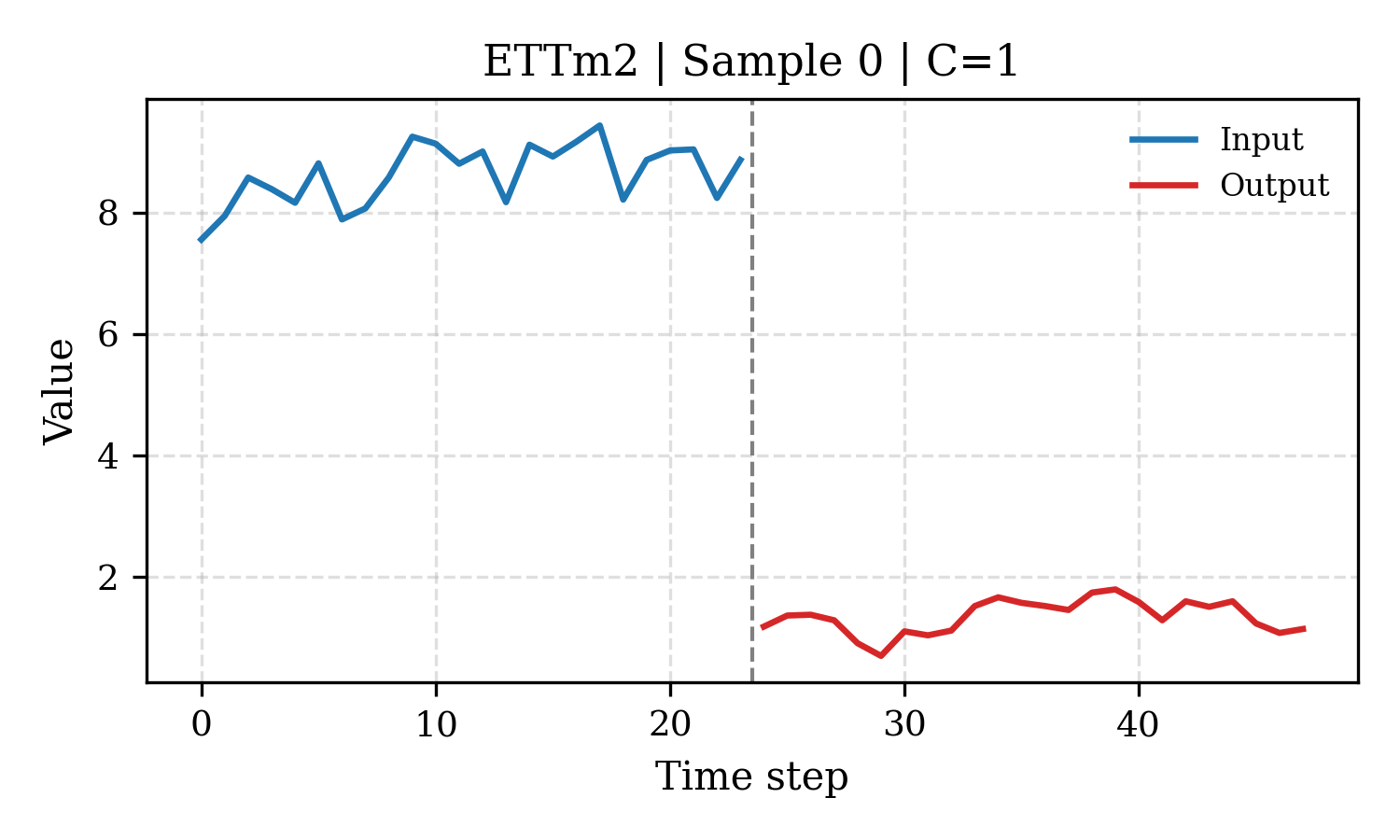} &
    \includegraphics[width=0.11\linewidth]{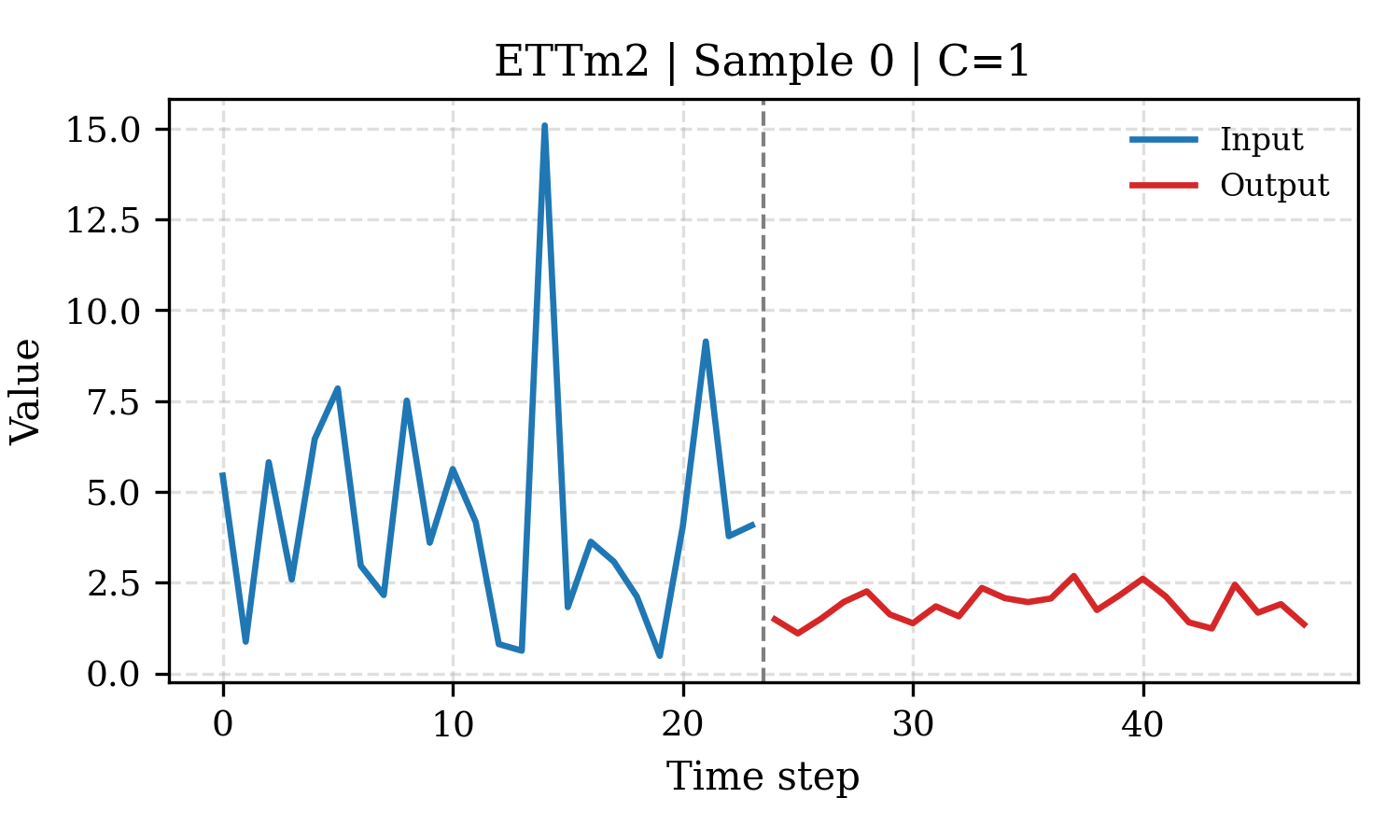} &
    \includegraphics[width=0.11\linewidth]{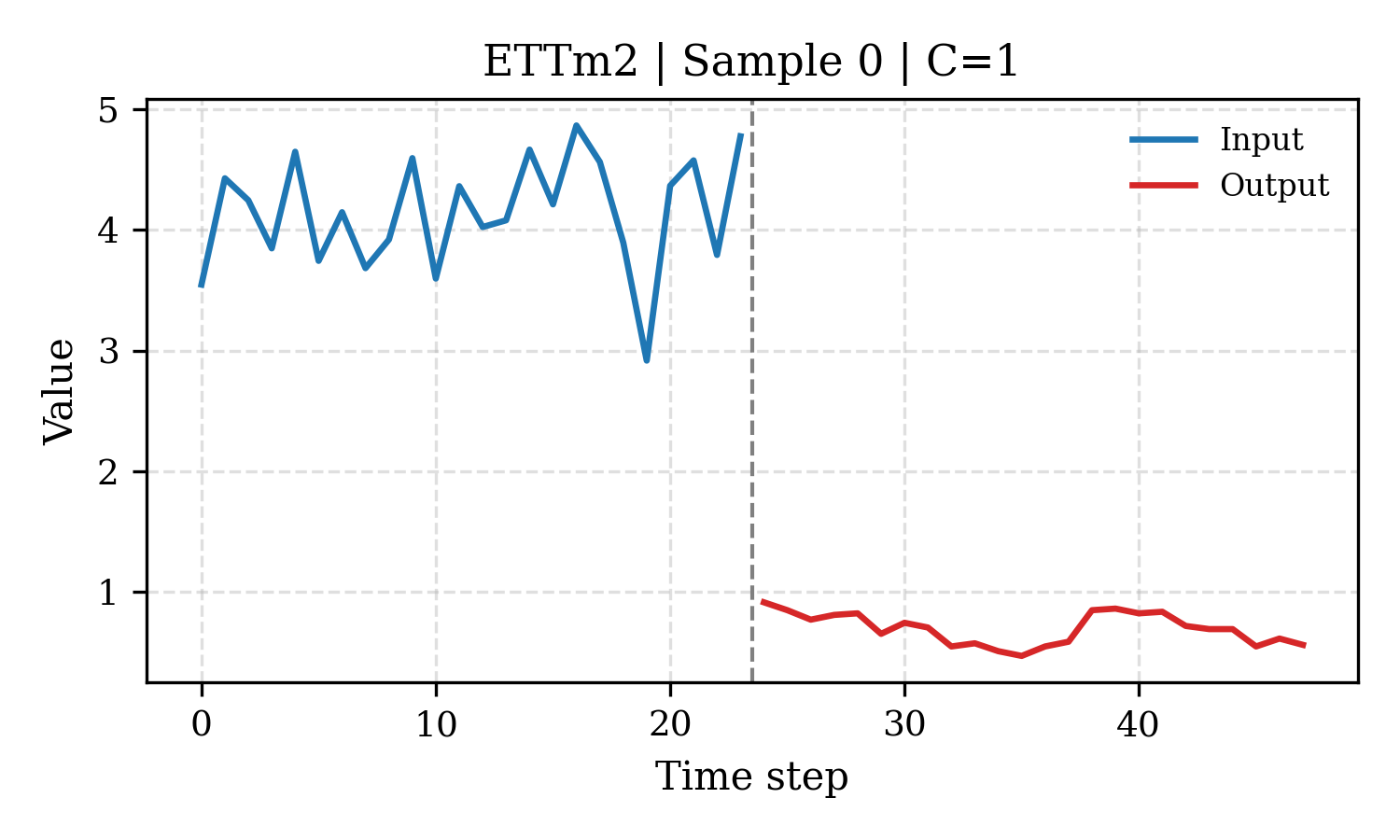} &
    \includegraphics[width=0.11\linewidth]{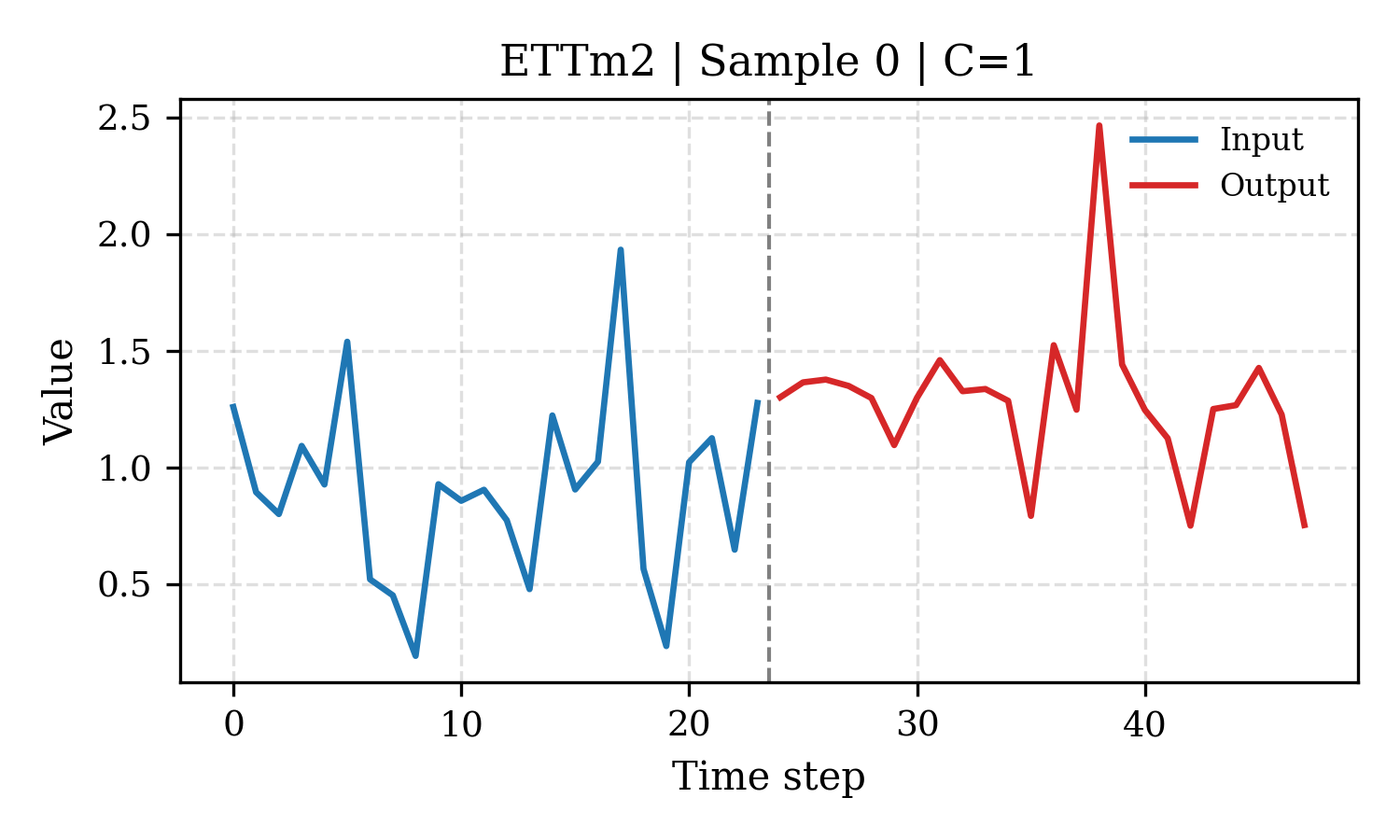} &
    \includegraphics[width=0.11\linewidth]{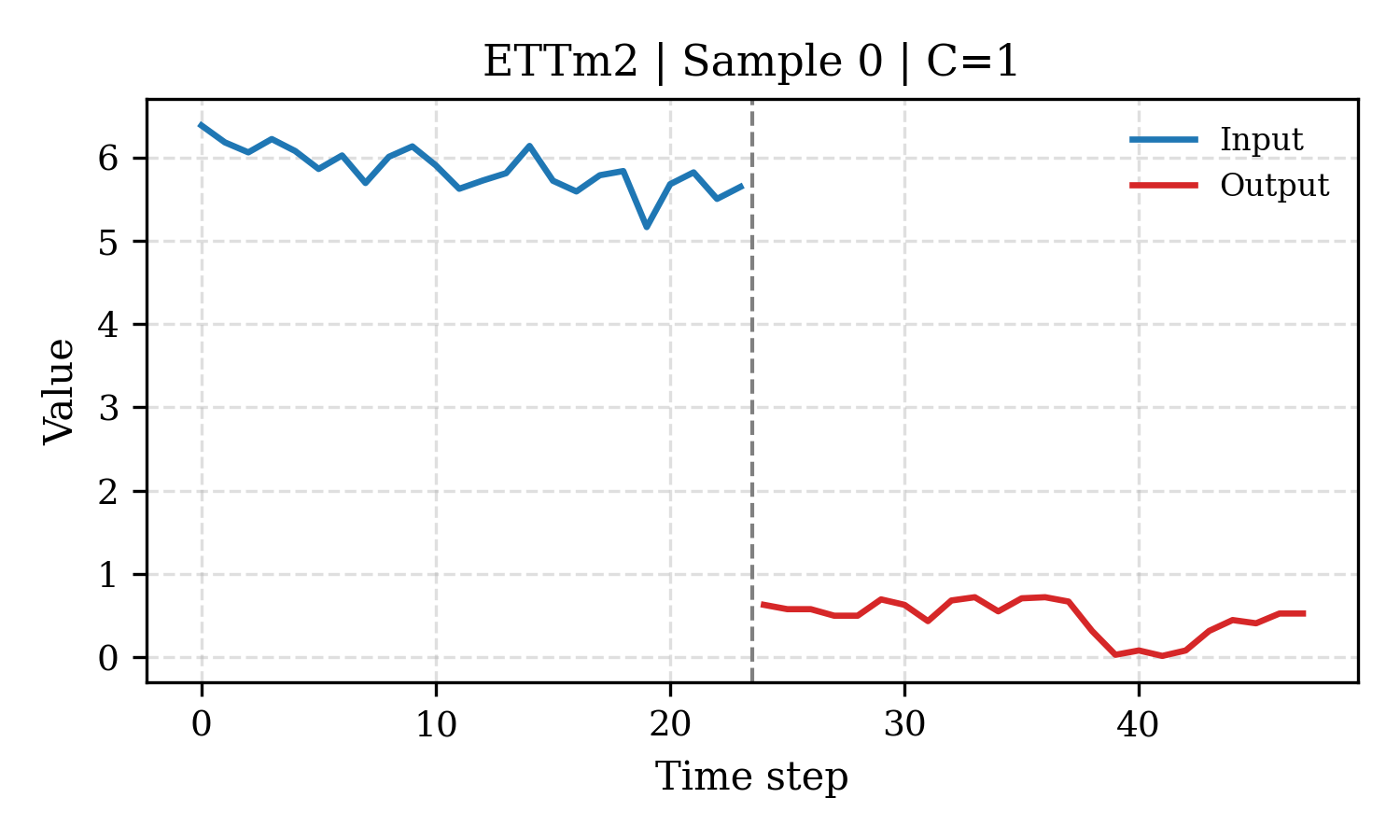} &
    \includegraphics[width=0.11\linewidth]{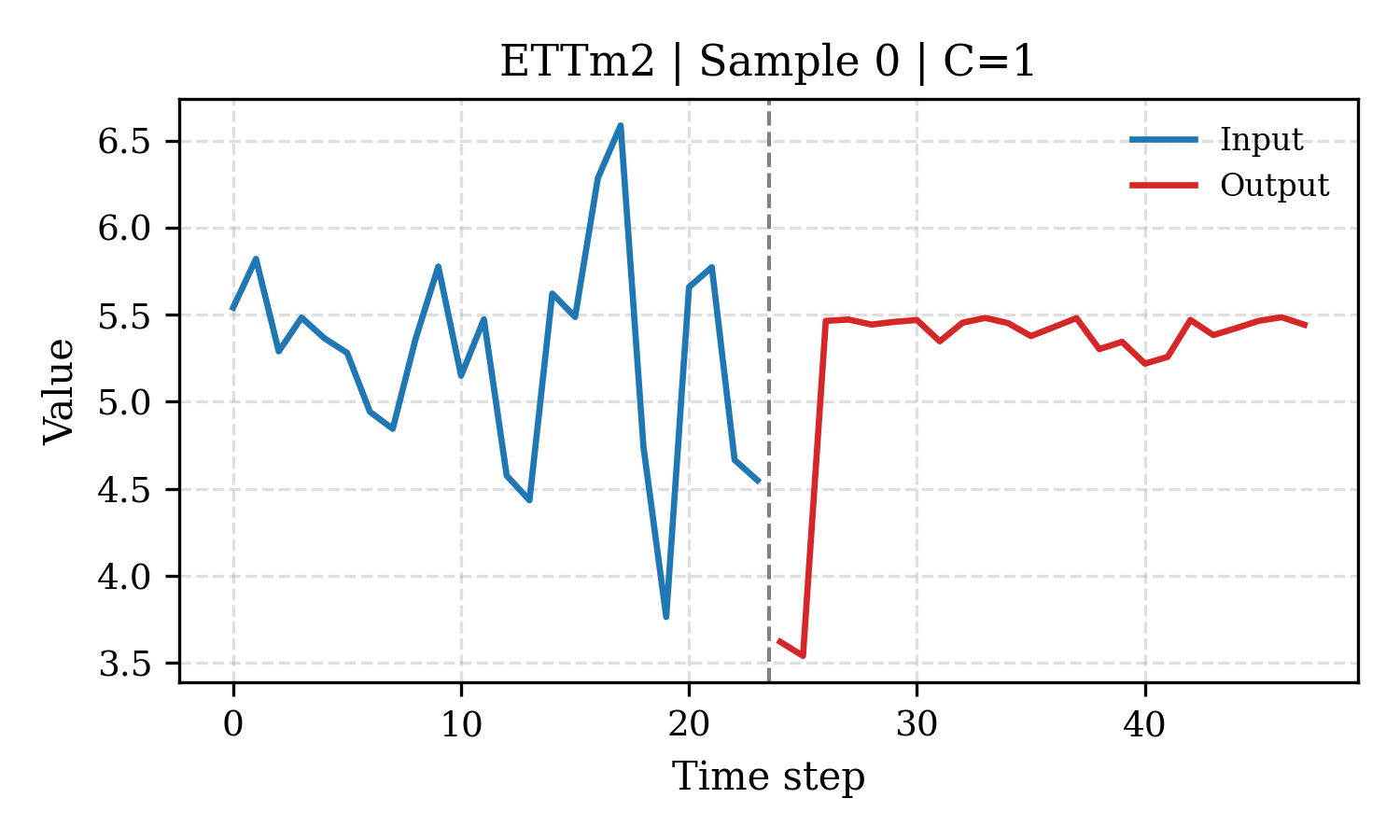} \\[0.3em]

    \rotatebox[origin=l]{90}{\textbf{Electricity}} &
    \includegraphics[width=0.11\linewidth]{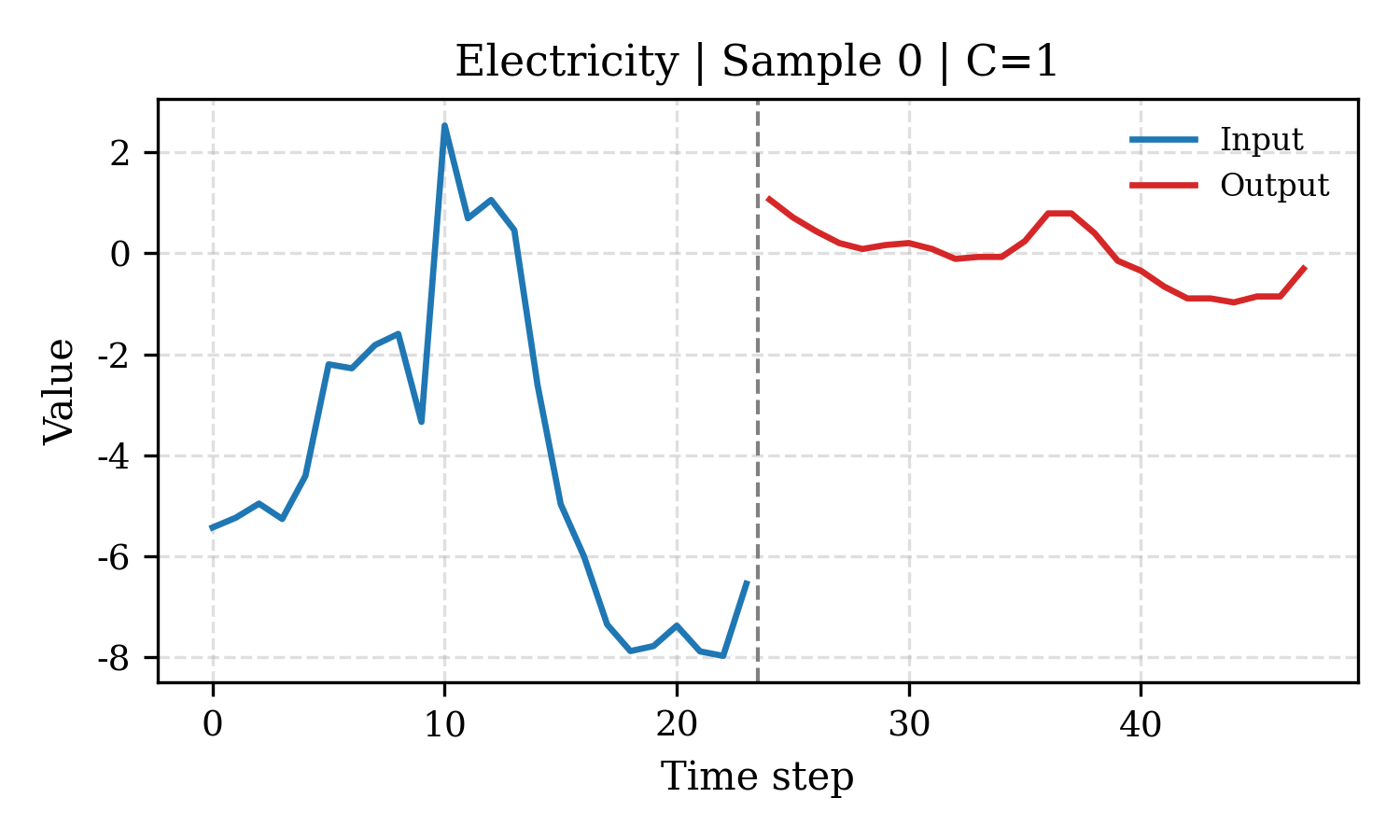} &
    \includegraphics[width=0.11\linewidth]{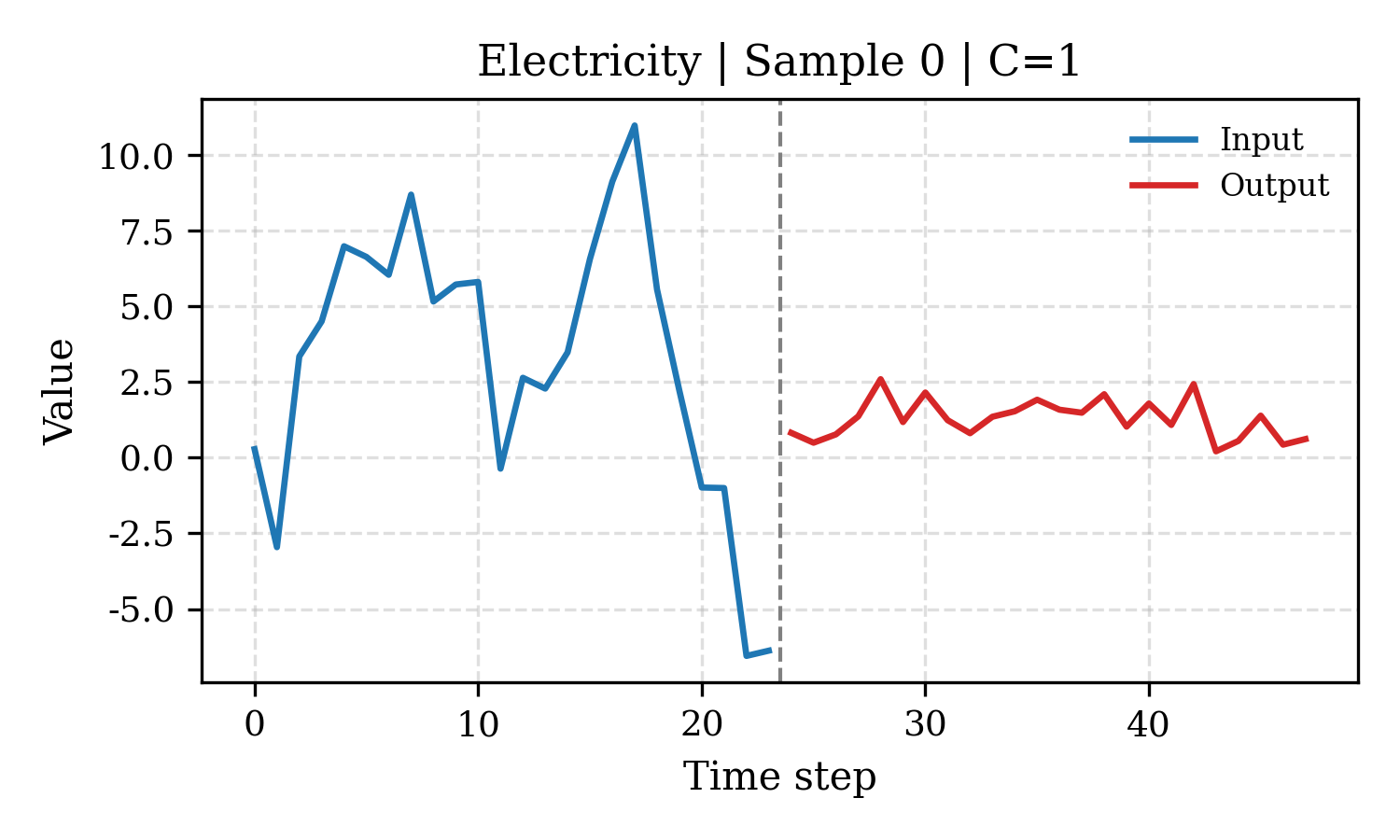} &
    \includegraphics[width=0.11\linewidth]{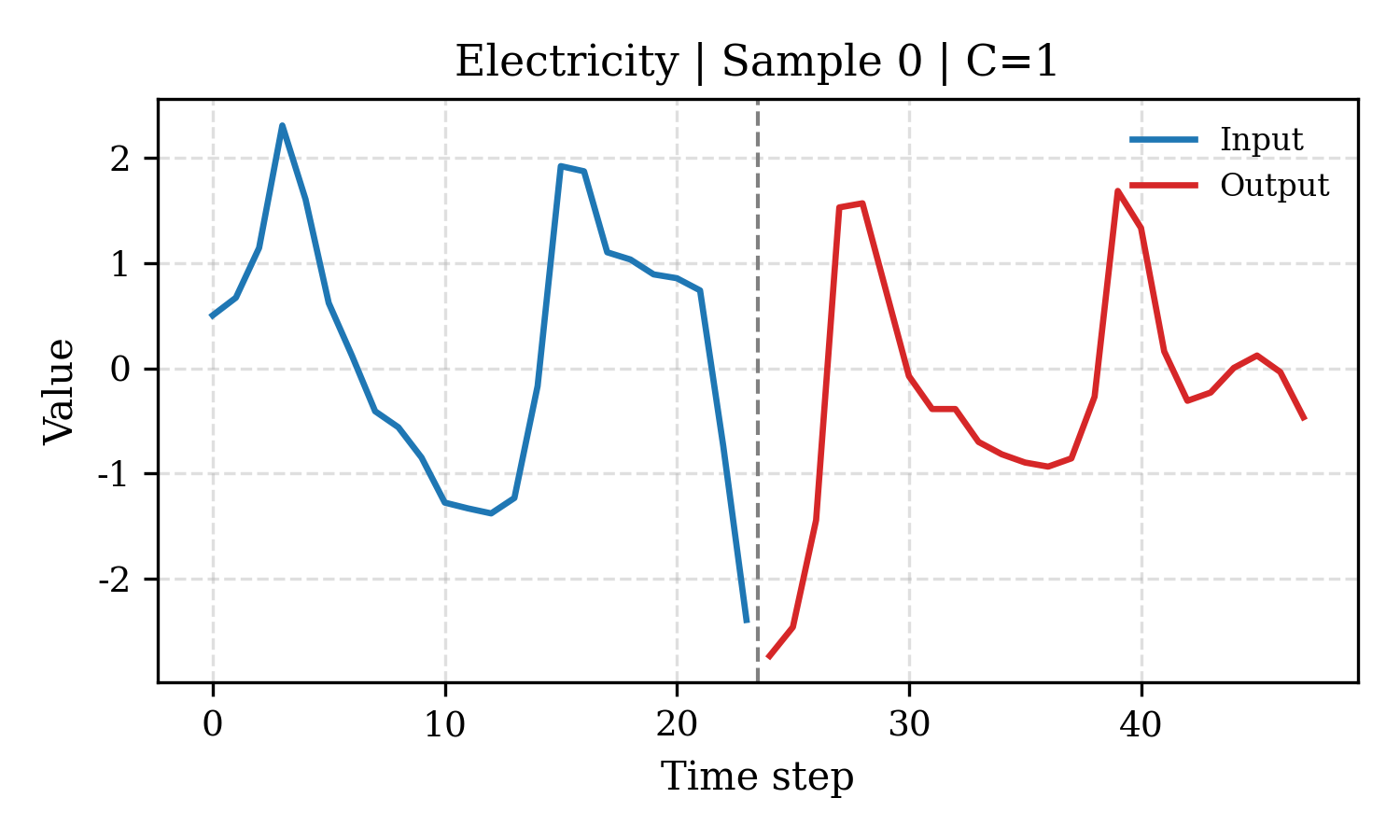} &
    \includegraphics[width=0.11\linewidth]{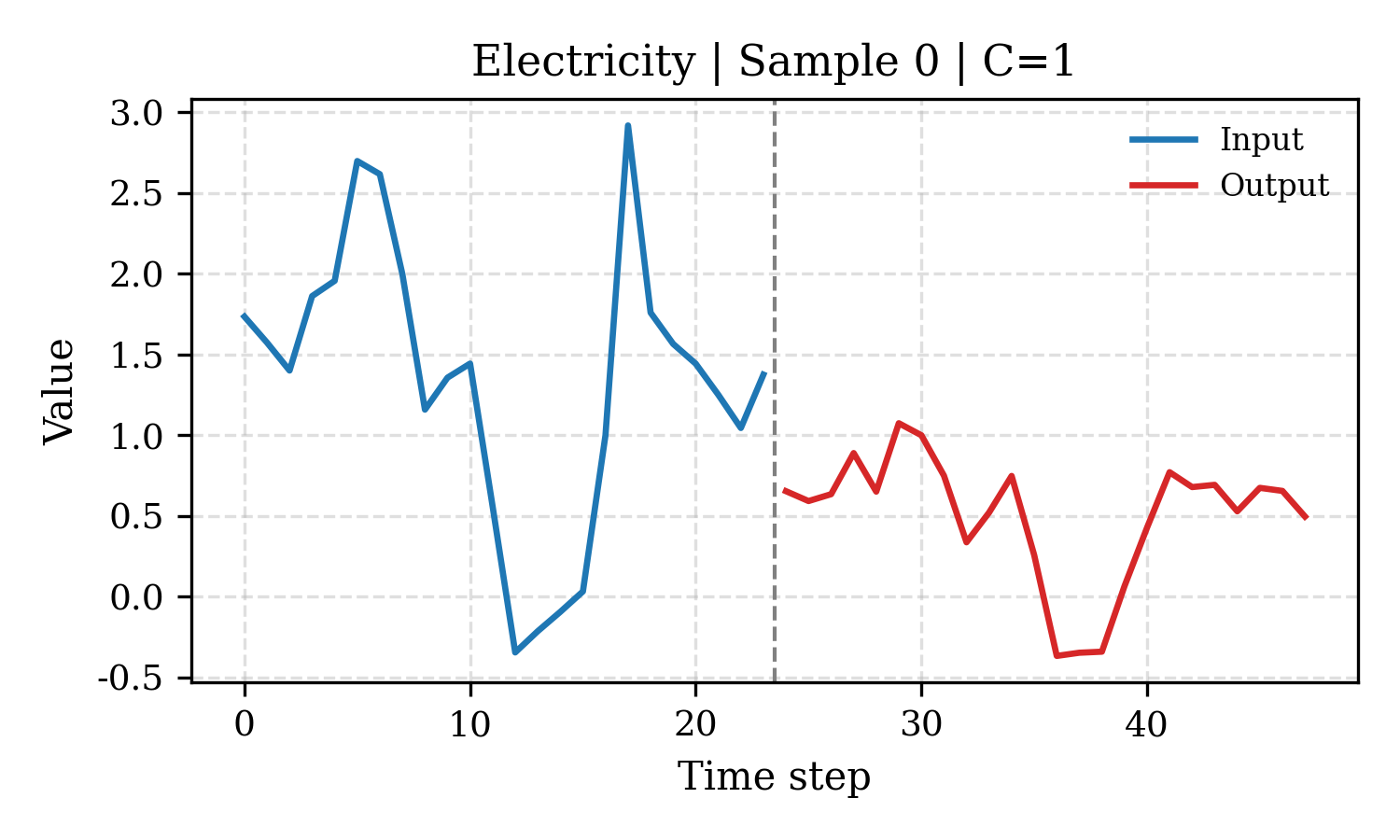} &
    \includegraphics[width=0.11\linewidth]{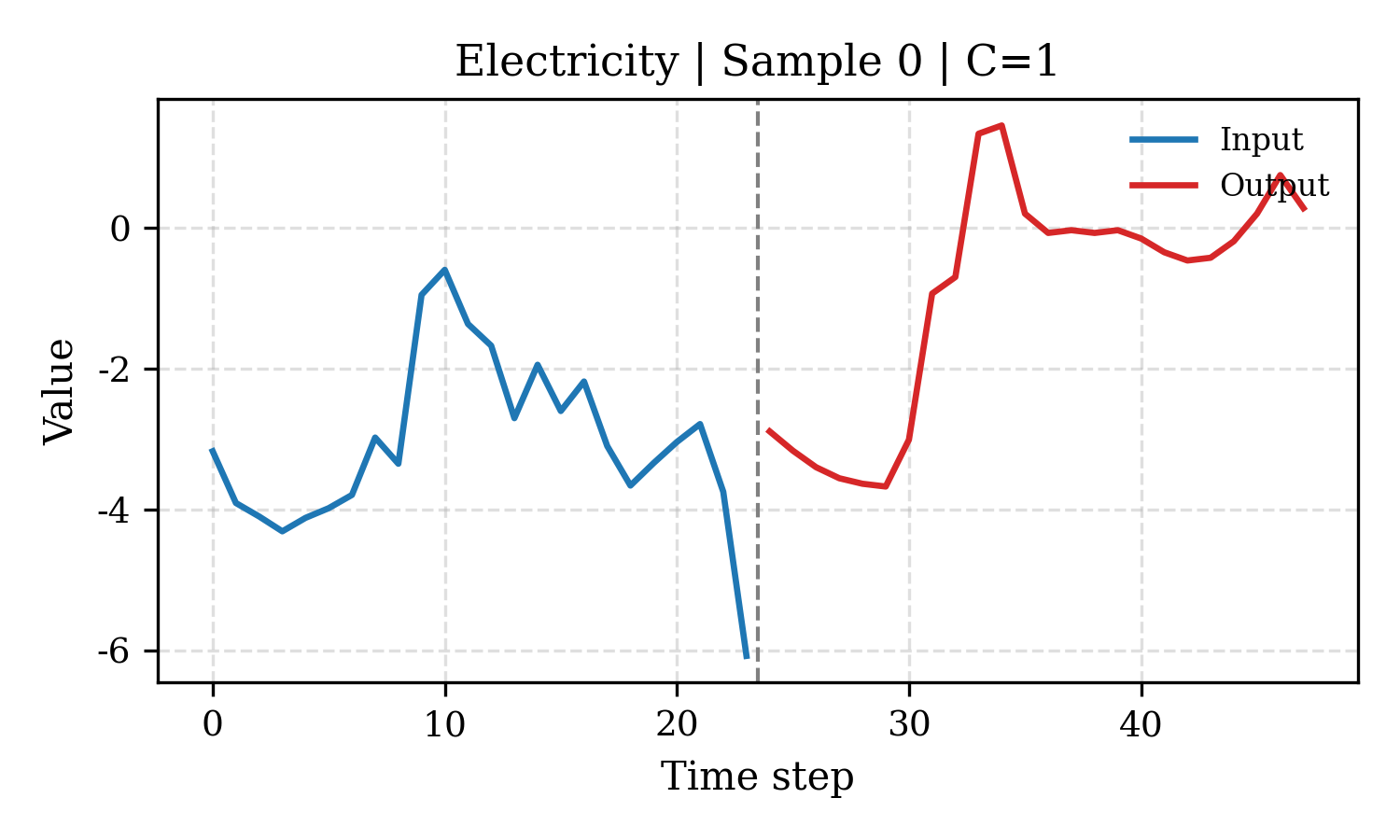} &
    \includegraphics[width=0.11\linewidth]{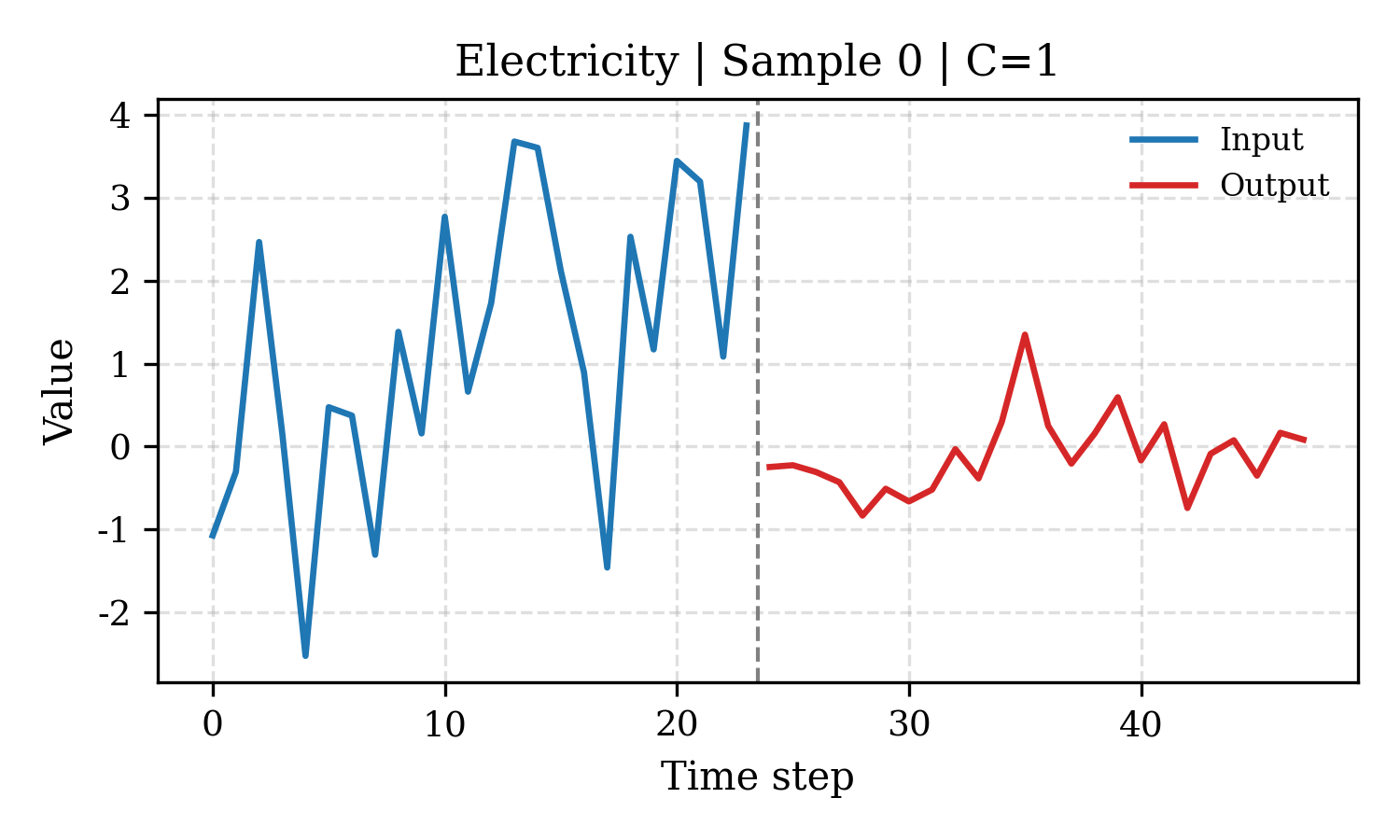} &
    \includegraphics[width=0.11\linewidth]{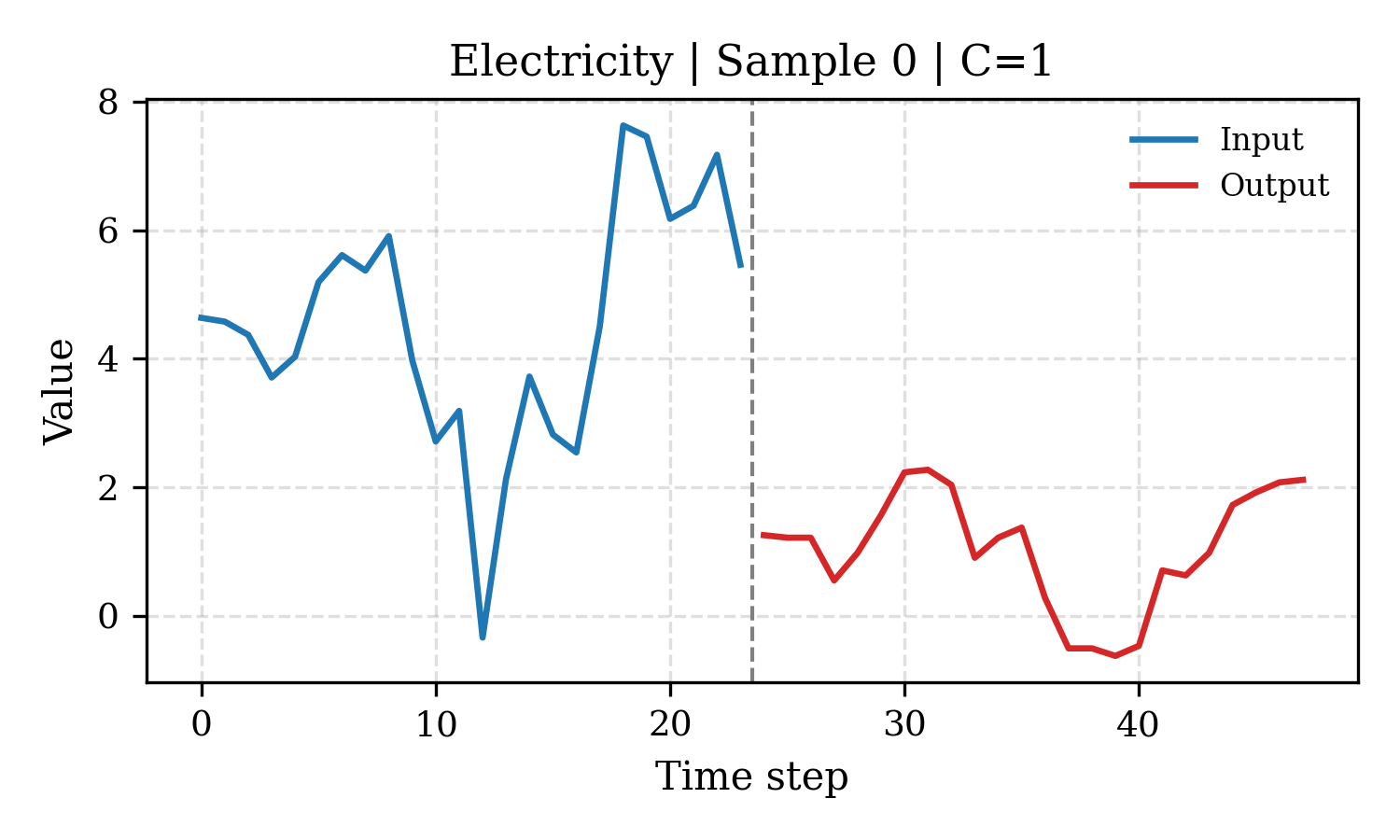} &
    \includegraphics[width=0.11\linewidth]{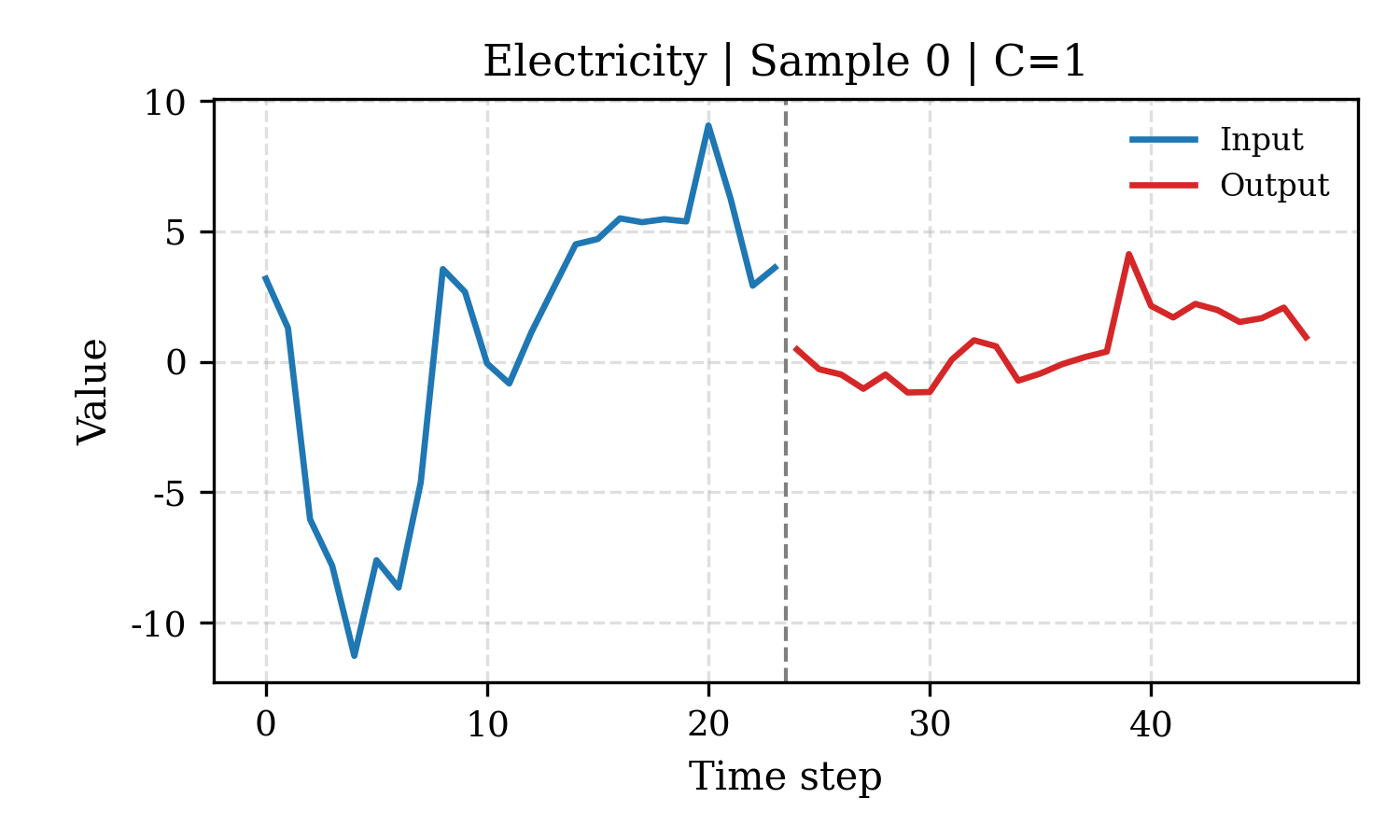} \\[0.3em]

    \rotatebox[origin=l]{90}{\textbf{Exchange}} &
    \includegraphics[width=0.11\linewidth]{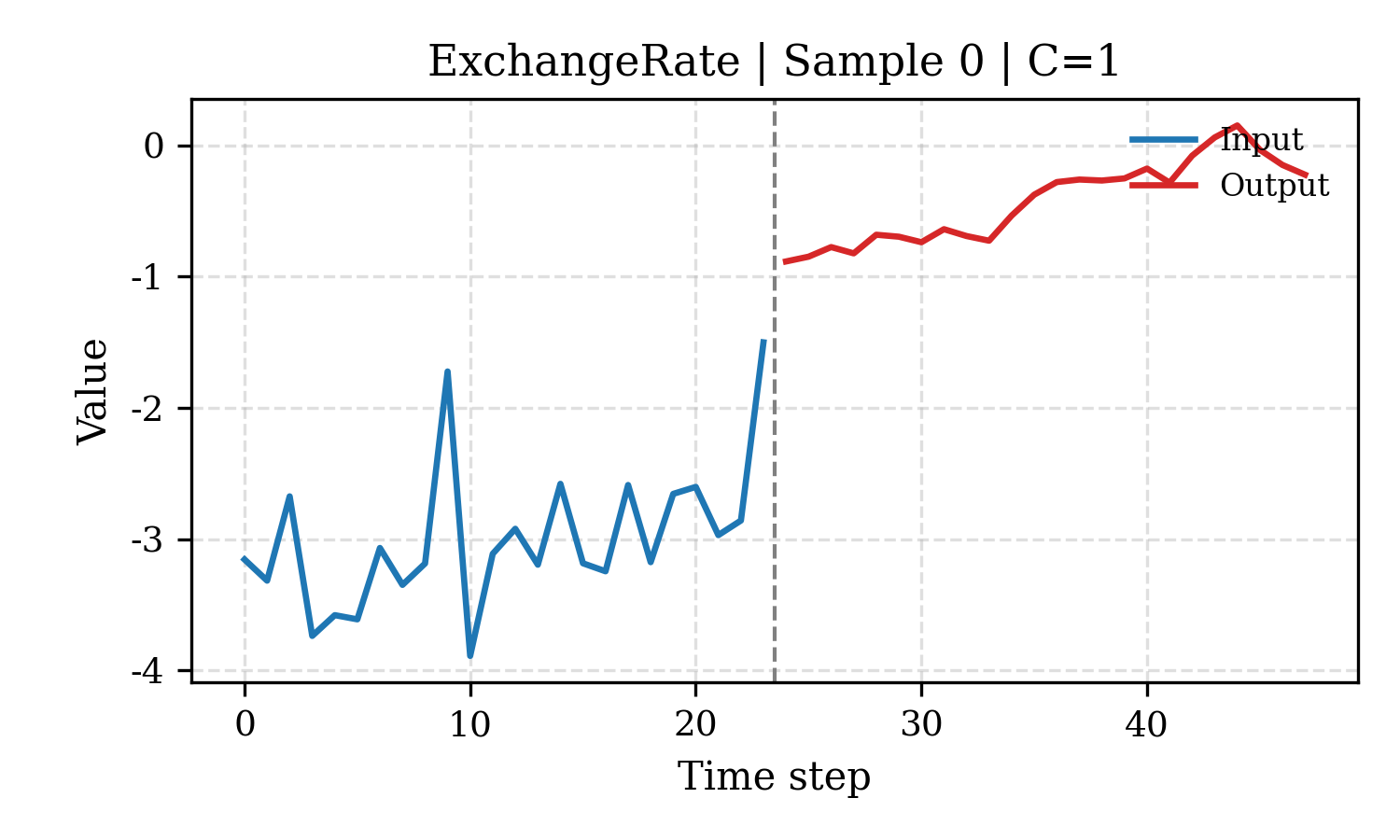} &
    \includegraphics[width=0.11\linewidth]{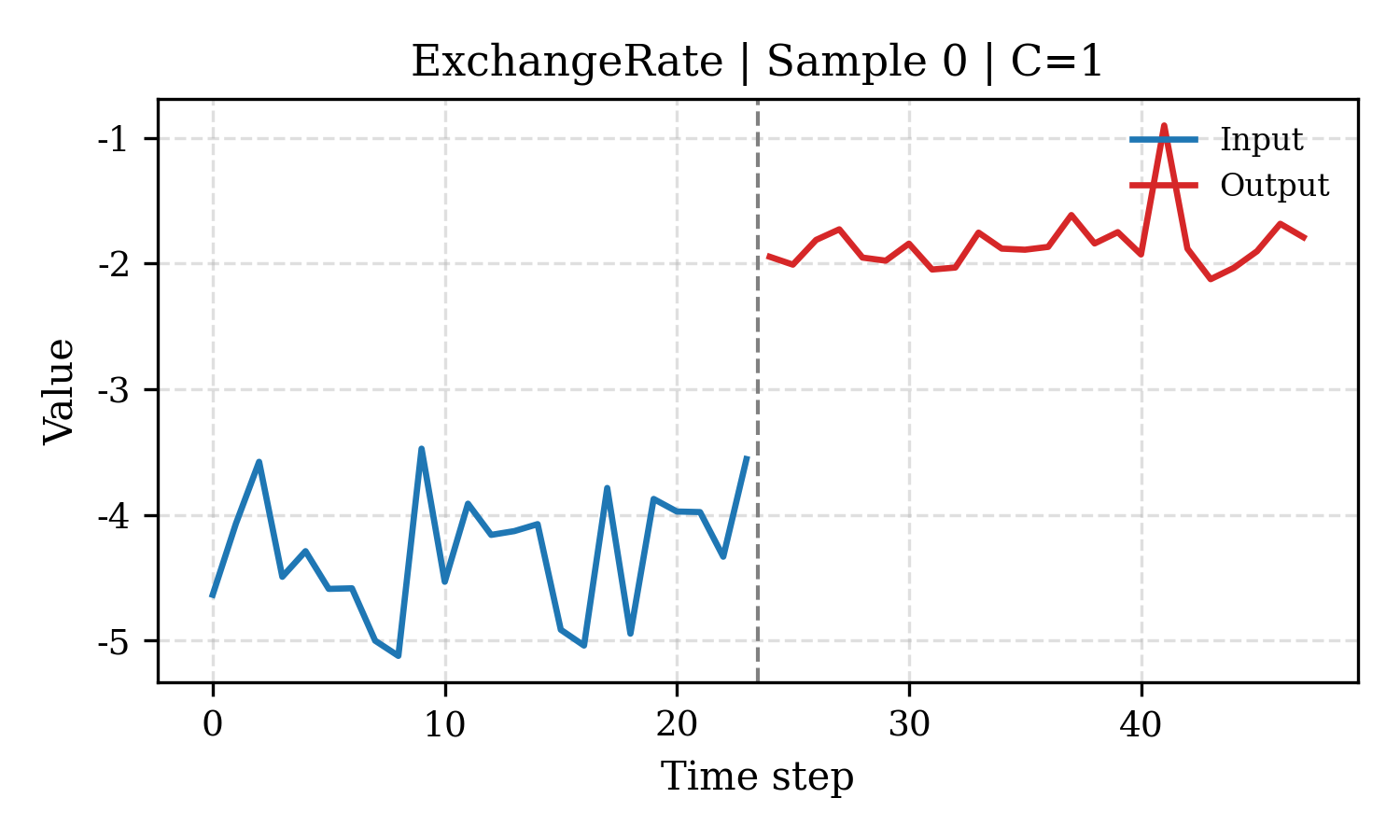} &
    \includegraphics[width=0.11\linewidth]{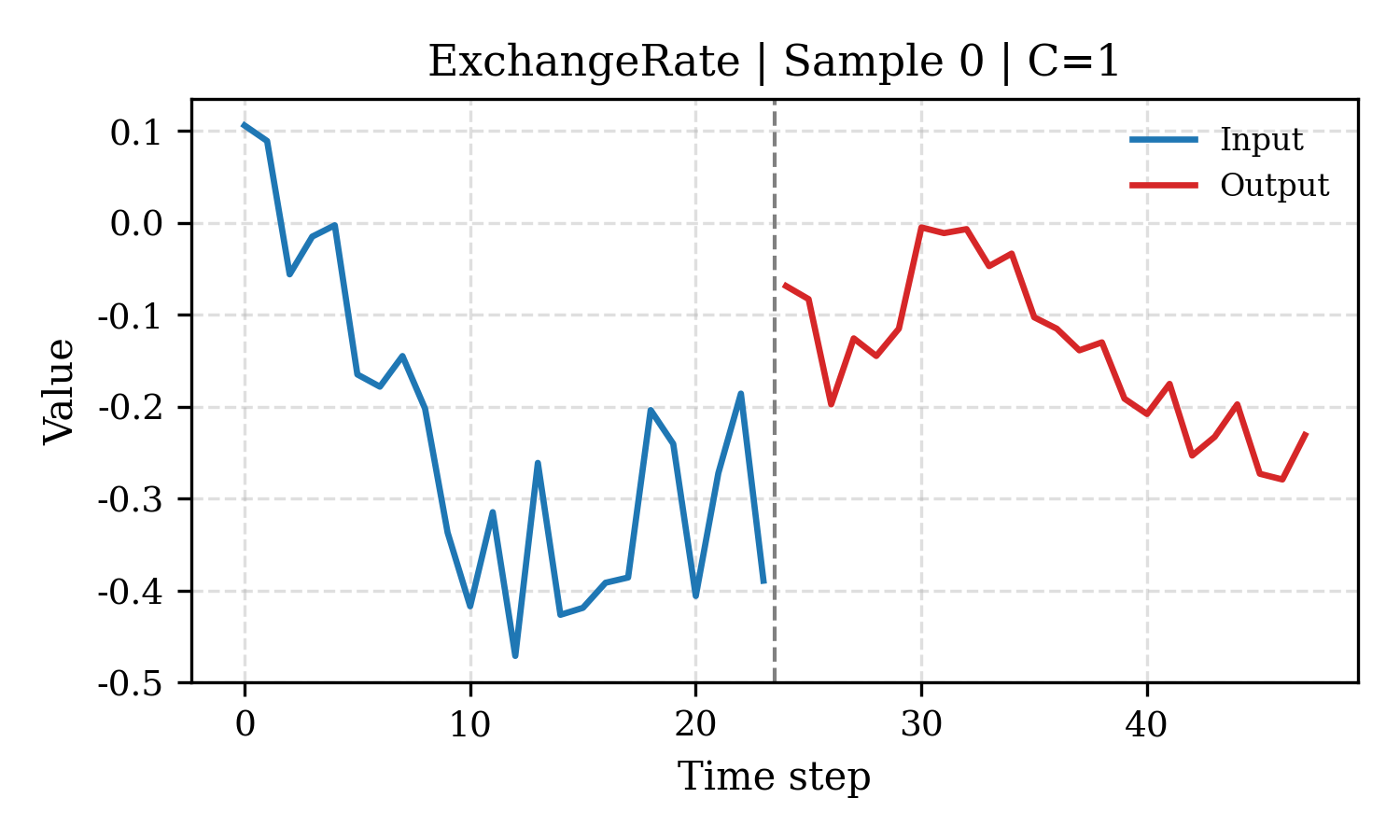} &
    \includegraphics[width=0.11\linewidth]{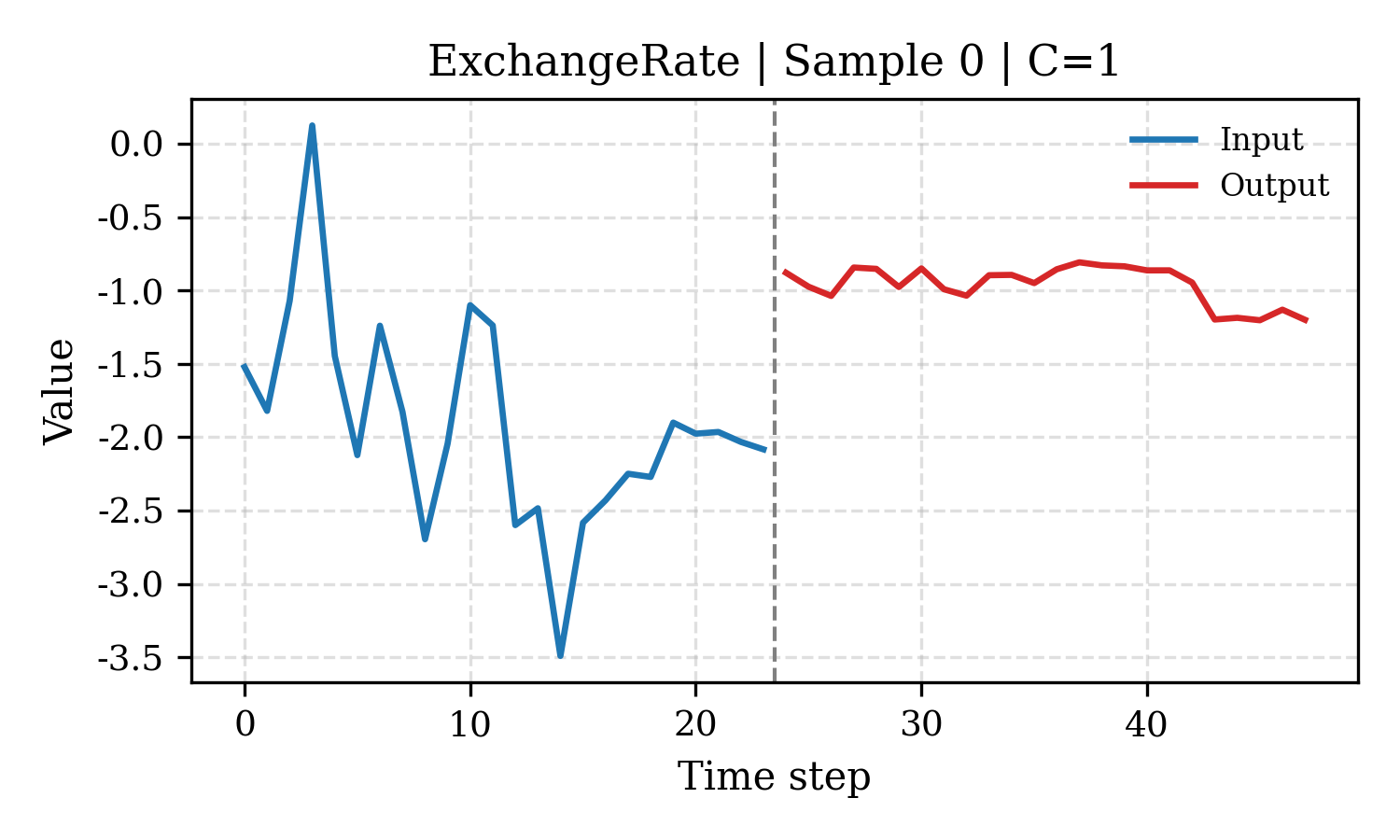} &
    \includegraphics[width=0.11\linewidth]{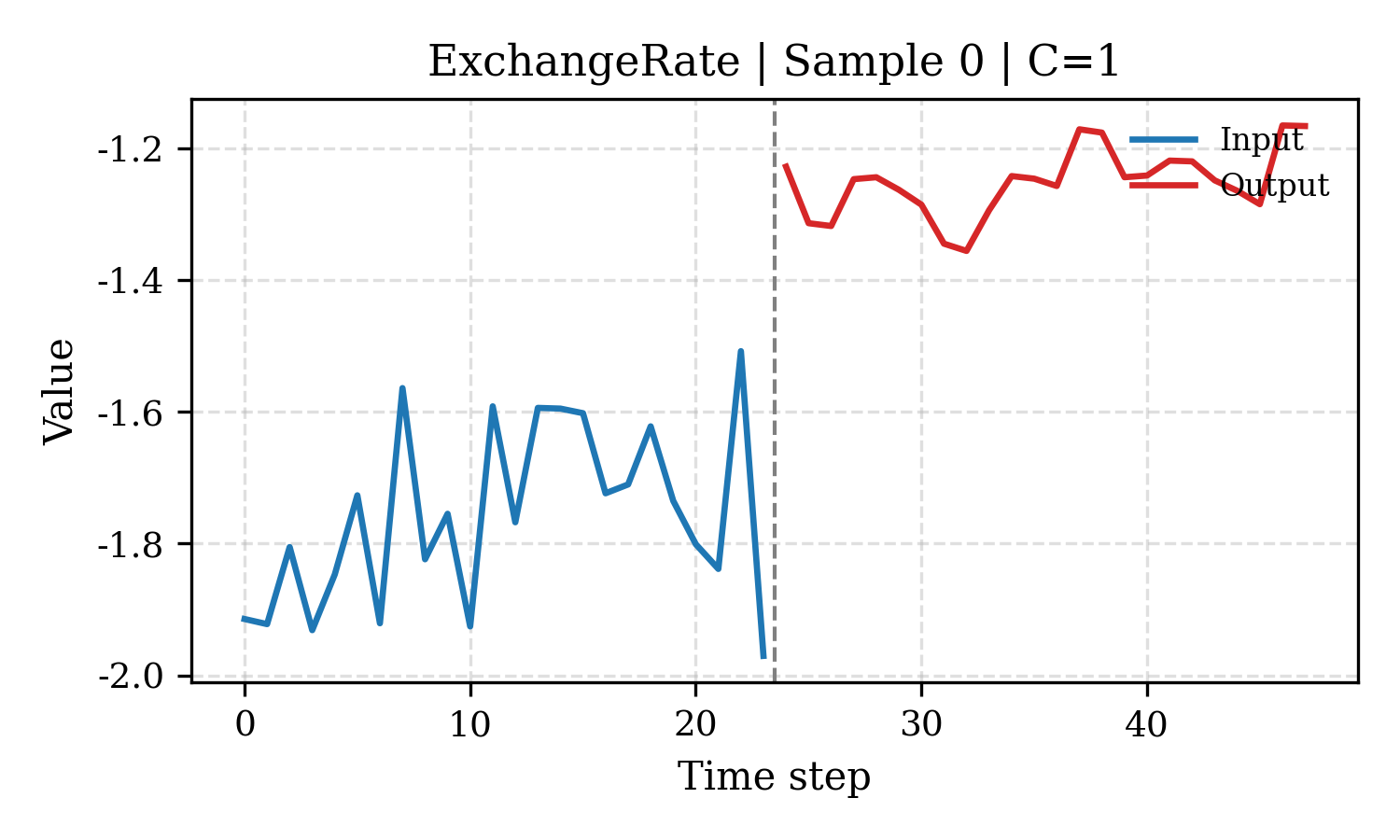} &
    \includegraphics[width=0.11\linewidth]{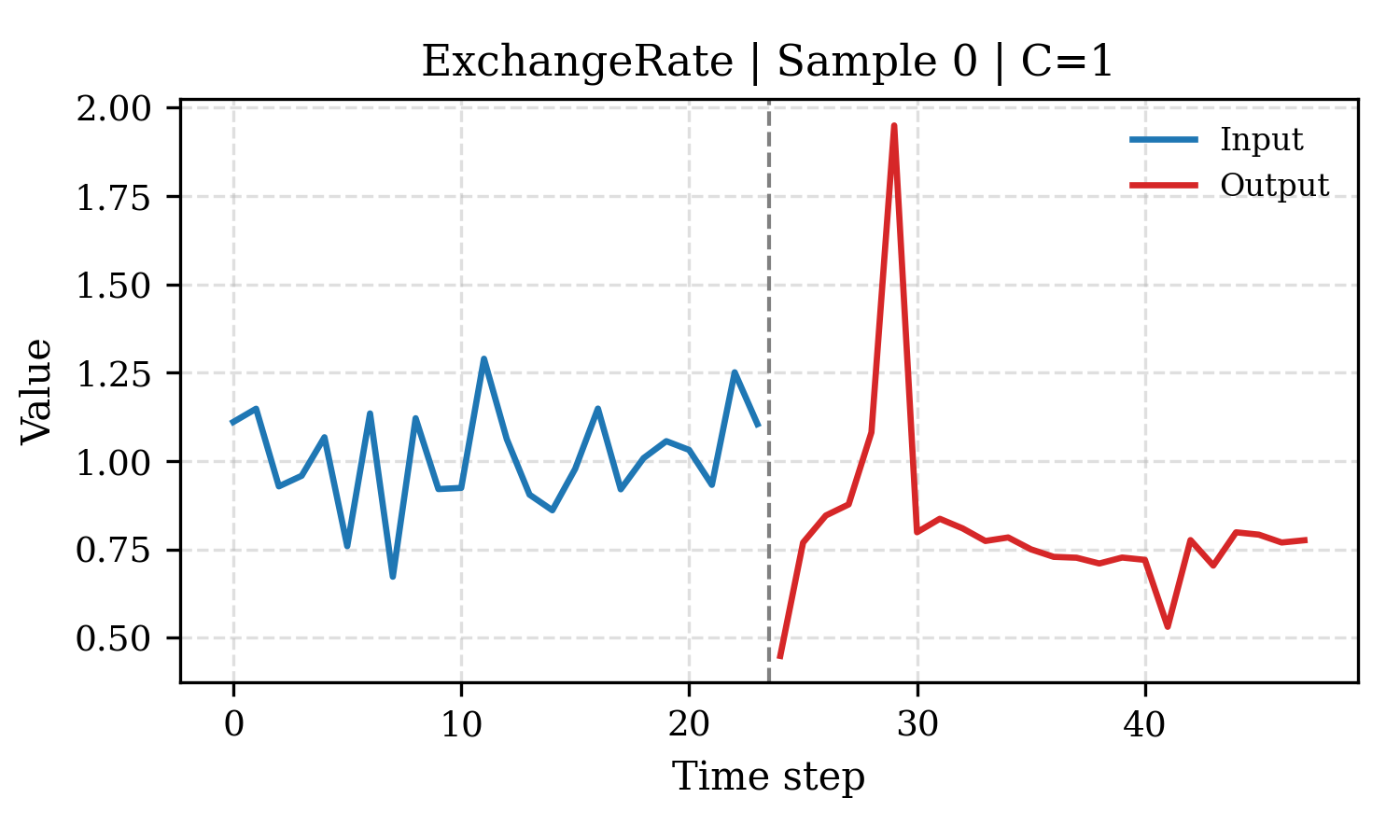} &
    \includegraphics[width=0.11\linewidth]{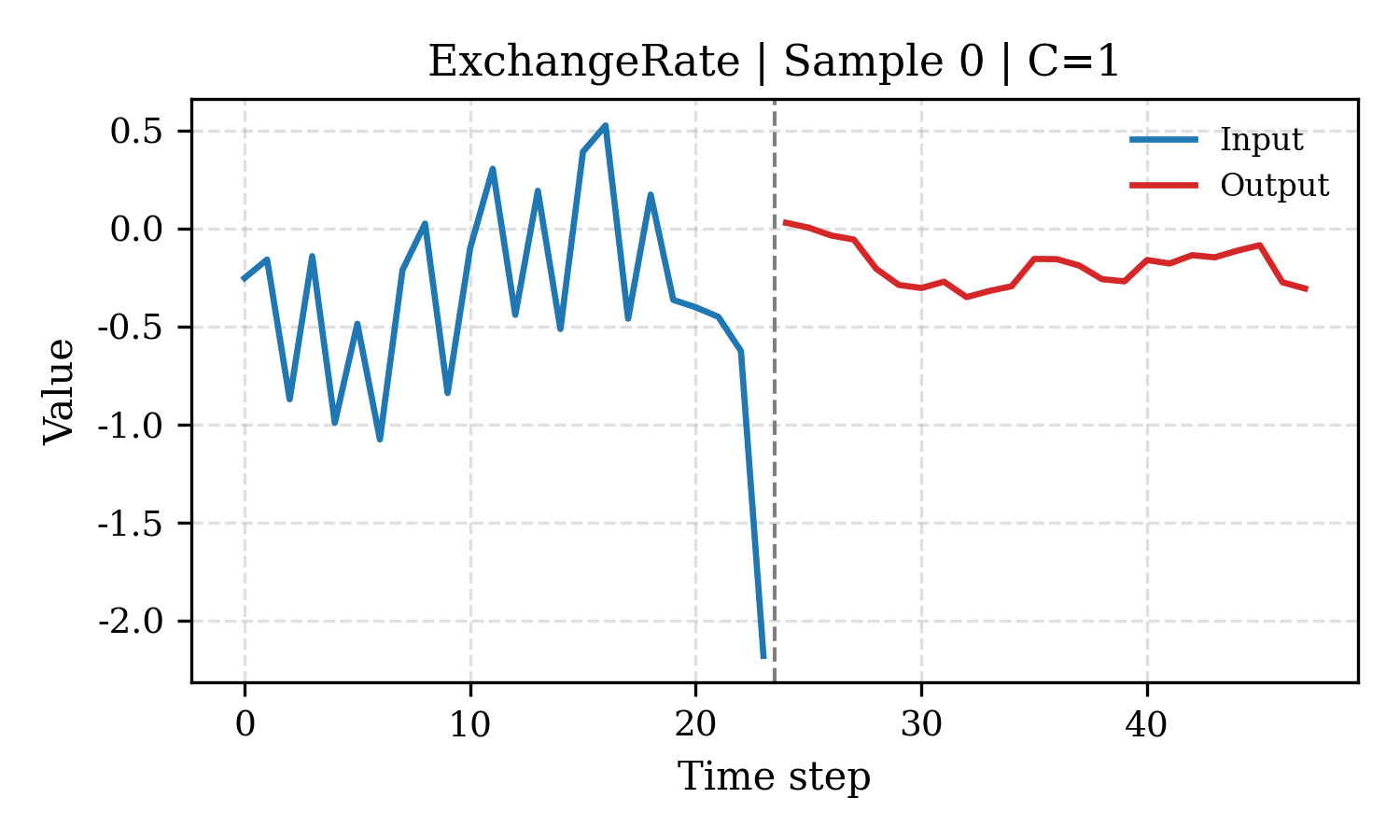} &
    \includegraphics[width=0.11\linewidth]{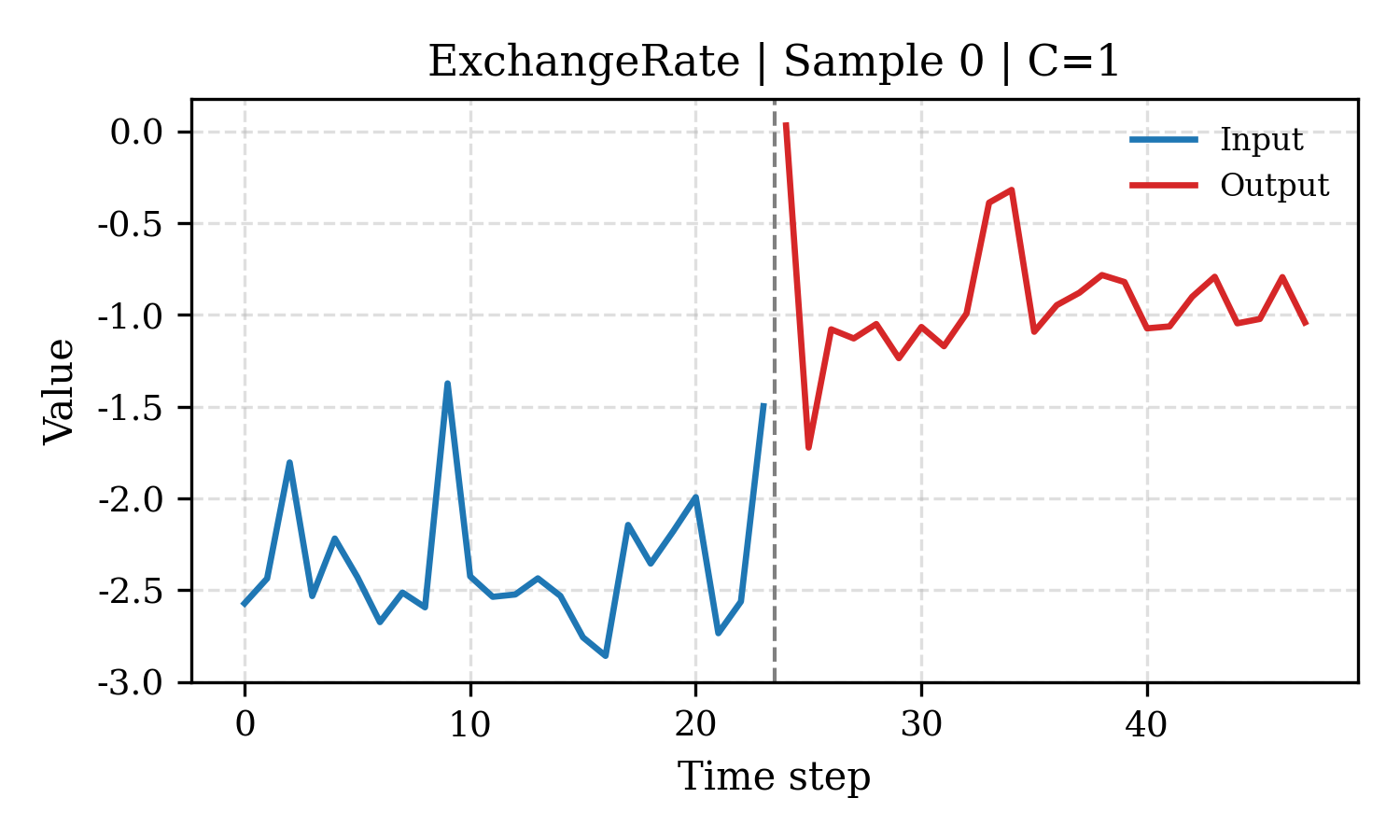} \\[0.3em]

    \rotatebox[origin=l]{90}{\textbf{Solar}} &
    \includegraphics[width=0.11\linewidth]{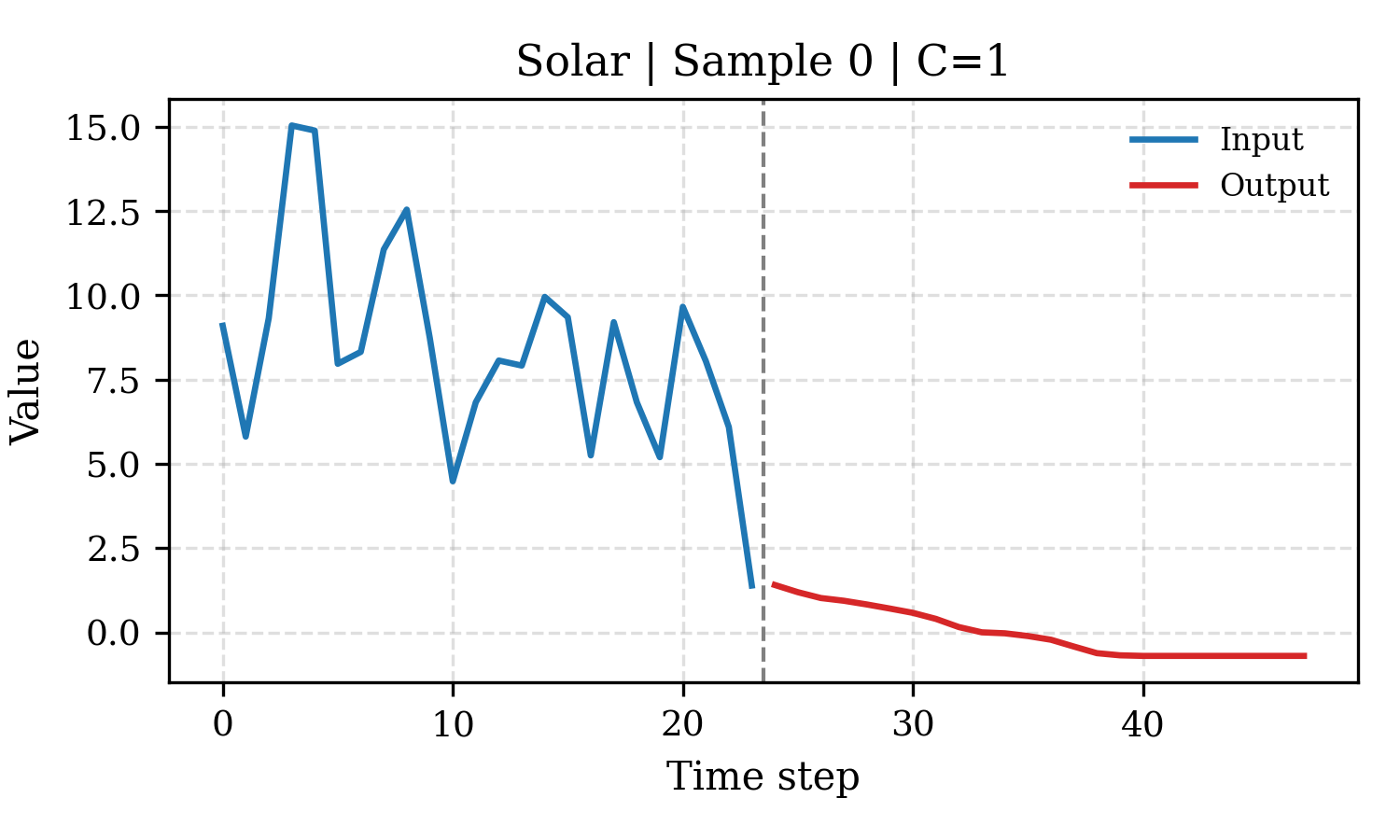} &
    \includegraphics[width=0.11\linewidth]{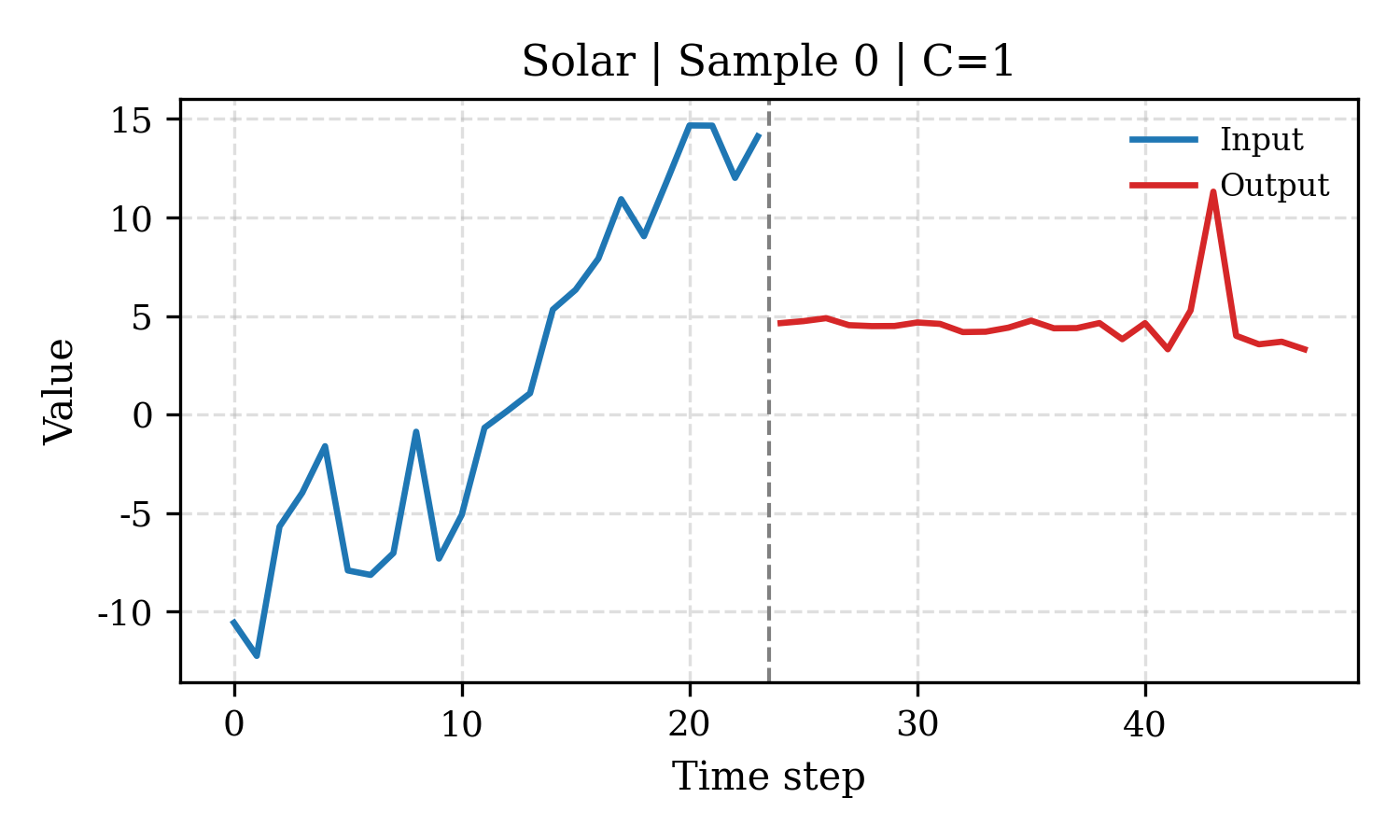} &
    \includegraphics[width=0.11\linewidth]{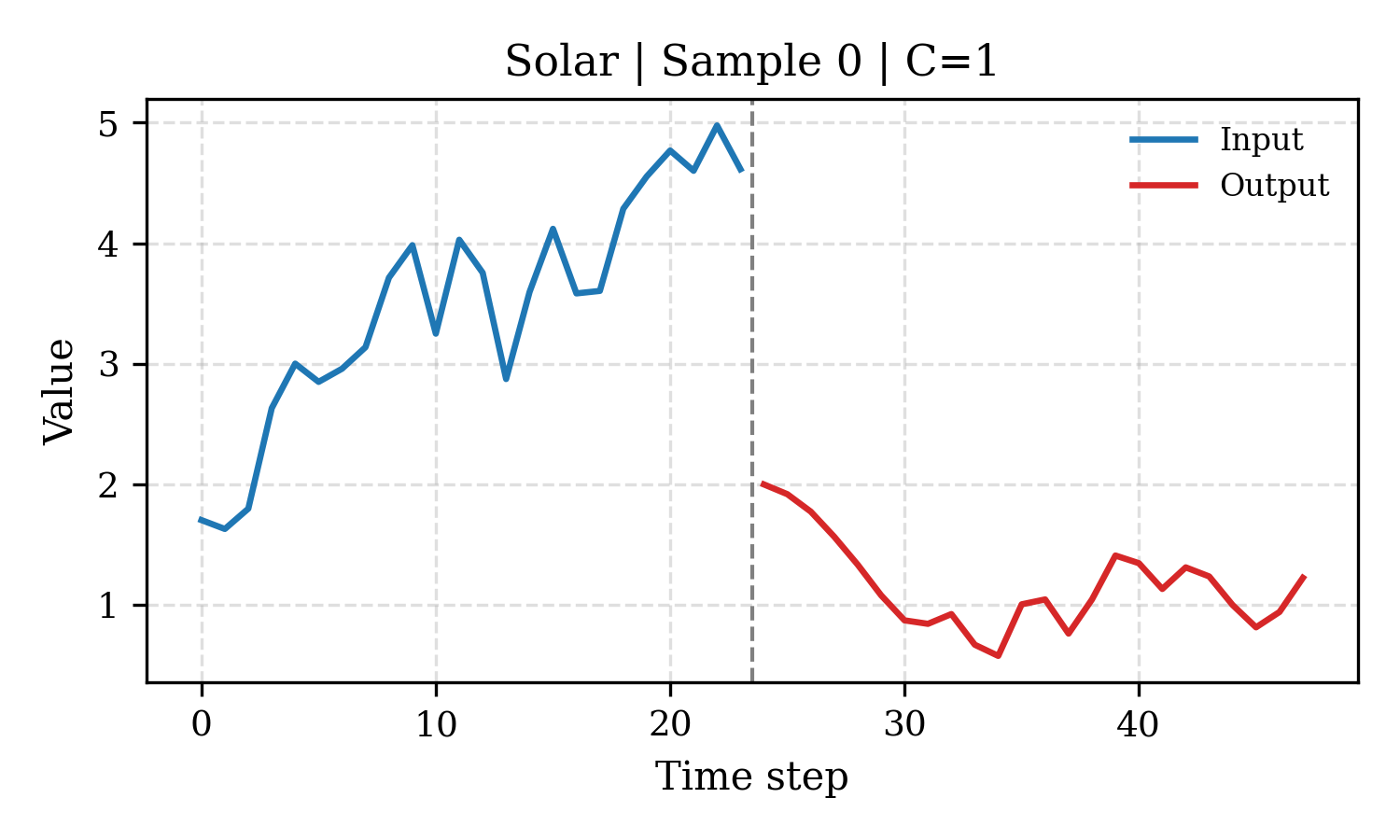} &
    \includegraphics[width=0.11\linewidth]{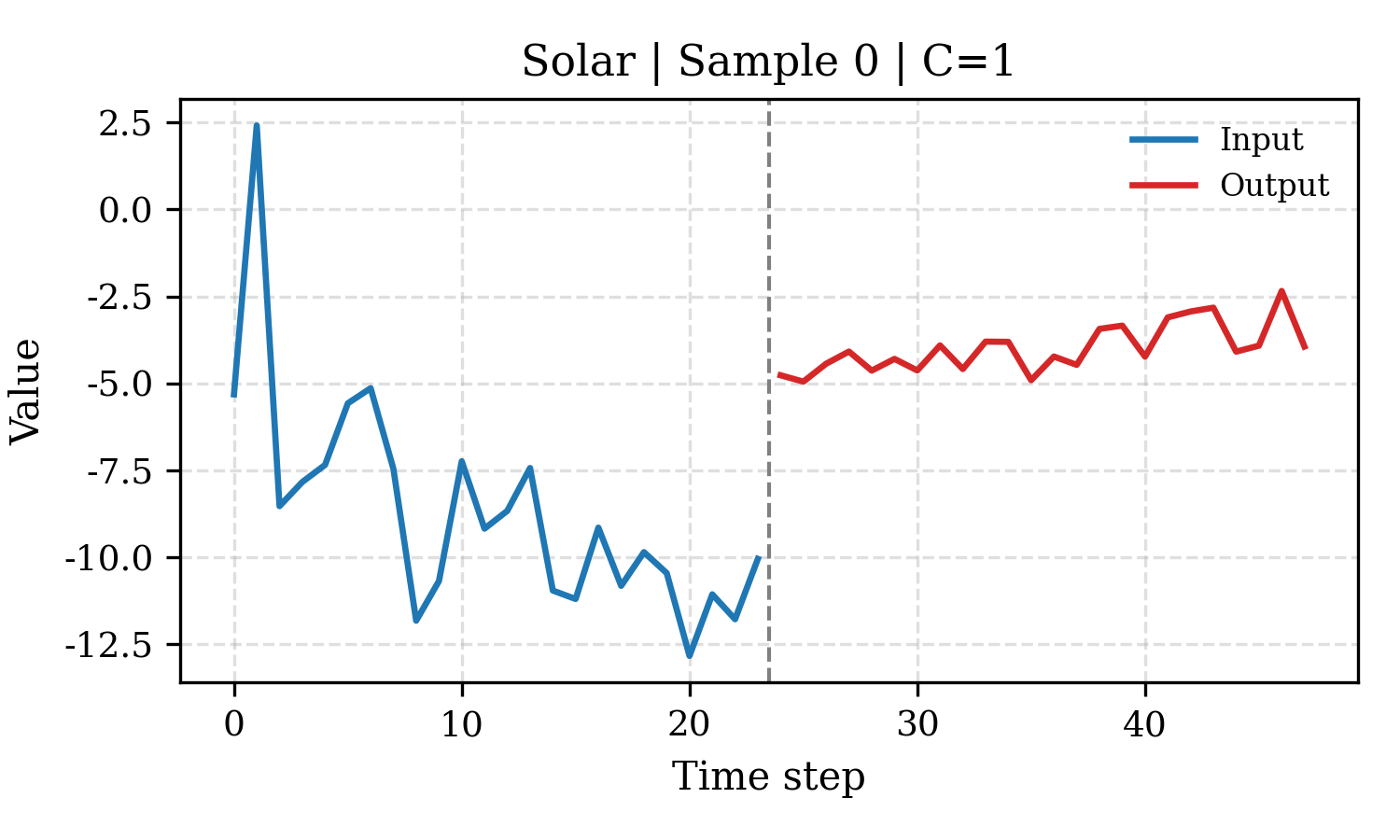} &
    \includegraphics[width=0.11\linewidth]{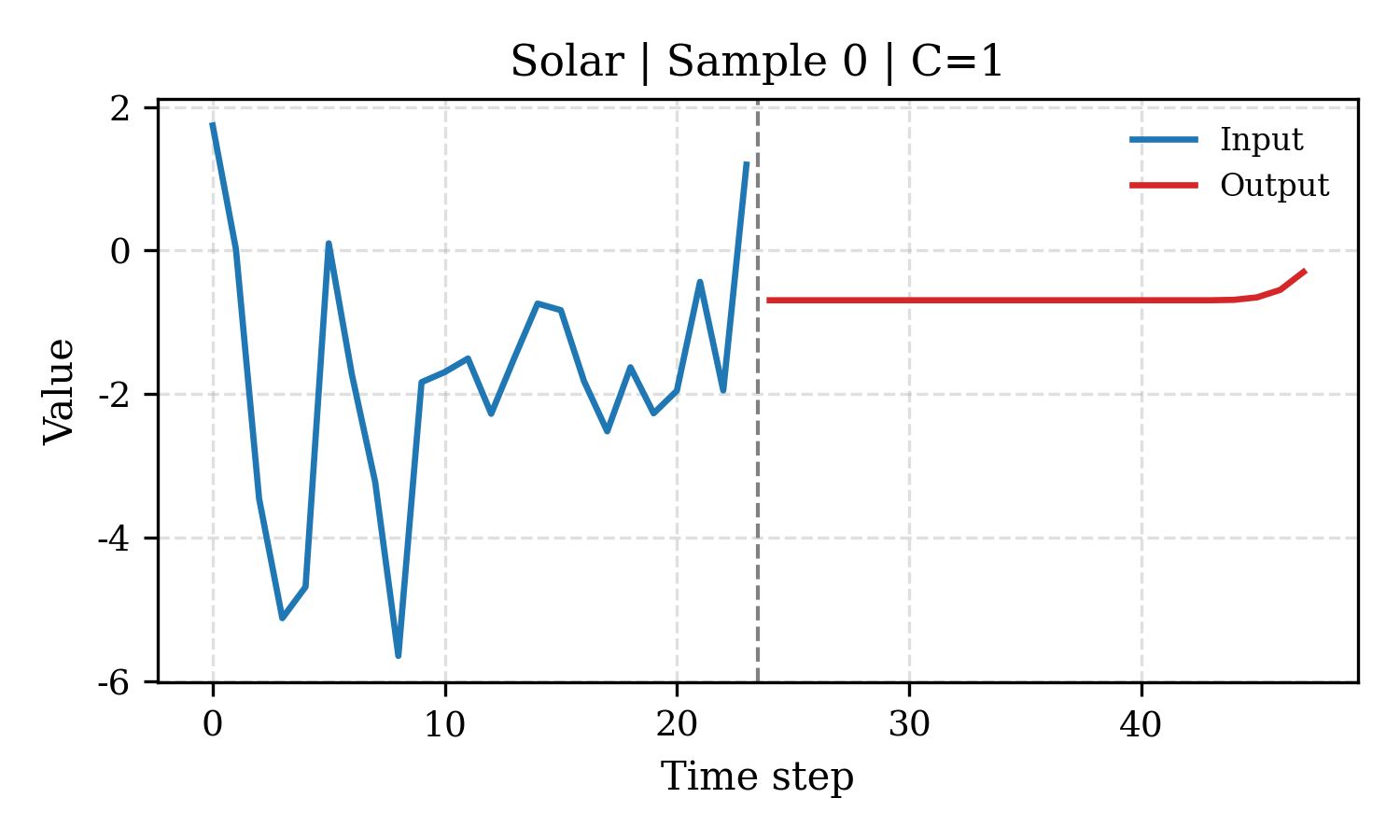} &
    \includegraphics[width=0.11\linewidth]{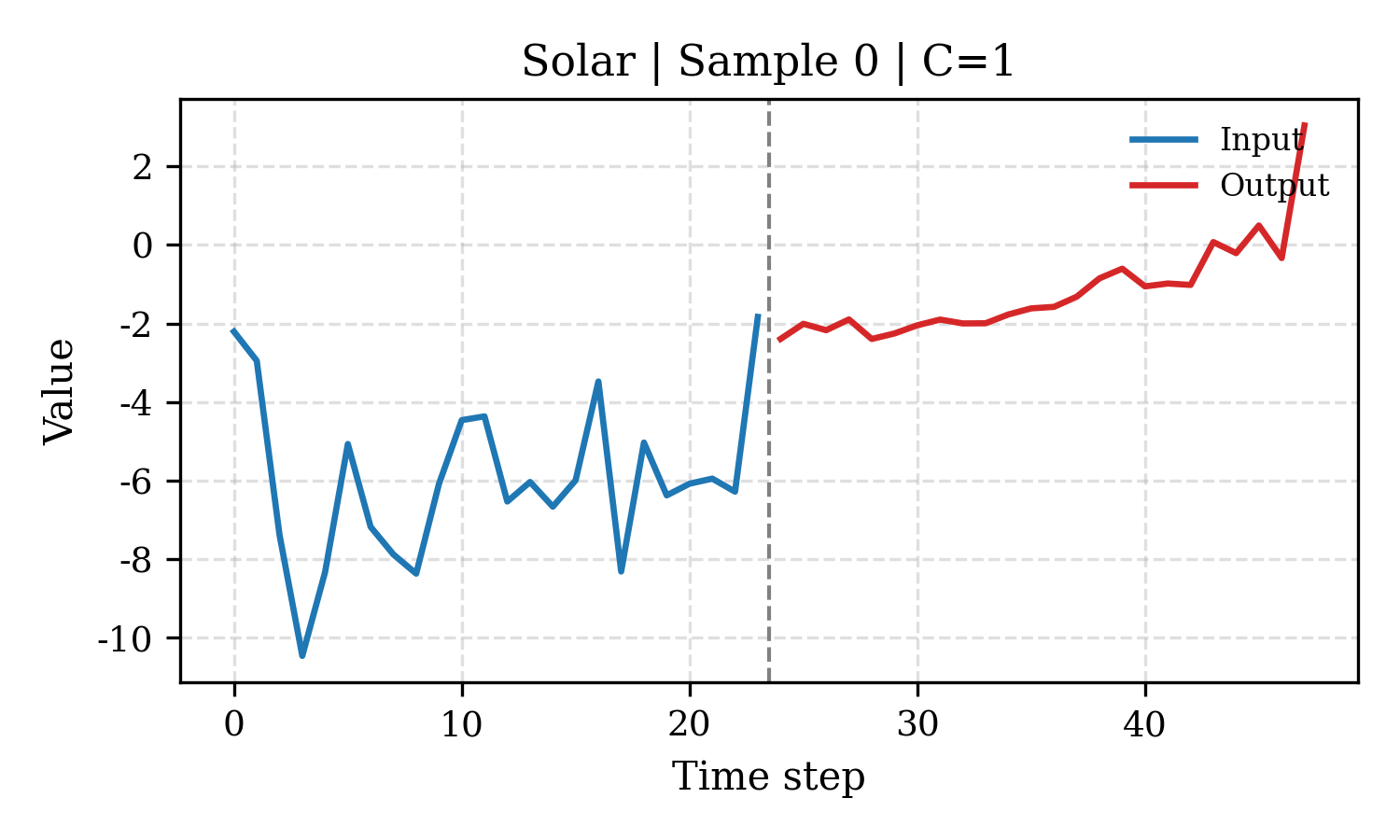} &
    \includegraphics[width=0.11\linewidth]{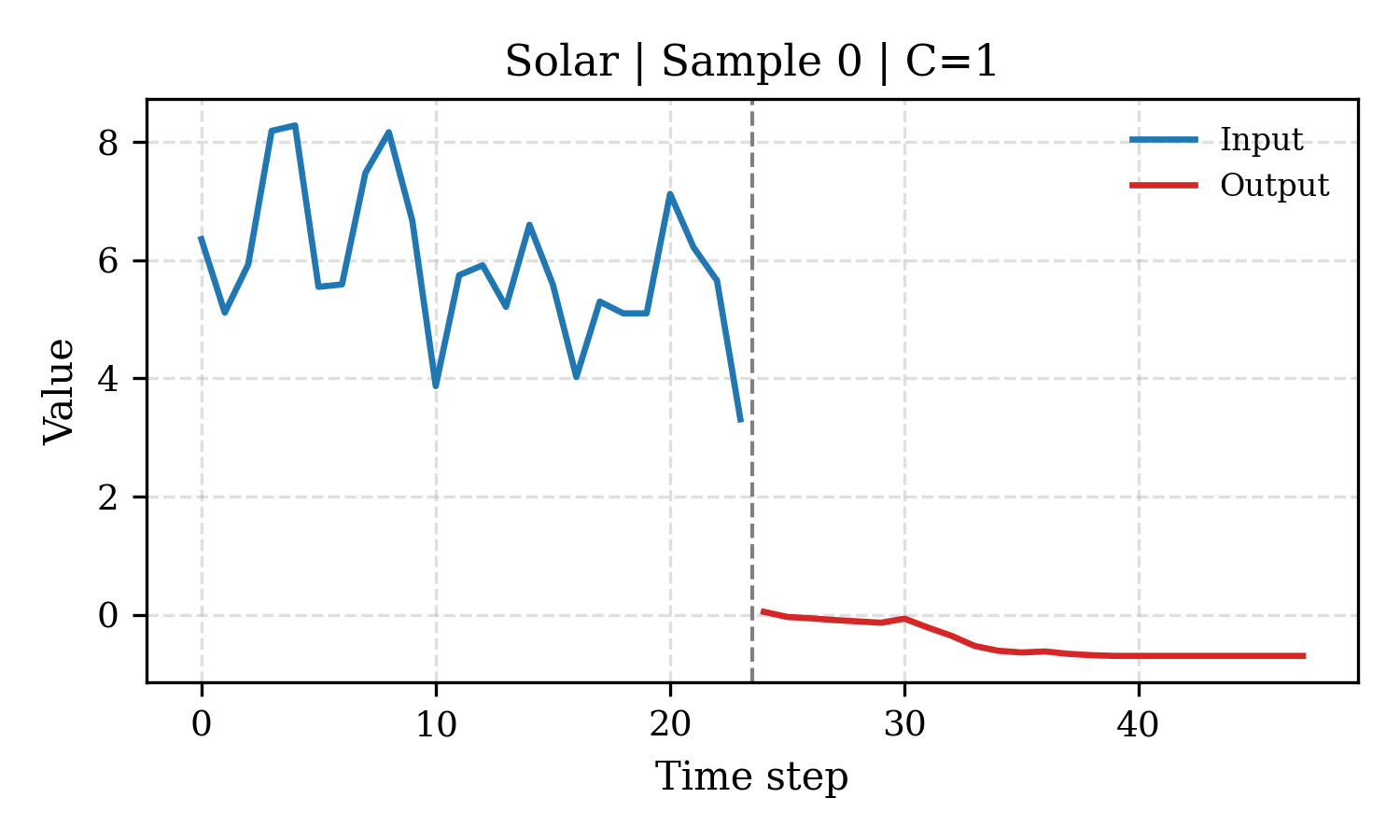} &
    \includegraphics[width=0.11\linewidth]{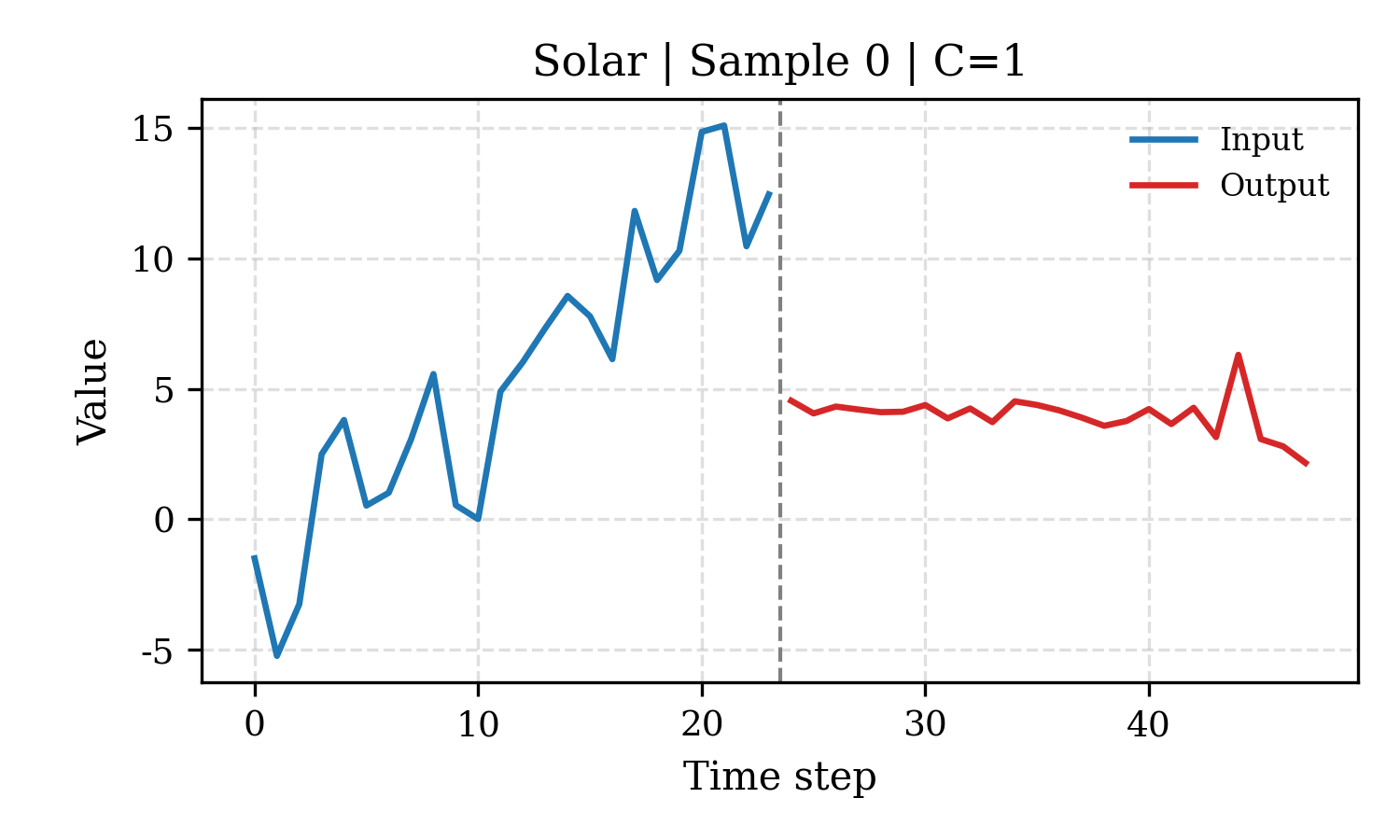} \\[0.3em]

    \rotatebox[origin=l]{90}{\textbf{Weather}} &
    \includegraphics[width=0.11\linewidth]{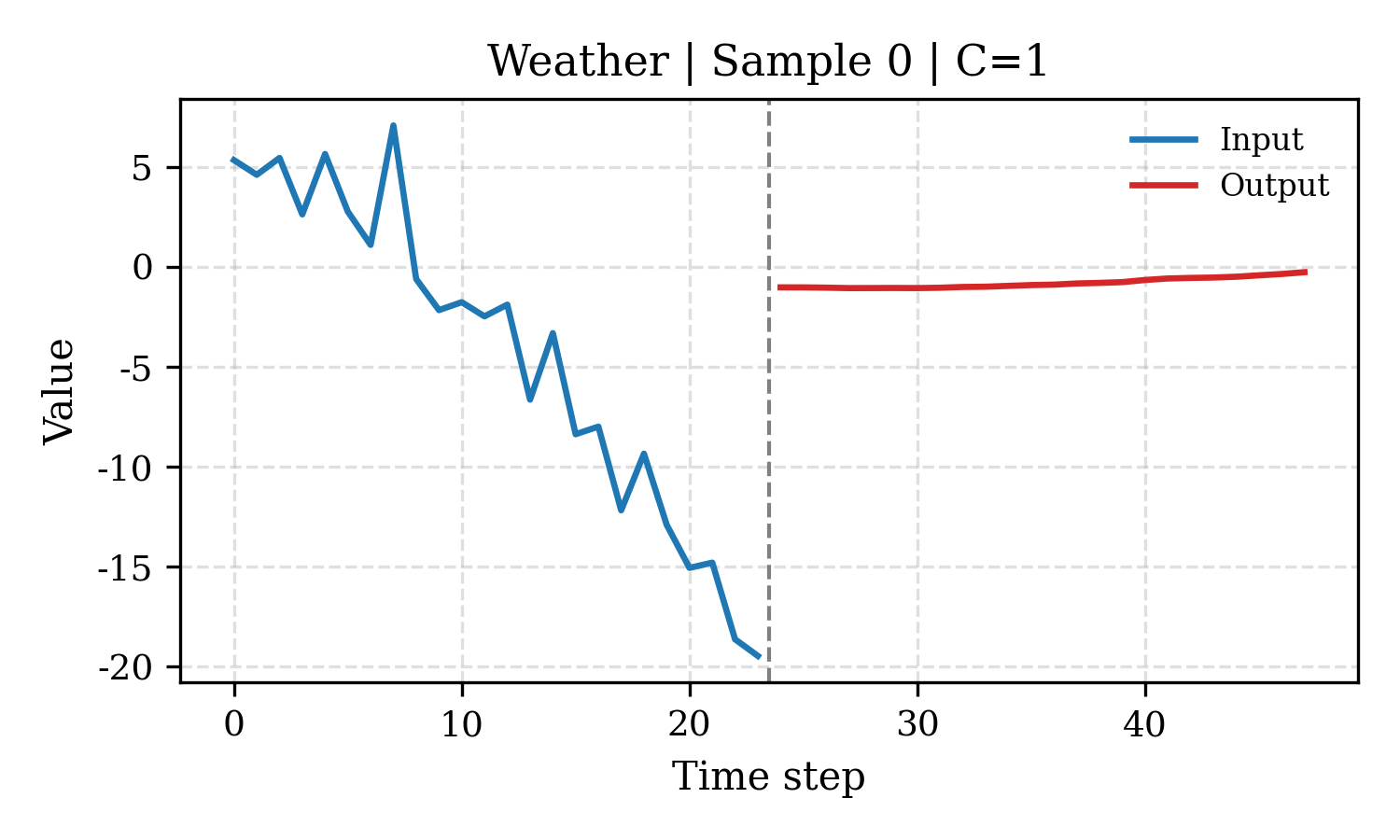} &
    \includegraphics[width=0.11\linewidth]{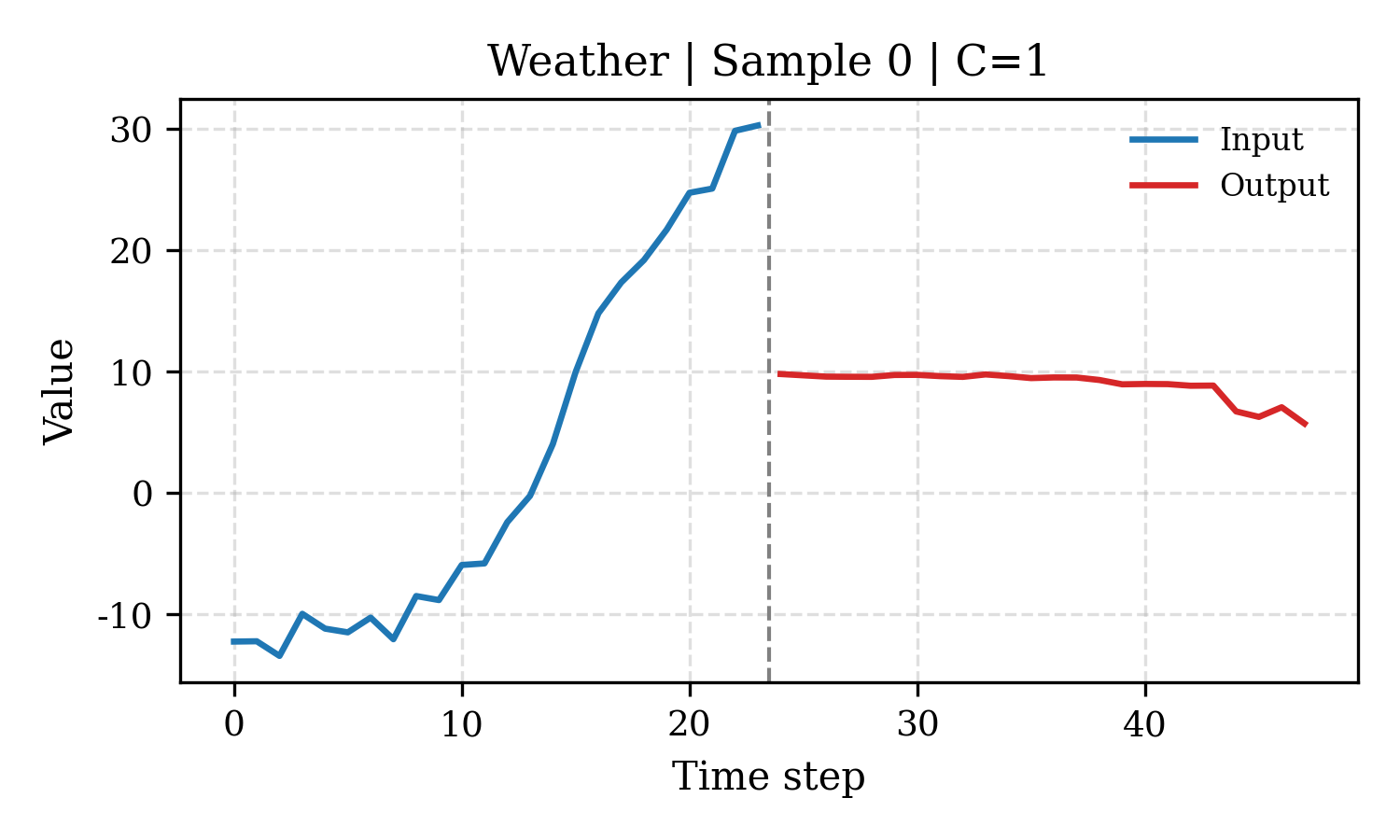} &
    \includegraphics[width=0.11\linewidth]{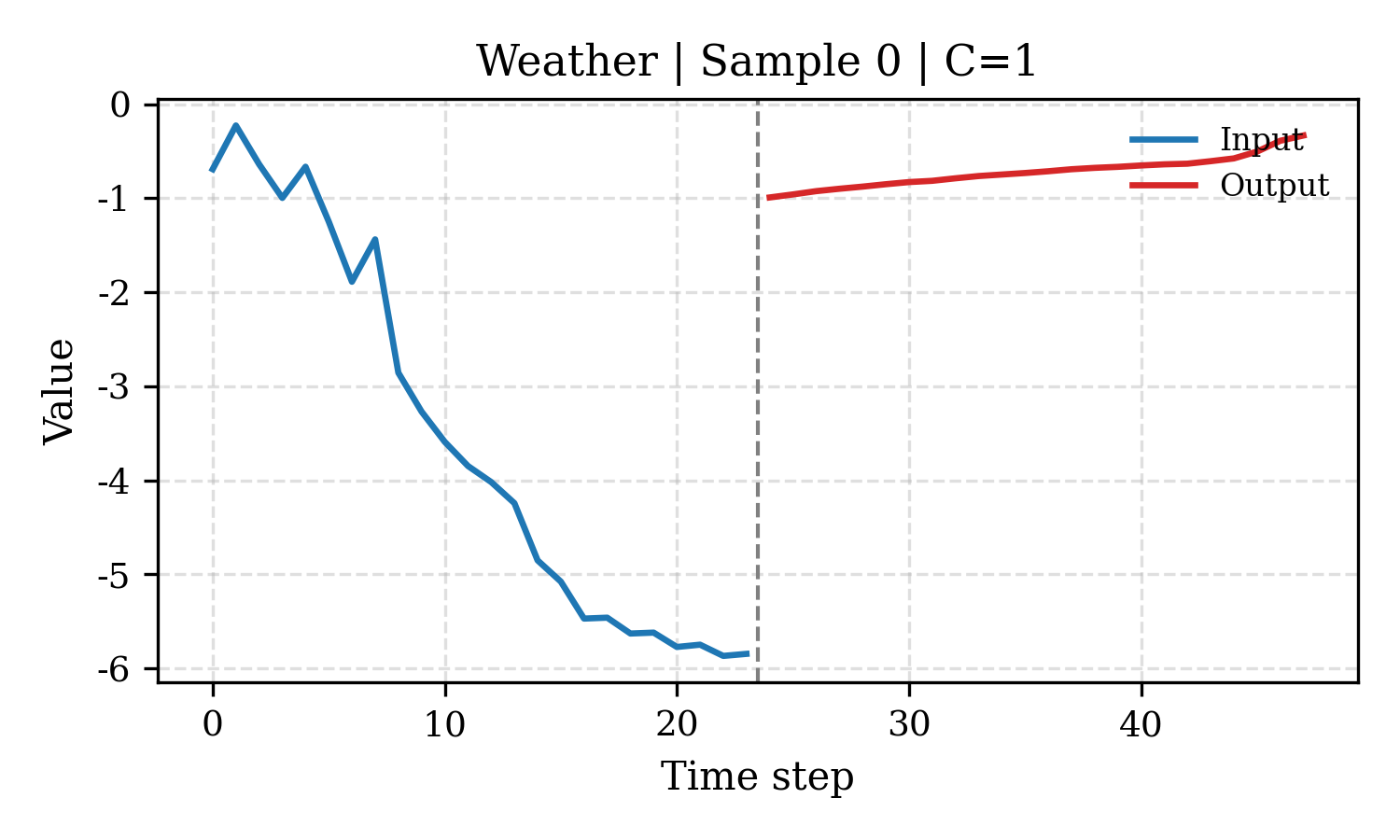} &
    \includegraphics[width=0.11\linewidth]{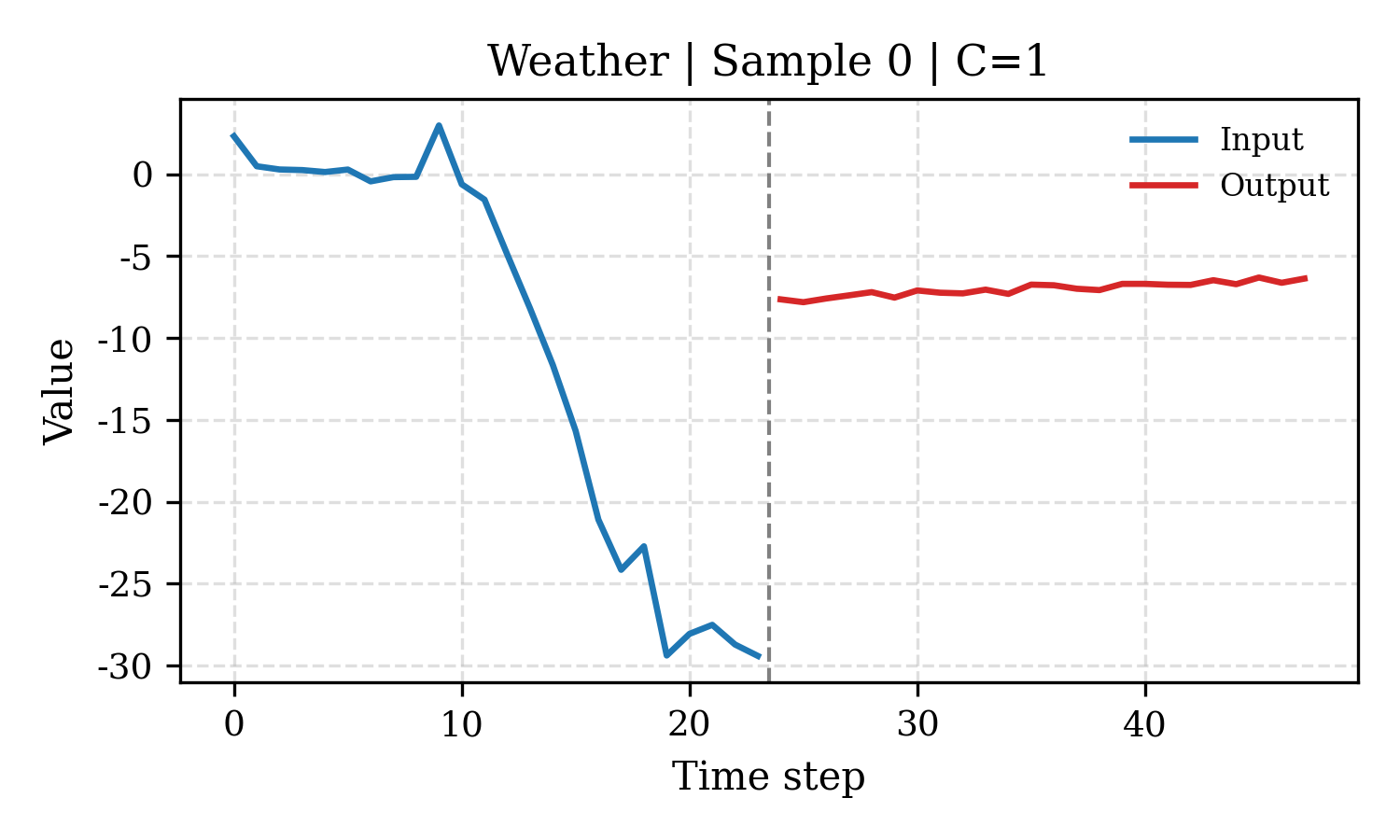} &
    \includegraphics[width=0.11\linewidth]{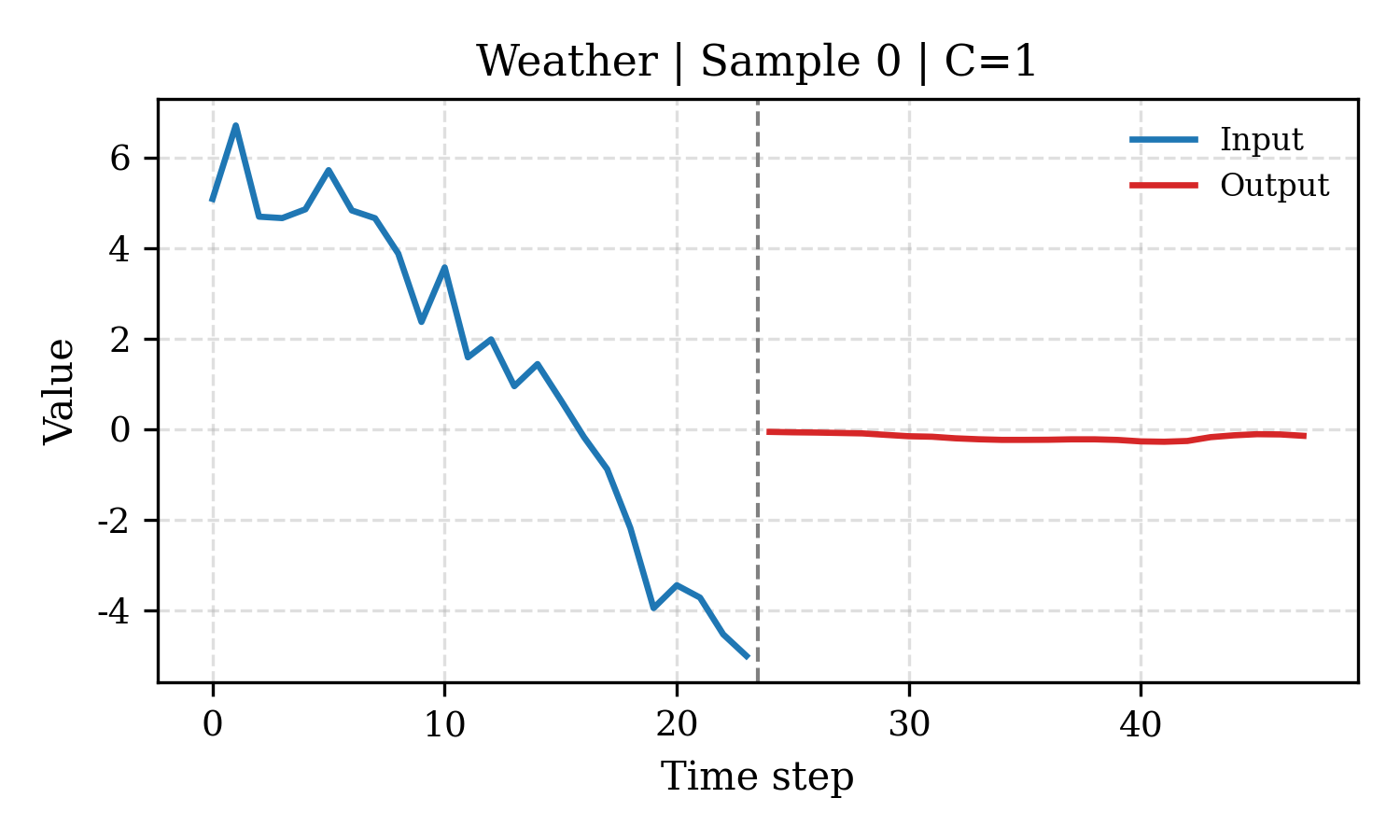} &
    \includegraphics[width=0.11\linewidth]{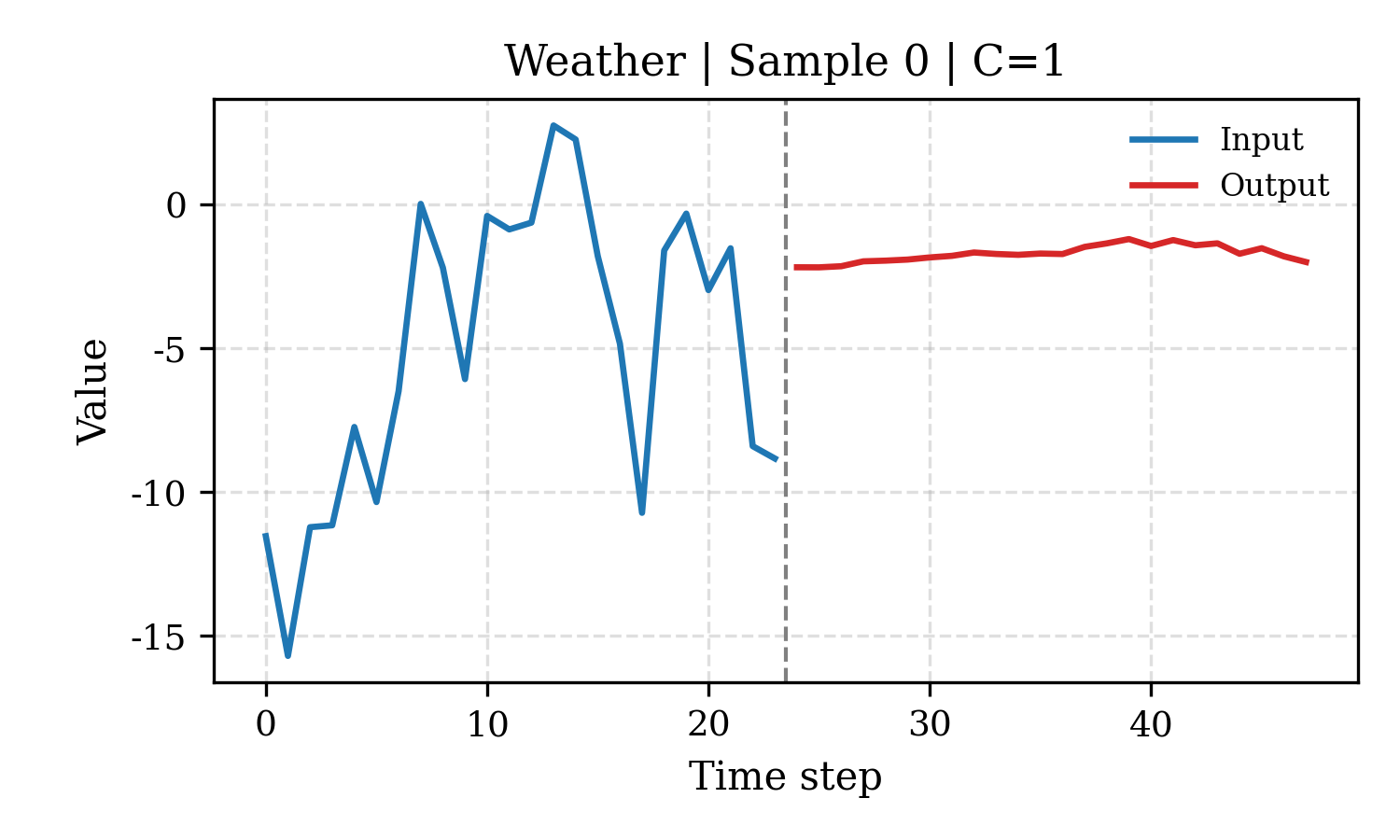} &
    \includegraphics[width=0.11\linewidth]{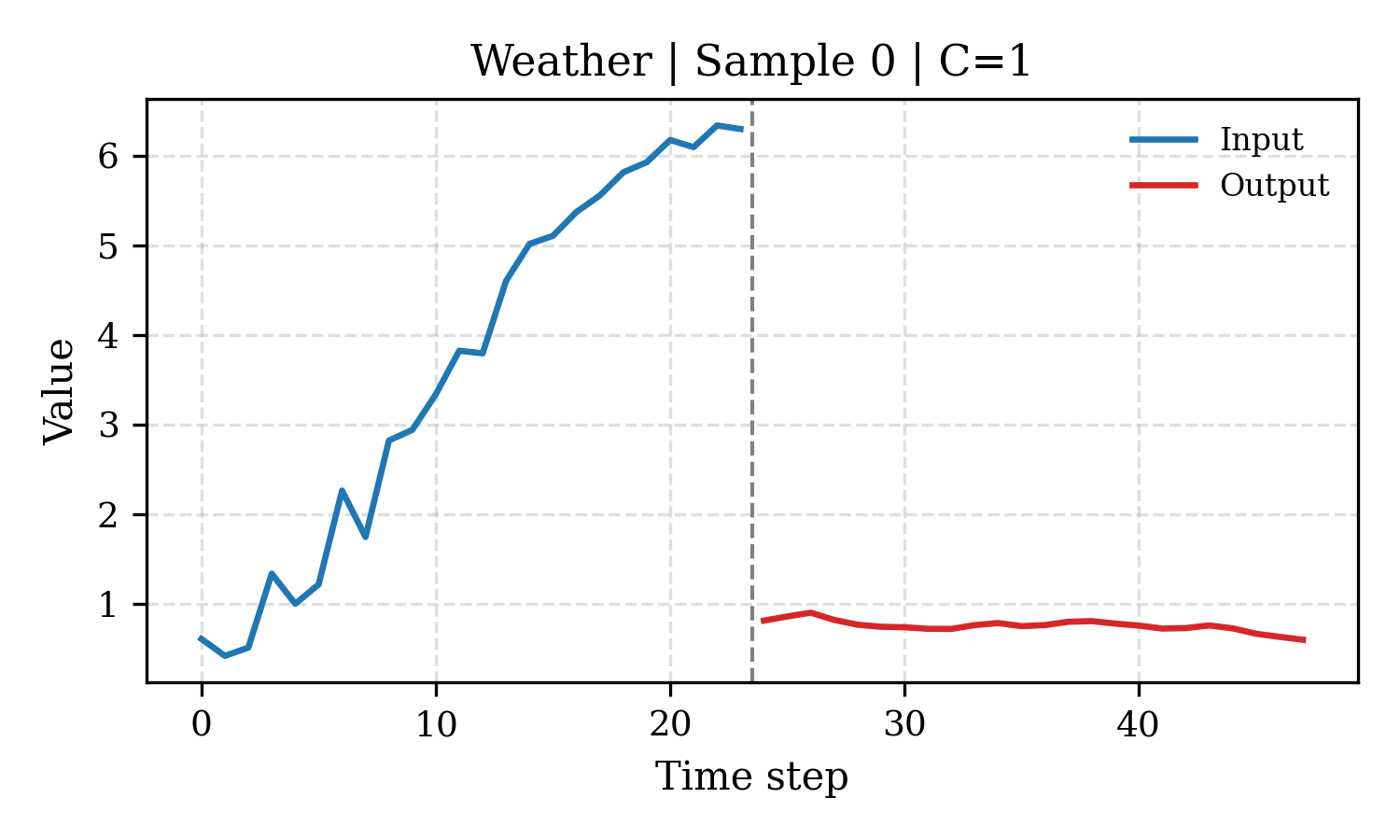} &
    \includegraphics[width=0.11\linewidth]{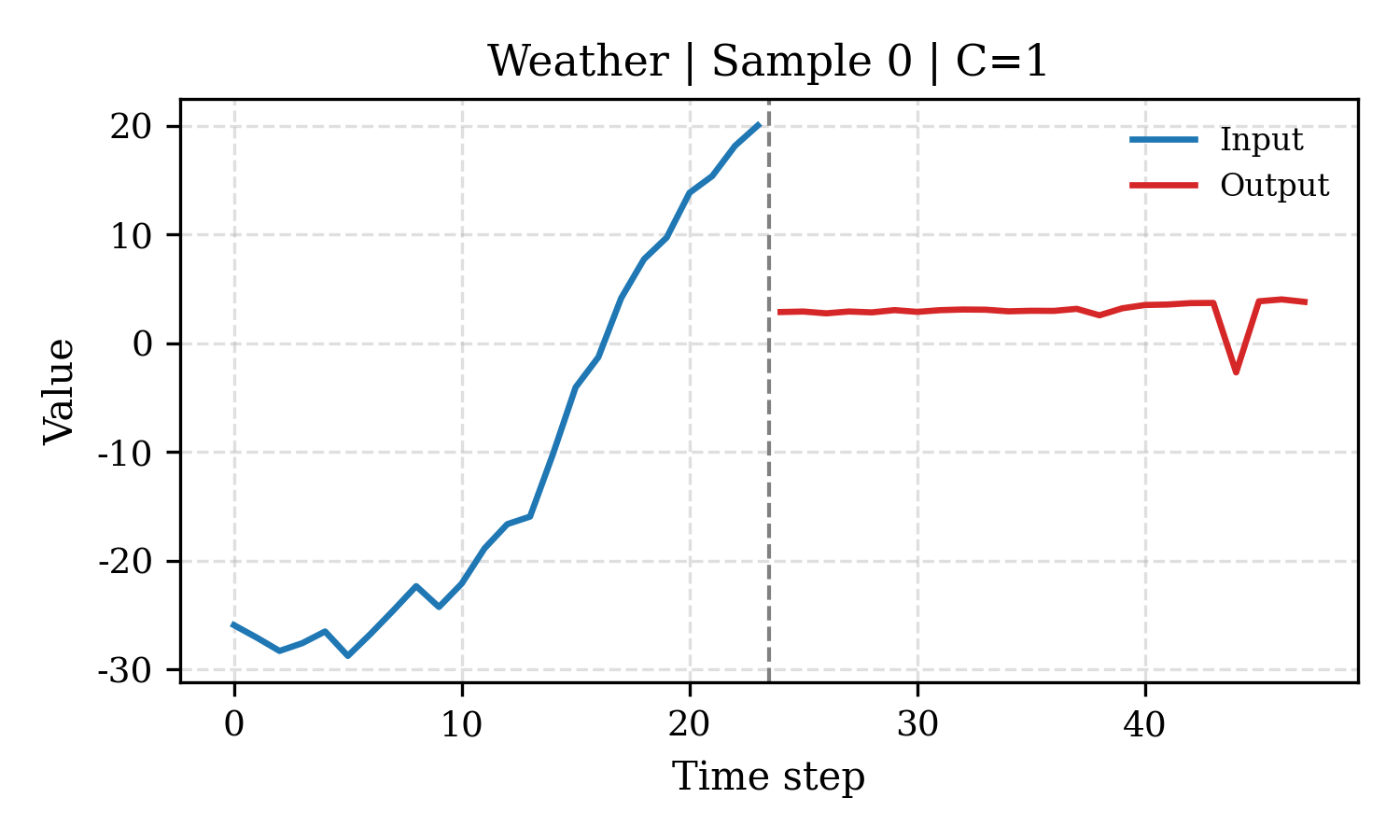} \\[0.3em]

    \rotatebox[origin=l]{90}{\textbf{Traffic}} &
    \includegraphics[width=0.11\linewidth]{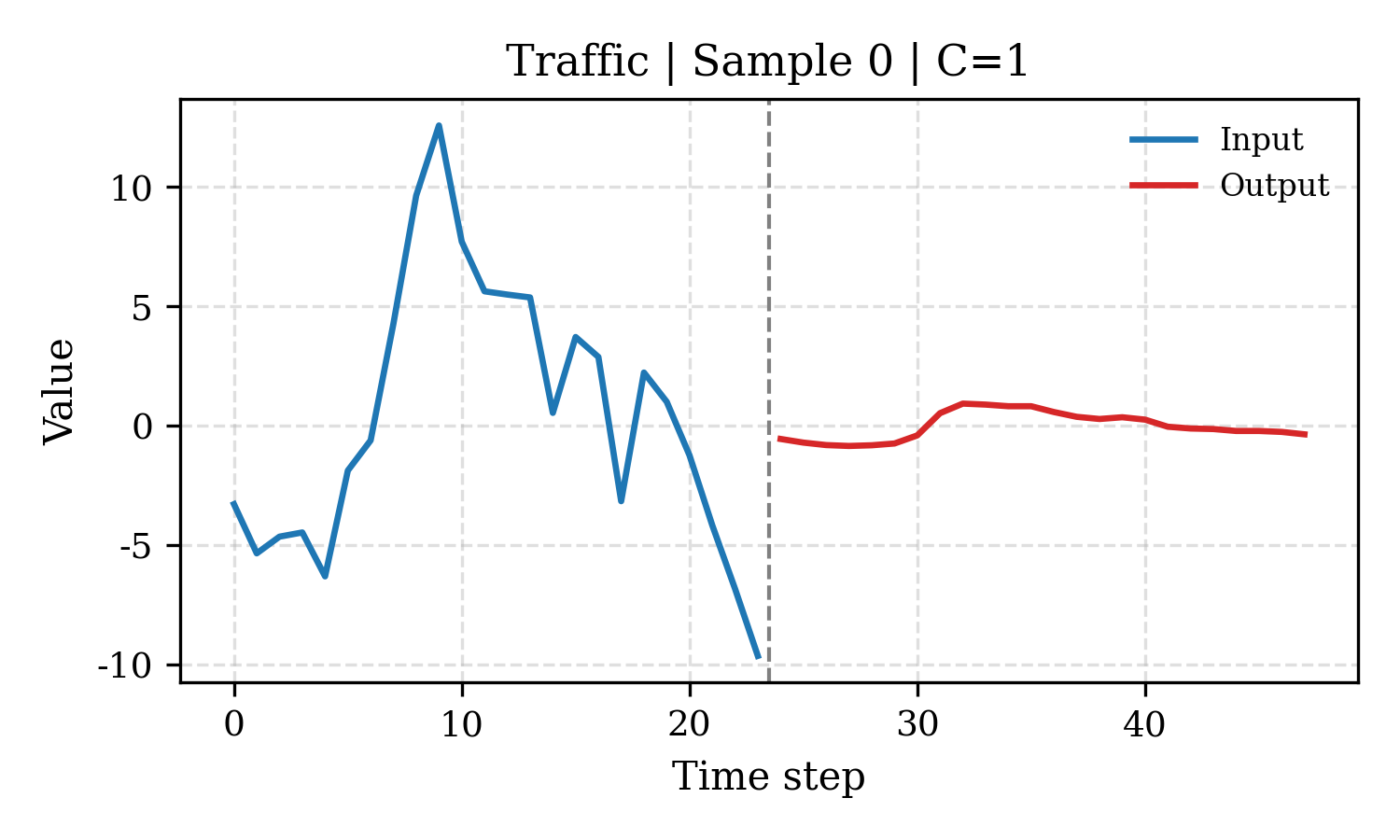} &
    \includegraphics[width=0.11\linewidth]{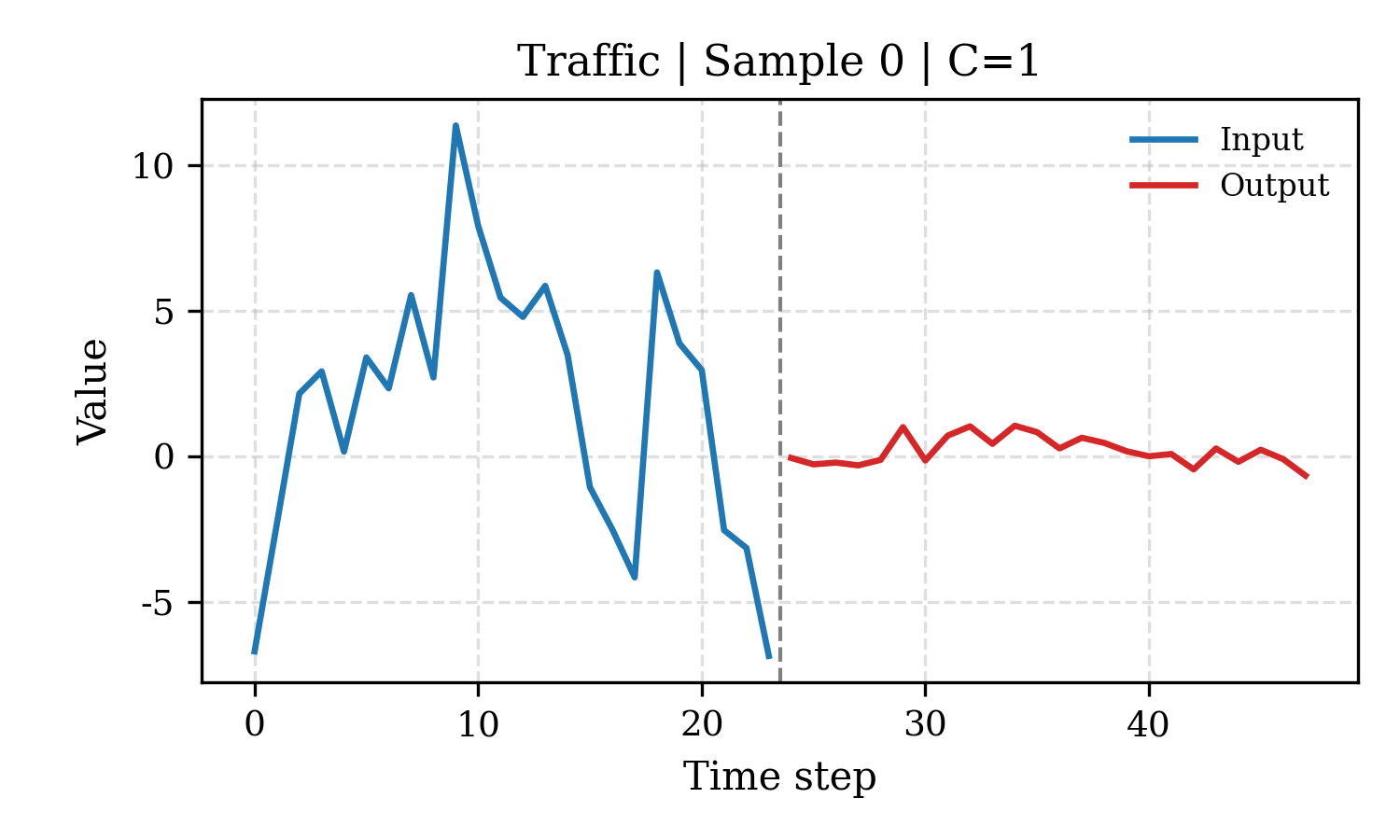} &
    \includegraphics[width=0.11\linewidth]{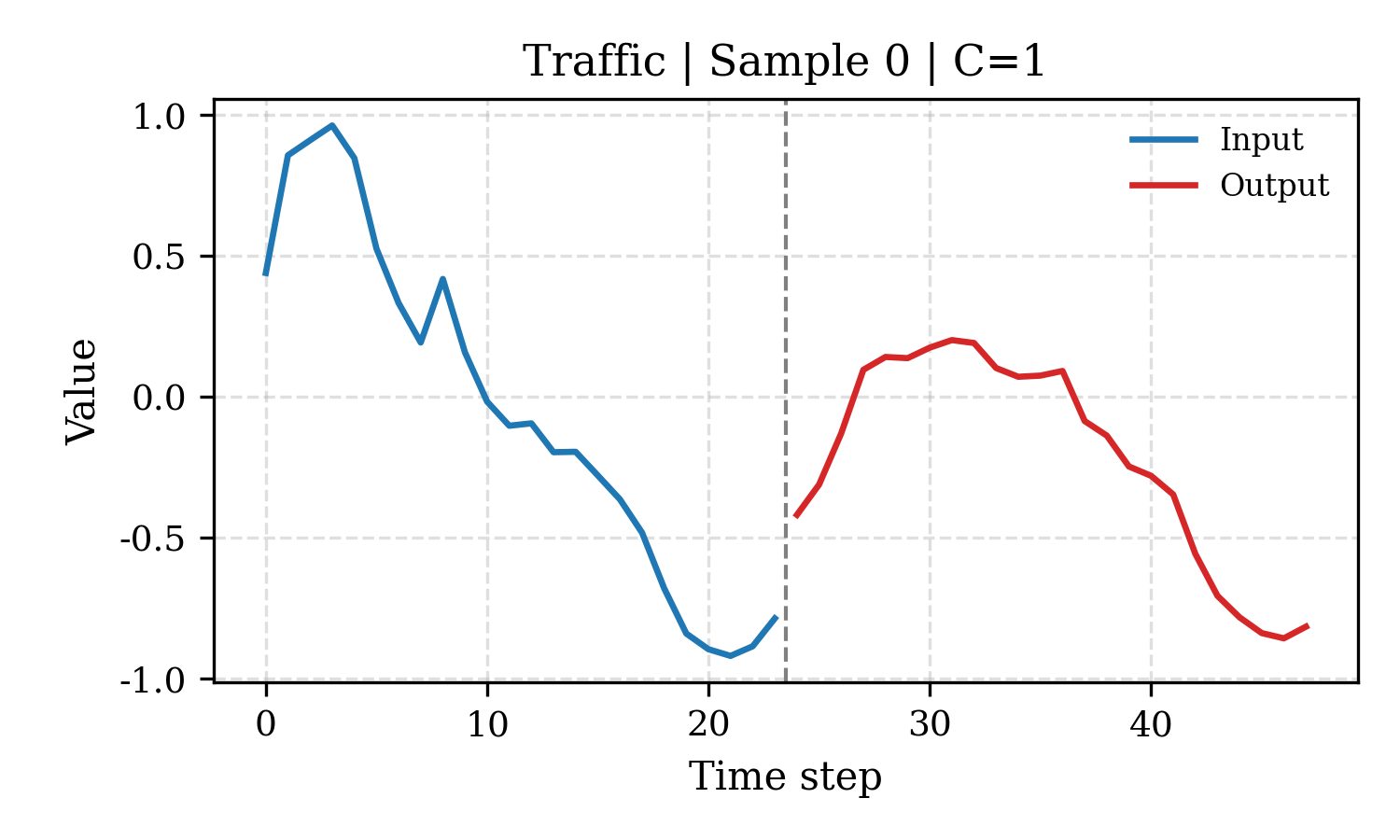} &
    \includegraphics[width=0.11\linewidth]{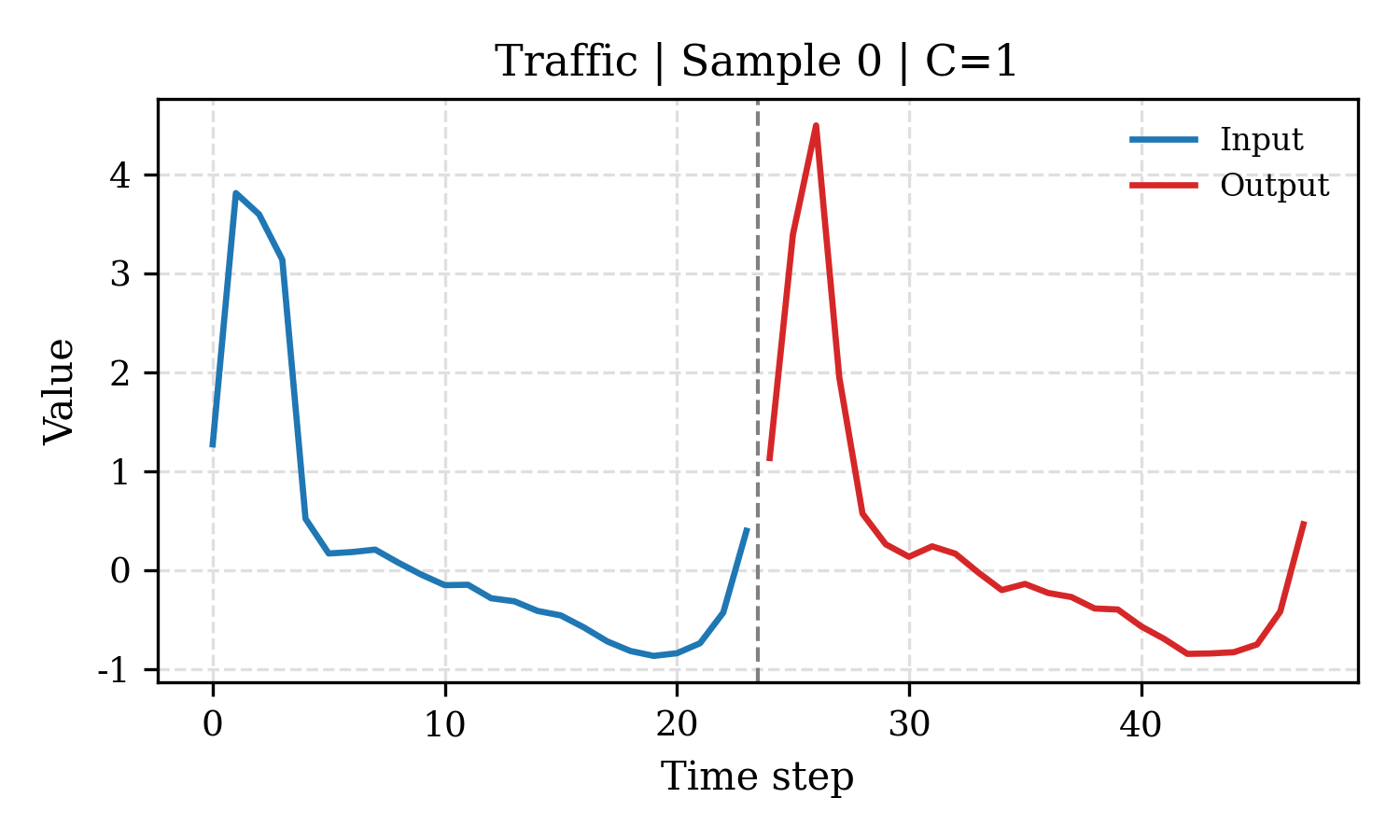} &
    \includegraphics[width=0.11\linewidth]{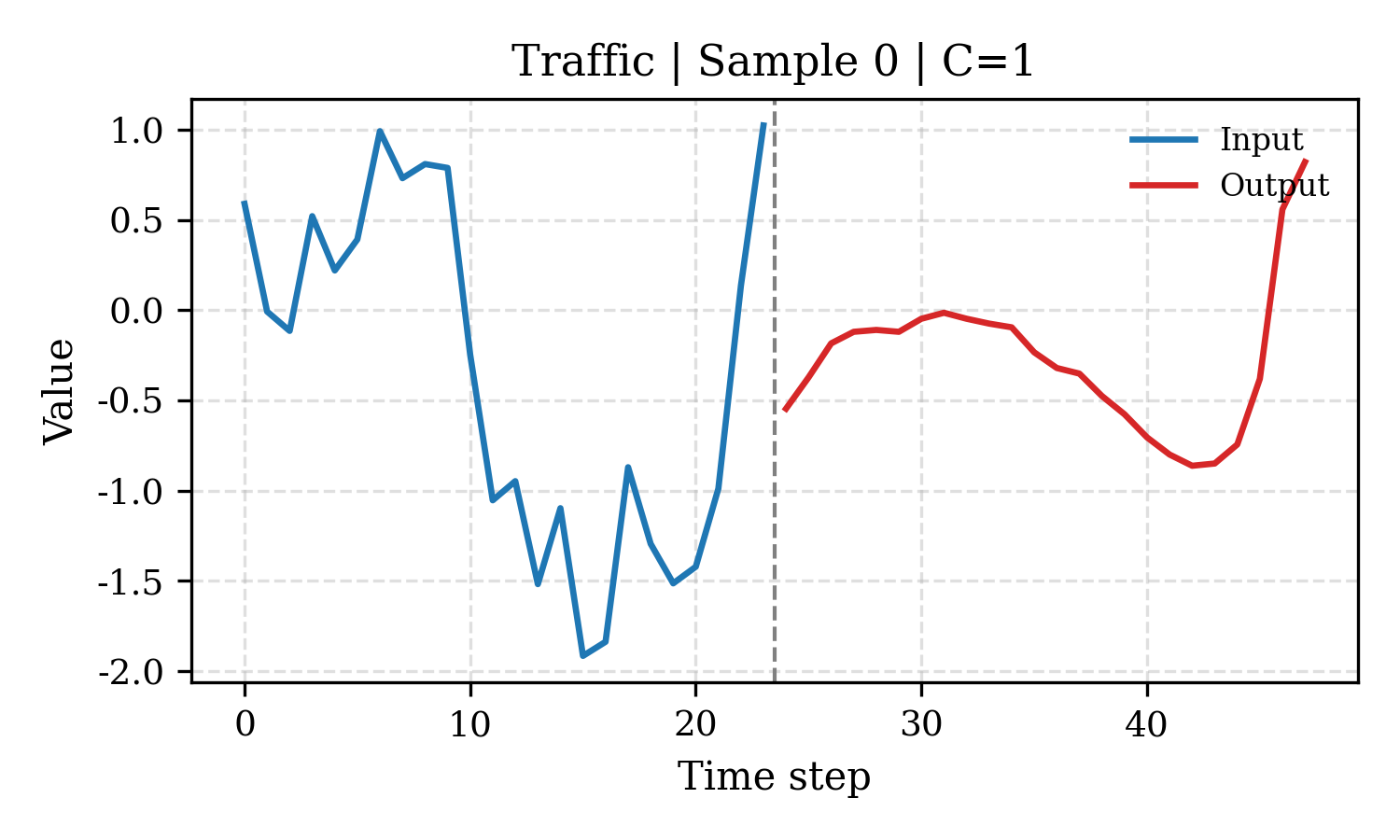} &
    \includegraphics[width=0.11\linewidth]{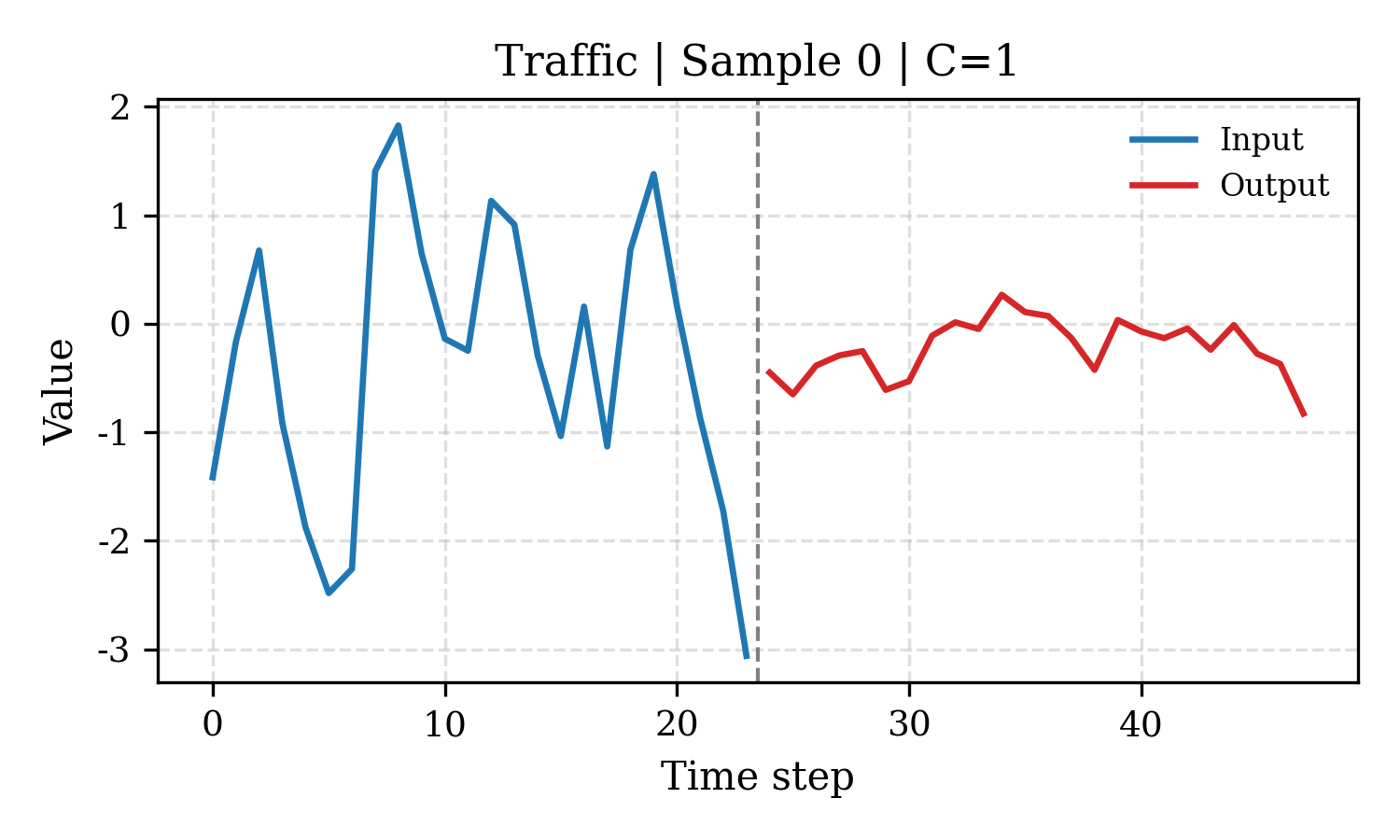} &
    \includegraphics[width=0.11\linewidth]{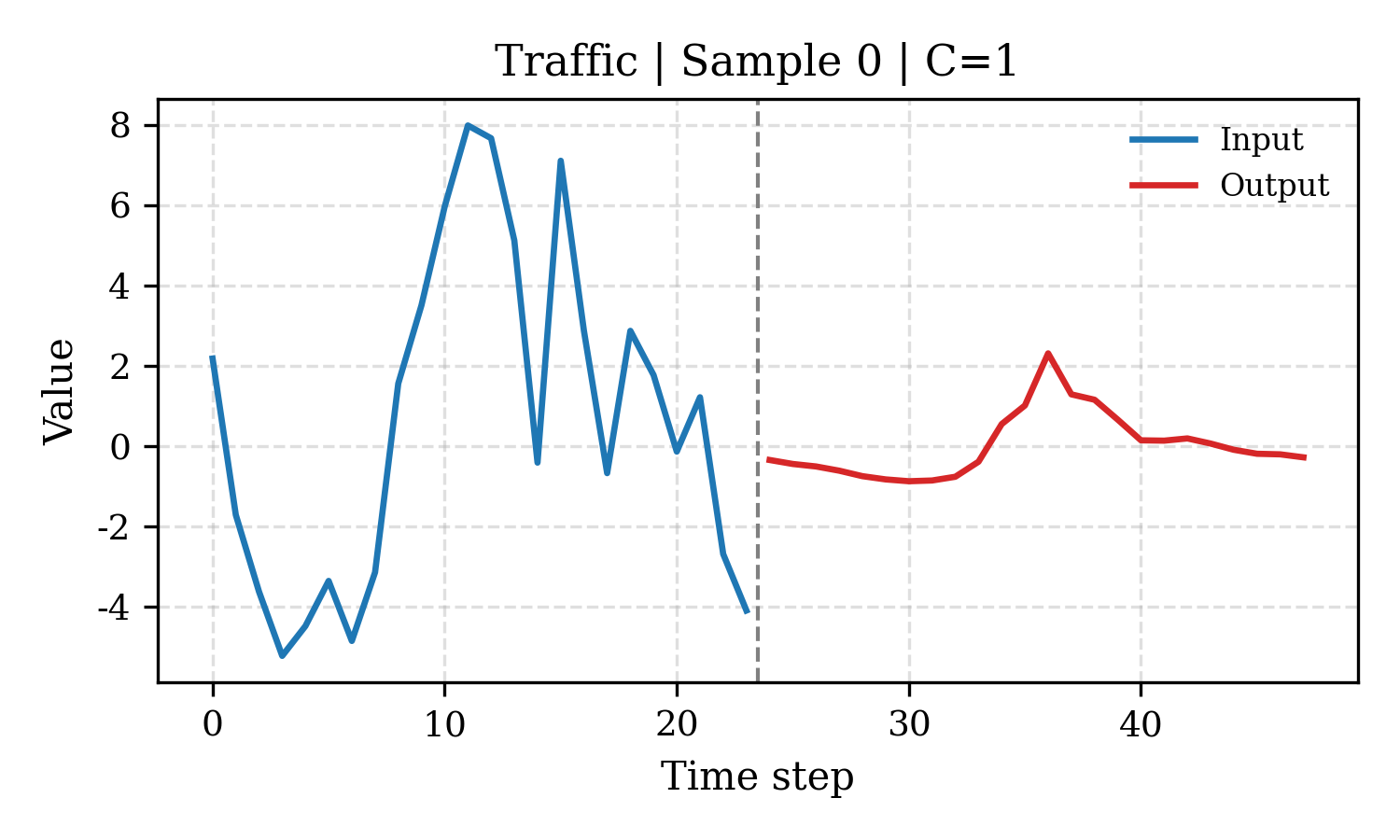} &
    \includegraphics[width=0.11\linewidth]{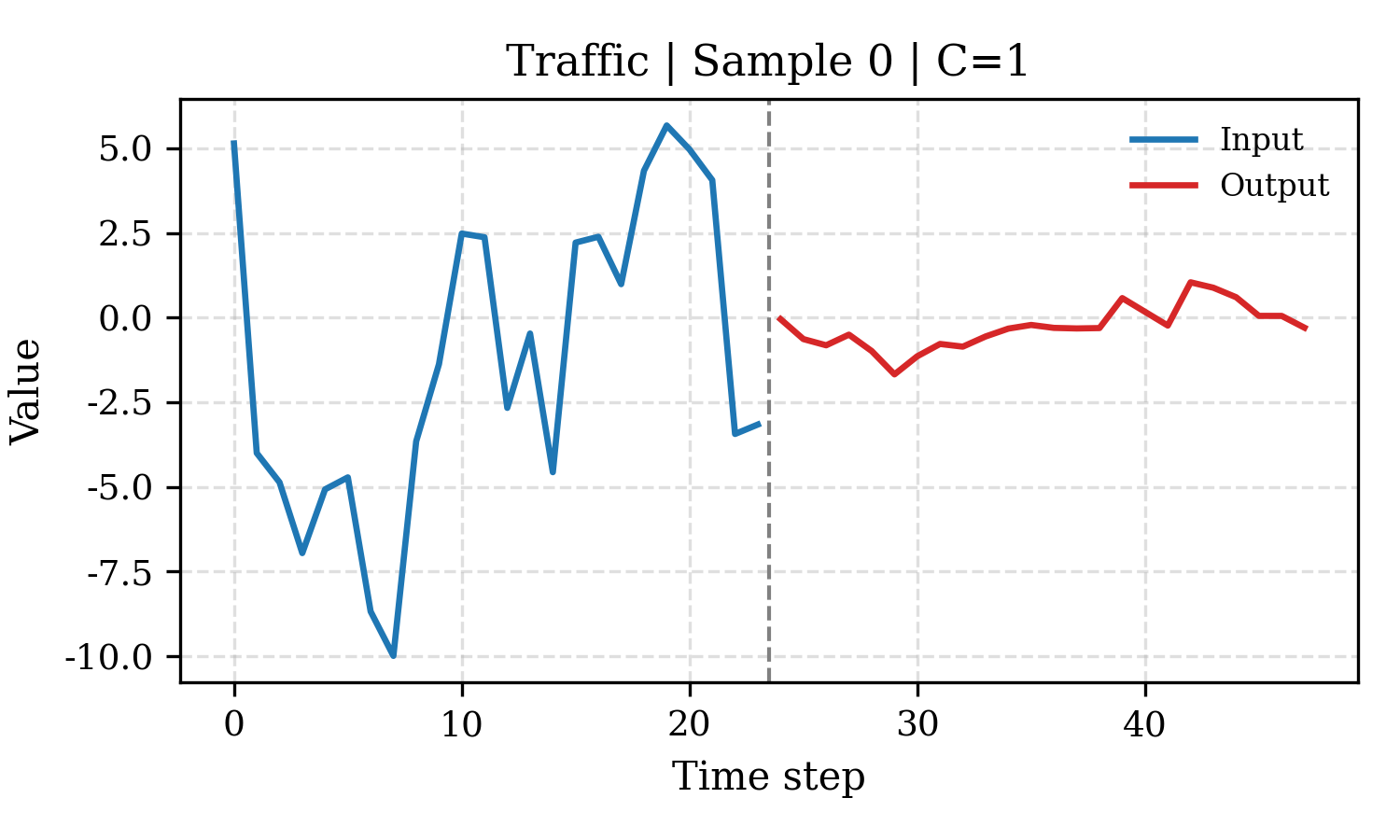} \\[0.3em]

    \rotatebox[origin=l]{90}{\textbf{PEMS03}} &
    \includegraphics[width=0.11\linewidth]{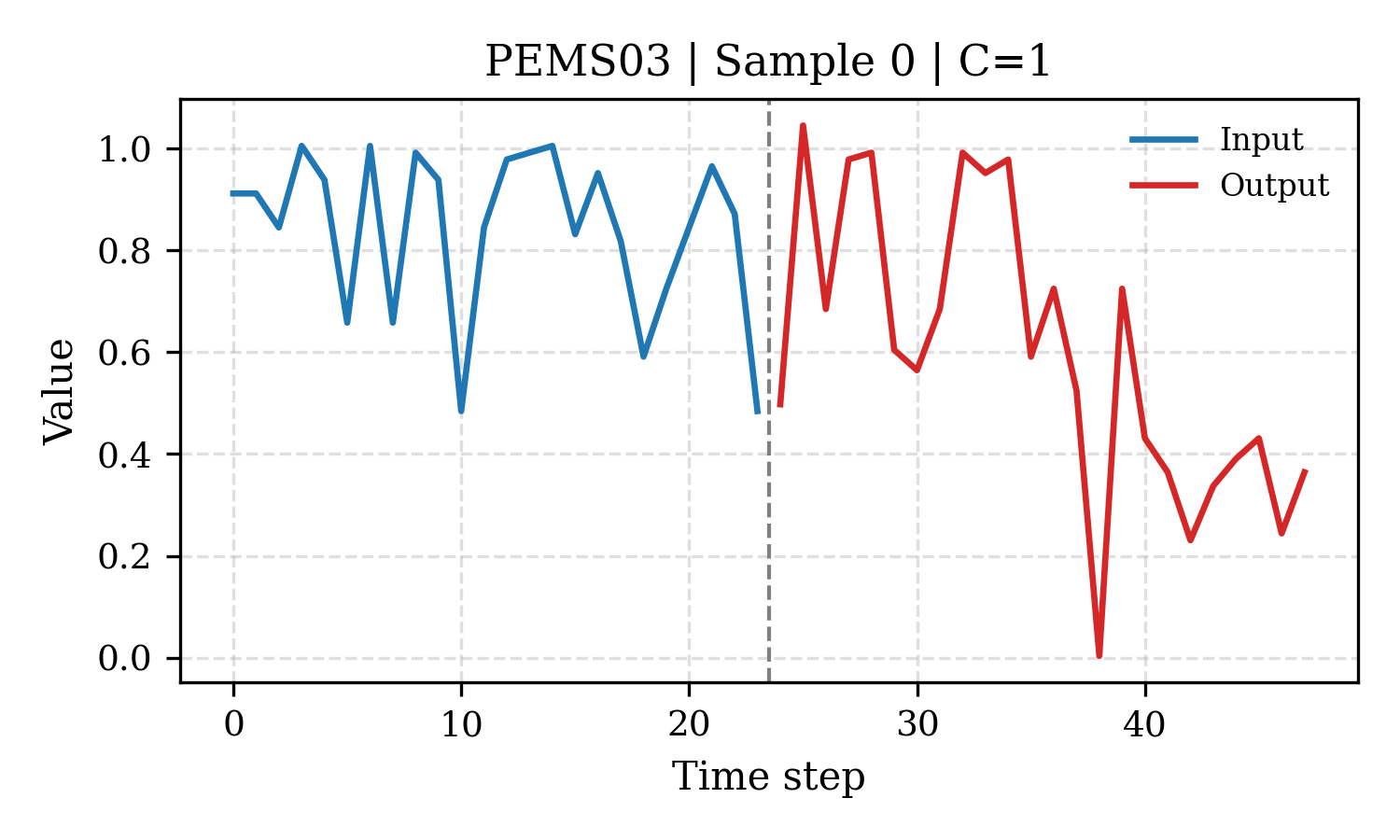} &
    \includegraphics[width=0.11\linewidth]{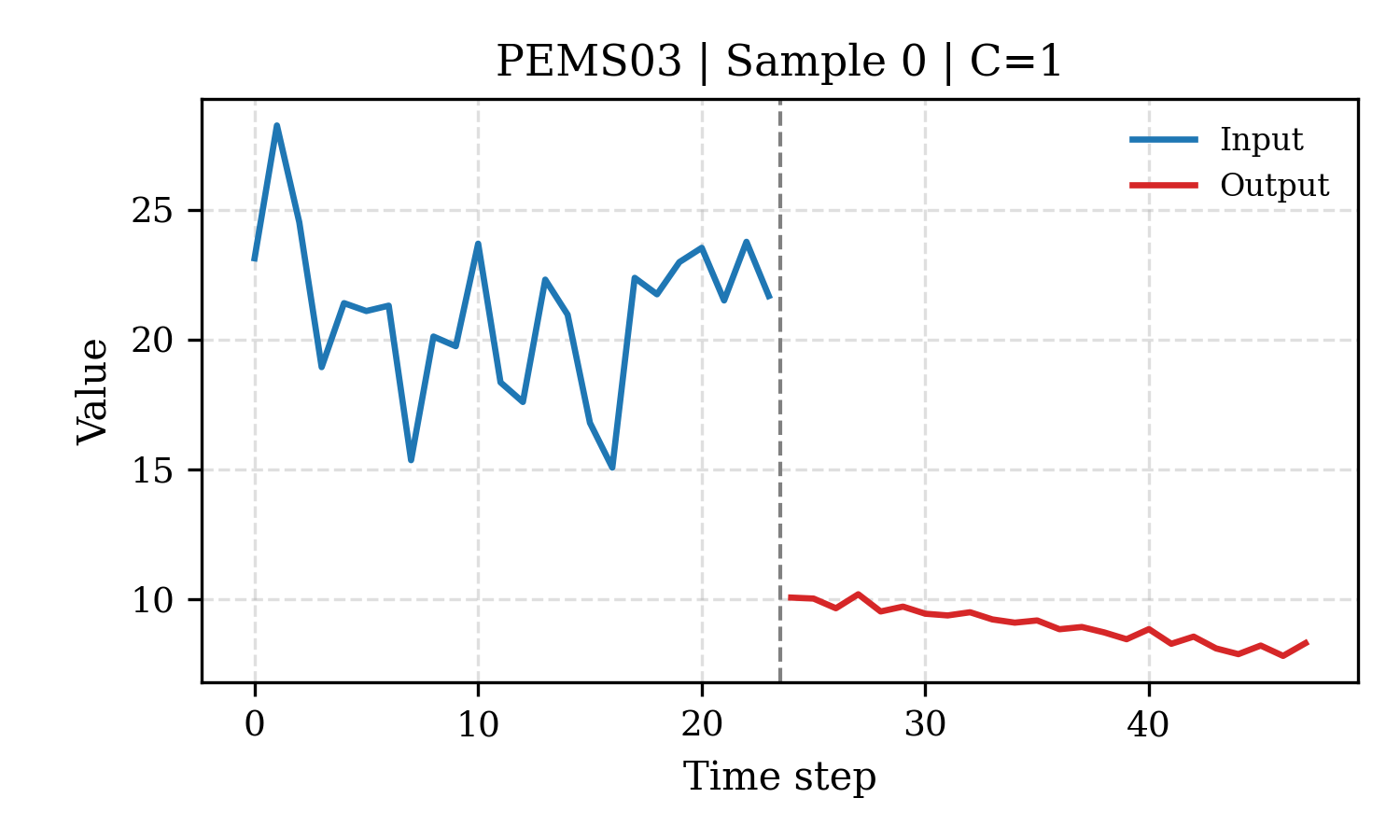} &
    \includegraphics[width=0.11\linewidth]{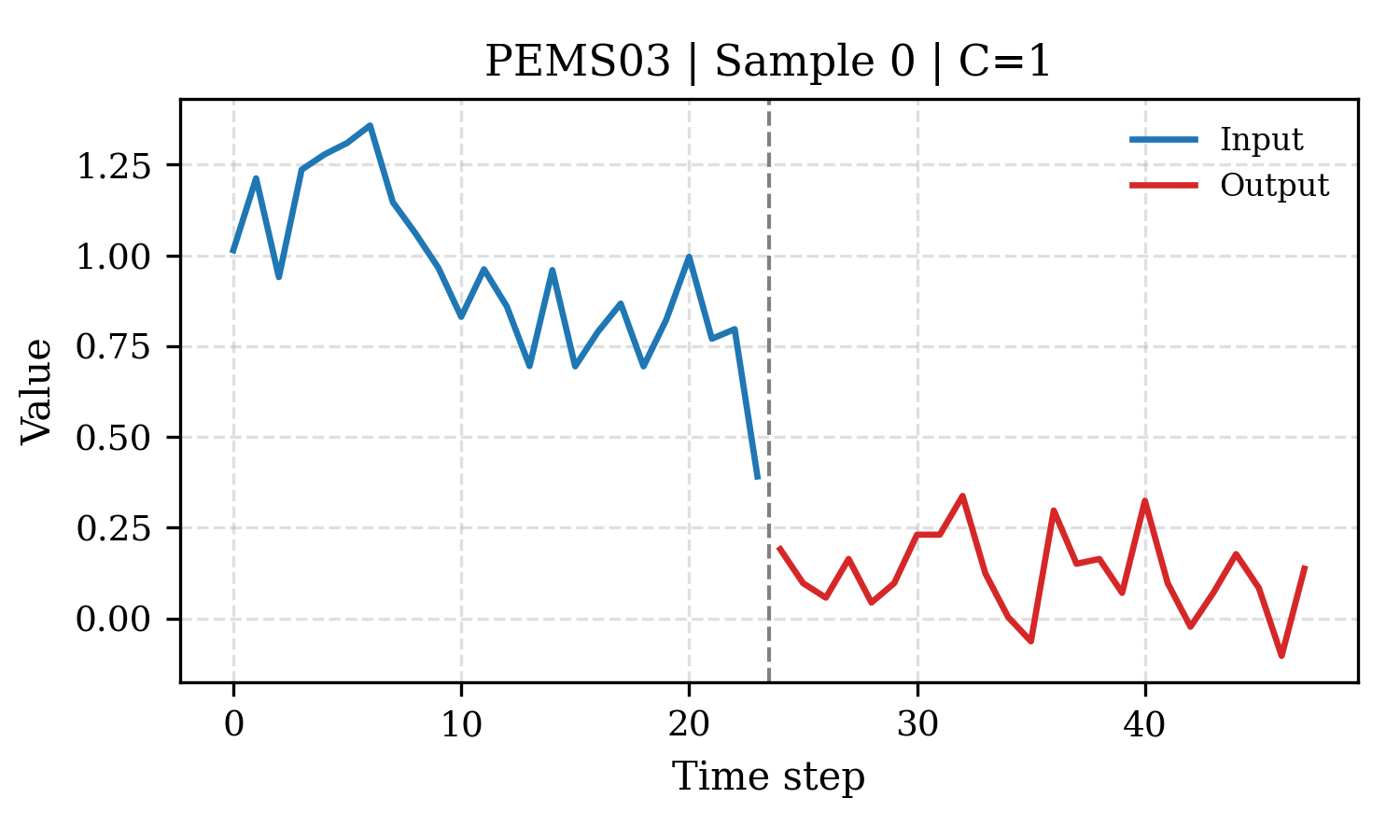} &
    \includegraphics[width=0.11\linewidth]{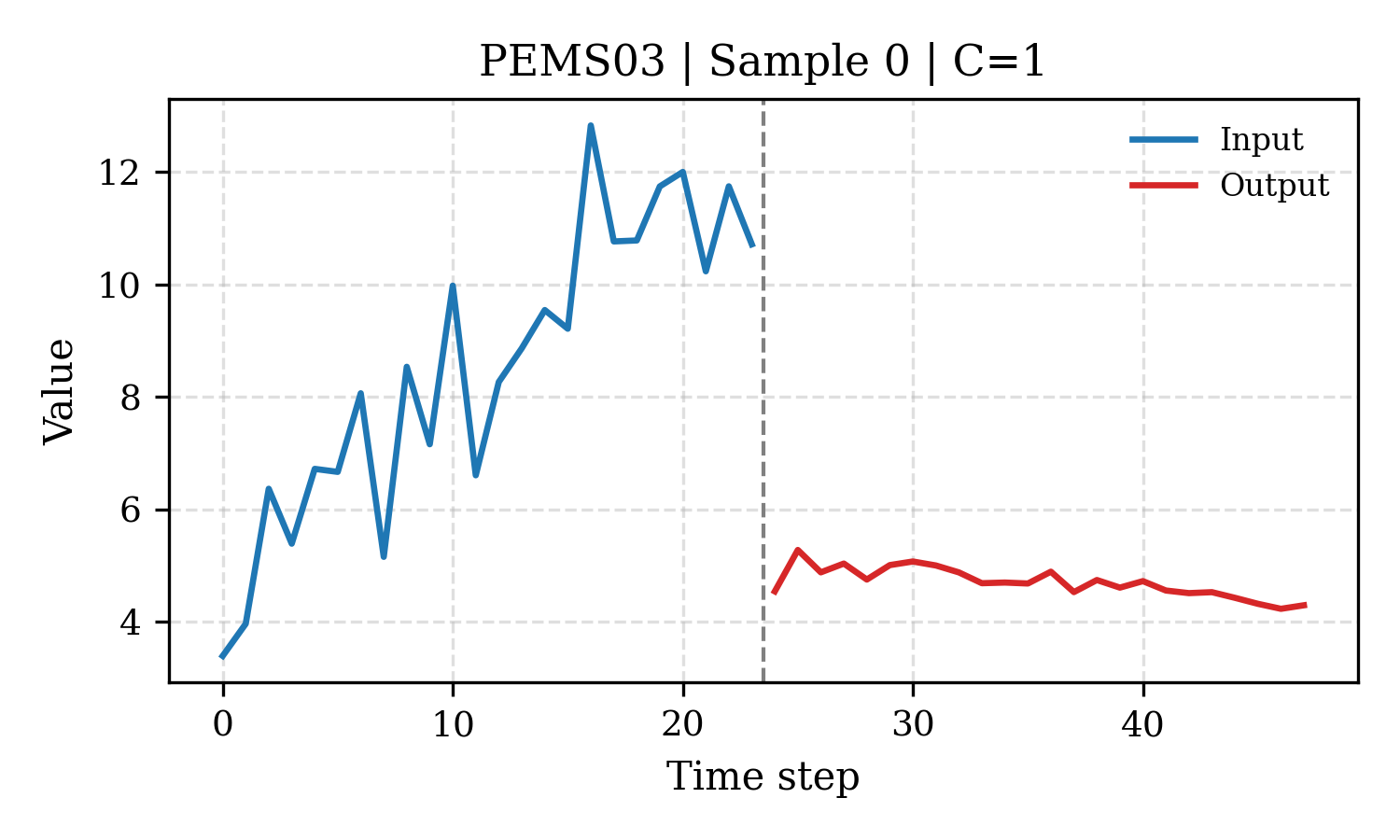} &
    \includegraphics[width=0.11\linewidth]{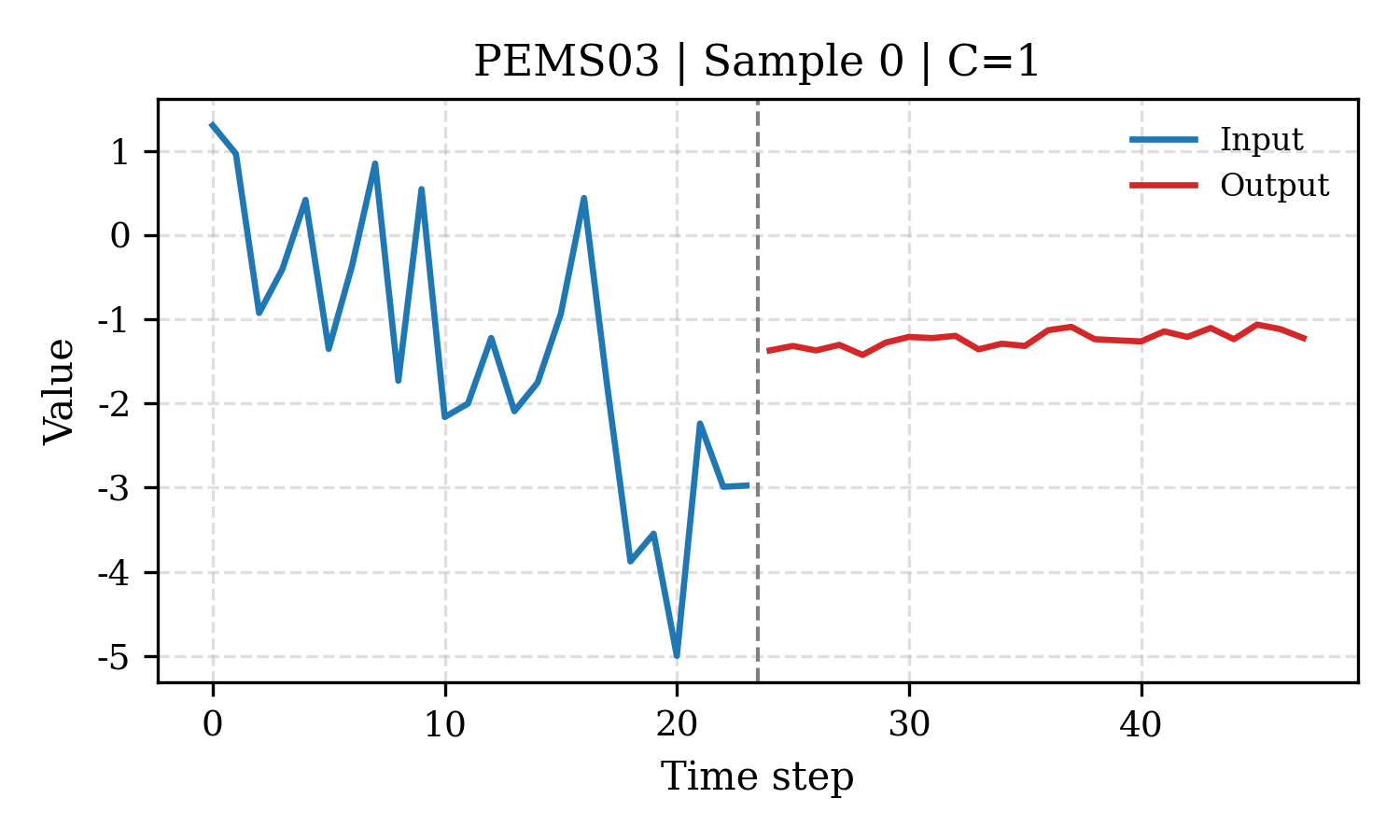} &
    \includegraphics[width=0.11\linewidth]{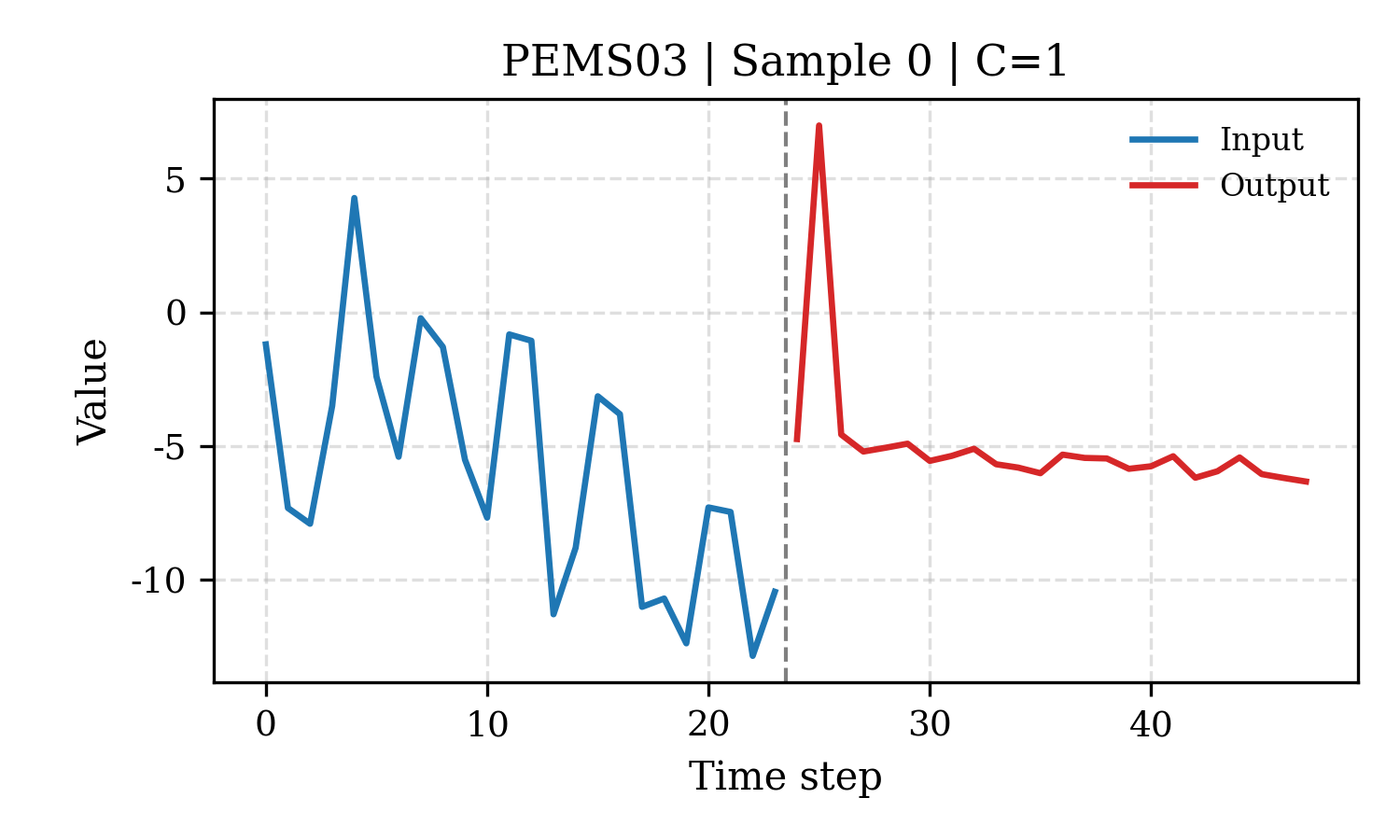} &
    \includegraphics[width=0.11\linewidth]{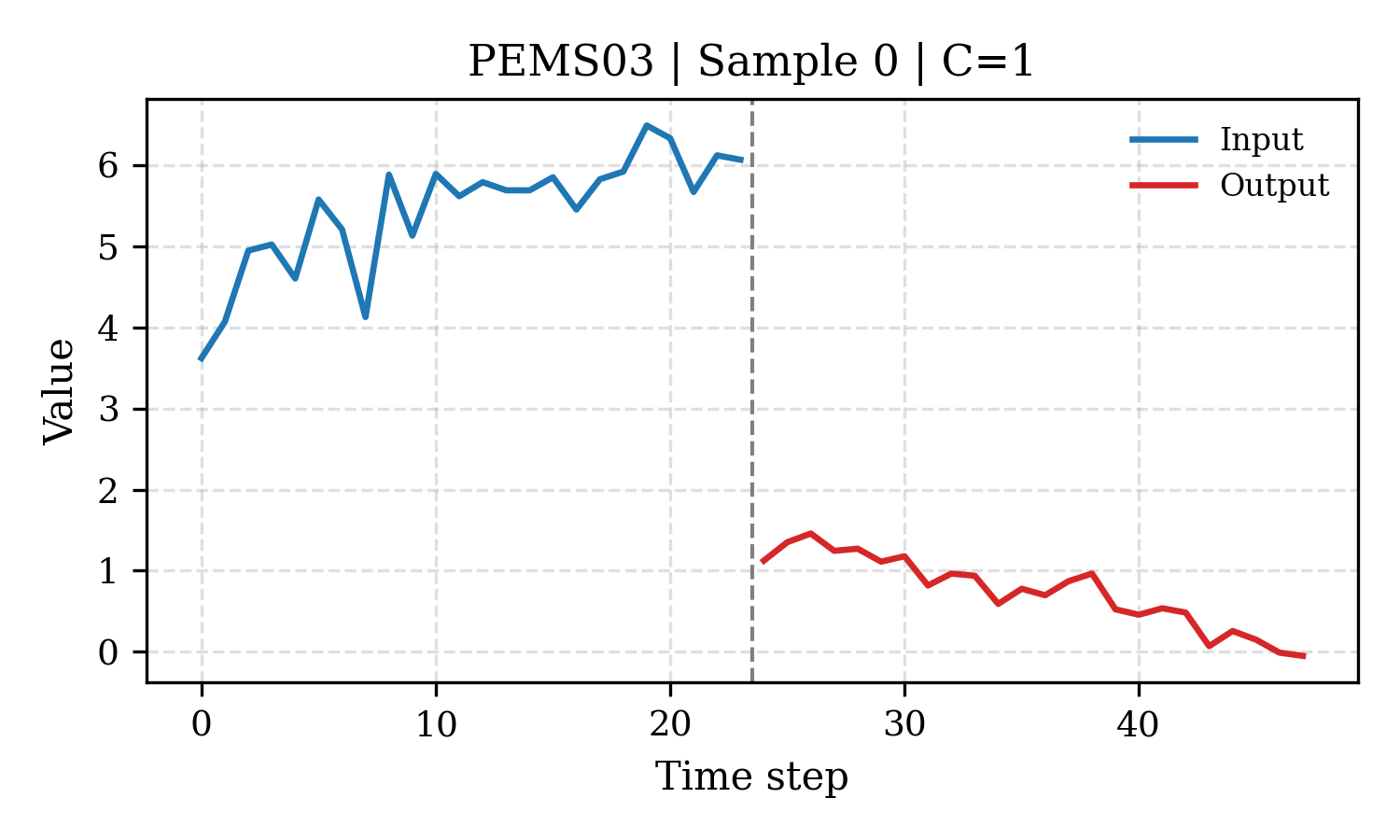} &
    \includegraphics[width=0.11\linewidth]{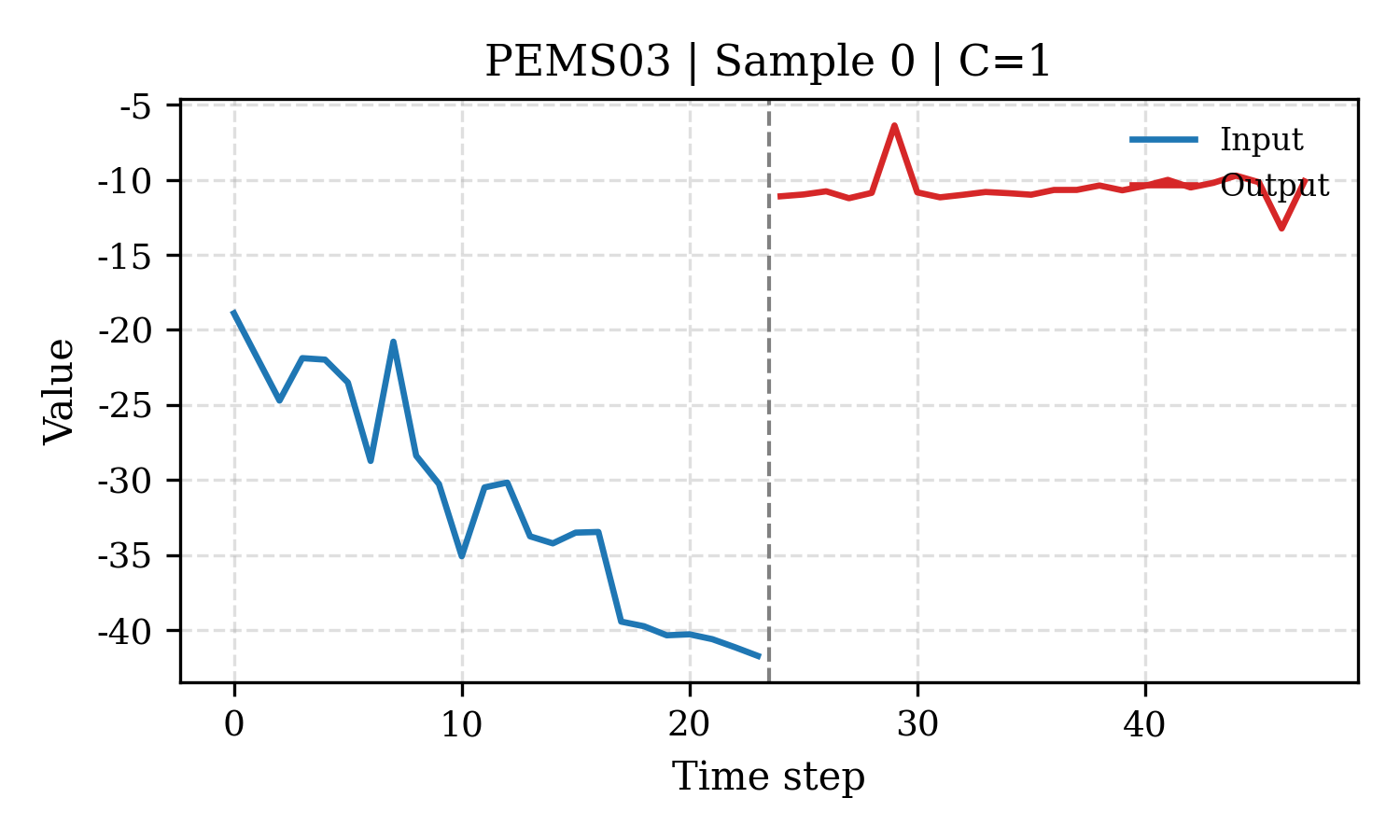} \\[0.3em]

    \rotatebox[origin=l]{90}{\textbf{PEMS04}} &
    \includegraphics[width=0.11\linewidth]{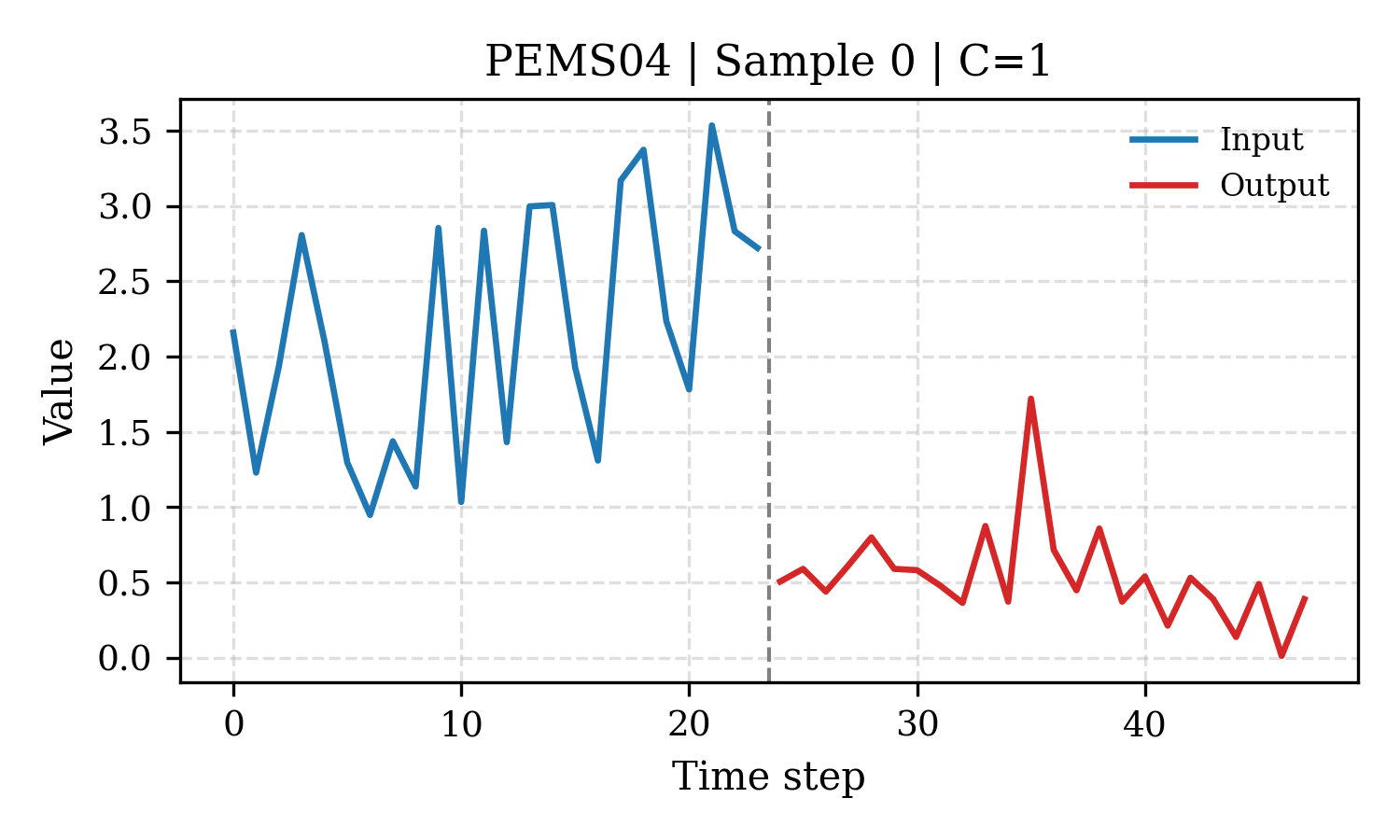} &
    \includegraphics[width=0.11\linewidth]{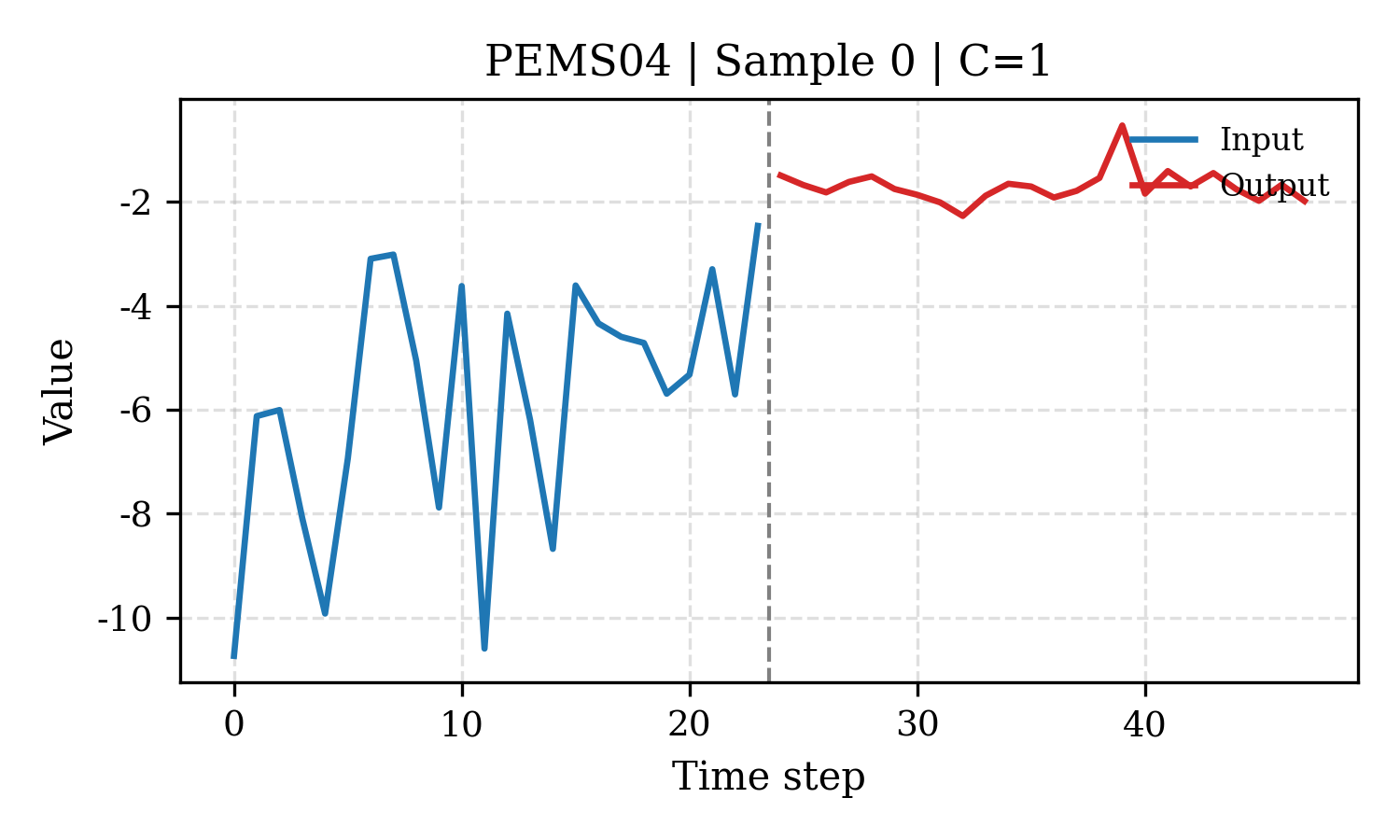} &
    \includegraphics[width=0.11\linewidth]{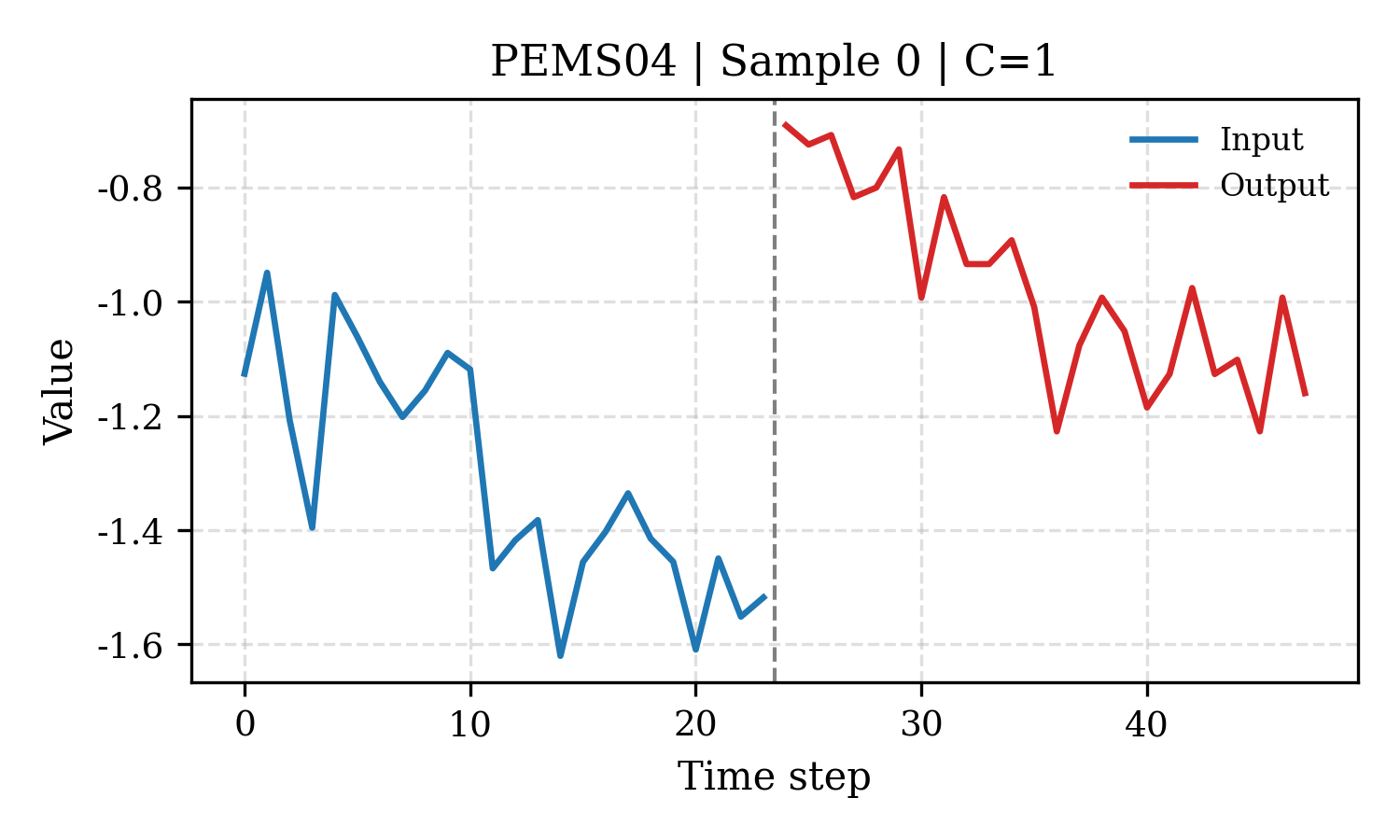} &
    \includegraphics[width=0.11\linewidth]{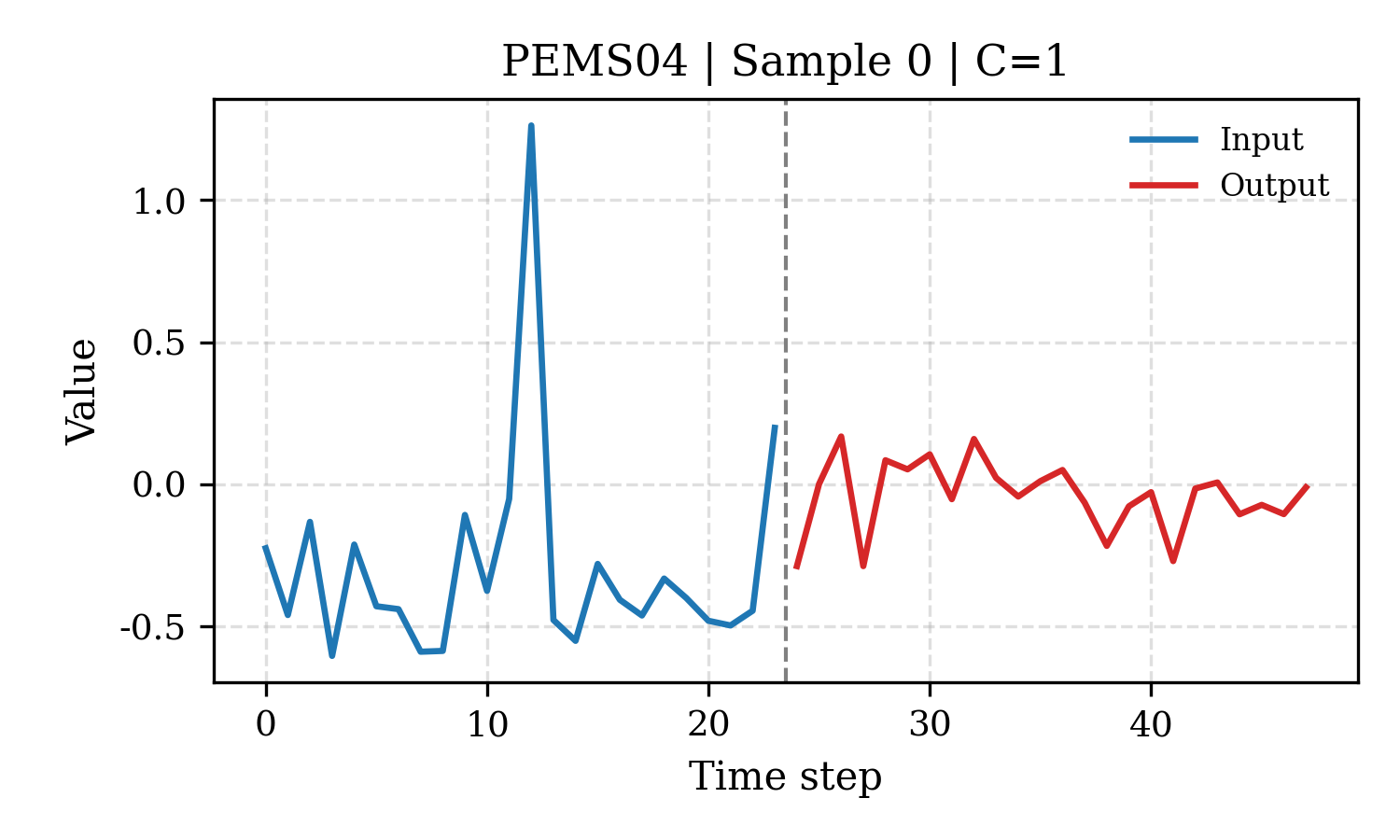} &
    \includegraphics[width=0.11\linewidth]{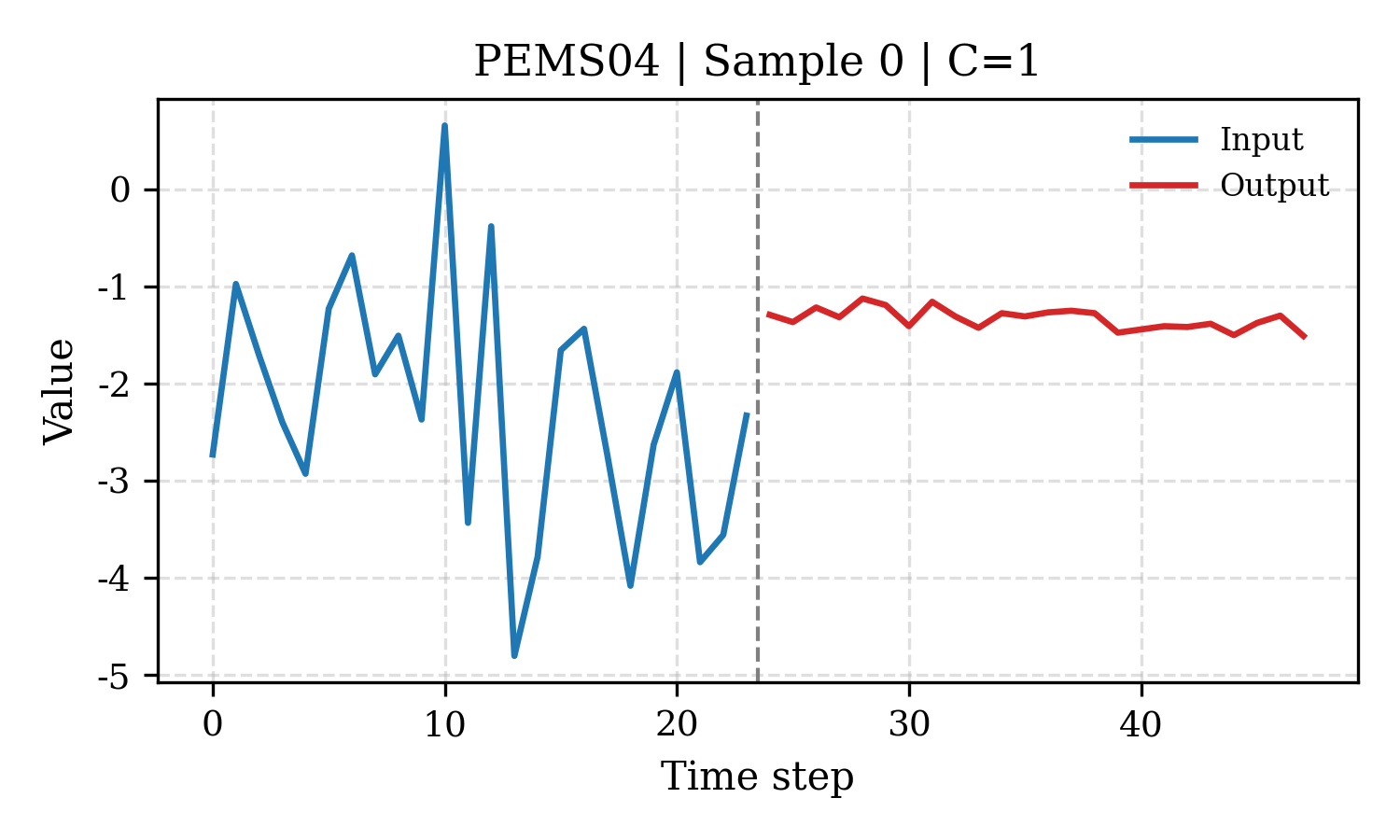} &
    \includegraphics[width=0.11\linewidth]{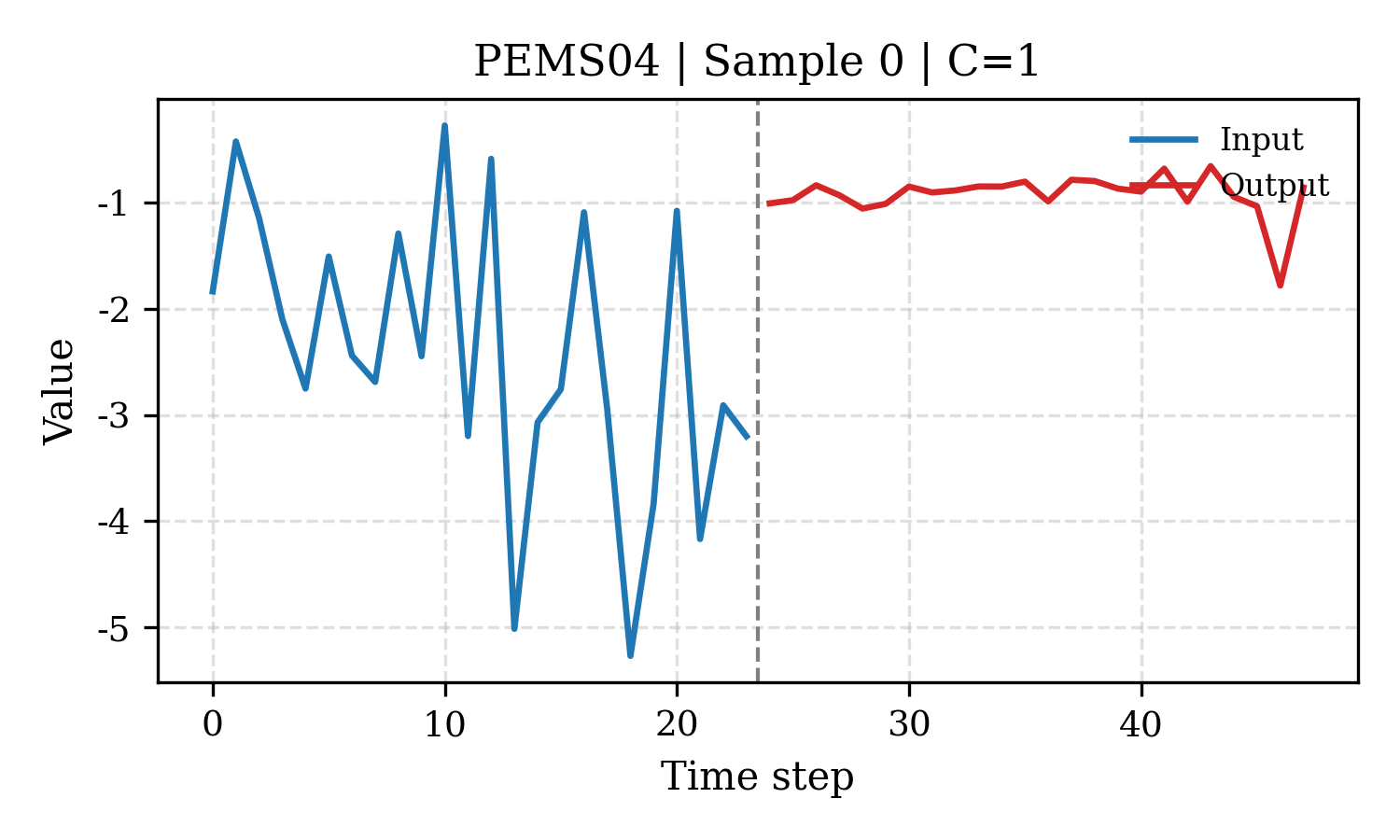} &
    \includegraphics[width=0.11\linewidth]{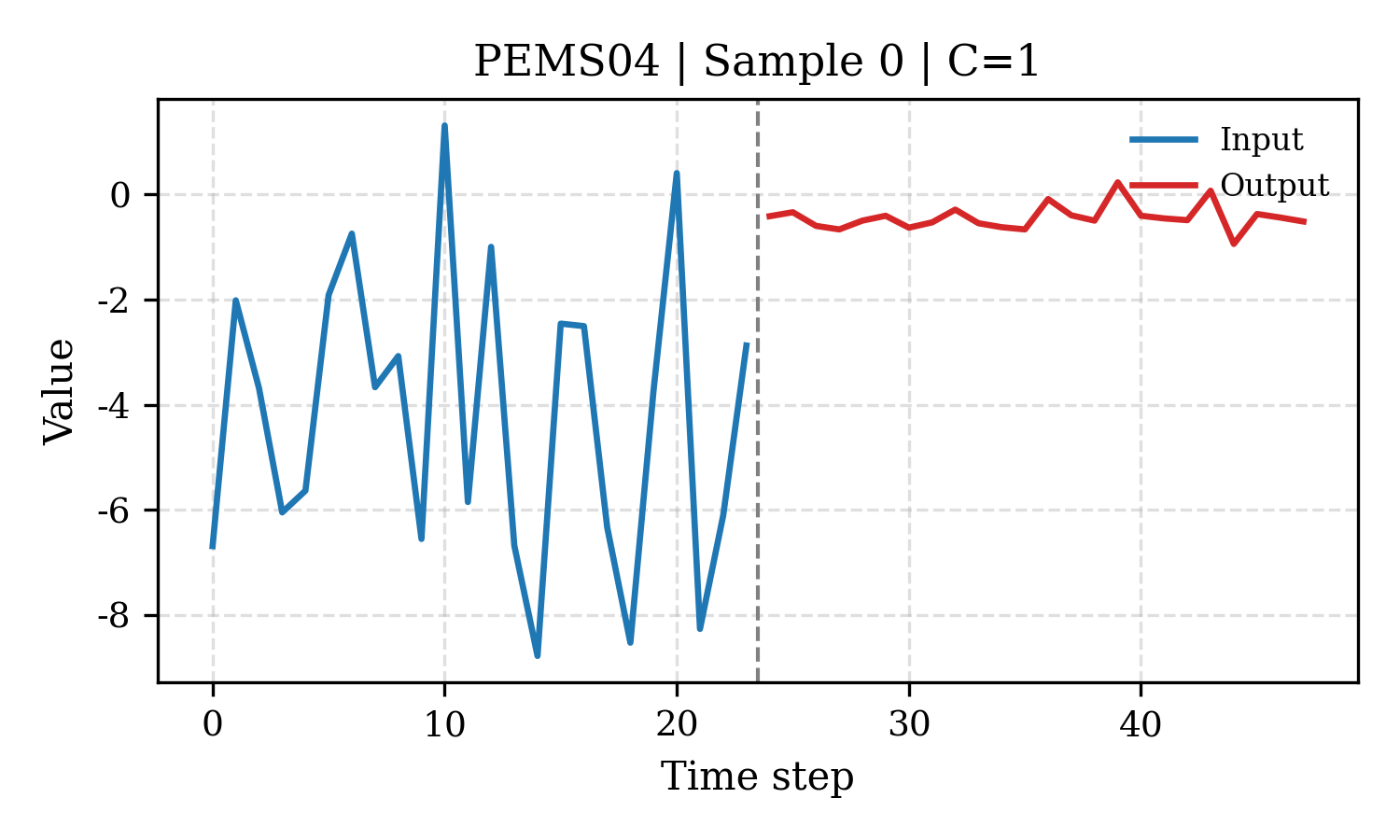} &
    \includegraphics[width=0.11\linewidth]{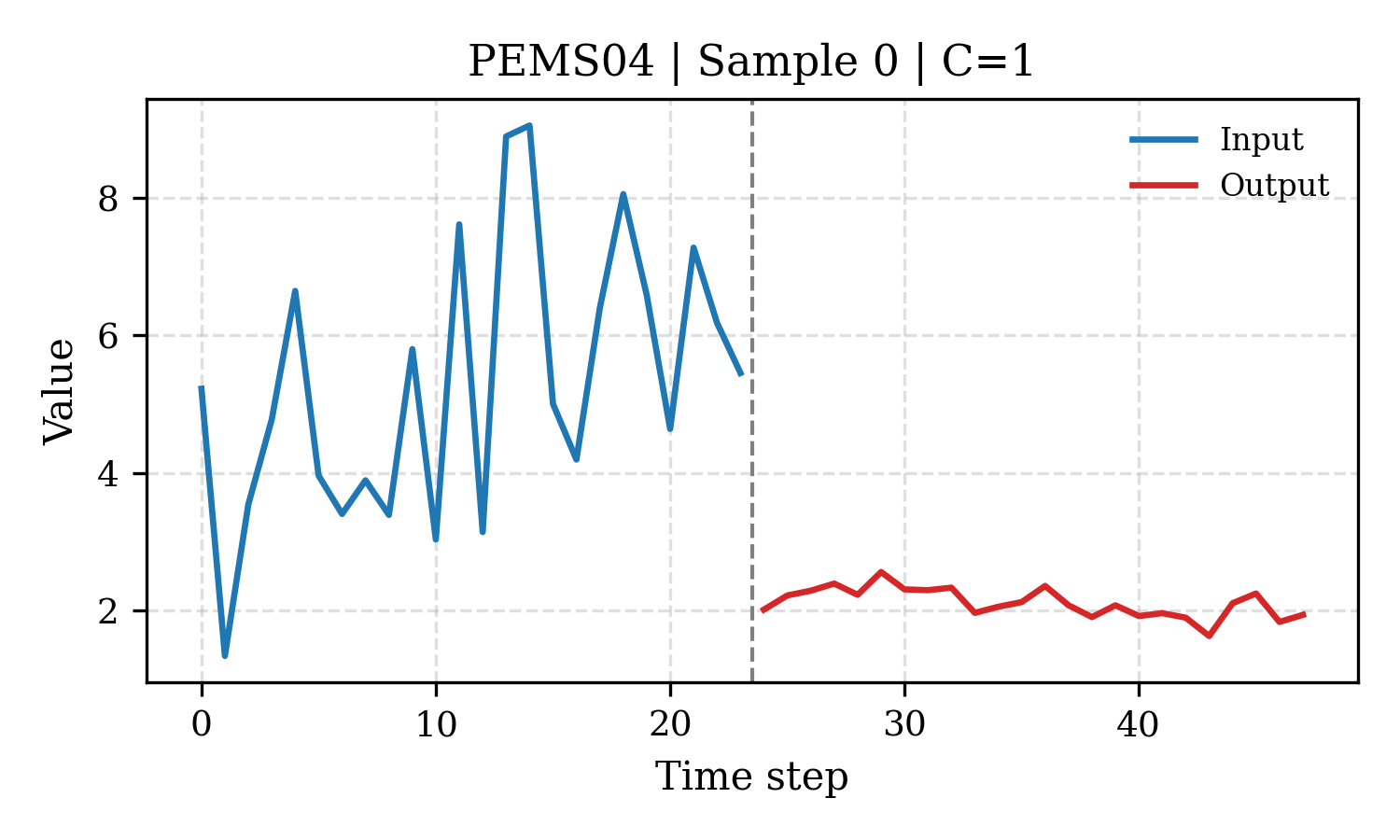} \\[0.3em]

    \rotatebox[origin=l]{90}{\textbf{PEMS07}} &
    \includegraphics[width=0.11\linewidth]{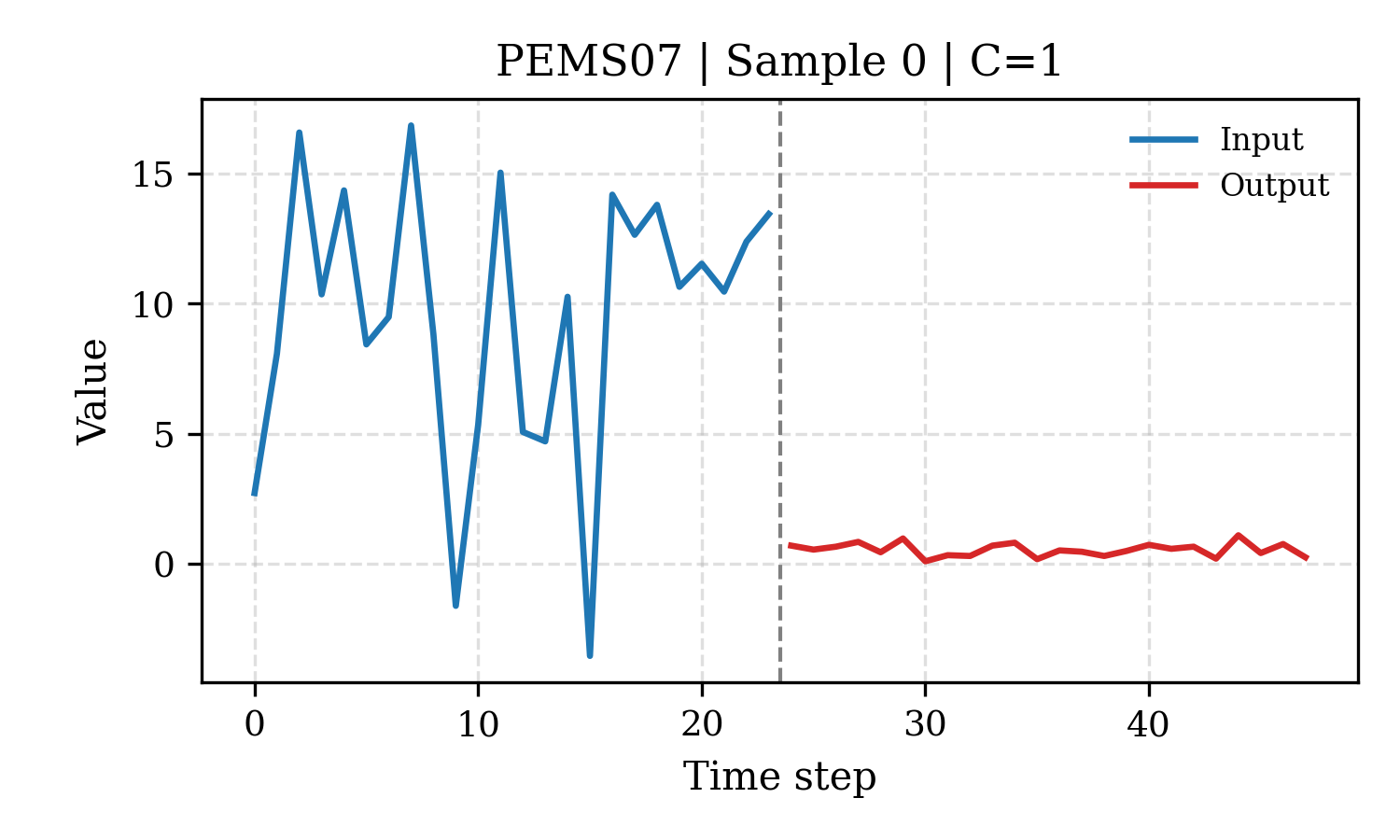} &
    \includegraphics[width=0.11\linewidth]{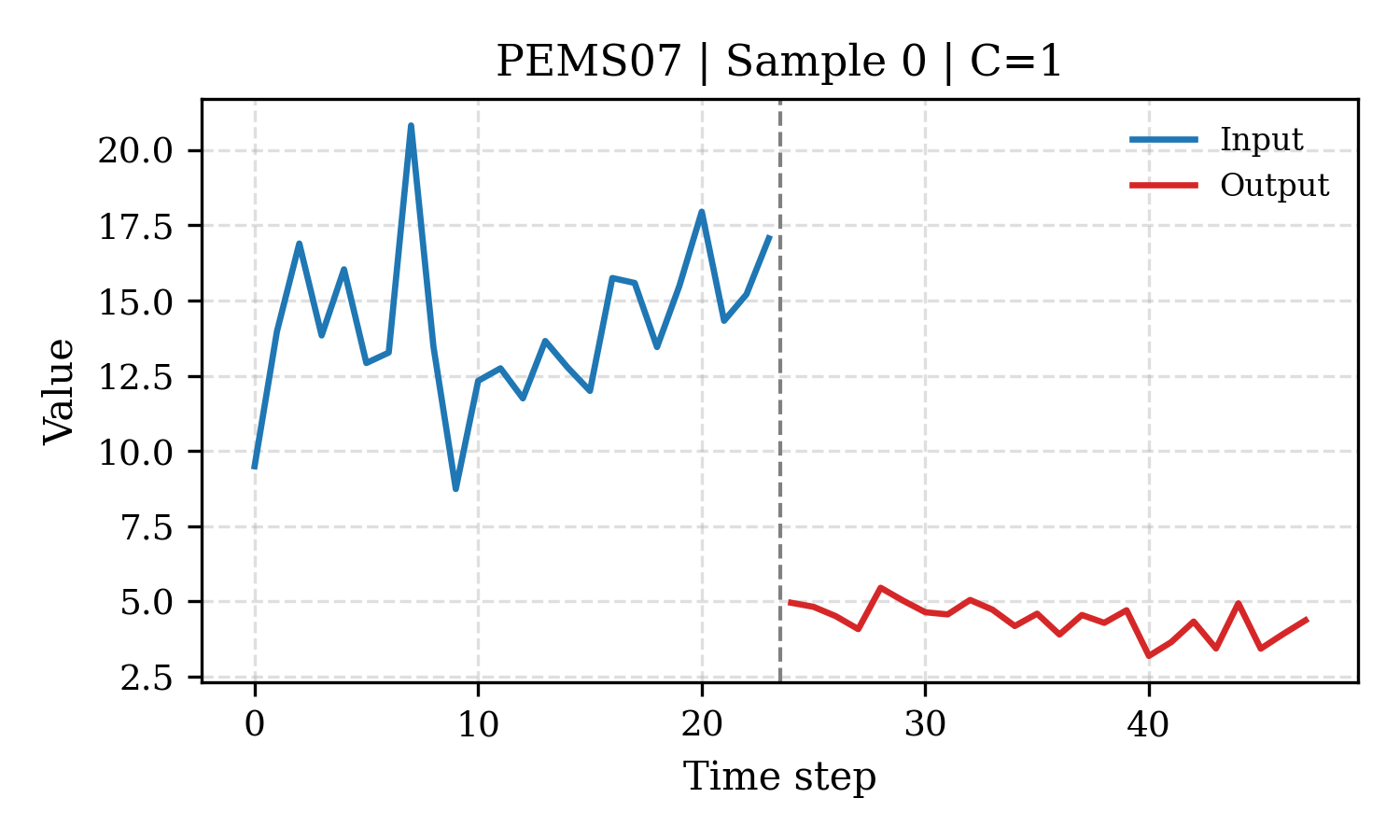} &
    \includegraphics[width=0.11\linewidth]{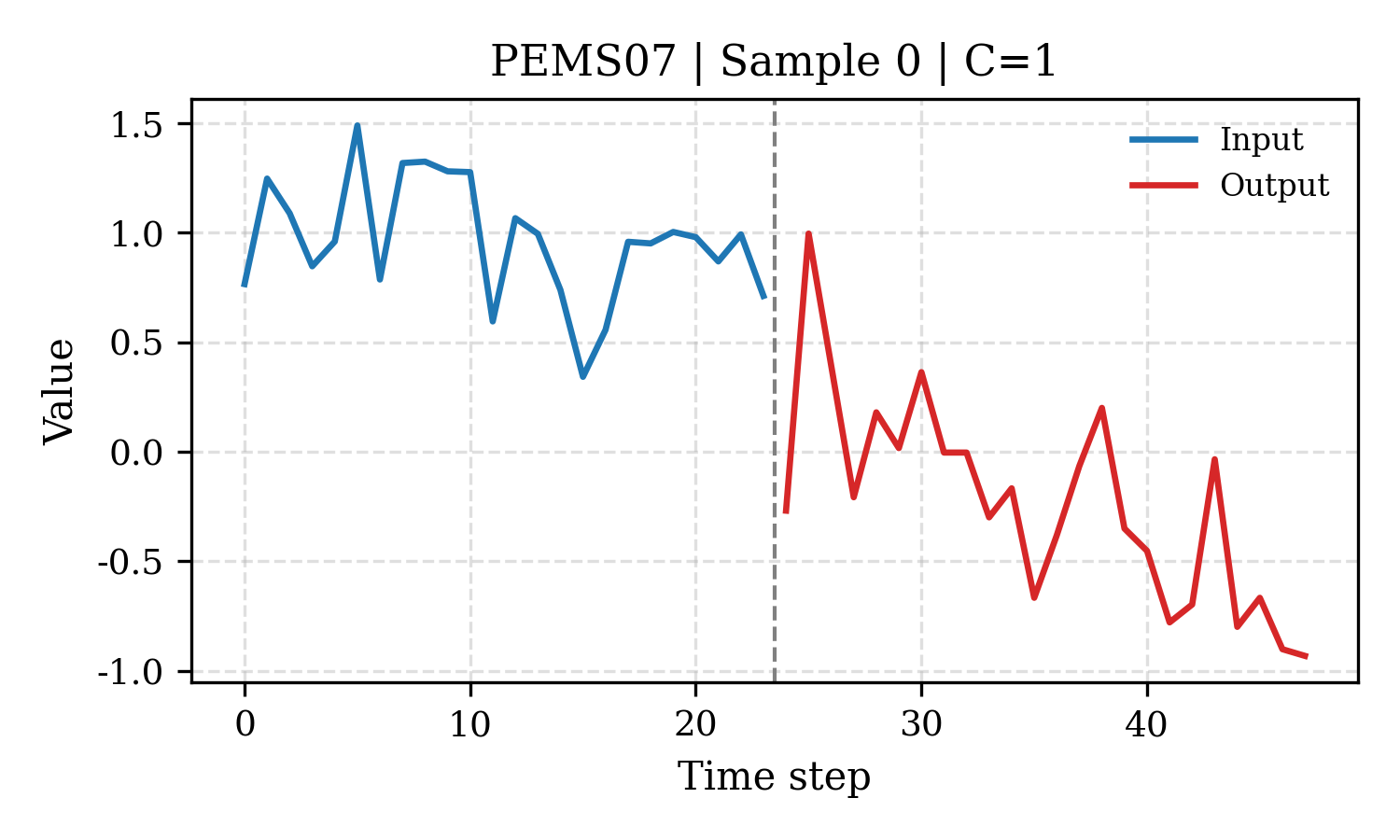} &
    \includegraphics[width=0.11\linewidth]{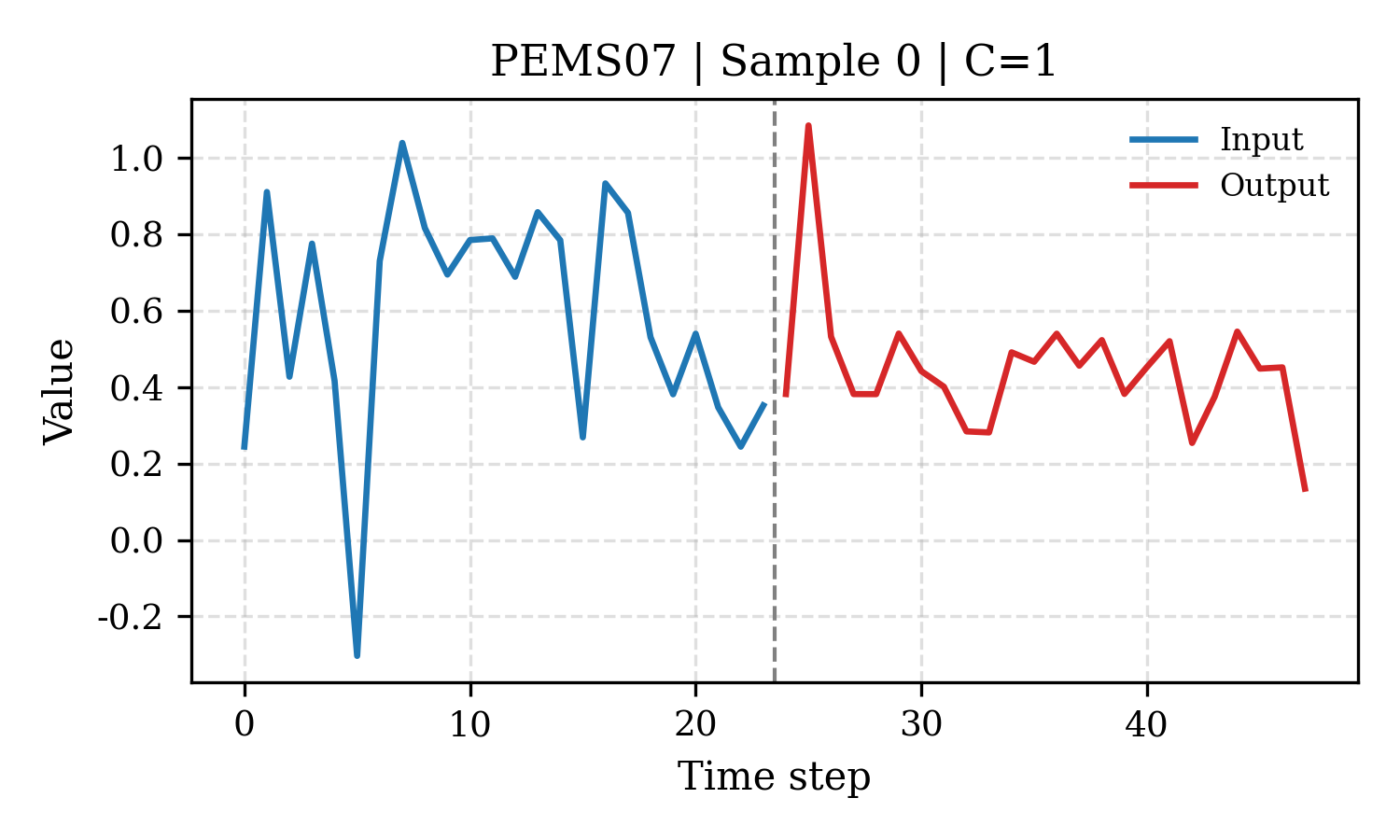} &
    \includegraphics[width=0.11\linewidth]{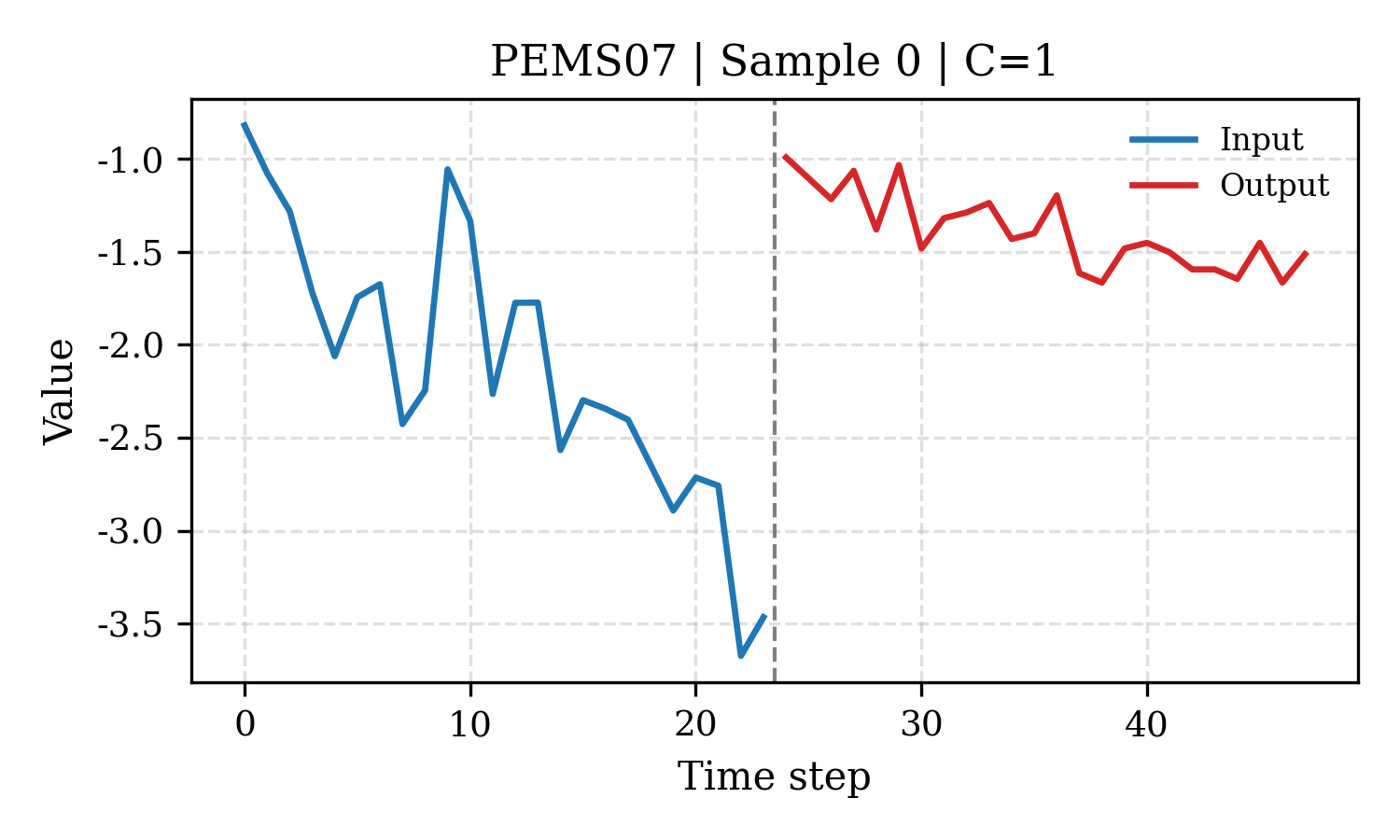} &
    \includegraphics[width=0.11\linewidth]{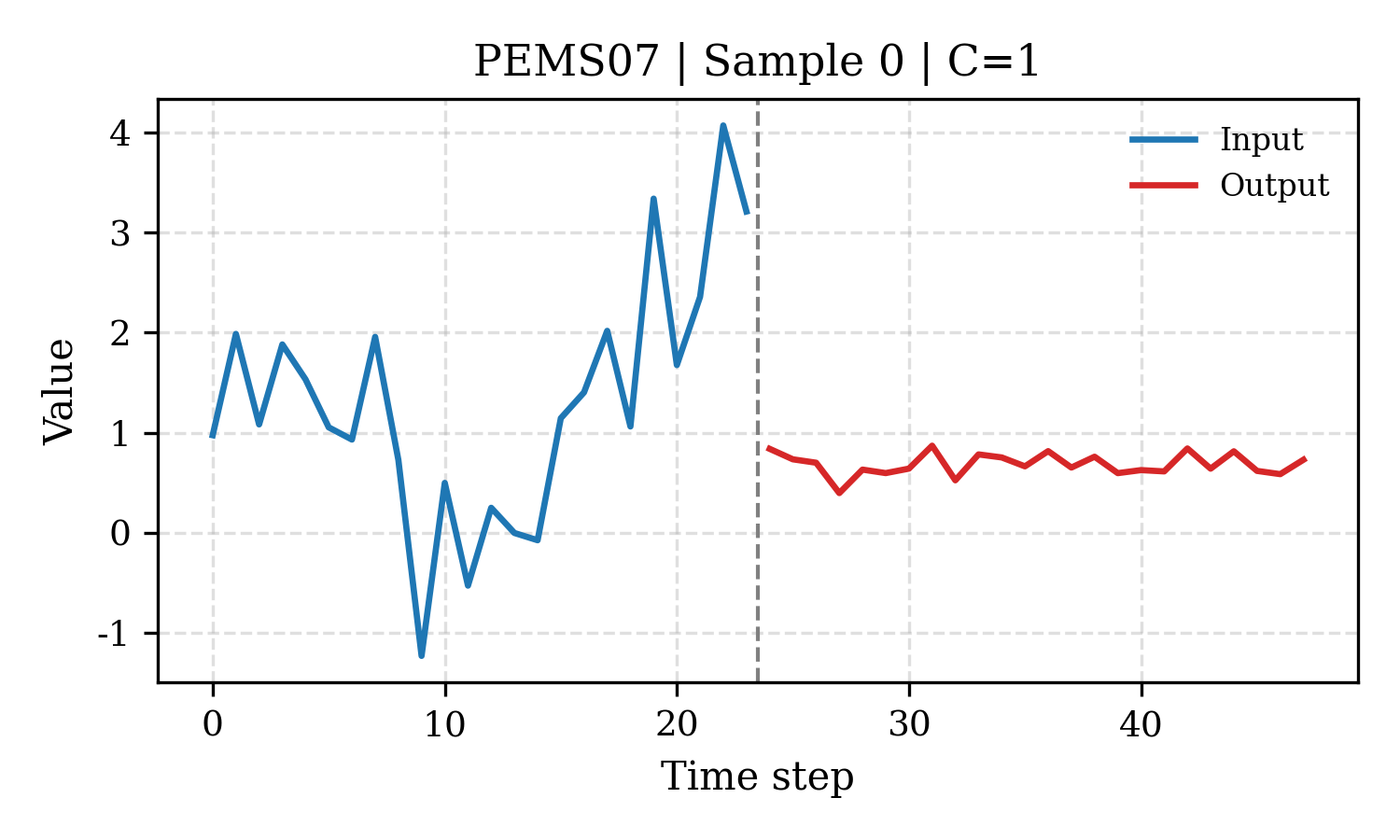} &
    \includegraphics[width=0.11\linewidth]{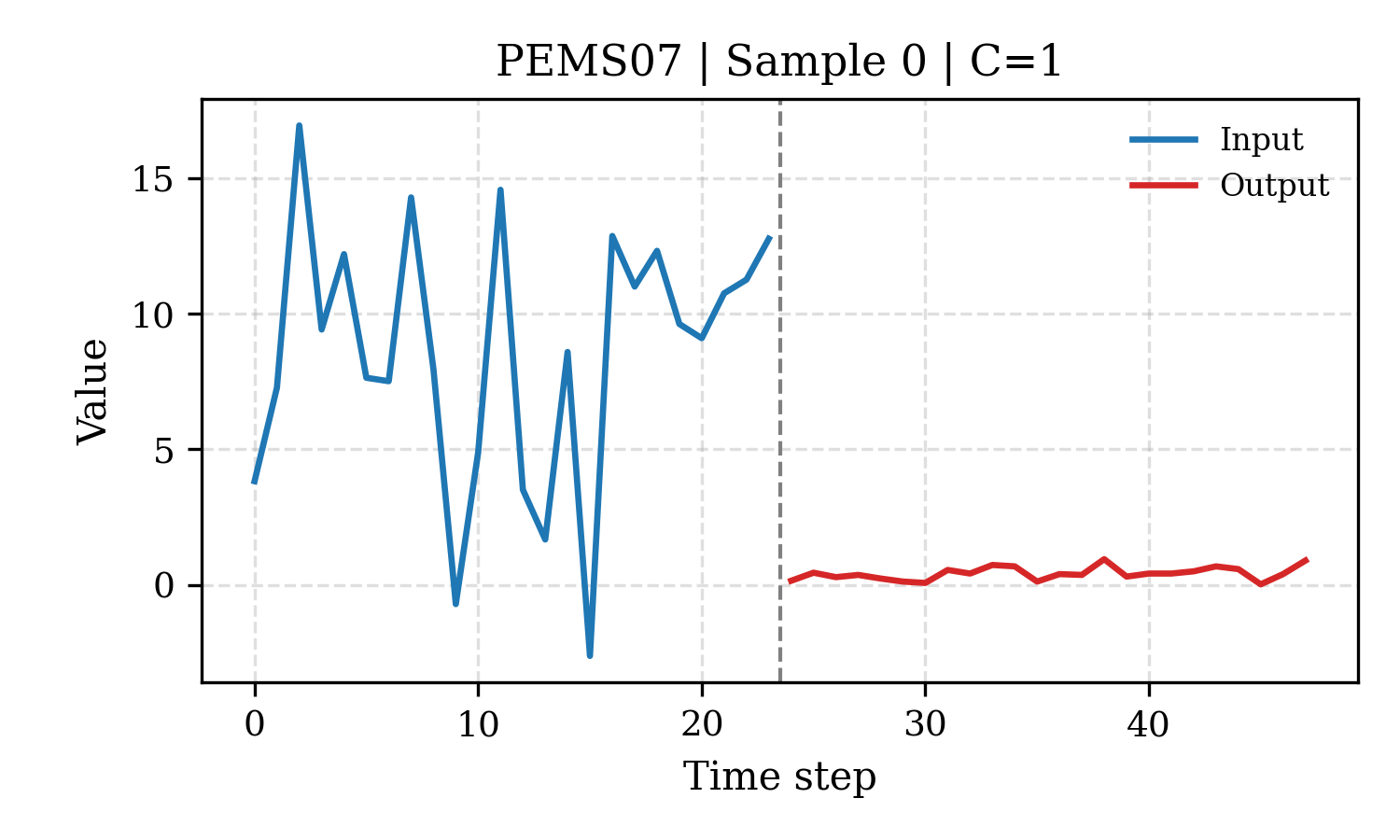} &
    \includegraphics[width=0.11\linewidth]{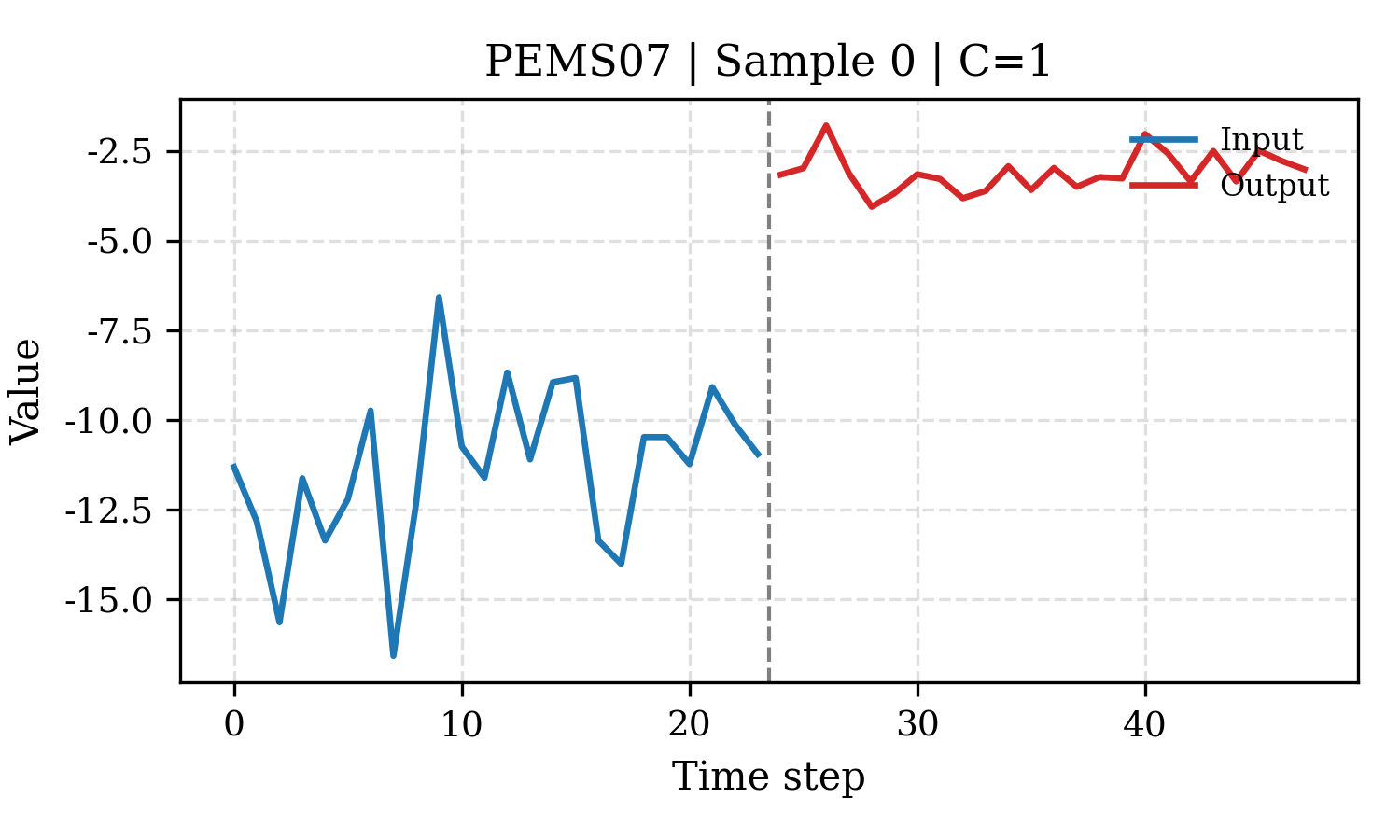} \\[0.3em]

    \rotatebox[origin=l]{90}{\textbf{PEMS08}} &
    \includegraphics[width=0.11\linewidth]{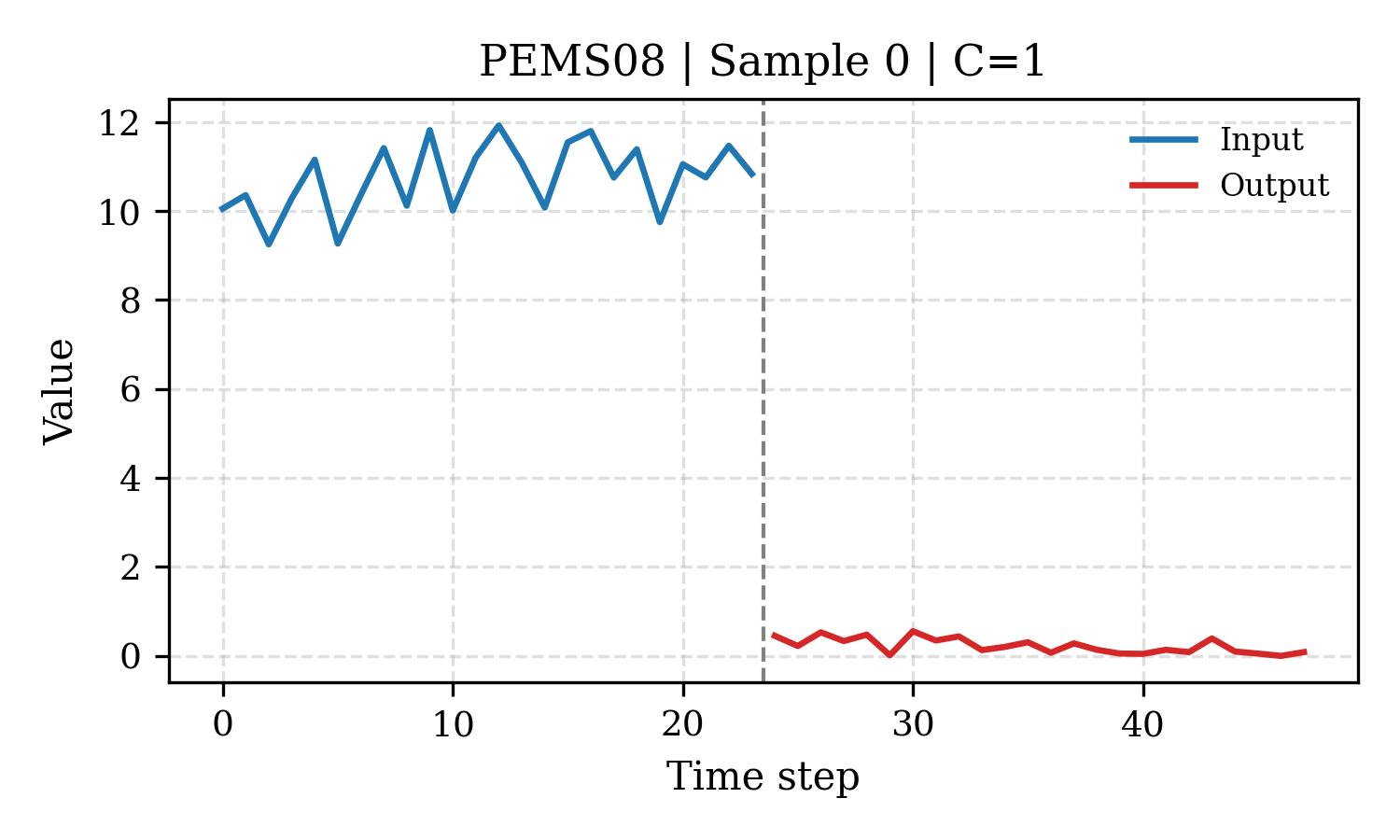} &
    \includegraphics[width=0.11\linewidth]{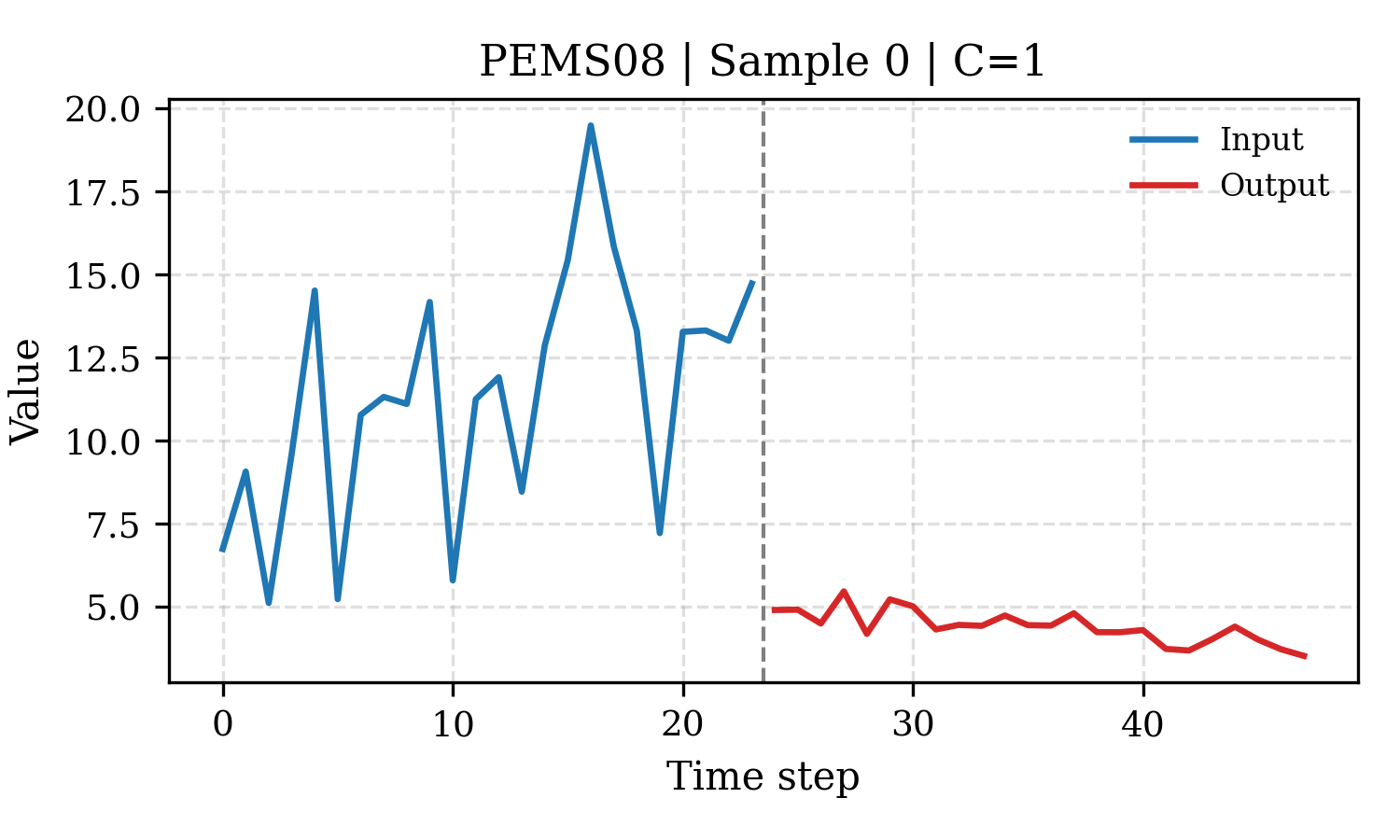} &
    \includegraphics[width=0.11\linewidth]{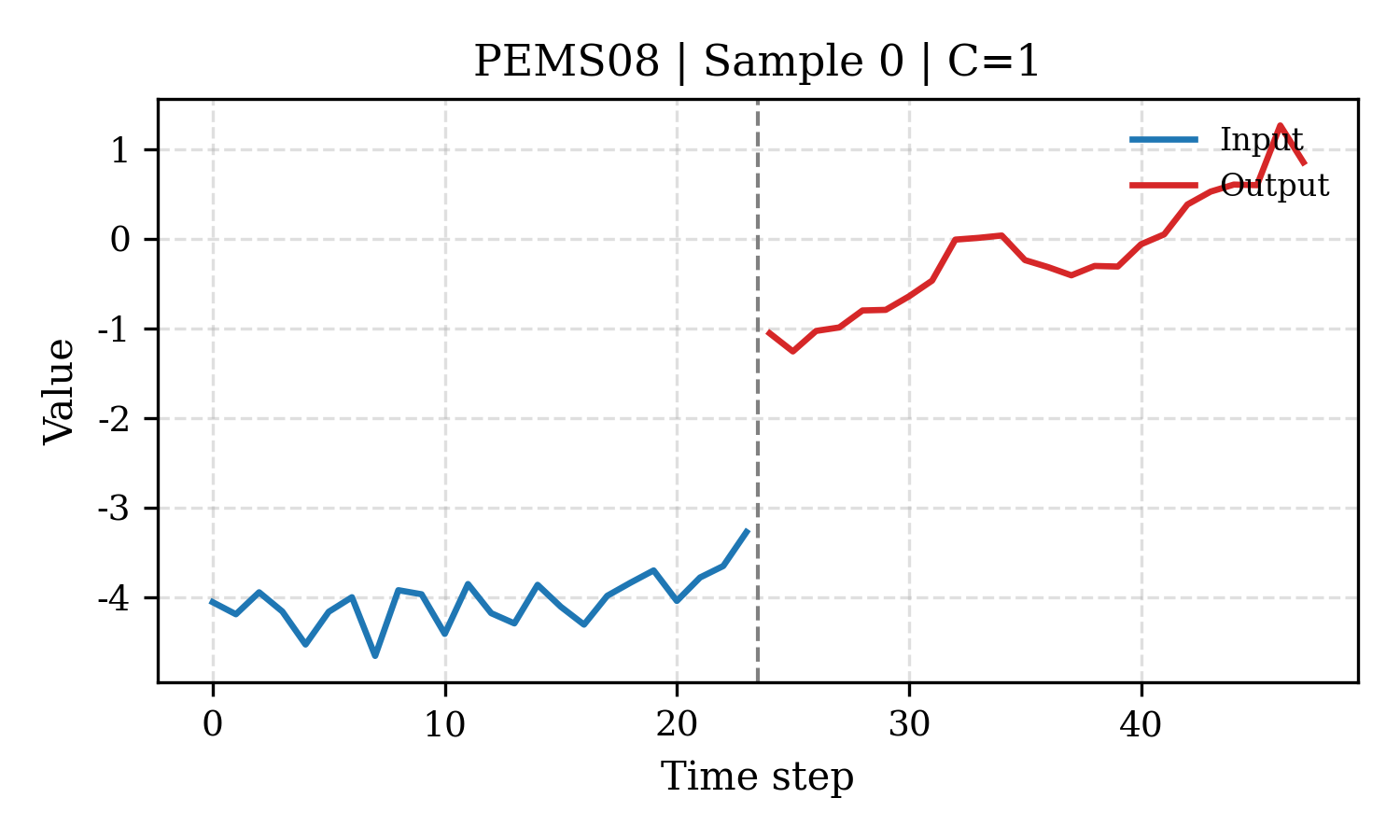} &
    \includegraphics[width=0.11\linewidth]{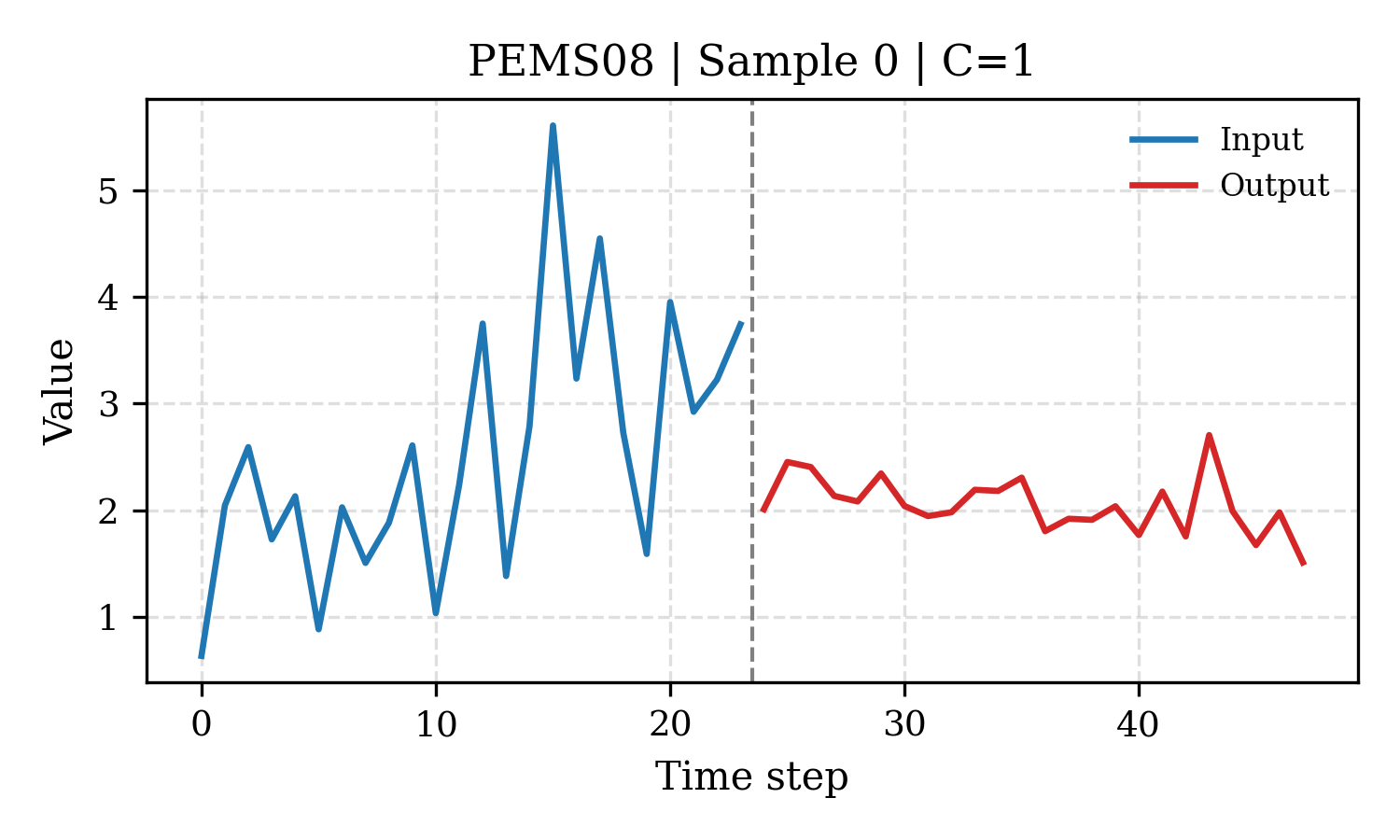} &
    \includegraphics[width=0.11\linewidth]{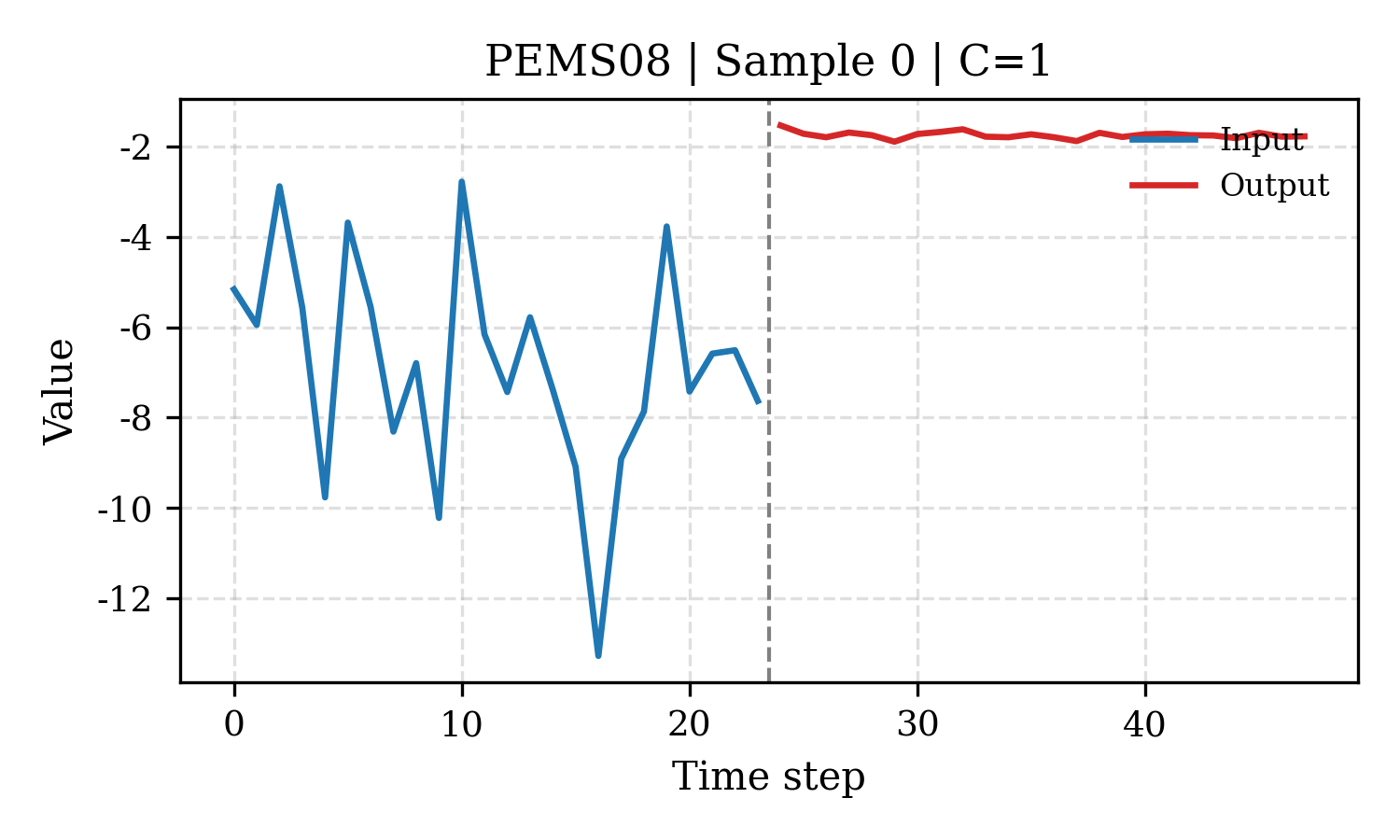} &
    \includegraphics[width=0.11\linewidth]{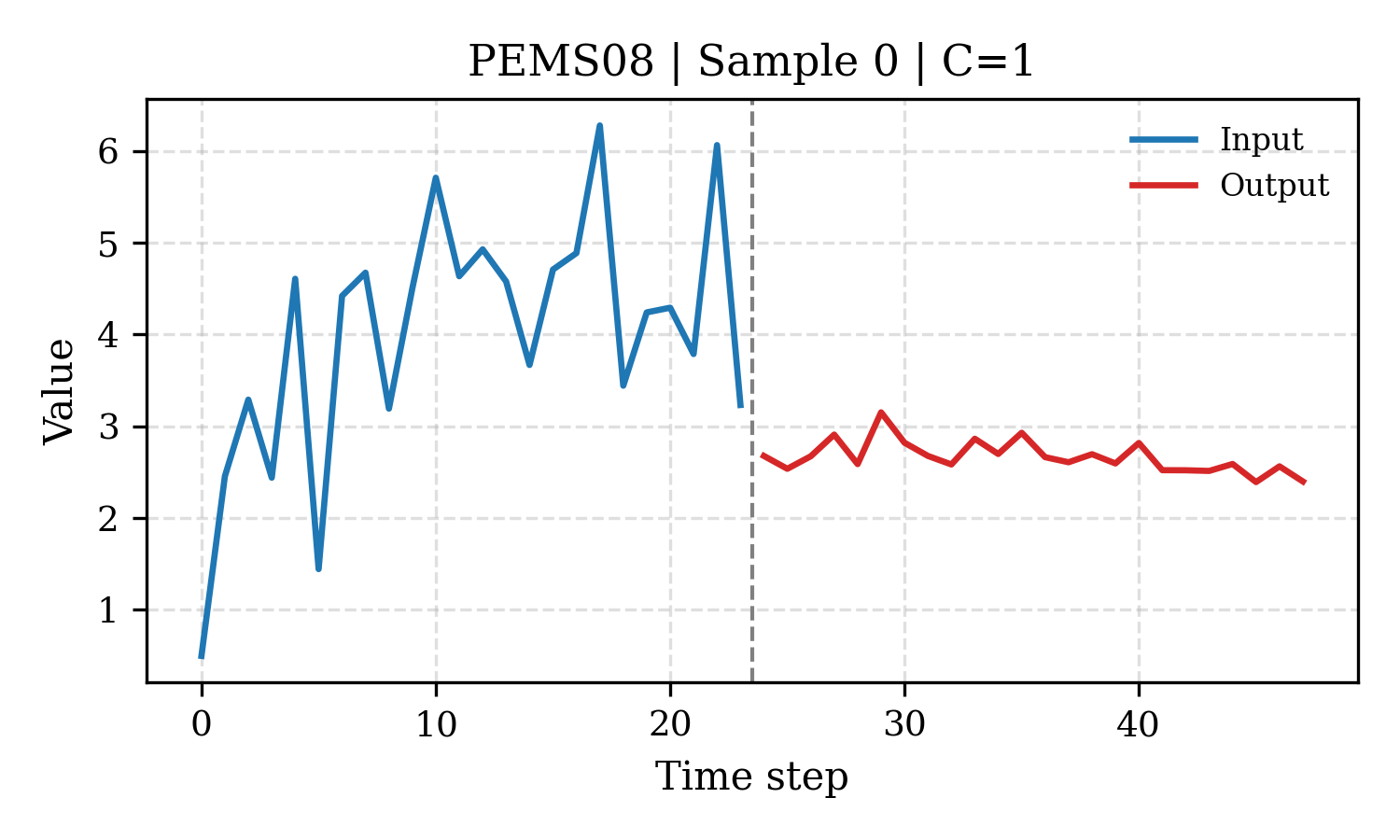} &
    \includegraphics[width=0.11\linewidth]{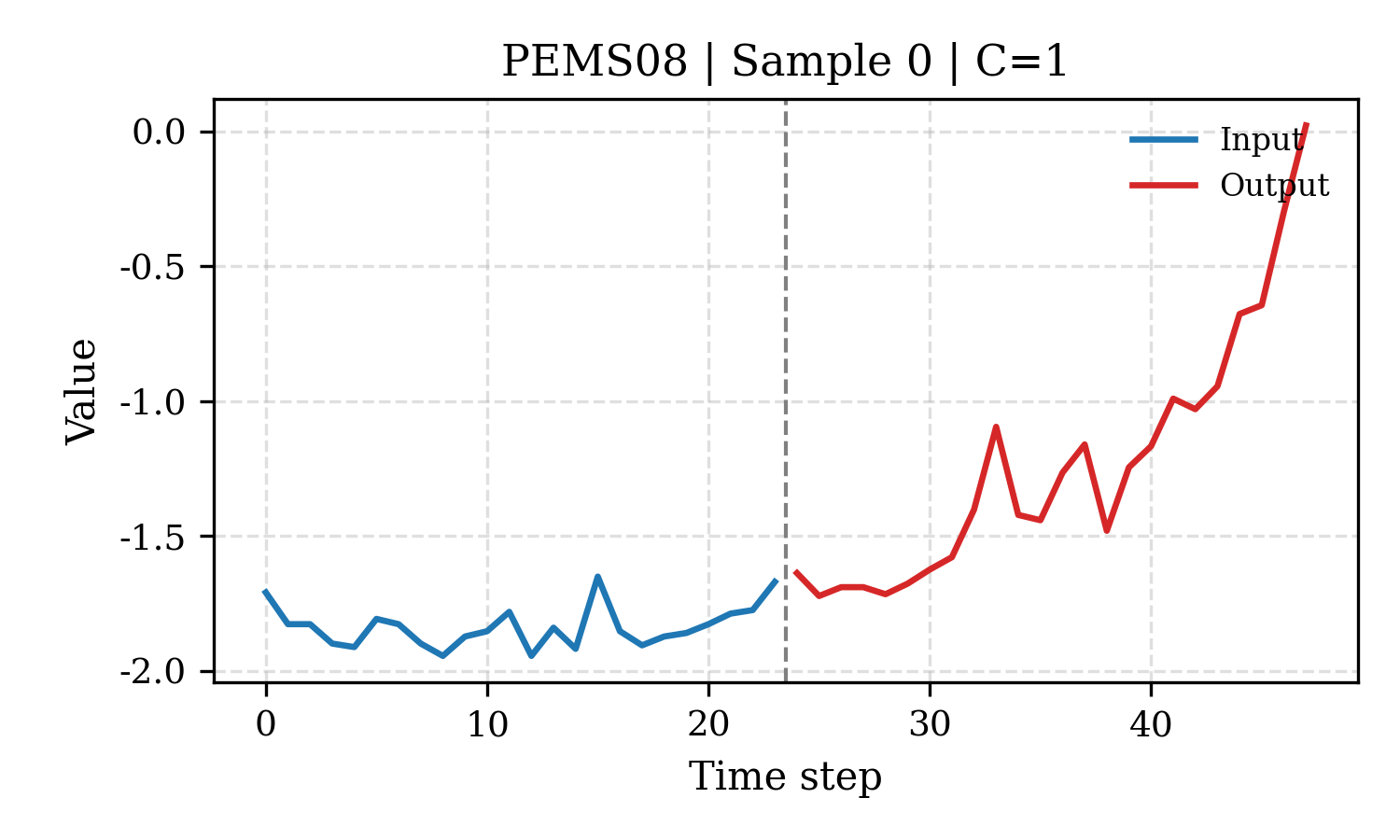} &
    \includegraphics[width=0.11\linewidth]{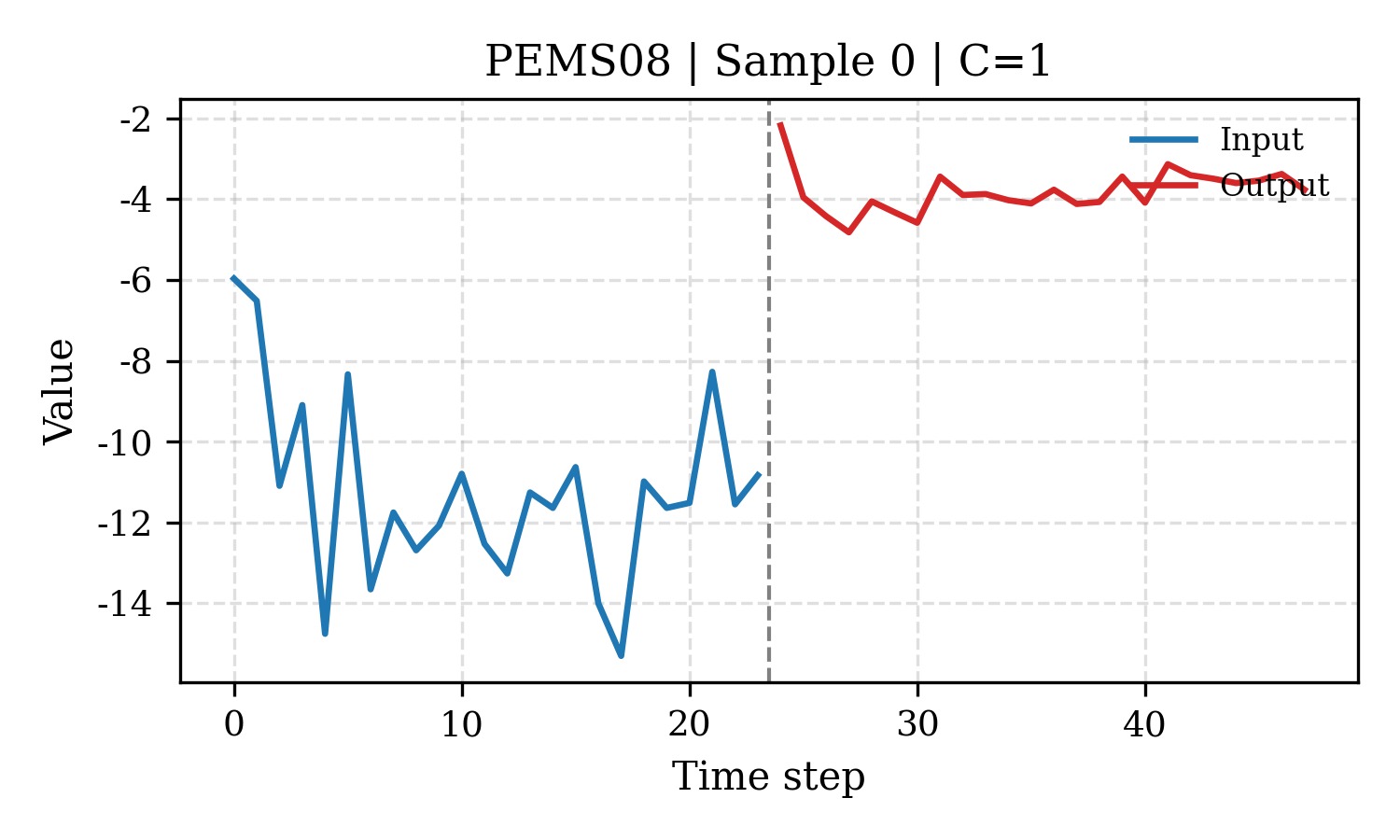} \\[0.3em]

    \rotatebox[origin=l]{90}{\textbf{Wike2000}} &
    \includegraphics[width=0.11\linewidth]{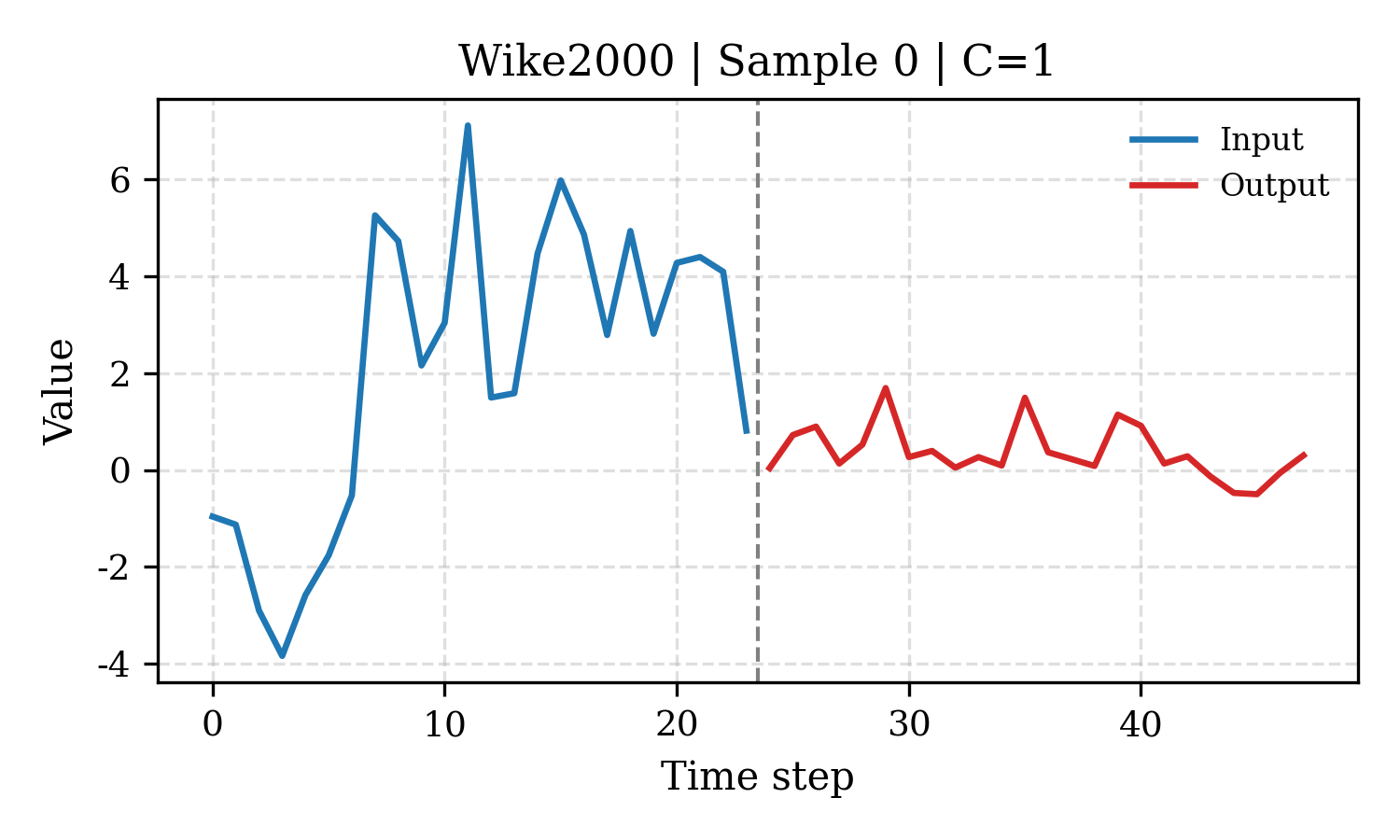} &
    \includegraphics[width=0.11\linewidth]{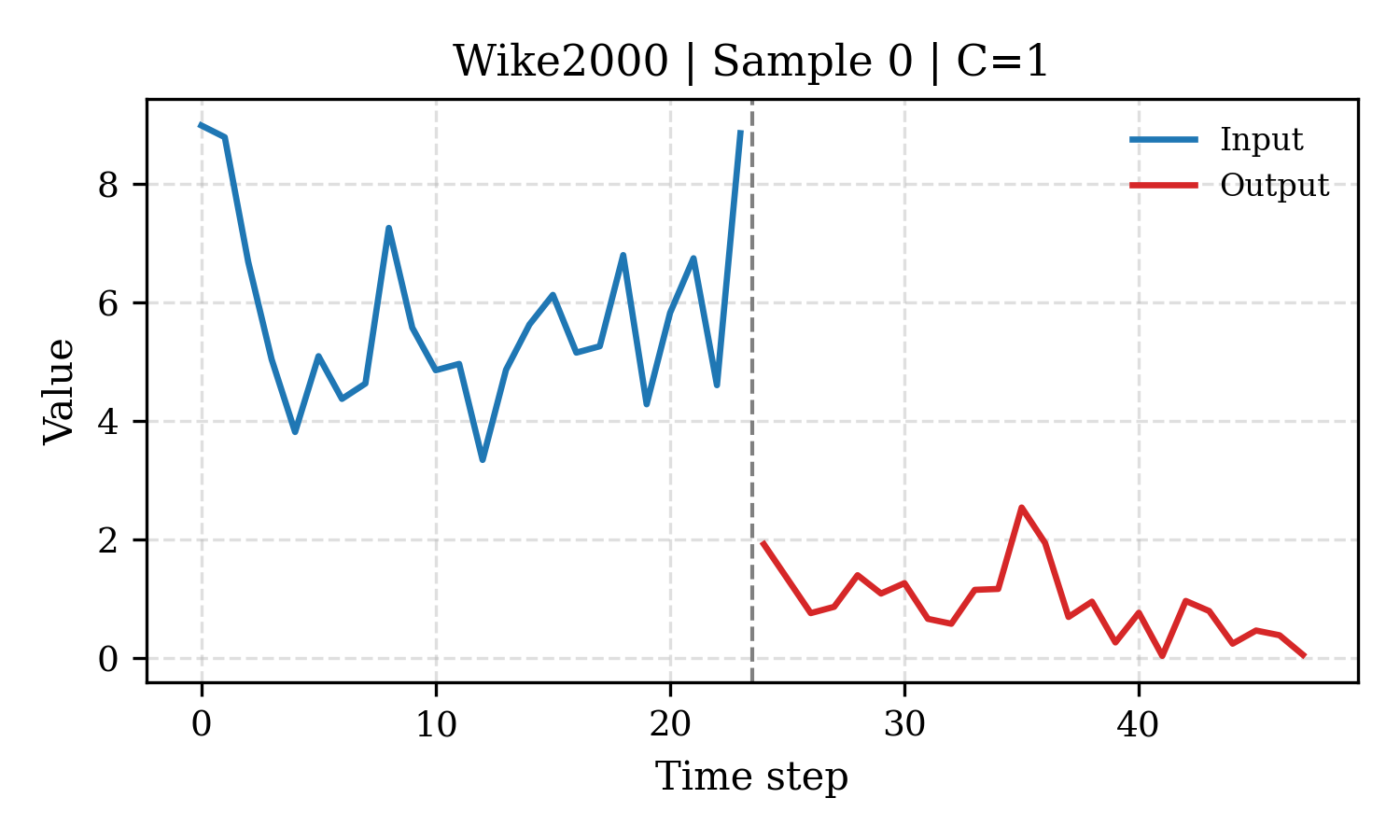} &
    \includegraphics[width=0.11\linewidth]{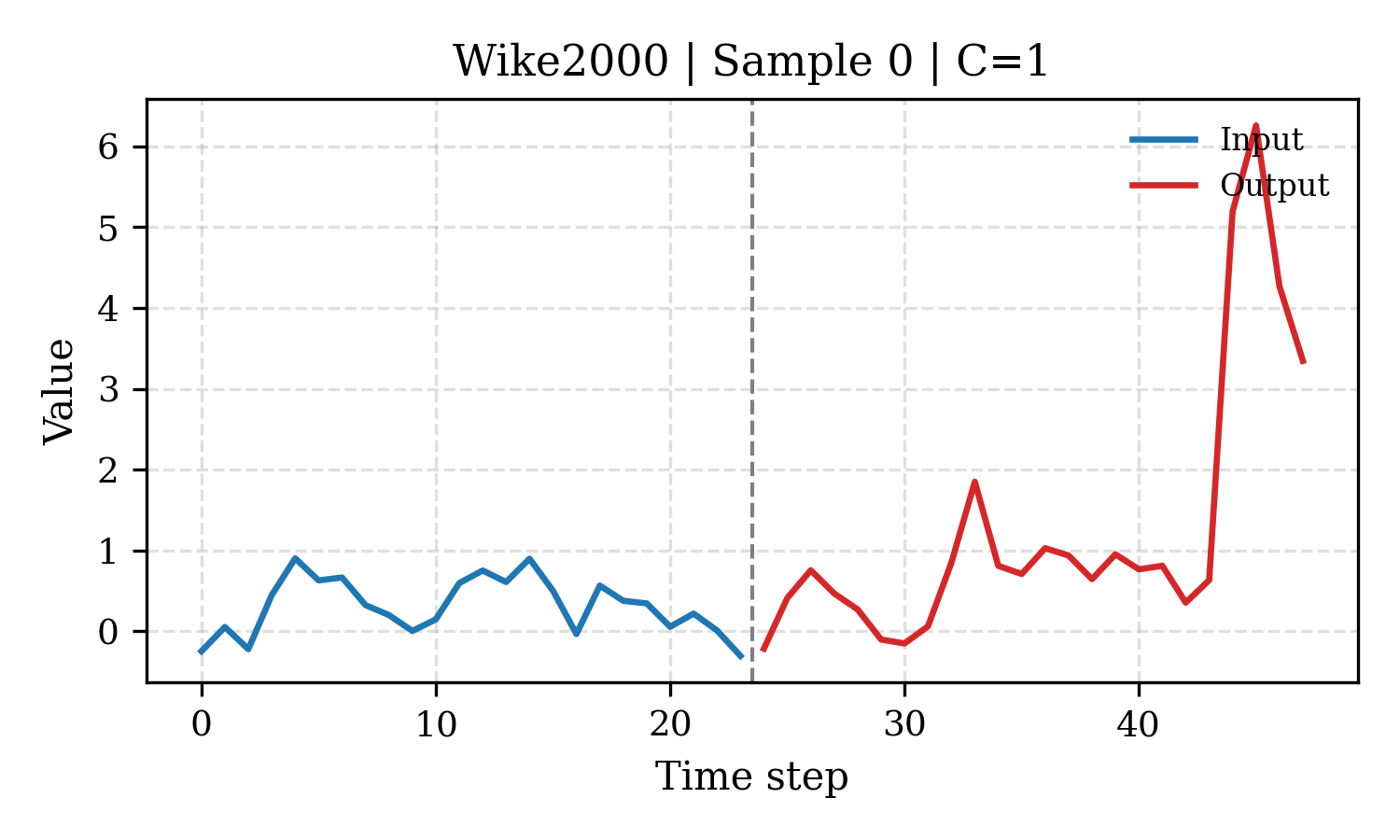} &
    \includegraphics[width=0.11\linewidth]{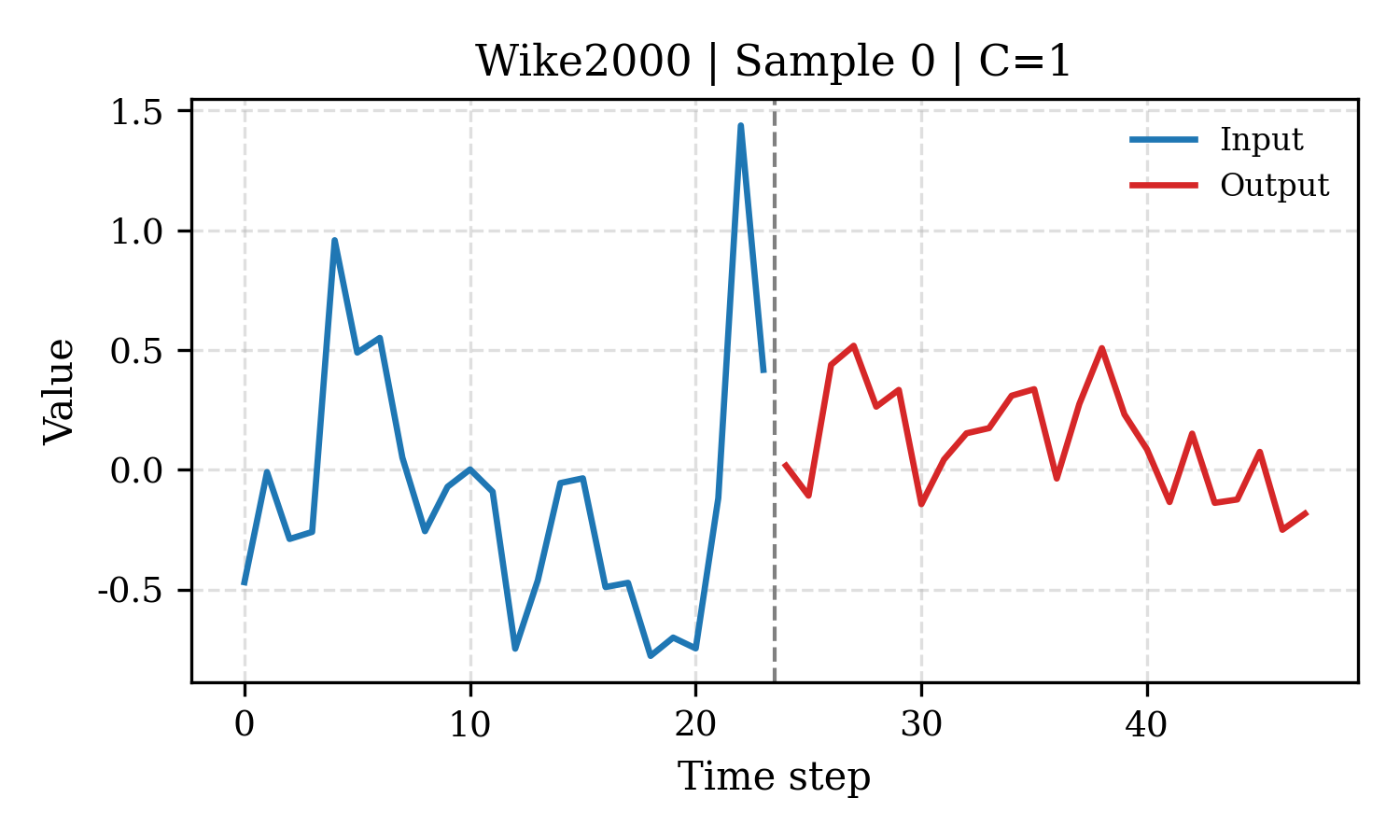} &
    \includegraphics[width=0.11\linewidth]{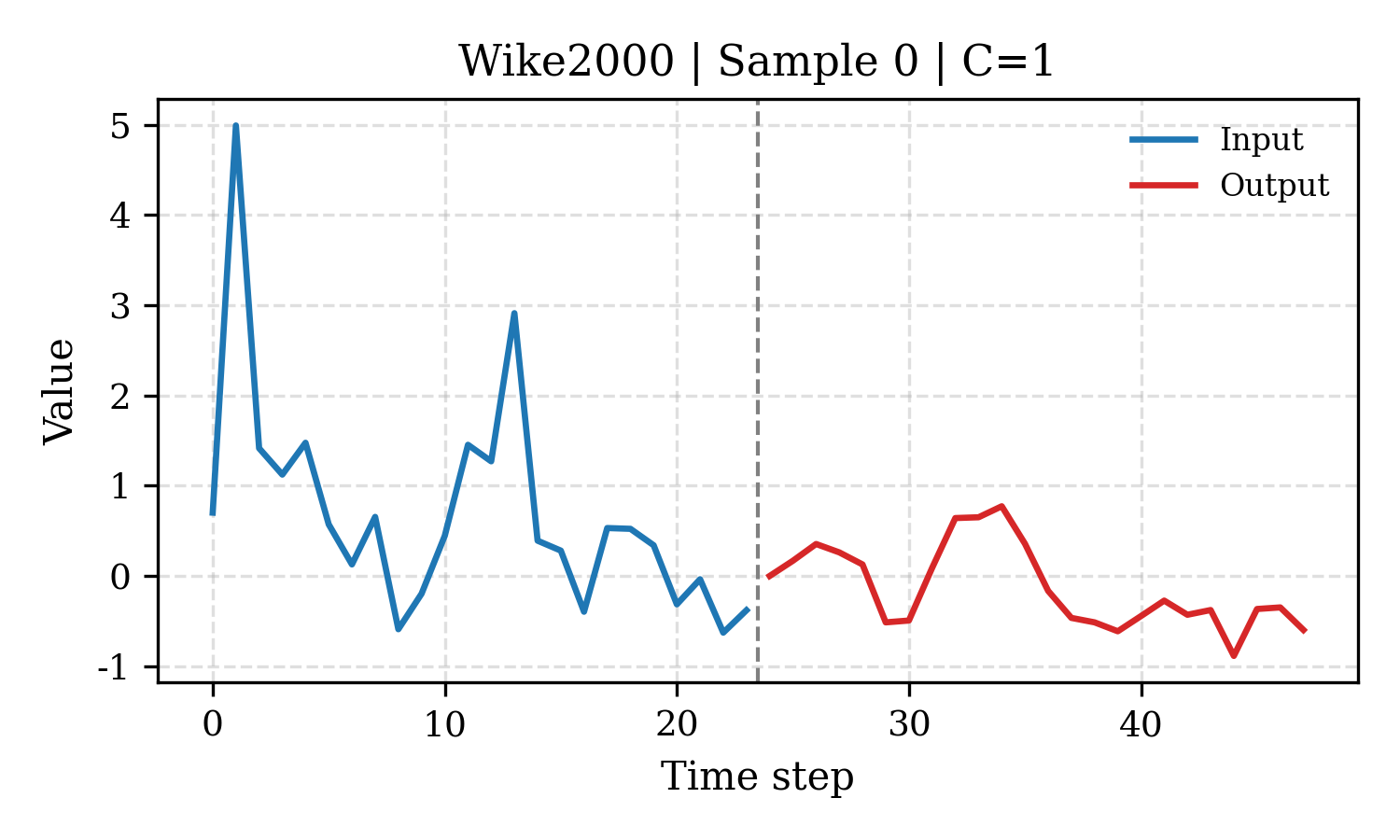} &
    \includegraphics[width=0.11\linewidth]{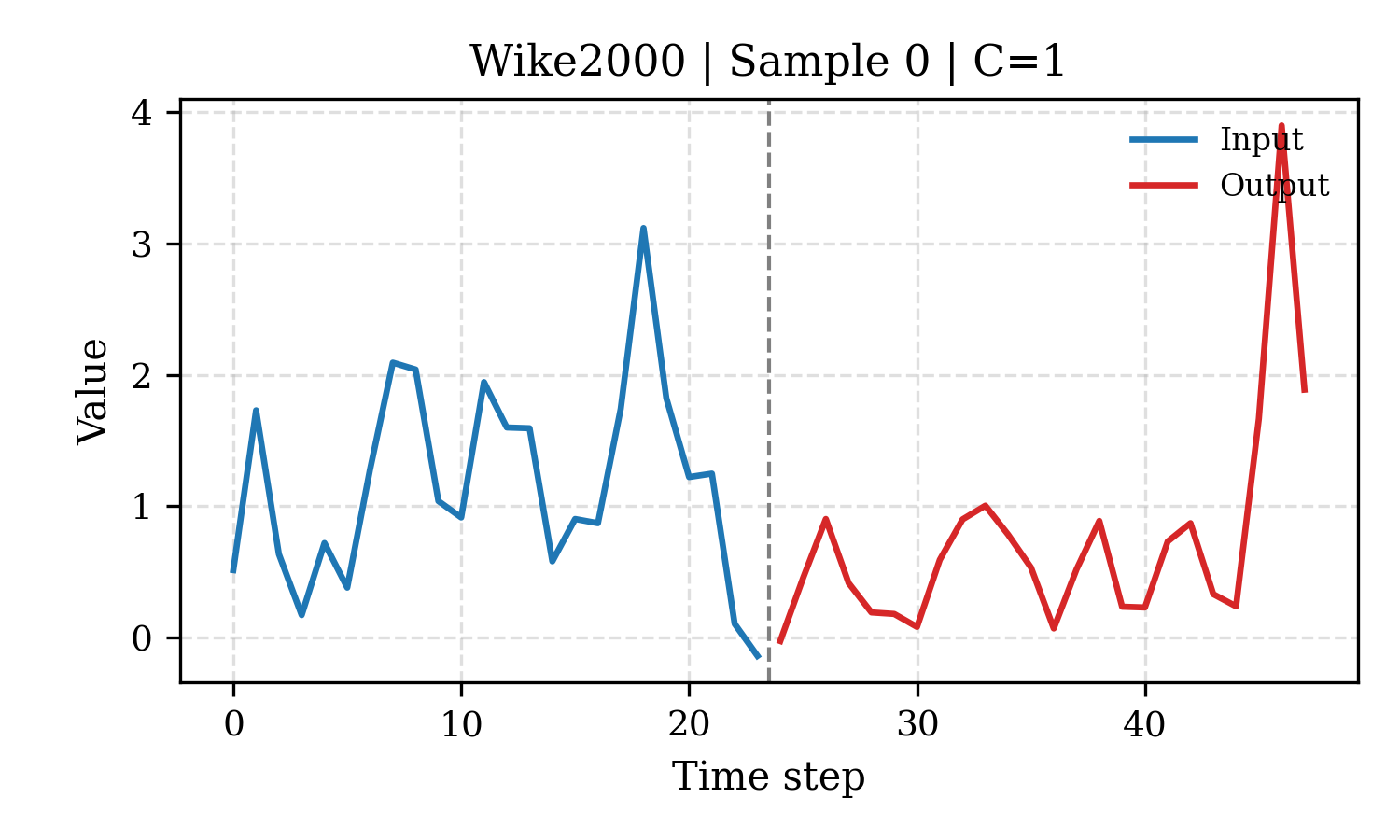} &
    \includegraphics[width=0.11\linewidth]{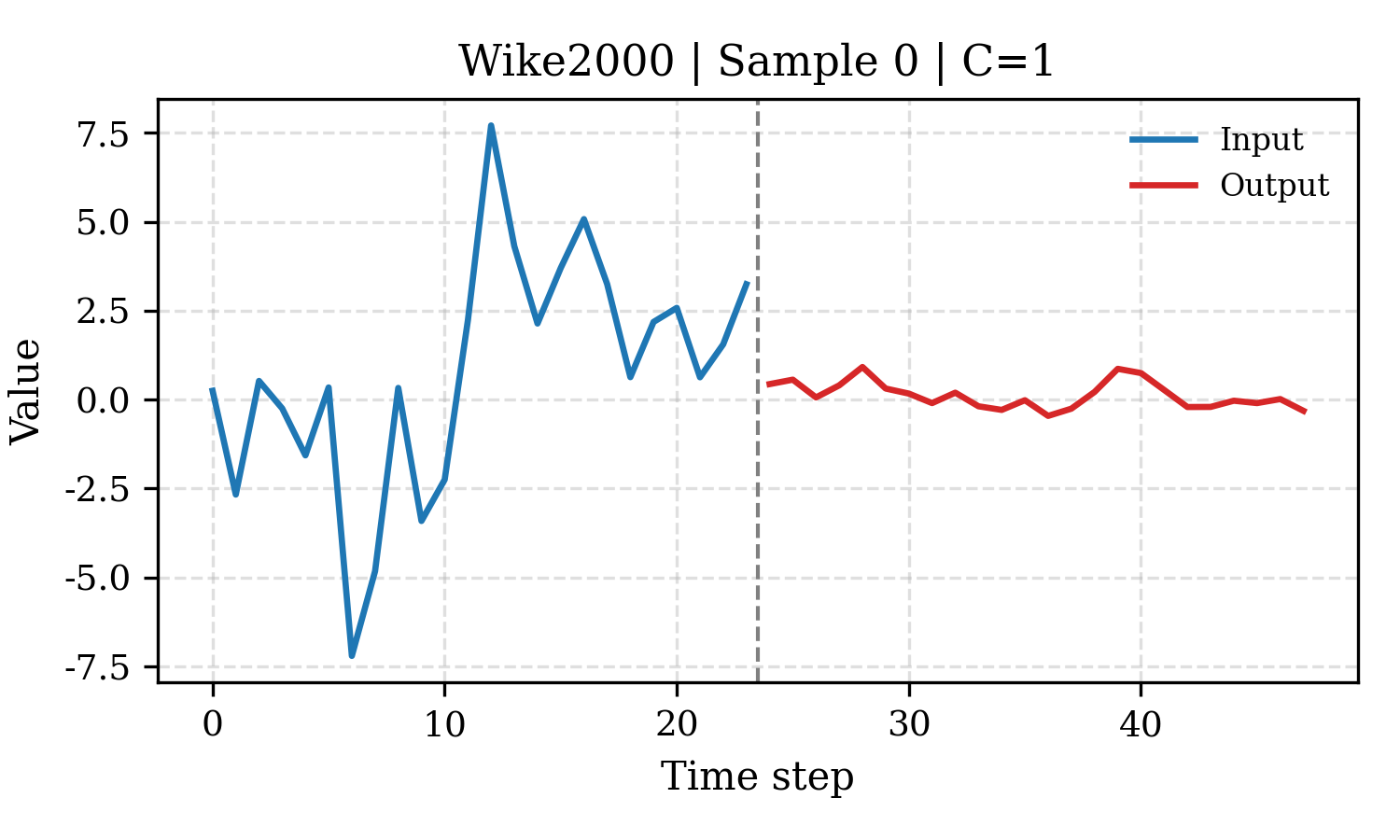} &
    \includegraphics[width=0.11\linewidth]{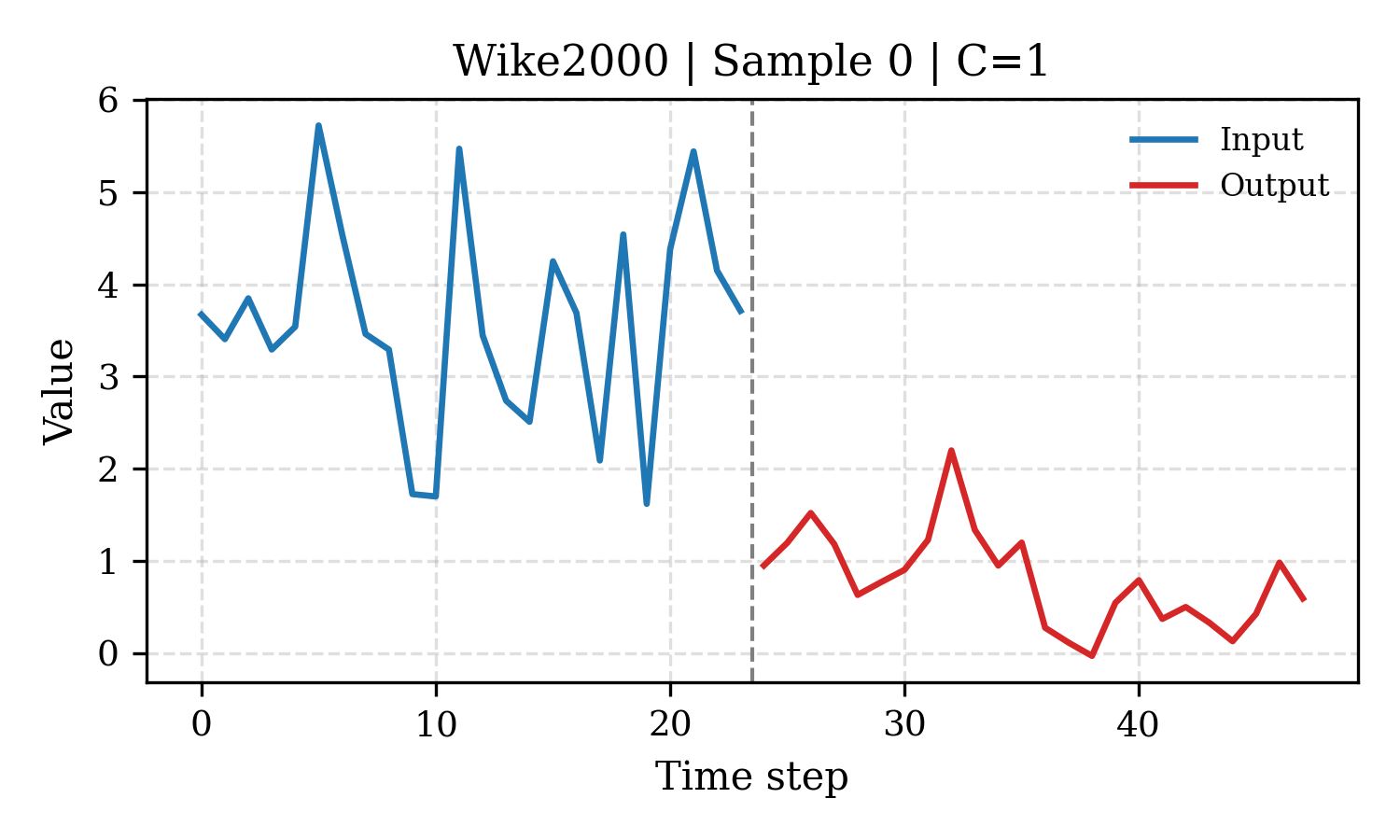} \\[0.3em]
    
  \end{tabular}

  \label{fig:vis_syn_full}
\end{figure*}

\section{Detailed Ablation Results}
\label{app:abla_big}

Table~\ref{tab::abla_big} presents the full ablation results of our method across multiple datasets and distillation baselines. 
Each block corresponds to a specific dataset (\textit{Weather}, \textit{Electricity}, \textit{Solar}, and \textit{Exchange}), with four synthetic sample settings ($3$, $5$, $10$, and $20$). 
For each distillation baseline, results are reported in terms of MSE and MAE, averaged over 5 independent runs. 

Our method consists of two plugin modules: the {Temporal–Frequency Alignment (TFA)} module and the {Inter-Sample Information Bottleneck (ISIB)} module. 
TFA enhances the temporal–frequency consistency between synthetic and real trajectories, while ISIB promotes inter-sample diversity by constraining redundancy in the condensed set. 
The results under ``+TFA’’ and ``+ISIB’’ indicate the effect of each module independently. 
Unless otherwise specified, the hyperparameters are fixed as $\alpha=0.8$ for TFA and $\lambda_{\mathrm{IS}}=0.6$ for ISIB to ensure consistency across all experiments.

The corresponding summarized averages in the main text (Table~\ref{tab::abla_final_avg_mae}) are obtained by averaging the MAE values across the four datasets and four sample settings. 
As discussed in Section~\ref{sec:ablation}, while EDF and MCT under $\alpha=0.8$ on \textit{Electricity} and \textit{Solar} are slightly higher than the baseline, adjusting $\alpha$ (see Appendix Tables~\ref{tab::ecl_fre} and~\ref{tab::solar_fre}) yields superior results beyond the baseline. 
Overall, the full results in Table~\ref{tab::abla_big} confirm that both TFA and ISIB consistently improve data efficiency across different trajectory-matching distillation frameworks. 

\section{Ablation under Different Architectures.}
\label{sec:arch}

We evaluate the DATM-based distillation on the ETTh1 dataset across diverse forecasting architectures. As shown in Table~\ref{tab:abla_arch}, our method consistently improves performance over the baseline for both Transformer-based and conventional models. Although the formulation in Eq.~\ref{eq:two-term} strictly holds under linear forecasting settings, the experimental results indicate that this distillation paradigm can be directly transferred to other forecasting architectures without modification.

\begin{table}[!ht]
\setlength{\tabcolsep}{6pt}
\small
\centering
\caption{
Distillation performance (MSE) across different forecasting architectures.  
}
\begin{tabular}{lccc}
\toprule
\textbf{Model} & \textbf{Whole Dataset} & \textbf{Baseline} & \textbf{+Ours} \\
\midrule
FreTS          & 0.347 & 0.515 & \textbf{0.488} \\
PatchTST       & 0.388 & 0.807 & \textbf{0.731} \\
xPatch         & 0.531 & 0.570 & \textbf{0.559} \\
SimpleTM       & 0.561 & 0.893 & \textbf{0.816} \\
iTransformer   & 0.577 & 0.875 & \textbf{0.812} \\
CNN            & 0.711 & 0.817 & \textbf{0.724} \\
MLP            & 0.727 & 0.825 & \textbf{0.661} \\
LSTM           & 0.890 & 0.899 & \textbf{0.891} \\
\bottomrule
\end{tabular}
\label{tab:abla_arch}
\end{table}

\section{Model Size and Storage Analysis}

Table~\ref{tab:buffer_size} summarizes the model buffer sizes (in MB) trained on the full original datasets across various forecasting architectures. 
These values reflect the actual storage footprint of each model after being trained on its corresponding real dataset, providing a reference for comparing the relative parameter and representation capacity among different architectures.
As shown, lightweight models such as \textit{DLinear} and \textit{MLP} occupy minimal space, whereas more complex architectures like \textit{FreTS} and large-scale CNN-based models exhibit substantially larger storage sizes, indicating higher model complexity and memory demand.
It is worth noting that models with identical architectures show consistent file sizes across datasets (e.g., the ETT family), since the storage size is primarily determined by the number and precision of learnable parameters rather than dataset-specific characteristics.

\begin{table*}[!ht]
\centering
\small
\setlength{\tabcolsep}{4pt}
\renewcommand{\arraystretch}{1.05}
\caption{Model buffer size (MB) across datasets.}
\begin{tabular}{lccccccccc}
\toprule
\textbf{Dataset} & \textbf{CNN} & \textbf{DLinear} & \textbf{FreTS} & \textbf{LSTM} & \textbf{MLP} & \textbf{PatchTST} & \textbf{SimpleTM} & \textbf{iTransformer} & \textbf{xPatch} \\
\midrule
ETTh1            & 3.05  & 0.45 & 265.71 & 3.65 & 1.06 & 5.90 & 5.01 & 4.90 & 3.71 \\
ETTh2            & 3.05  & 0.45 & 265.71 & 3.65 & 1.06 & 5.90 & 5.01 & 4.90 & 3.71 \\
ETTm1            & 3.05  & 0.45 & 265.71 & 3.65 & 1.06 & 5.90 & 5.01 & 4.90 & 3.71 \\
ETTm2            & 3.05  & 0.45 & 265.71 & 3.65 & 1.06 & 5.90 & 5.01 & 4.90 & 3.71 \\
Electricity      & 4077.25 & 0.45 & 265.71 & 3.65 & 1.06 & 5.90 & 5.78 & 4.90 & 3.91 \\
ExchangeRate     & 3.65  & 0.45 & 265.71 & 3.65 & 1.06 & 5.90 & 5.01 & 4.90 & 3.71 \\
Global\_Temp     & 39554.36 & 0.45 & 265.71 & 3.65 & 1.06 & 29.49 & 7.46 & 4.90 & 21.64 \\
ILI              & 3.05  & 0.45 & 265.71 & 3.65 & 1.06 & 5.90 & 5.01 & 4.90 & 3.71 \\
PEMS03           & 5070.97 & 0.45 & 265.71 & 3.65 & 1.06 & 5.90 & 5.86 & 4.90 & 3.93 \\
PEMS04           & 3729.48 & 0.45 & 265.71 & 3.65 & 1.06 & 5.90 & 5.74 & 4.90 & 3.90 \\
PEMS07           & 30840.60 & 0.45 & 265.71 & 3.65 & 1.06 & 5.90 & 7.16 & 4.90 & 4.26 \\
PEMS08           & 1144.54 & 0.45 & 265.71 & 3.65 & 1.06 & 5.90 & 5.40 & 4.90 & 3.81 \\
Solar            & 743.77 & 0.45 & 265.71 & 3.65 & 1.06 & 5.90 & 5.32 & 4.90 & 3.79 \\
Traffic          & 29391.21 & 2.25 & 265.71 & 3.65 & 5.30 & 29.49 & 35.52 & 24.50 & 21.24 \\
Weather          & 18.59 & 2.25 & 1328.53 & 18.24 & 5.30 & 29.49 & 25.14 & 4.90 & 3.72 \\
Wike2000         & 277.50 & 2.25 & 1328.53 & 18.24 & 5.30 & 29.49 & 49.66 & 4.90 & 4.95 \\
WindL1           & 4.32  & 0.45 & 265.71 & 3.65 & 1.06 & 5.90 & 5.01 & 4.90 & 3.71 \\
WindL2           & 4.32  & 0.45 & 265.71 & 3.65 & 1.06 & 5.90 & 5.01 & 4.90 & 3.71 \\
WindL3           & 4.32  & 0.45 & 265.71 & 3.65 & 1.06 & 5.90 & 5.01 & 4.90 & 3.71 \\
WindL4           & 4.32  & 0.45 & 265.71 & 3.65 & 1.06 & 5.90 & 5.01 & 4.90 & 3.71 \\
\bottomrule
\end{tabular}
\label{tab:buffer_size}
\end{table*}

\section{Runtime and Memory Overhead Analysis}
\label{app:runtime_memory}

To evaluate the computational efficiency of the proposed framework, 
we measure the wall-clock runtime and GPU memory consumption of each distillation framework
under both the baseline and the enhanced (\textit{ours}) settings across all datasets.
Table~\ref{tab:runtime_memory_total} reports the results in detail.

Overall, the proposed improvements incur only marginal computational overhead.
When aggregated over all datasets and frameworks, the total runtime of the \textit{ours} version
is \textbf{2.59\% faster} than the baseline on average, while the overall GPU memory usage increases by only \textbf{2.49\%}.
This demonstrates that our method introduces negligible additional cost in training and inference,
yet provides consistent performance gains reported in the main experiments. The acceleration mainly stems from faster convergence, as the introduced frequency-domain alignment stabilizes optimization and the diversity regularization provides more informative synthetic samples, enabling efficient learning with minimal additional cost.

\begin{table}[!ht] 

	\setlength{\tabcolsep}{2.5pt}

	\scriptsize

	\centering
\caption{
Sensitivity analysis of model performance to the number of synthetic samples. The observation that 3–5 synthetic sequences suffice to drive meaningful performance suggests that existing TSF architectures may not fully exploit the whole information present in real data.
}
	\begin{threeparttable}

		\begin{tabular}{c|c|c c|c c}

			\toprule

			\multicolumn{2}{c}{\multirow{2}{*}{\scalebox{1.1}{Models}}}& \multicolumn{2}{c}{DATM+Ours} & \multicolumn{2}{c}{FTD+Ours} \\

			\multicolumn{2}{c}{} & \multicolumn{2}{c}{\scalebox{0.8}{\citeyearpar{guo2024datm}}} & \multicolumn{2}{c}{\scalebox{0.8}{\citeyearpar{du2023ftd}}}\\

			\cmidrule(lr){3-4} \cmidrule(lr){5-6} 

			\multicolumn{2}{c}{Metric}& MSE & MAE & MSE & MAE \\

\toprule

\multirow{10}{*}{\rotatebox[origin=c]{90}{Electricity}}

& 2 & 0.72{\tiny$\pm$0.01} & 0.69{\tiny$\pm$0.01} & 0.73{\tiny$\pm$0.01} & 0.69{\tiny$\pm$0.01}\\
& 3 & 0.72{\tiny$\pm$0.02} & 0.69{\tiny$\pm$0.01} & 0.72{\tiny$\pm$0.02} & 0.68{\tiny$\pm$0.01}\\
& 5 & 0.69{\tiny$\pm$0.02} & 0.68{\tiny$\pm$0.01} & 0.70{\tiny$\pm$0.02} & 0.68{\tiny$\pm$0.01}\\
& 10 & 0.71{\tiny$\pm$0.01} & 0.69{\tiny$\pm$0.01} & 0.70{\tiny$\pm$0.03} & 0.68{\tiny$\pm$0.02}\\
& 20 & 0.72{\tiny$\pm$0.02} & 0.69{\tiny$\pm$0.01} & 0.73{\tiny$\pm$0.02} & 0.69{\tiny$\pm$0.01}\\
& 30 & 0.74{\tiny$\pm$0.03} & 0.70{\tiny$\pm$0.01} & 0.74{\tiny$\pm$0.04} & 0.70{\tiny$\pm$0.02}\\
& 40 & 0.76{\tiny$\pm$0.03} & 0.71{\tiny$\pm$0.02} & 0.75{\tiny$\pm$0.03} & 0.70{\tiny$\pm$0.02}\\
& 50 & 0.76{\tiny$\pm$0.03} & 0.71{\tiny$\pm$0.01} & 0.75{\tiny$\pm$0.02} & 0.70{\tiny$\pm$0.01}\\
& 80 & 0.80{\tiny$\pm$0.03} & 0.72{\tiny$\pm$0.01} & 0.76{\tiny$\pm$0.02} & 0.70{\tiny$\pm$0.01}\\
& 100 & 0.78{\tiny$\pm$0.02} & 0.72{\tiny$\pm$0.01} & 0.77{\tiny$\pm$0.06} & 0.71{\tiny$\pm$0.03}\\

\midrule
\multirow{10}{*}{\rotatebox[origin=c]{90}{ExchangeRate}}

& 2 & 0.55{\tiny$\pm$0.02} & 0.62{\tiny$\pm$0.01} & 0.47{\tiny$\pm$0.03} & 0.57{\tiny$\pm$0.02}\\
& 3 & 0.54{\tiny$\pm$0.05} & 0.61{\tiny$\pm$0.03} & 0.44{\tiny$\pm$0.06} & 0.54{\tiny$\pm$0.04}\\
& 5 & 0.39{\tiny$\pm$0.03} & 0.52{\tiny$\pm$0.02} & 0.27{\tiny$\pm$0.03} & 0.42{\tiny$\pm$0.02}\\
& 10 & 0.33{\tiny$\pm$0.03} & 0.47{\tiny$\pm$0.02} & 0.32{\tiny$\pm$0.01} & 0.47{\tiny$\pm$0.01}\\
& 20 & 0.43{\tiny$\pm$0.04} & 0.55{\tiny$\pm$0.02} & 0.44{\tiny$\pm$0.03} & 0.55{\tiny$\pm$0.02}\\
& 30 & 0.46{\tiny$\pm$0.03} & 0.56{\tiny$\pm$0.02} & 0.48{\tiny$\pm$0.03} & 0.57{\tiny$\pm$0.02}\\
& 40 & 0.47{\tiny$\pm$0.02} & 0.57{\tiny$\pm$0.02} & 0.41{\tiny$\pm$0.05} & 0.53{\tiny$\pm$0.03}\\
& 50 & 0.46{\tiny$\pm$0.03} & 0.56{\tiny$\pm$0.02} & 0.44{\tiny$\pm$0.02} & 0.55{\tiny$\pm$0.01}\\
& 80 & 0.46{\tiny$\pm$0.04} & 0.56{\tiny$\pm$0.03} & 0.40{\tiny$\pm$0.03} & 0.53{\tiny$\pm$0.02}\\
& 100 & 0.48{\tiny$\pm$0.04} & 0.58{\tiny$\pm$0.02} & 0.48{\tiny$\pm$0.03} & 0.58{\tiny$\pm$0.02}\\

\midrule
\multirow{10}{*}{\rotatebox[origin=c]{90}{Solar}}

& 2 & 0.65{\tiny$\pm$0.01} & 0.66{\tiny$\pm$0.01} & 0.62{\tiny$\pm$0.01} & 0.65{\tiny$\pm$0.01}\\
& 3 & 0.57{\tiny$\pm$0.01} & 0.61{\tiny$\pm$0.01} & 0.61{\tiny$\pm$0.00} & 0.64{\tiny$\pm$0.01}\\
& 5 & 0.52{\tiny$\pm$0.01} & 0.58{\tiny$\pm$0.01} & 0.58{\tiny$\pm$0.01} & 0.62{\tiny$\pm$0.02}\\
& 10 & 0.56{\tiny$\pm$0.01} & 0.61{\tiny$\pm$0.02} & 0.56{\tiny$\pm$0.01} & 0.60{\tiny$\pm$0.02}\\
& 20 & 0.56{\tiny$\pm$0.01} & 0.60{\tiny$\pm$0.01} & 0.60{\tiny$\pm$0.01} & 0.63{\tiny$\pm$0.02}\\
& 30 & 0.60{\tiny$\pm$0.01} & 0.62{\tiny$\pm$0.02} & 0.59{\tiny$\pm$0.01} & 0.62{\tiny$\pm$0.02}\\
& 40 & 0.58{\tiny$\pm$0.01} & 0.63{\tiny$\pm$0.01} & 0.60{\tiny$\pm$0.01} & 0.63{\tiny$\pm$0.01}\\
& 50 & 0.59{\tiny$\pm$0.01} & 0.62{\tiny$\pm$0.01} & 0.60{\tiny$\pm$0.01} & 0.63{\tiny$\pm$0.01}\\
& 80 & 0.60{\tiny$\pm$0.01} & 0.63{\tiny$\pm$0.01}  & 0.69{\tiny$\pm$0.02} & 0.67{\tiny$\pm$0.01}\\
& 100 & 0.65{\tiny$\pm$0.01} & 0.66{\tiny$\pm$0.01} & 0.66{\tiny$\pm$0.01} & 0.67{\tiny$\pm$0.01}\\

\midrule
\multirow{10}{*}{\rotatebox[origin=c]{90}{Weather}}

& 2 & 0.36{\tiny$\pm$0.01} & 0.41{\tiny$\pm$0.01} & 0.44{\tiny$\pm$0.01} & 0.47{\tiny$\pm$0.01}\\
& 3 & 0.28{\tiny$\pm$0.01} & 0.31{\tiny$\pm$0.01} & 0.38{\tiny$\pm$0.01} & 0.43{\tiny$\pm$0.01}\\
& 5 & 0.32{\tiny$\pm$0.01} & 0.37{\tiny$\pm$0.01} & 0.42{\tiny$\pm$0.01} & 0.45{\tiny$\pm$0.01}\\
& 10 & 0.31{\tiny$\pm$0.00} & 0.36{\tiny$\pm$0.01} & 0.40{\tiny$\pm$0.01} & 0.44{\tiny$\pm$0.01}\\
& 20 & 0.31{\tiny$\pm$0.01} & 0.36{\tiny$\pm$0.02} & 0.40{\tiny$\pm$0.01} & 0.44{\tiny$\pm$0.01}\\
& 30 & 0.31{\tiny$\pm$0.01} & 0.37{\tiny$\pm$0.01} & 0.39{\tiny$\pm$0.01} & 0.43{\tiny$\pm$0.01}\\
& 40 & 0.32{\tiny$\pm$0.01} & 0.36{\tiny$\pm$0.01} & 0.42{\tiny$\pm$0.01} & 0.45{\tiny$\pm$0.01}\\
& 50 & 0.32{\tiny$\pm$0.01} & 0.37{\tiny$\pm$0.01} & 0.66{\tiny$\pm$0.06} & 0.56{\tiny$\pm$0.03}\\
& 80 & 0.34{\tiny$\pm$0.02} & 0.38{\tiny$\pm$0.02} & 0.66{\tiny$\pm$0.06} & 0.56{\tiny$\pm$0.03}\\
& 100 & 0.36{\tiny$\pm$0.00} & 0.40{\tiny$\pm$0.01} & 0.66{\tiny$\pm$0.05} & 0.56{\tiny$\pm$0.02}\\

			\toprule

		\end{tabular}
	\end{threeparttable}

	\label{tab::sample_num}

\end{table}

\section{Discussion: Image Distillation vs. Time-Series Distillation}
\label{sec:discussion_dd}

Most existing dataset distillation methods are developed for image classification, where inputs are static, approximately i.i.d.\ samples and labels are discrete class indices. In this regime, the distilled dataset is mainly required to preserve decision boundaries in a high-dimensional feature space, and the evaluation metric is usually top-$1$ accuracy with a clear upper bound given by training on the full dataset. In contrast, time-series forecasting involves continuous-valued outputs, long-range temporal dependencies, and multi-scale patterns, which fundamentally change both the learning objective and the role of the distilled data.

A first key difference lies in the label space and diversity priors. In image classification, categorical labels naturally partition the data into classes, and most distillation objectives operate under an explicit per-class structure, which implicitly regularizes diversity within and across classes. For time-series forecasting, however, the targets are multi-step real-valued trajectories without explicit categorical priors. As discussed in our main text, synthetic sequences can easily collapse to a few redundant temporal patterns, leading to low information density even when the condensation ratio is fixed. This motivates the inter-sample regularization in our framework, which explicitly penalizes redundancy among synthetic trajectories instead of relying on class labels.

A second difference concerns temporal structure and label autocorrelation. Image distillation methods typically assume weak correlations across samples and do not model ordered dependencies within a sample. Time-series data, by contrast, exhibit strong autocorrelation and multi-scale temporal structures, so directly matching teacher--student predictions in the temporal domain can mix trend, seasonal, and high-frequency components in a single loss and amplify the bias of direct forecasting. Our frequency-domain reformulation of the value term is precisely designed to alleviate this label autocorrelation bias by aligning predictions in a decorrelated spectral space while preserving temporal fidelity.

Finally, the notion of ``optimal'' performance is less clear in time-series distillation than in image classification. In classification, distilled datasets are often judged by how closely they recover full-data accuracy, and in many cases aiming for $100\%$ accuracy is a reasonable conceptual target. For forecasting, achieving zero prediction error is unrealistic due to intrinsic stochasticity and noise in real-world signals. Our experiments show that carefully optimized synthetic datasets can sometimes even surpass models trained on the full data, reflecting the presence of redundancy and noise in modern time-series benchmarks. This observation suggests that, beyond simply mimicking full-data training, time-series distillation should be viewed as a tool for constructing compact, diverse, and denoised supervision signals tailored to temporal dynamics, rather than a direct extension of image-based distillation objectives.

\section{Practical Guidelines for Applying \mymethod}

This section summarizes practical recommendations derived from the experiments and ablations already reported in the paper, without introducing new empirical claims.

\textbf{Synthetic Sample Size.}
Across all benchmarks, using only $3$--$5$ synthetic sequences per dataset already provides strong performance at condensation ratios below $1\%$. As shown in Table~1, increasing the number of synthetic samples brings diminishing returns. We therefore suggest starting with $3$ or $5$ sequences and increasing the budget only when necessary.

\textbf{Temporal--Frequency Trade-off $\alpha$.}
The coefficient $\alpha$ controls the strength of frequency-domain alignment. Our reported results indicate that activating the frequency term ($\alpha>0$) consistently stabilizes learning and improves forecasting, while overly large values may overly bias spectral matching. Values in $[0.3, 0.8]$ work robustly, and we adopt $\alpha=0.8$ as a universal default.

\textbf{Diversity Regularization $\lambda_{\mathrm{IS}}$.}
ISIB encourages diverse synthetic trajectories by penalizing redundant patterns. We use a fixed $\lambda_{\mathrm{IS}}=0.6$ in all experiments, which provides stable improvements without requiring dataset-specific tuning. Adjustments are only necessary if synthetic sequences visibly collapse to similar shapes.

\textbf{When DDTime Helps Most.}
DDTime is compatible with many distillation frameworks, but the improvements are most pronounced when combined with trajectory-matching methods, where Parameter and Value terms are jointly optimized. For value-only distillation, DDTime still offers benefits, though typically smaller, reflecting the complementary role of multi-term optimization in TSF.

\textbf{Recommended Default Setup.}
For new forecasting tasks, a practical configuration is:
$3$--$5$ synthetic sequences per dataset,
$\alpha=0.8$ for balanced temporal--frequency alignment,
$\lambda_{\mathrm{IS}}=0.6$ for stable diversity,
and integration with a trajectory-matching framework when possible.
This setup matches all reported experiments and requires no additional tuning.

\begin{table*}[!ht] 
	\setlength{\tabcolsep}{1.35pt}
	\scriptsize
	\centering
    \caption{Extended ablation results across four benchmark datasets(PEMS03, PEMS04, PEMS07, PEMS08) under different synthetic sample sizes. \textcolor{green!40!black}{Green rows} denote results of each method augmented with \textit{CondTSF}, and the \textcolor{blue!50!black}{blue rows} denote results augmented with our proposed method. All values are averaged over three runs; lower MSE/MAE indicate better performance.}
    \label{tab::pems_abla}
	\begin{threeparttable}
    \resizebox{1\linewidth}{!}{
\begin{tabular}{c|c|c c|c c|c c|c c|c c|c c|c c|c c|c c|c c|c c}

			\toprule
			\multicolumn{2}{c}{\multirow{2}{*}{\scalebox{1.1}{Models}}}& \multicolumn{2}{c}{EDF} & \multicolumn{2}{c}{MCT} & \multicolumn{2}{c}{DATM} & \multicolumn{2}{c}{FTD} & \multicolumn{2}{c}{PP} & \multicolumn{2}{c}{TESLA} & \multicolumn{2}{c}{IDM} & \multicolumn{2}{c}{FRePo}  & \multicolumn{2}{c}{MTT} & \multicolumn{2}{c}{KIP} & \multicolumn{2}{c}{DC}\\
			\multicolumn{2}{c}{} & \multicolumn{2}{c}{\scalebox{0.8}{\citeyearpar{wang2025edf}}} & \multicolumn{2}{c}{\scalebox{0.8}{\citeyearpar{zhong2025mct}}} & \multicolumn{2}{c}{\scalebox{0.8}{\citeyearpar{guo2024datm}}} & \multicolumn{2}{c}{\scalebox{0.8}{\citeyearpar{du2023ftd}}} & \multicolumn{2}{c}{\scalebox{0.8}{\citeyearpar{li2023pp}}} & \multicolumn{2}{c}{\scalebox{0.8}{\citeyearpar{cui2023tesla}}} & \multicolumn{2}{c}{\scalebox{0.8}{\citeyearpar{zhao2023idm}}} & \multicolumn{2}{c}{\scalebox{0.8}{\citeyearpar{FRePo}}} & \multicolumn{2}{c}{\scalebox{0.8}{\citeyearpar{cazenavette2022mtt}}} & \multicolumn{2}{c}{\scalebox{0.8}{\citeyearpar{kip}}} & \multicolumn{2}{c}{\scalebox{0.8}{\citeyearpar{dc2021}}}\\

			\cmidrule(lr){3-4} \cmidrule(lr){5-6} \cmidrule(lr){7-8} \cmidrule(lr){9-10} \cmidrule(lr){11-12} \cmidrule(lr){13-14} \cmidrule(lr){15-16} \cmidrule(lr){17-18} \cmidrule(lr){19-20} \cmidrule(lr){21-22} \cmidrule(lr){23-24} 
			\multicolumn{2}{c}{Metric}& MSE & MAE & MSE & MAE & MSE & MAE & MSE & MAE & MSE & MAE & MSE & MAE & MSE & MAE & MSE & MAE & MSE & MAE & MSE & MAE & MSE & MAE \\
\end{tabular}
        }
	\end{threeparttable}
\end{table*}

\begin{table*}[!ht] 
	\setlength{\tabcolsep}{1.35pt}
	\scriptsize
	\centering
    \caption{Extended ablation results across four benchmark datasets(ETTh1, ETTh2, ETTm1, ETTm2) under different synthetic sample sizes. \textcolor{green!40!black}{Green rows} denote results of each method augmented with \textit{CondTSF}, and the \textcolor{blue!50!black}{blue rows} denote results augmented with our proposed method. All values are averaged over three runs; lower MSE/MAE indicate better performance.}
	\begin{threeparttable}
    \resizebox{1\linewidth}{!}{
\begin{tabular}{c|c|c c|c c|c c|c c|c c|c c|c c|c c|c c|c c|c c}

			\toprule
			\multicolumn{2}{c}{\multirow{2}{*}{\scalebox{1.1}{Models}}}& \multicolumn{2}{c}{EDF} & \multicolumn{2}{c}{MCT} & \multicolumn{2}{c}{DATM} & \multicolumn{2}{c}{FTD} & \multicolumn{2}{c}{PP} & \multicolumn{2}{c}{TESLA} & \multicolumn{2}{c}{IDM} & \multicolumn{2}{c}{FRePo}  & \multicolumn{2}{c}{MTT} & \multicolumn{2}{c}{KIP} & \multicolumn{2}{c}{DC}\\
			\multicolumn{2}{c}{} & \multicolumn{2}{c}{\scalebox{0.8}{\citeyearpar{wang2025edf}}} & \multicolumn{2}{c}{\scalebox{0.8}{\citeyearpar{zhong2025mct}}} & \multicolumn{2}{c}{\scalebox{0.8}{\citeyearpar{guo2024datm}}} & \multicolumn{2}{c}{\scalebox{0.8}{\citeyearpar{du2023ftd}}} & \multicolumn{2}{c}{\scalebox{0.8}{\citeyearpar{li2023pp}}} & \multicolumn{2}{c}{\scalebox{0.8}{\citeyearpar{cui2023tesla}}} & \multicolumn{2}{c}{\scalebox{0.8}{\citeyearpar{zhao2023idm}}} & \multicolumn{2}{c}{\scalebox{0.8}{\citeyearpar{FRePo}}} & \multicolumn{2}{c}{\scalebox{0.8}{\citeyearpar{cazenavette2022mtt}}} & \multicolumn{2}{c}{\scalebox{0.8}{\citeyearpar{kip}}} & \multicolumn{2}{c}{\scalebox{0.8}{\citeyearpar{dc2021}}}\\

			\cmidrule(lr){3-4} \cmidrule(lr){5-6} \cmidrule(lr){7-8} \cmidrule(lr){9-10} \cmidrule(lr){11-12} \cmidrule(lr){13-14} \cmidrule(lr){15-16} \cmidrule(lr){17-18} \cmidrule(lr){19-20} \cmidrule(lr){21-22} \cmidrule(lr){23-24}
			\multicolumn{2}{c}{Metric} & MSE & MAE & MSE & MAE & MSE & MAE & MSE & MAE & MSE & MAE & MSE & MAE & MSE & MAE & MSE & MAE & MSE & MAE & MSE & MAE & MSE & MAE \\
\end{tabular}
        }
	\end{threeparttable}
	\label{tab::ett_abla}
\end{table*}

\begin{table*}[!ht] 
	\setlength{\tabcolsep}{1.35pt}
	\scriptsize
	\centering
    \caption{Extended ablation results across four benchmark datasets(WindL1, WindL2, WindL3, WindL4) under different synthetic sample sizes. \textcolor{green!40!black}{Green rows} denote results of each method augmented with \textit{CondTSF}, and the \textcolor{blue!50!black}{blue rows} denote results augmented with our proposed method. All values are averaged over three runs; lower MSE/MAE indicate better performance.}
	\begin{threeparttable}
    \resizebox{1\linewidth}{!}{
\begin{tabular}{c|c|c c|c c|c c|c c|c c|c c|c c|c c|c c|c c|c c}

			\toprule
			\multicolumn{2}{c}{\multirow{2}{*}{\scalebox{1.1}{Models}}} & \multicolumn{2}{c}{EDF} & \multicolumn{2}{c}{MCT} & \multicolumn{2}{c}{DATM} & \multicolumn{2}{c}{FTD} & \multicolumn{2}{c}{PP} & \multicolumn{2}{c}{TESLA} & \multicolumn{2}{c}{IDM} & \multicolumn{2}{c}{FRePo}  & \multicolumn{2}{c}{MTT} & \multicolumn{2}{c}{KIP} & \multicolumn{2}{c}{DC}\\
			\multicolumn{2}{c}{} & \multicolumn{2}{c}{\scalebox{0.8}{\citeyearpar{wang2025edf}}} & \multicolumn{2}{c}{\scalebox{0.8}{\citeyearpar{zhong2025mct}}} & \multicolumn{2}{c}{\scalebox{0.8}{\citeyearpar{guo2024datm}}} & \multicolumn{2}{c}{\scalebox{0.8}{\citeyearpar{du2023ftd}}} & \multicolumn{2}{c}{\scalebox{0.8}{\citeyearpar{li2023pp}}} & \multicolumn{2}{c}{\scalebox{0.8}{\citeyearpar{cui2023tesla}}} & \multicolumn{2}{c}{\scalebox{0.8}{\citeyearpar{zhao2023idm}}} & \multicolumn{2}{c}{\scalebox{0.8}{\citeyearpar{FRePo}}} & \multicolumn{2}{c}{\scalebox{0.8}{\citeyearpar{cazenavette2022mtt}}} & \multicolumn{2}{c}{\scalebox{0.8}{\citeyearpar{kip}}} & \multicolumn{2}{c}{\scalebox{0.8}{\citeyearpar{dc2021}}}\\

			\cmidrule(lr){3-4} \cmidrule(lr){5-6} \cmidrule(lr){7-8} \cmidrule(lr){9-10} \cmidrule(lr){11-12} \cmidrule(lr){13-14} \cmidrule(lr){15-16} \cmidrule(lr){17-18} \cmidrule(lr){19-20} \cmidrule(lr){21-22} \cmidrule(lr){23-24} 
			\multicolumn{2}{c}{Metric}& MSE & MAE & MSE & MAE & MSE & MAE & MSE & MAE & MSE & MAE & MSE & MAE & MSE & MAE & MSE & MAE & MSE & MAE & MSE & MAE & MSE & MAE\\
\end{tabular}
        }
	\end{threeparttable}
	\label{tab::wind_abla}
\end{table*}

\begin{table*}[!ht] 
	\setlength{\tabcolsep}{1.35pt}
	\scriptsize
	\centering
    \caption{Extended ablation results across four benchmark datasets(Wike2000, Weather, Electricity, Traffic) under different synthetic sample sizes. \textcolor{green!40!black}{Green rows} denote results of each method augmented with \textit{CondTSF}, and the \textcolor{blue!50!black}{blue rows} denote results augmented with our proposed method. All values are averaged over three runs; lower MSE/MAE indicate better performance.}
	\begin{threeparttable}
    \resizebox{1\linewidth}{!}{
\begin{tabular}{c|c|c c|c c|c c|c c|c c|c c|c c|c c|c c|c c|c c}

			\toprule
			\multicolumn{2}{c}{\multirow{2}{*}{\scalebox{1.1}{Models}}} & \multicolumn{2}{c}{EDF} & \multicolumn{2}{c}{MCT} & \multicolumn{2}{c}{DATM} & \multicolumn{2}{c}{FTD} & \multicolumn{2}{c}{PP} & \multicolumn{2}{c}{TESLA} & \multicolumn{2}{c}{IDM} & \multicolumn{2}{c}{FRePo}  & \multicolumn{2}{c}{MTT} & \multicolumn{2}{c}{KIP} & \multicolumn{2}{c}{DC}\\
			\multicolumn{2}{c}{} & \multicolumn{2}{c}{\scalebox{0.8}{\citeyearpar{wang2025edf}}} & \multicolumn{2}{c}{\scalebox{0.8}{\citeyearpar{zhong2025mct}}} & \multicolumn{2}{c}{\scalebox{0.8}{\citeyearpar{guo2024datm}}} & \multicolumn{2}{c}{\scalebox{0.8}{\citeyearpar{du2023ftd}}} & \multicolumn{2}{c}{\scalebox{0.8}{\citeyearpar{li2023pp}}} & \multicolumn{2}{c}{\scalebox{0.8}{\citeyearpar{cui2023tesla}}} & \multicolumn{2}{c}{\scalebox{0.8}{\citeyearpar{zhao2023idm}}} & \multicolumn{2}{c}{\scalebox{0.8}{\citeyearpar{FRePo}}} & \multicolumn{2}{c}{\scalebox{0.8}{\citeyearpar{cazenavette2022mtt}}} & \multicolumn{2}{c}{\scalebox{0.8}{\citeyearpar{kip}}} & \multicolumn{2}{c}{\scalebox{0.8}{\citeyearpar{dc2021}}}\\

			\cmidrule(lr){3-4} \cmidrule(lr){5-6} \cmidrule(lr){7-8} \cmidrule(lr){9-10} \cmidrule(lr){11-12} \cmidrule(lr){13-14} \cmidrule(lr){15-16} \cmidrule(lr){17-18} \cmidrule(lr){19-20} \cmidrule(lr){21-22} \cmidrule(lr){23-24}
			\multicolumn{2}{c}{Metric}& MSE & MAE & MSE & MAE & MSE & MAE & MSE & MAE & MSE & MAE & MSE & MAE & MSE & MAE & MSE & MAE & MSE & MAE & MSE & MAE & MSE & MAE\\
\end{tabular}
    }
	\end{threeparttable}

	\label{tab::other1_abla}
\end{table*}

\begin{table*}[!ht] 
	\setlength{\tabcolsep}{1.35pt}
	\scriptsize
	\centering
    \caption{Extended ablation results across four benchmark datasets(Solar, Global Temp, ExchangeRate, ILI) under different synthetic sample sizes. \textcolor{green!40!black}{Green rows} denote results of each method augmented with \textit{CondTSF}, and the \textcolor{blue!50!black}{blue rows} denote results augmented with our proposed method. All values are averaged over three runs; lower MSE/MAE indicate better performance.}
	\begin{threeparttable}
    \resizebox{1\linewidth}{!}{
\begin{tabular}{c|c|c c|c c|c c|c c|c c|c c|c c|c c|c c|c c|c c}

			\toprule
			\multicolumn{2}{c}{\multirow{2}{*}{\scalebox{1.1}{Models}}} & \multicolumn{2}{c}{EDF} & \multicolumn{2}{c}{MCT} & \multicolumn{2}{c}{DATM} & \multicolumn{2}{c}{FTD} & \multicolumn{2}{c}{PP} & \multicolumn{2}{c}{TESLA} & \multicolumn{2}{c}{IDM} & \multicolumn{2}{c}{FRePo}  & \multicolumn{2}{c}{MTT} & \multicolumn{2}{c}{KIP} & \multicolumn{2}{c}{DC}\\
			\multicolumn{2}{c}{} & \multicolumn{2}{c}{\scalebox{0.8}{\citeyearpar{wang2025edf}}} & \multicolumn{2}{c}{\scalebox{0.8}{\citeyearpar{zhong2025mct}}} & \multicolumn{2}{c}{\scalebox{0.8}{\citeyearpar{guo2024datm}}} & \multicolumn{2}{c}{\scalebox{0.8}{\citeyearpar{du2023ftd}}} & \multicolumn{2}{c}{\scalebox{0.8}{\citeyearpar{li2023pp}}} & \multicolumn{2}{c}{\scalebox{0.8}{\citeyearpar{cui2023tesla}}} & \multicolumn{2}{c}{\scalebox{0.8}{\citeyearpar{zhao2023idm}}} & \multicolumn{2}{c}{\scalebox{0.8}{\citeyearpar{FRePo}}} & \multicolumn{2}{c}{\scalebox{0.8}{\citeyearpar{cazenavette2022mtt}}} & \multicolumn{2}{c}{\scalebox{0.8}{\citeyearpar{kip}}} & \multicolumn{2}{c}{\scalebox{0.8}{\citeyearpar{dc2021}}}\\

			\cmidrule(lr){3-4} \cmidrule(lr){5-6} \cmidrule(lr){7-8} \cmidrule(lr){9-10} \cmidrule(lr){11-12} \cmidrule(lr){13-14} \cmidrule(lr){15-16} \cmidrule(lr){17-18} \cmidrule(lr){19-20} \cmidrule(lr){21-22} \cmidrule(lr){23-24}
			\multicolumn{2}{c}{Metric}& MSE & MAE & MSE & MAE & MSE & MAE & MSE & MAE & MSE & MAE & MSE & MAE & MSE & MAE & MSE & MAE & MSE & MAE & MSE & MAE & MSE & MAE\\
\end{tabular}
    }
	\end{threeparttable}
	\label{tab::other2_abla}
\end{table*}

\begin{table*}[!ht] 
	\setlength{\tabcolsep}{1.35pt}
	\scriptsize
	\centering
    \caption{Comparison of trajectory-matching dataset distillation frameworks on 20 datasets and synthetic-set sizes $\{3,\,5,\,10,\,20\}$, evaluated by MSE and MAE. The best and second best outcomes are highlighted in \best{best} and \second{second}, respectively. The notation "$1^{\text{st}}$ \textit{Count}" denotes the frequency of each method achieving the top results.}
	\begin{threeparttable}
    \resizebox{1\linewidth}{!}{
\begin{tabular}{c|c|c c|c c|c c|c c|c c|c c|c c|c c|c c|c c|c c|c c}

			\toprule
			\multicolumn{2}{c}{\multirow{2}{*}{\scalebox{1.1}{Models}}}& \multicolumn{2}{c}{\mymethod} & \multicolumn{2}{c}{EDF} & \multicolumn{2}{c}{MCT} & \multicolumn{2}{c}{DATM} & \multicolumn{2}{c}{FTD} & \multicolumn{2}{c}{PP} & \multicolumn{2}{c}{TESLA} & \multicolumn{2}{c}{IDM} & \multicolumn{2}{c}{FRePo}  & \multicolumn{2}{c}{MTT} & \multicolumn{2}{c}{KIP} & \multicolumn{2}{c}{DC}\\
			\multicolumn{2}{c}{} & \multicolumn{2}{c}{\scalebox{0.8}{\textbf{Ours}}} & \multicolumn{2}{c}{\scalebox{0.8}{\citeyearpar{wang2025edf}}} & \multicolumn{2}{c}{\scalebox{0.8}{\citeyearpar{zhong2025mct}}} & \multicolumn{2}{c}{\scalebox{0.8}{\citeyearpar{guo2024datm}}} & \multicolumn{2}{c}{\scalebox{0.8}{\citeyearpar{du2023ftd}}} & \multicolumn{2}{c}{\scalebox{0.8}{\citeyearpar{li2023pp}}} & \multicolumn{2}{c}{\scalebox{0.8}{\citeyearpar{cui2023tesla}}} & \multicolumn{2}{c}{\scalebox{0.8}{\citeyearpar{zhao2023idm}}} & \multicolumn{2}{c}{\scalebox{0.8}{\citeyearpar{FRePo}}} & \multicolumn{2}{c}{\scalebox{0.8}{\citeyearpar{cazenavette2022mtt}}} & \multicolumn{2}{c}{\scalebox{0.8}{\citeyearpar{kip}}} & \multicolumn{2}{c}{\scalebox{0.8}{\citeyearpar{dc2021}}}\\

			\cmidrule(lr){3-4} \cmidrule(lr){5-6} \cmidrule(lr){7-8} \cmidrule(lr){9-10} \cmidrule(lr){11-12} \cmidrule(lr){13-14} \cmidrule(lr){15-16} \cmidrule(lr){17-18} \cmidrule(lr){19-20} \cmidrule(lr){21-22} \cmidrule(lr){23-24} \cmidrule(lr){25-26}
			\multicolumn{2}{c}{Metric}& MSE & MAE & MSE & MAE & MSE & MAE & MSE & MAE & MSE & MAE & MSE & MAE & MSE & MAE & MSE & MAE & MSE & MAE & MSE & MAE & MSE & MAE & MSE & MAE\\
\end{tabular}
    }
	\end{threeparttable}

	\label{tab::long-term}
\end{table*}

\begin{table*}[!ht] 

	\setlength{\tabcolsep}{1.35pt}

	\scriptsize

	\centering
    \caption{
Comparison of synthetic data distillation methods on the {Weather} dataset. 
{MSE} and {MAE} results are shown for $S\in\{3,5,10,20\}$. Our frequency domain label loss alignment approach yields consistent improvements in predictive accuracy and robustness against noise across meteorological variables.
}
	\begin{threeparttable}

    \resizebox{1\linewidth}{!}{

\begin{tabular}{c|c|c c|c c|c c|c c|c c|c c|c c|c c|c c|c c|c c}

			\toprule

			\multicolumn{2}{c}{\multirow{2}{*}{\scalebox{1.1}{Models}}}& \multicolumn{2}{c}{EDF} & \multicolumn{2}{c}{MCT} & \multicolumn{2}{c}{DATM} & \multicolumn{2}{c}{FTD} & \multicolumn{2}{c}{PP} & \multicolumn{2}{c}{TESLA} & \multicolumn{2}{c}{IDM} & \multicolumn{2}{c}{FRePo}  & \multicolumn{2}{c}{MTT} & \multicolumn{2}{c}{KIP} & \multicolumn{2}{c}{DC}\\

			\multicolumn{2}{c}{} & \multicolumn{2}{c}{\scalebox{0.8}{\citeyearpar{wang2025edf}}} & \multicolumn{2}{c}{\scalebox{0.8}{\citeyearpar{zhong2025mct}}} & \multicolumn{2}{c}{\scalebox{0.8}{\citeyearpar{guo2024datm}}} & \multicolumn{2}{c}{\scalebox{0.8}{\citeyearpar{du2023ftd}}} & \multicolumn{2}{c}{\scalebox{0.8}{\citeyearpar{li2023pp}}} & \multicolumn{2}{c}{\scalebox{0.8}{\citeyearpar{cui2023tesla}}} & \multicolumn{2}{c}{\scalebox{0.8}{\citeyearpar{zhao2023idm}}} & \multicolumn{2}{c}{\scalebox{0.8}{\citeyearpar{FRePo}}} & \multicolumn{2}{c}{\scalebox{0.8}{\citeyearpar{cazenavette2022mtt}}} & \multicolumn{2}{c}{\scalebox{0.8}{\citeyearpar{kip}}} & \multicolumn{2}{c}{\scalebox{0.8}{\citeyearpar{dc2021}}}\\

			\cmidrule(lr){3-4} \cmidrule(lr){5-6} \cmidrule(lr){7-8} \cmidrule(lr){9-10} \cmidrule(lr){11-12} \cmidrule(lr){13-14} \cmidrule(lr){15-16} \cmidrule(lr){17-18} \cmidrule(lr){19-20} \cmidrule(lr){21-22} \cmidrule(lr){23-24}

			\multicolumn{2}{c}{Metric}& MSE & MAE & MSE & MAE & MSE & MAE & MSE & MAE & MSE & MAE & MSE & MAE & MSE & MAE & MSE & MAE & MSE & MAE & MSE & MAE & MSE & MAE\\
\end{tabular}

    }

	\end{threeparttable}

	\label{tab::weather_fre}

\end{table*}
\begin{table*}[!ht] 

	\setlength{\tabcolsep}{1.35pt}

	\scriptsize

	\centering
        \caption{Forecasting performance of dataset distillation methods on the \textbf{Electricity} dataset. Each cell shows \textbf{MSE} / \textbf{MAE} for various $S$ values. Our method demonstrates clear advantages over prior baselines, maintaining stable accuracy improvements through frequency-domain teacher–student label-loss alignment.}
	\begin{threeparttable}

    \resizebox{1\linewidth}{!}{

		\begin{tabular}{c|c|c c|c c|c c|c c|c c|c c|c c|c c|c c|c c|c c}

			\toprule

			\multicolumn{2}{c}{\multirow{2}{*}{\scalebox{1.1}{Models}}}& \multicolumn{2}{c}{EDF} & \multicolumn{2}{c}{MCT} & \multicolumn{2}{c}{DATM} & \multicolumn{2}{c}{FTD} & \multicolumn{2}{c}{PP} & \multicolumn{2}{c}{TESLA} & \multicolumn{2}{c}{IDM} & \multicolumn{2}{c}{FRePo}  & \multicolumn{2}{c}{MTT} & \multicolumn{2}{c}{KIP} & \multicolumn{2}{c}{DC}\\

			\multicolumn{2}{c}{} & \multicolumn{2}{c}{\scalebox{0.8}{\citeyearpar{wang2025edf}}} & \multicolumn{2}{c}{\scalebox{0.8}{\citeyearpar{zhong2025mct}}} & \multicolumn{2}{c}{\scalebox{0.8}{\citeyearpar{guo2024datm}}} & \multicolumn{2}{c}{\scalebox{0.8}{\citeyearpar{du2023ftd}}} & \multicolumn{2}{c}{\scalebox{0.8}{\citeyearpar{li2023pp}}} & \multicolumn{2}{c}{\scalebox{0.8}{\citeyearpar{cui2023tesla}}} & \multicolumn{2}{c}{\scalebox{0.8}{\citeyearpar{zhao2023idm}}} & \multicolumn{2}{c}{\scalebox{0.8}{\citeyearpar{FRePo}}} & \multicolumn{2}{c}{\scalebox{0.8}{\citeyearpar{cazenavette2022mtt}}} & \multicolumn{2}{c}{\scalebox{0.8}{\citeyearpar{kip}}} & \multicolumn{2}{c}{\scalebox{0.8}{\citeyearpar{dc2021}}}\\

			\cmidrule(lr){3-4} \cmidrule(lr){5-6} \cmidrule(lr){7-8} \cmidrule(lr){9-10} \cmidrule(lr){11-12} \cmidrule(lr){13-14} \cmidrule(lr){15-16} \cmidrule(lr){17-18} \cmidrule(lr){19-20} \cmidrule(lr){21-22} \cmidrule(lr){23-24}

			\multicolumn{2}{c}{Metric}& MSE & MAE & MSE & MAE & MSE & MAE & MSE & MAE & MSE & MAE & MSE & MAE & MSE & MAE & MSE & MAE & MSE & MAE & MSE & MAE & MSE & MAE\\
\end{tabular}

    }

	\end{threeparttable}


	\label{tab::ecl_fre}

\end{table*}
\begin{table*}[!ht] 

	\setlength{\tabcolsep}{1.35pt}

	\scriptsize

	\centering
    \caption{
Comprehensive comparison on the {Solar} dataset. 
Each cell reports {MSE} / {MAE} under different synthetic sample sizes. 
Our approach achieves strong generalization in periodic solar generation data, demonstrating the effectiveness of frequency domain label loss matching for preserving temporal periodicity.
}
	\begin{threeparttable}

    \resizebox{1\linewidth}{!}{

\begin{tabular}{c|c|c c|c c|c c|c c|c c|c c|c c|c c|c c|c c|c c}

			\toprule

			\multicolumn{2}{c}{\multirow{2}{*}{\scalebox{1.1}{Models}}}& \multicolumn{2}{c}{EDF} & \multicolumn{2}{c}{MCT} & \multicolumn{2}{c}{DATM} & \multicolumn{2}{c}{FTD} & \multicolumn{2}{c}{PP} & \multicolumn{2}{c}{TESLA} & \multicolumn{2}{c}{IDM} & \multicolumn{2}{c}{FRePo}  & \multicolumn{2}{c}{MTT} & \multicolumn{2}{c}{KIP} & \multicolumn{2}{c}{DC}\\

			\multicolumn{2}{c}{} & \multicolumn{2}{c}{\scalebox{0.8}{\citeyearpar{wang2025edf}}} & \multicolumn{2}{c}{\scalebox{0.8}{\citeyearpar{zhong2025mct}}} & \multicolumn{2}{c}{\scalebox{0.8}{\citeyearpar{guo2024datm}}} & \multicolumn{2}{c}{\scalebox{0.8}{\citeyearpar{du2023ftd}}} & \multicolumn{2}{c}{\scalebox{0.8}{\citeyearpar{li2023pp}}} & \multicolumn{2}{c}{\scalebox{0.8}{\citeyearpar{cui2023tesla}}} & \multicolumn{2}{c}{\scalebox{0.8}{\citeyearpar{zhao2023idm}}} & \multicolumn{2}{c}{\scalebox{0.8}{\citeyearpar{FRePo}}} & \multicolumn{2}{c}{\scalebox{0.8}{\citeyearpar{cazenavette2022mtt}}} & \multicolumn{2}{c}{\scalebox{0.8}{\citeyearpar{kip}}} & \multicolumn{2}{c}{\scalebox{0.8}{\citeyearpar{dc2021}}}\\

			\cmidrule(lr){3-4} \cmidrule(lr){5-6} \cmidrule(lr){7-8} \cmidrule(lr){9-10} \cmidrule(lr){11-12} \cmidrule(lr){13-14} \cmidrule(lr){15-16} \cmidrule(lr){17-18} \cmidrule(lr){19-20} \cmidrule(lr){21-22} \cmidrule(lr){23-24}

			\multicolumn{2}{c}{Metric}& MSE & MAE & MSE & MAE & MSE & MAE & MSE & MAE & MSE & MAE & MSE & MAE & MSE & MAE & MSE & MAE & MSE & MAE & MSE & MAE & MSE & MAE\\
\end{tabular}

    }

	\end{threeparttable}


	\label{tab::solar_fre}

\end{table*}
\begin{table*}[!ht] 

	\setlength{\tabcolsep}{1.35pt}

	\scriptsize

	\centering

    \caption{Comparison of dataset distillation methods on the \textbf{Exchange} dataset. The table reports \textbf{MSE} and \textbf{MAE} under different synthetic sample sizes ($S\!=\!3,5,10,20$). Our frequency domain label loss alignment method achieves consistently lower errors and higher proportions of top-ranked (Best/Second) results across settings, confirming its robustness in capturing financial temporal dynamics.}

	\begin{threeparttable}

    \resizebox{1\linewidth}{!}{

\begin{tabular}{c|c|c c|c c|c c|c c|c c|c c|c c|c c|c c|c c|c c}

			\toprule

			\multicolumn{2}{c}{\multirow{2}{*}{\scalebox{1.1}{Models}}}& \multicolumn{2}{c}{EDF} & \multicolumn{2}{c}{MCT} & \multicolumn{2}{c}{DATM} & \multicolumn{2}{c}{FTD} & \multicolumn{2}{c}{PP} & \multicolumn{2}{c}{TESLA} & \multicolumn{2}{c}{IDM} & \multicolumn{2}{c}{FRePo}  & \multicolumn{2}{c}{MTT} & \multicolumn{2}{c}{KIP} & \multicolumn{2}{c}{DC}\\

			\multicolumn{2}{c}{} & \multicolumn{2}{c}{\scalebox{0.8}{\citeyearpar{wang2025edf}}} & \multicolumn{2}{c}{\scalebox{0.8}{\citeyearpar{zhong2025mct}}} & \multicolumn{2}{c}{\scalebox{0.8}{\citeyearpar{guo2024datm}}} & \multicolumn{2}{c}{\scalebox{0.8}{\citeyearpar{du2023ftd}}} & \multicolumn{2}{c}{\scalebox{0.8}{\citeyearpar{li2023pp}}} & \multicolumn{2}{c}{\scalebox{0.8}{\citeyearpar{cui2023tesla}}} & \multicolumn{2}{c}{\scalebox{0.8}{\citeyearpar{zhao2023idm}}} & \multicolumn{2}{c}{\scalebox{0.8}{\citeyearpar{FRePo}}} & \multicolumn{2}{c}{\scalebox{0.8}{\citeyearpar{cazenavette2022mtt}}} & \multicolumn{2}{c}{\scalebox{0.8}{\citeyearpar{kip}}} & \multicolumn{2}{c}{\scalebox{0.8}{\citeyearpar{dc2021}}}\\

			\cmidrule(lr){3-4} \cmidrule(lr){5-6} \cmidrule(lr){7-8} \cmidrule(lr){9-10} \cmidrule(lr){11-12} \cmidrule(lr){13-14} \cmidrule(lr){15-16} \cmidrule(lr){17-18} \cmidrule(lr){19-20} \cmidrule(lr){21-22} \cmidrule(lr){23-24}

			\multicolumn{2}{c}{Metric}& MSE & MAE & MSE & MAE & MSE & MAE & MSE & MAE & MSE & MAE & MSE & MAE & MSE & MAE & MSE & MAE & MSE & MAE & MSE & MAE & MSE & MAE\\
\end{tabular}

    }

	\end{threeparttable}


	\label{tab::ExchangeRate_fre}

\end{table*}

\begin{table*}[!ht] 
	\setlength{\tabcolsep}{1.35pt}
	\scriptsize
	\centering
    \caption{Ablation results of our method on multiple datasets using different distillation baselines.}
	\begin{threeparttable}
\begin{tabular}{c|c|c c|c c|c c|c c|c c|c c}
			\toprule
			\multicolumn{2}{c}{\multirow{2}{*}{\scalebox{1.1}{Models}}}
            & \multicolumn{2}{c}{EDF} 
            & \multicolumn{2}{c}{MCT} 
            & \multicolumn{2}{c}{DATM} 
            & \multicolumn{2}{c}{FTD} 
            & \multicolumn{2}{c}{PP} 
            & \multicolumn{2}{c}{MTT} \\
			\multicolumn{2}{c}{} 
            & \multicolumn{2}{c}{\scalebox{0.8}{\citeyearpar{wang2025edf}}} 
            & \multicolumn{2}{c}{\scalebox{0.8}{\citeyearpar{zhong2025mct}}} 
            & \multicolumn{2}{c}{\scalebox{0.8}{\citeyearpar{guo2024datm}}} 
            & \multicolumn{2}{c}{\scalebox{0.8}{\citeyearpar{du2023ftd}}} 
            & \multicolumn{2}{c}{\scalebox{0.8}{\citeyearpar{li2023pp}}}
            & \multicolumn{2}{c}{\scalebox{0.8}{\citeyearpar{cazenavette2022mtt}}} \\
			\cmidrule(lr){3-4} \cmidrule(lr){5-6} \cmidrule(lr){7-8} \cmidrule(lr){9-10} \cmidrule(lr){11-12} \cmidrule(lr){13-14}
			\multicolumn{2}{c}{Metric} 
            & MSE & MAE & MSE & MAE & MSE & MAE & MSE & MAE & MSE & MAE & MSE & MAE \\
			\toprule

\multirow{4}{*}{\rotatebox[origin=c]{90}{\textbf{Weather}}}
& 3 & 0.50{\tiny$\pm$0.01} & 0.50{\tiny$\pm$0.01} & 0.64{\tiny$\pm$0.05} & 0.57{\tiny$\pm$0.03} & 0.51{\tiny$\pm$0.01} & 0.50{\tiny$\pm$0.01} & 0.53{\tiny$\pm$0.02} & 0.52{\tiny$\pm$0.02} & 0.49{\tiny$\pm$0.01} & 0.49{\tiny$\pm$0.01} & 0.56{\tiny$\pm$0.01} & 0.54{\tiny$\pm$0.01}\\
& 5 & 0.59{\tiny$\pm$0.01} & 0.55{\tiny$\pm$0.01} & 0.51{\tiny$\pm$0.03} & 0.49{\tiny$\pm$0.01} & 0.48{\tiny$\pm$0.02} & 0.47{\tiny$\pm$0.01} & 0.55{\tiny$\pm$0.01} & 0.53{\tiny$\pm$0.01} & 0.57{\tiny$\pm$0.00} & 0.54{\tiny$\pm$0.00} & 0.56{\tiny$\pm$0.01} & 0.53{\tiny$\pm$0.01}\\
& 10 & 0.51{\tiny$\pm$0.00} & 0.50{\tiny$\pm$0.00} & 0.57{\tiny$\pm$0.04} & 0.53{\tiny$\pm$0.02} & 0.57{\tiny$\pm$0.01} & 0.54{\tiny$\pm$0.01} & 0.52{\tiny$\pm$0.01} & 0.51{\tiny$\pm$0.01} & 0.56{\tiny$\pm$0.01} & 0.53{\tiny$\pm$0.01} & 0.58{\tiny$\pm$0.01} & 0.54{\tiny$\pm$0.01}\\
\bottomrule
		\end{tabular}
	\end{threeparttable}
	\label{tab::abla_big}
\end{table*}

\clearpage

\onecolumn
\scriptsize
\setlength{\tabcolsep}{6pt}
\begin{longtable}{l l l l r r}
\caption{Runtime and GPU memory comparison of baseline and ours across all frameworks and datasets.}
\label{tab:runtime_memory_total} \\
\toprule
Dataset & Framework & Num\_Samples & Mode & Runtime (s) & Max GPU Memory (MB) \\
\midrule
\endfirsthead
\toprule
Dataset & Framework & Num\_Samples & Mode & Runtime (s) & Max GPU Memory (MB) \\
\midrule
\endhead
   ETTh1 &       EDF &           5 &                  baseline &         73 &                637 \\
   ETTh1 &       EDF &           5 &                      ours &         76 &                645 \\
   ETTh1 &       MCT &           5 &                  baseline &         63 &                637 \\
   ETTh1 &       MCT &           5 &                      ours &         70 &                645 \\
   ETTh1 &        DC &           5 &                  baseline &        257 &                637 \\
   ETTh1 &        DC &           5 &                      ours &        228 &                645 \\
   ETTh1 &       MTT &           5 &                  baseline &         77 &                637 \\
   ETTh1 &       MTT &           5 &                      ours &         77 &                645 \\
\bottomrule
\end{longtable}

\end{document}